\pdfoutput=1 
\documentclass[10pt,twocolumn,letterpaper]{article}

\PassOptionsToPackage{table}{xcolor}

\usepackage[pagenumbers]{cvpr} 

\def\confName{CVPR}
\def\confYear{2027}

\usepackage[utf8]{inputenc}
\usepackage[T1]{fontenc}
\usepackage{microtype}
\usepackage{textcomp}

\usepackage{amsmath,amsfonts,amssymb,amsthm}
\usepackage{mathtools}
\usepackage{nicefrac}
\usepackage{xfrac}

\usepackage{xcolor}
\usepackage{colortbl}
\definecolor{cvprblue}{rgb}{0.21,0.49,0.74}
\definecolor{gold}{HTML}{FFD966}
\definecolor{silver}{HTML}{D9D9D9}
\definecolor{bronze}{HTML}{E8C39E}
\definecolor{bgTeach}{RGB}{247,245,253}
\definecolor{bgPart}{RGB}{248,248,248}
\definecolor{bgZoneB}{RGB}{254,245,245}
\definecolor{bgZoneC}{RGB}{254,249,241}
\definecolor{bgZoneA}{RGB}{243,247,254}
\definecolor{bgZoneD}{RGB}{243,251,246}
\colorlet{dsgrey}{teal!10}
\definecolor{task1color}{RGB}{0,128,0}
\definecolor{task2color}{RGB}{255,165,0}
\definecolor{task3color}{RGB}{0,255,255}
\colorlet{myred}{red!80!black}
\colorlet{myblue}{blue!80!black}
\colorlet{mygreen}{green!60!black}
\colorlet{mydarkblue}{blue!40!black}
\definecolor{ourcolor}{RGB}{220,237,200}
\definecolor{bestcolor}{RGB}{255,215,0}

\usepackage{graphicx}
\graphicspath{{./Images/}} 
\usepackage{caption}
\usepackage{subcaption}
\usepackage{wrapfig}
\usepackage{float}
\usepackage{adjustbox}
\usepackage{tikz}
\usetikzlibrary{tikzmark,arrows.meta,backgrounds,fit,decorations.pathreplacing,patterns,shapes.geometric,3d}
\usepackage{tkz-euclide}

\usepackage{booktabs}
\usepackage{array}
\usepackage{multirow}
\usepackage{makecell}
\usepackage{longtable}
\usepackage{threeparttable}
\usepackage{tabularx}
\usepackage{stackengine}

\usepackage{balance}
\usepackage{placeins}
\usepackage{pdflscape} 
\usepackage{comment}
\usepackage{enumitem}
\usepackage{etoolbox}
\usepackage{pifont}
\usepackage{fontawesome5}
\usepackage{url}
\usepackage{cite}

\usepackage[pagebackref,breaklinks,colorlinks,allcolors=cvprblue]{hyperref}

\usepackage{algorithm}
\usepackage{algpseudocode}
\usepackage[capitalize,noabbrev]{cleveref}

\theoremstyle{plain}

\theoremstyle{definition}

\theoremstyle{remark}

\newcommand{\band}[3]{%
  \iftikzmark{#1}{\iftikzmark{#2}{%
  \begin{tikzpicture}[remember picture,overlay]
    \fill[#3] ([xshift=-0.95cm,yshift=0.30cm]pic cs:#1)
      rectangle ([xshift=\dimexpr\linewidth-0.95cm\relax,yshift=0.30cm]pic cs:#2);
  \end{tikzpicture}}{}}{}}
\newcommand{\bandlast}[3]{%
  \iftikzmark{#1}{\iftikzmark{#2}{%
  \begin{tikzpicture}[remember picture,overlay]
    \fill[#3] ([xshift=-0.95cm,yshift=0.30cm]pic cs:#1)
      rectangle ([xshift=\dimexpr\linewidth-0.95cm\relax,yshift=-0.16cm]pic cs:#2);
  \end{tikzpicture}}{}}{}}

\algrenewcommand\algorithmicrequire{\textbf{Input:}}
\algrenewcommand\algorithmicensure{\textbf{Output:}}
\newcommand{\zhead}[2]{\Statex\tikzmark{#1}\textbf{#2}}
\newcommand{\ZA}{\mathcal{Z}_A}
\newcommand{\ZB}{\mathcal{Z}_B}
\newcommand{\ZC}{\mathcal{Z}_C}
\newcommand{\ZD}{\mathcal{Z}_D}
\newcommand{\Df}{\mathcal{D}_f}
\newcommand{\Drp}{\mathcal{D}'_r}

\title{Learning with Unlearning: Synaptic Role Partitioning for Simultaneous Continual Learning and Machine Unlearning}

\author{%
  Mohammad Ali Vahedifar \quad Behzad Bozorgtabar \quad Qi Zhang\\
  DIGIT and Department of Electrical and Computer Engineering, Aarhus University, Denmark\\
  {\tt\small \{av, behzad, qz\}@ece.au.dk}
}

\begin{document}
\maketitle
\begin{figure*}[h!]
    \centering
    \includegraphics[width=.9\linewidth]{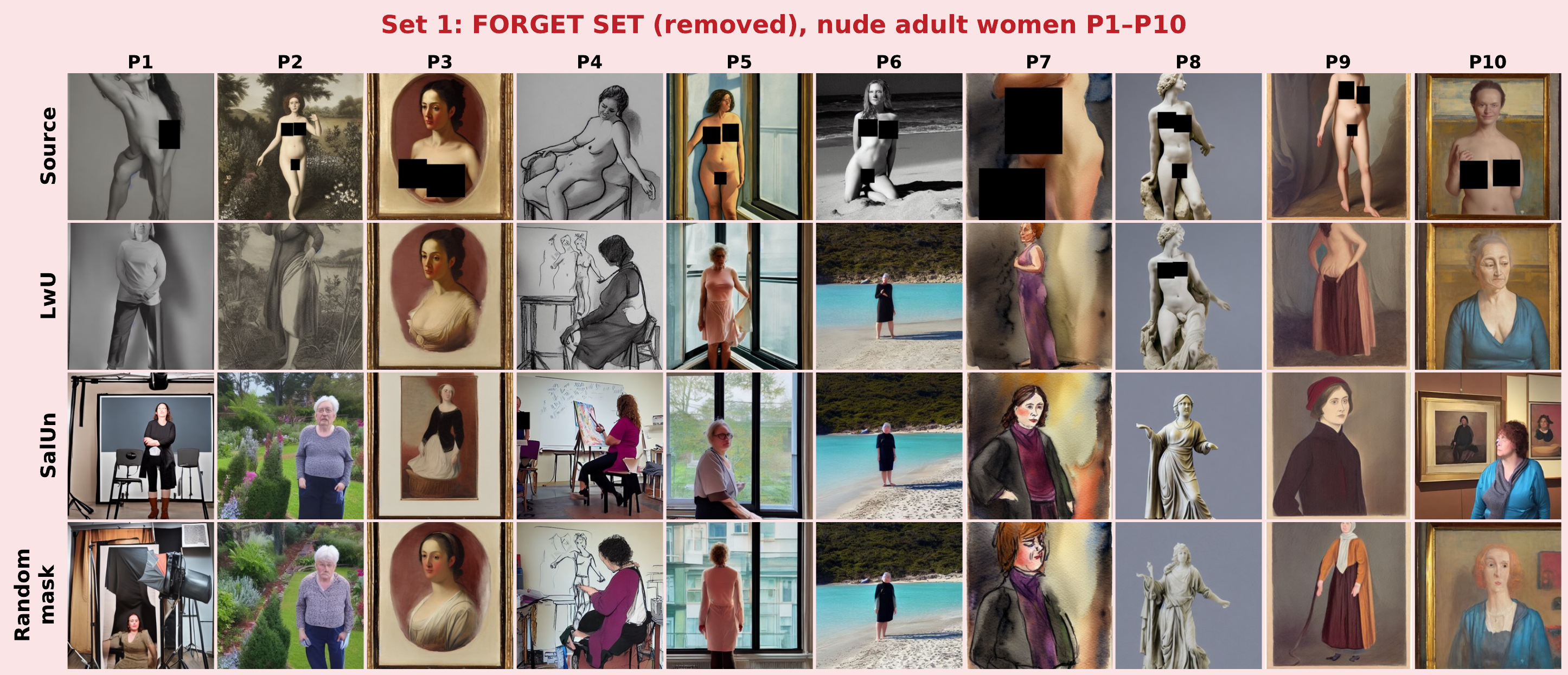}
 \caption{Suppressing nudity in text-to-image generation. Our proposed LwU is compared with the original Stable Diffusion, SalUn, and a random-mask baseline. Each column shares the same prompt across methods. Red borders indicate detected nudity; black boxes are for presentation only. The rest of this set of images is provided in Fig.~\ref{fig:fig2,3,4}.}
    \label{fig:Nudefig1}
\end{figure*}
\begin{abstract}
Continual Learning (CL) asks a model to acquire new knowledge without forgetting what it already knows. Machine Unlearning (MU) asks it to destroy specific knowledge without damaging the rest. The challenge is integrating them in the same system; the two objectives pull on the same weights, and an edit made for one harms the other. To address this, we propose Learning with Unlearning (LwU), a unified framework that resolves the conflict where it occurs: at the individual synapse. LwU scores every weight twice, once against the data to keep and once against the data to delete, and partitions the network into four zones with separate update rules. Weights that only require retention are frozen. Weights that only require deletion are cleared and returned to the network as free capacity. Contested weights take a gradient-ascent step projected orthogonally to the retention gradient. The remaining capacity learns the incoming task by a privileged teacher via self-distillation. Deletion is served post hoc, and no retraining is required.  Across both CL and MU jobs, LwU outperforms or matches the strongest baseline. Its real-world practicality has been demonstrated on the Physical Artificial Intelligence (PAI) dataset we conducted. Additionally, in the image generation task and in removing dangerous content, it outperformed the closest baseline, as shown in Fig.~\ref{fig:Nudefig1}. \textbf{\href{https://github.com/Ali-Vahedifar/Learning-with-Unlearning}{GitHub Code}}
\end{abstract}

\section{Introduction}

Neural networks are typically trained under the assumption that all training data is available upfront and remains valid indefinitely. Real deployments violate both parts of this assumption. First, data arrives sequentially, causing models to overwrite previously acquired knowledge through \emph{catastrophic forgetting}~\citep{delange2021continual}. Second, some training data may later need to be removed, requiring the model to eliminate its influence after training~\citep{ginart2019making,xu2023machineunlearningsurvey}. Continual Learning (CL) addresses the first requirement by acquiring new knowledge while preserving previously obtained knowledge~\citep{vahedifar_survey}. Machine Unlearning (MU) addresses the second requirement by handling deletion requests after training is completed, when retraining from scratch is prohibitively expensive~\citep{xu2023machineunlearningsurvey}. Consider a Physical Artificial Intelligence (PAI) system that learns sequentially from different surgeons. The system must adapt to each new operating style without losing earlier skills, while also being able to remove a patient's influence months later. By then, however, the parameters shaped by that patient may have been reused by several subsequent tasks. Supporting continual learning and data deletion in isolation is therefore insufficient: a deployed system must handle both throughout its lifetime. CL and MU cannot be combined reliably by simply applying one method after the other. Every unlearning operation changes the starting point of the next continual update, while subsequent learning may reactivate representations that the deletion step was intended to remove. Existing joint CL--MU approaches primarily manage this interaction at the task level with buffers~\citep{chatterjee2024uniclun} or at the module level~\citep {liu2022clpu}. However, retained and forgotten knowledge are often entangled within the same parameters. The central conflict, therefore, arises at the level of individual synapses: a parameter may support information that must be preserved, information that must be removed, or both.

\begin{figure*}[t]
    \centering
    \includegraphics[width=\linewidth]{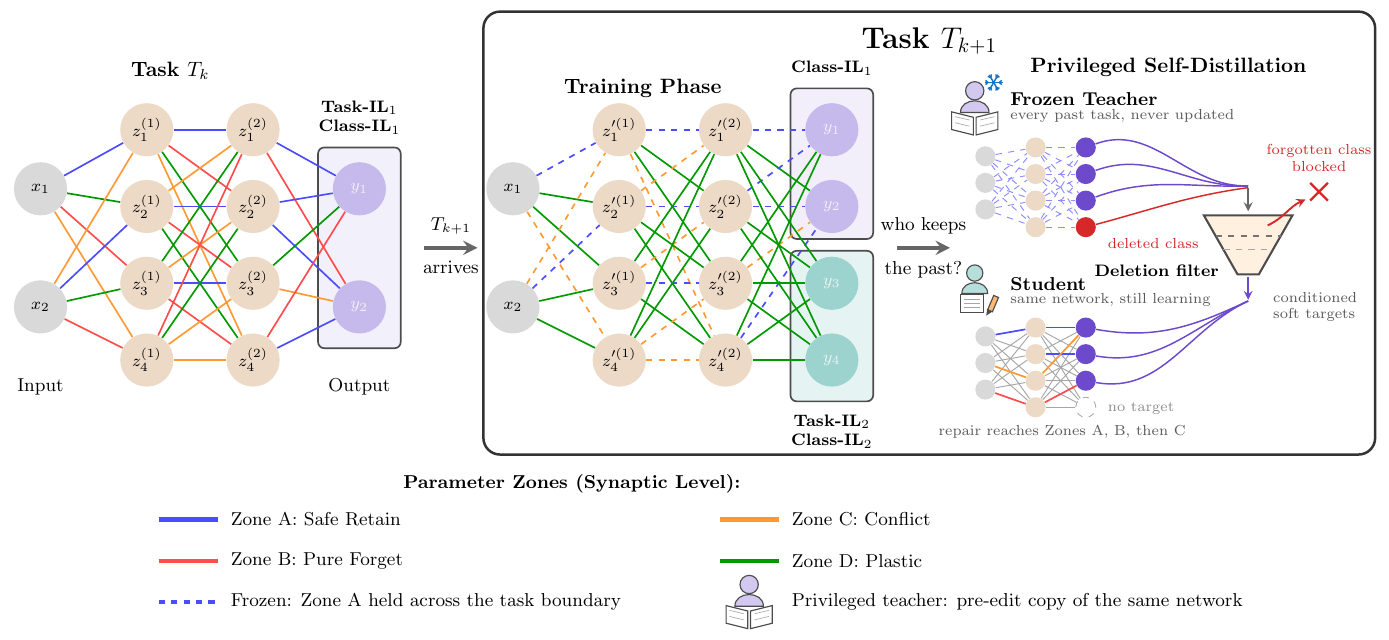}
    \caption{The LwU pipeline. \textbf{Left ($T_k$):} synapses, not neurons, are partitioned by a pair of Fisher scores into Zone A (blue, safe retain), Zone B (red, pure forget), Zone C (orange, conflict), and Zone D (green, plastic). \textbf{Right ($T_{k+1}$):} Zone A is held by the cumulative freeze mask $\mathbf{A}_{k-1}$ (dashed blue), Zone B is reinitialized and returned to the plastic pool, Zone C receives the retain-orthogonal ascent, and Zone D absorbs the new task and its new output classes.}
\label{fig:LwU_architecture}
\end{figure*}

To address the need to integrate learning and unlearning, we introduce \textit{Learning with Unlearning} (LwU), a unified framework that enables both processes within a single architecture. LwU decomposes the network into four functional subnetworks (See Fig.\ref{fig:LwU_architecture}): \textit{(i) Zone A (Safe Retain)}, which preserves parameters critical for retained data and previously learned tasks; \textit{(ii) Zone B (Pure Forget)}, which isolates parameters associated with data targeted for unlearning; \textit{(iii) Zone C (Conflict)}, where shared parameters are selectively retained or suppressed based on their importance; and \textit{(iv) Zone D (Plastic)}, which remains fully adaptable to accommodate new knowledge.
This partition separates preservation, deletion, conflict resolution, and adaptation, rather than forcing all four objectives into a single global update. We evaluate LwU against buffer-free CL methods; we examine whether its frozen and plastic zones improve the balance between learning new tasks and preserving earlier ones. Additionally, we evaluate LwU against established MU methods, to measure whether its edits remove the requested. We also compare LwU with joint CL--MU frameworks over long sequences of interleaved learning and deletion requests. Finally, we apply the same protocol to real haptic recordings from twenty participants in a PAI scenario, where surgeon-specific tasks arrive sequentially, and patient-level deletion requests occur over time. Our contributions are:
\begin{itemize}[leftmargin=*,noitemsep,topsep=0pt]
    \item \textbf{A unified framework for continual learning and unlearning.}
    LwU uses a single fixed-capacity model and a synapse-level partition for both objectives. Dual importance scores assign each parameter to one of preservation, deletion, conflict resolution, or adaptation, allowing these operations to be controlled independently.

    \item \textbf{Buffer-free continual learning.}
    LwU stores no examples from previous tasks. A frozen privileged teacher transfers permitted prior behavior to the plastic parameters, while a persistent causal memory supplies contextual information through confidence-gated pseudo-label statistics.

    \item \textbf{Post hoc, retrain-free unlearning.}
    LwU handles deletion requests after training and between continual tasks through a single model edit. Reinitializing forget-dominant parameters removes their stored information and restores their capacity, while retain-orthogonal updates reduce the influence of the forget set on contested parameters without a first-order change in the retain loss.
\end{itemize}

\textbf{Related work.} Due to space constraints, we defer our related work to Appendix~\ref{RW}.

\section{Learning with Unlearning}\label{sec:method}
\textbf{Problem formulation.} Tasks $T=\{T_k\}_{k=1}^{K}$ arrive in sequence with data $\mathcal{D}=\{\mathcal{D}_k\}_{k=1}^{K}$, where $\mathcal{D}_k=\{(x^k_i,y^k_i)\}_{i=1}^{n_k}$. We want a model $f_\theta=Wh_\theta$, with parameters $\theta$, penultimate feature $h_\theta(x)\in\mathbb{R}^d$ and classifier $W\in\mathbb{R}^{C\times d}$, that performs well on everything it has seen, removes a forget set $\mathcal{D}_f\subset\mathcal{D}$ on request, and does not catastrophically forget $\mathcal{D}_r=\mathcal{D}\setminus\mathcal{D}_f$. Deletion is post hoc: a user supplies $\mathcal{D}_f$ and a small retain subset $\mathcal{D}'_r\subset\mathcal{D}_r$, and nothing else. Buffer-free means we store \textbf{no} episodic exemplars from any previous task at any point. We define a partition at the synaptic (weights) level, \textbf{not} the neuron level.

\subsection{Zone Identification}\label{sec:zone_id}

Every zone decision answers one question. Does this weight serve the data we keep, or the data we delete? We score each parameter twice, once against each side, using the diagonal empirical Fisher information. For $s\in\{r,f\}$,
\begin{equation}\label{eq:fisher}
    \phi^i_{s}\;=\;\frac{1}{N_s}\sum_{b\subset\mathcal{D}_s}|b|
    \left(\frac{\partial\ell_b(\theta)}{\partial\theta_i}\right)^{\!2},
    \qquad N_s=\sum_{b\subset\mathcal{D}_s}|b| .
\end{equation}
The score is the expected squared gradient of the log-likelihood, so it measures how sharply the loss on $\mathcal{D}_s$ responds to a perturbation of $\theta_i$.

\subsection{Zone Construction}\label{sec:zone_construction}

\textbf{Thresholds.} Cutoffs are quantiles of the two score maps, $\tau_r=Q_{q_r}(\phi_r)$ and $\tau_f=Q_{q_f}(\phi_f)$, recomputed once per request. Quantiles rather than absolute cutoffs keep zone sizes stable across architectures and datasets.

\textbf{Dominance.} The two maps live on different scales. The forget set is small, and its gradients are large, so comparing $\phi^i_r$ against $\phi^i_f$ raw would hand almost everything to the forget side. We therefore compare scores normalized by their own thresholds,
\begin{equation}\label{eq:rho}
    \rho^i_s=\phi^i_s/\tau_s,\qquad s\in\{r,f\},
\end{equation}
so $\rho^i_s>1$ exactly when $\phi^i_s$ clears its quantile. A weight belongs to one side only when its normalized score there exceeds $m\geq1$ times its normalized score on the other.

\textbf{Zones.} Combining the thresholds with dominance gives
\begin{align}
&\textbf{Safe Retain}\Rightarrow \mathcal{Z}_A =
  \{i:\ \rho_r^i>1,\ \ \rho_r^i>m\,\rho_f^i\}, \label{eq:zone_a}\\
&\textbf{Pure Forget}\Rightarrow \mathcal{Z}_B =
  \{i:\ \rho_f^i>1,\ \ \rho_f^i>m\,\rho_r^i\}, \label{eq:zone_b}\\
&\textbf{Conflict}\Rightarrow \mathcal{Z}_C =
  \{i:\ \max(\rho_r^i,\rho_f^i)>1\}\setminus(\mathcal{Z}_A\cup\mathcal{Z}_B), \label{eq:zone_c}\\
&\textbf{Plastic}\Rightarrow \mathcal{Z}_D =
  \{i:\ \max(\rho_r^i,\rho_f^i)\leq1\}. \label{eq:zone_d}
\end{align}
The scores are nonnegative and $m\geq1$, so no weight can dominate on both sides and $\mathcal{Z}_A\cap\mathcal{Z}_B=\emptyset$; $\mathcal{Z}_C$ is whatever else clears a threshold and $\mathcal{Z}_D$ is what clears neither, so the four sets partition $\theta=\theta_A\cup\theta_B\cup\theta_C\cup\theta_D$. Zone A is what only retention needs, Zone B is what only deletion needs, Zone C is contested, and Zone D is spare capacity.
\subsection{Zone-Based Optimization}\label{sec:zone_opt}

A method that runs CL and MU at once must meet three requirements. Hold on to $\mathcal{D}_r$. Remove $\mathcal{D}_f$. Keep learning.

\textbf{Zone A: retention under deletion.} Unlearning is not scored on forget accuracy alone. Retaining accuracy is the other half, and it is the half that collapses when an edit is careless, because an aggressive unlearning step takes the rest of the model down with the deleted class. Zone A is the answer to that half. Its weights are excluded from every destructive update: no reinitialization, no ascent, and no gradient from the incoming task, enforced across task boundaries by the cumulative freeze mask $\mathbf{A}_{k-1}$. The one update Zone A does receive is the distillation repair of Section~\ref{sec:teacher}, whose target is the pre-edit model itself. Zone A can therefore be pulled back toward prior behavior, never away from it.

\textbf{Zone B: forgetting, made cheap by the continual setting (catastrophic forgetting).} Post-hoc unlearning usually cannot afford a hard edit, so it dampens instead of deleting. The continual setting removes that constraint, because deletion is not the last thing that happens to the model. A next task is already on its way, and Zone B is handed straight back to the plastic pool, so we take the hard edit and clear the weights outright with He initialization~\citep {He_2015_ICCV}:
\begin{equation}\label{eq:zone_b_reinit}
    \theta_B\leftarrow\mathrm{He}(\theta_B),\qquad \theta_B^{(\text{1-D})}\leftarrow 0 .
\end{equation}
Two forces finish what the reset starts. Training on $T_{k+1}$ writes unrelated features into those slots, so the erased capacity is occupied rather than merely blank. And ordinary catastrophic forgetting pushes whatever residue survives further away from $\mathcal{D}_f$ with each subsequent task; however, this is not guaranteed and can be seen as beneficial forgetting. The open door that makes CL hard is what makes unlearning cheap. No damping coefficient appears, because nothing has to be held in place.

\textbf{Zone C: the contested weights.} These serve both objectives and belong to neither. They have to increase the forget loss without reducing retention performance. With $g^C_r=\nabla_{\theta_C}\ell(\mathcal{D}'_r)$ and $g^C_f=\nabla_{\theta_C}\ell(\mathcal{D}_f)$, we project the ascent direction onto the orthogonal complement of the retain gradient,
\begin{equation}\label{eq:ogd_projection}
  \hspace{-2mm}  g_u^{\perp}=g^C_f-\frac{\langle g^C_f,g^C_r\rangle}{\lVert g^C_r\rVert^2+\epsilon}\,g^C_r,
    \quad
    \theta_C^{(s+1)}=\theta_C^{(s)}+\eta_C\,g_u^{\perp},
\end{equation}

and clip the cumulative displacement to $\lVert\theta_C^{(s)}-\theta_C^{(0)}\rVert\leq\delta_C$, with $\delta_C$ proportional to the mean importance in Zone C. The projected step
satisfies
\begin{equation}\label{eq:ogd_residual}
    \langle g_u^{\perp},g^C_r\rangle
    =\frac{\epsilon}{\lVert g^C_r\rVert^2+\epsilon}\,\langle g^C_f,g^C_r\rangle ,
\end{equation}
so it retains only a fraction $\epsilon/(\lVert g^C_r\rVert^2+\epsilon)$ of the raw ascent's first-order effect on the retain loss, and this fraction vanishes as $\epsilon\to0$. The stabilizer also keeps $\lVert g_u^{\perp}\rVert\leq\lVert g^C_f\rVert$, so projection can only shrink the step, never amplify it. The guarantee is local: it holds per step, to first order, with respect to the gradient on $\mathcal{D}'_r$, and while the displacement budget is inactive; when the clip binds, the rescaling toward $\theta_C^{(0)}$ can reintroduce a small retain-gradient component. Within these limits, conflict resolution needs no trade-off coefficient between the forget and retain objectives.

\textbf{Zone D: room to keep learning.} Before training on $T_k$ we release the recycled capacity, $\mathbf{A}_{k-1}\leftarrow\mathbf{A}_{k-1}\setminus\mathcal{Z}_B$, and apply the masked update
\begin{equation}\label{eq:zone_d_update}
    \theta^{(s+1)}=\theta^{(s)}-\eta\,\nabla_{\theta}\ell(\mathcal{D}_k)\odot(1-\mathbf{A}_{k-1}).
\end{equation}
Gradients reach Zones B, C, and D. Because Zone B rejoins the pool with every request, the plastic subspace grows each time something is deleted: forgetting buys plasticity rather than spending it. The three requirements now have owners. $\mathcal{D}_f$ is cleared by Zones B and C and finished off by the overwriting and drift that follow. $\mathcal{D}_r$ is protected by Zone A. New classes are learned in Zone D. What is still missing is whatever keeps the old behavior alive while Zone D writes the new one.

\subsection{The Privileged Teacher for Self Distillation}\label{sec:teacher}

Capacity alone does not give continual learning. Zone D says where new knowledge may be written; it says nothing about what survives the writing. To preserve earlier behavior without replaying earlier data, the model needs a teacher. We use its previous self. A network trained through $T_{k-1}$ already carries a compressed record of its experience in its parameters. Self-distillation (SD) makes that record a teaching signal: we freeze $\theta^T=\theta^{(k-1)}$ and let it guide the model as its parameters $\theta$ change. The teacher's privilege is having learned from data the student can no longer access. Then forcing the student to imitate the teacher's behavior without access to the teacher's privileged knowledge is self-distillation. Teacher and student share the same architecture; their difference is the knowledge each must preserve.

For continual learning, the previous model teaches what it knows. For unlearning, it must teach selectively. We therefore mask the forgotten classes $\mathcal{F}$ by setting their logits to $-\infty$ before softmax. At temperature $\kappa$, this gives, for every $c\notin\mathcal{F}$, \begin{equation}
\begin{split}
    \label{eq:teacher_mask}
    q^{\mathcal{F}}_\kappa(c\mid x;\theta^0) =\frac{q_\kappa(c\mid x;\theta^0)} {1-\sum_{c'\in\mathcal{F}}q_\kappa(c'\mid x;\theta^0)} =\\q_\kappa\big(c\mid x,\ y\notin\mathcal{F};\theta^0\big),\end{split} \end{equation} 
with zero probability for $c\in\mathcal{F}$, where $\theta^0$ is the frozen pre-edit copy. The renormalization has a precise meaning: it conditions the teacher's prediction on the label belonging to a retained class. The teacher still carries its history, but the target preserves only its relative preferences among the remaining classes. Masking removes the deleted classes' output mass, not their training influence: $\theta^0$ was trained on $\mathcal{D}_f$, so its preferences among retained classes still reflect it. For instance, in subclass requests, the deleted samples carry retained labels, so $\mathcal{F}$ does not grow, and the request adds no mask. In both cases, distillation is evaluated only on $\mathcal{D}'_r$, so the teacher never supplies a target on $\mathcal{D}_f$ itself; removal is carried by the Zone B reset and the Zone C ascent, and the teacher's role is to repair retained behavior around them. On the supplied retain subset, the student minimizes 
\begin{equation}\label{eq:sdft} 
\begin{split}
\mathcal{L}_{\mathrm{SD}}= \kappa^{2}D_{\mathrm{KL}}\!\big( q^{\mathcal{F}}_\kappa(\cdot\mid x;\theta^0) \,\big\|\,p_\kappa(\cdot\mid x;\theta)\big) \\+\lambda_{\mathrm{CE}}\, \mathrm{CE}\big(p_1(\cdot\mid x;\theta),y\big),
\end{split}
\end{equation}
where $D_{\mathrm{KL}}$ is the Kullback--Leibler divergence and $\mathrm{CE}$ is cross-entropy. Repair runs in two passes on $\mathcal{D}'_r$. The first restricts gradients to $\mathcal{Z}_A\cup\mathcal{Z}_B$ and distills from the pre-edit copy $\theta^0$: Zone A repairs collateral damage caused by the edit, while Zone B relearns the retained structure from its fresh initialization. The second, run when $\lambda_C>0$, updates Zone C alone and distills from $\theta^R$ in place of $\theta^0$, the model as repaired by the first pass, so the contested weights settle against a teacher whose Zones A and B already reflect the edit rather than against the pre-edit model. Both teachers are masked to $\mathcal{F}$. Task training reuses the teacher with a different target. After each task, and again after each deletion, the current model is frozen as $\theta^T$; the first task has no teacher and uses cross-entropy alone. While learning $T_k$, the student minimizes
\begin{equation}\label{eq:task_loss}
\begin{split}
    \mathcal{L}_{\mathrm{task}}
    =\mathrm{CE}\big(p(\cdot\mid x;\theta)_{\mathcal{O}_k},y\big)+
    \\\hspace{-6mm}\lambda_T\,\kappa_T^{2}\,
    D_{\mathrm{KL}}\!\big(q_{\kappa_T}(\cdot\mid x;\theta^T)_{\mathcal{O}^{\mathrm{old}}}
    \,\big\|\,p_{\kappa_T}(\cdot\mid x;\theta)_{\mathcal{O}^{\mathrm{old}}}\big),
    \end{split}
\end{equation}
where a subscript set means the softmax is taken over those logits only, $\mathcal{O}_k=\mathcal{Y}_k\setminus\mathcal{F}$ are the active classes, $\mathcal{O}^{\mathrm{old}}=\mathcal{Y}_{k-1}\setminus\mathcal{F}$ the retained old classes, and gradients are confined to the unfrozen weights $1-\mathbf{A}_{k-1}$. Distillation therefore covers only old classes; the new classes of $T_k$ are learned from their labels alone. The deleted set $\mathcal{F}$ is never reset at a task boundary: once a class is deleted, it is excluded from every later cross-entropy and distillation target. Self-distillation thus provides the thread connecting learning and unlearning: the previous model teaches the next one, with its targets constrained to what is allowed to remain.

\subsection{Test-Time Memorization}
The teacher helps knowledge survive changes to the model. Once deployed, however, the parameters are fixed, while the test stream may continue to change. We therefore add a lightweight fast-weight memory: each batch reads what earlier batches have written, then leaves information for those that follow. After editing and repair, we freeze the parameters $\theta^\star$, discard the teacher, and initialize $\mathbf{M}_0=\mathbf{0}\in\mathbb{R}^{C\times d}$ at the start of each evaluation stream. For a batch $X_t$, normalized feature keys $K_t=\operatorname{rowNorm}(h_{\theta^\star}(X_t))$ read the existing memory to augment the base logits $Z_t=f_{\theta^\star}(X_t)$: \begin{equation}\label{eq:predict_before_write} \widehat{Z}_t = Z_t+\beta K_t\mathbf{M}_{t-1}^{\top}, \end{equation} where $\beta$ controls the memory contribution. Only after prediction do high-confidence outputs from the base logits, selected using threshold $\rho$, produce centered pseudo-label targets. These update $\mathbf{M}_t$ through a closed-form least-squares step with momentum, decay, and row-norm projection. This predict-then-write order makes adaptation causal: information from $X_t$ can influence only future batches. The memory stores aggregate statistics rather than examples, requires no ground-truth labels, and has storage cost independent of stream length. The slow model remains unchanged, and deleted rows are zeroed after every update so that memory adds no logit to forgotten classes. With zero initialization, setting $\eta_M=0$ recovers the stored slow model's predictions.

\section{Results and Discussion}\label{Experiments}
\subsection{Continual Learning framework}
\textbf{Datasets and Architecture.} We introduced a new dataset, the Robotic Tactile Internet (RTI) dataset, containing haptic tele-manipulation for PAI scenarios, with details explained in Appendix~\ref{sec:real-world}. Additionally, we used Seq-CIFAR-100~\citep{krizhevsky2009learning} and Seq-Tiny-ImageNet~\citep{Tinyimagenet}. All methods were trained with the Adam optimizer~\citep{kingma2015adam} on ResNet-18~\citep{He_2016_CVPR} and evaluated on both Task-Incremental Learning (Task-IL) (Classifying only between the last classes, with task identity given to the head) and Class-Incremental Learning (Class-IL) (Classifying between all classes, without task identity given to the head).  We follow the Generalizable Two-phase Evaluation Protocol (GTEP)~\citep{cha2025hyperparameters}: hyperparameters are selected on half of the classes and then transferred, without further tuning, to a disjoint set of the remaining evaluation classes; phase results are averaged over 3 task orderings. Previously, the GTEP paper found that the CL method is cheating by selecting hyperparameters from a validation set that comes from the entire dataset. It is sensible, in a way, that the main assumption of CL is that we do not have access to all data and that using future data to tune hyperparameters gives your method an unfair advantage. In other words, tasks in the real world are not known beforehand.

\textbf{Baselines and metrics.} We compare against Stochastic Gradient Descent (SGD) and joint-training lower and upper bounds, respectively, as well as Elastic Weight Consolidation (EWC)~\citep{kirkpatrick2017overcoming}, Synaptic Intelligence (SI)~\citep{zenke2017continual}, Learning without Forgetting (LwF)~\citep{li2017learning}, SpaceNet~\citep{sokar2021spacenet}, Winning SubNetworks (WSN)~\citep{haeyong}, Neuro-Inspired Stability-Plasticity Adaptation (NISPA)~\citep{gurbuz2022nispa}, Prediction Error-based Classification (PEC)~\citep{PEC}, Unified Continual Learning and Machine Unlearning (UniCLUN)$^\dagger$~\citep{chatterjee2024uniclun}, and Shapley Neuron Value (SNV)~\citep{SNV}. WSN is evaluated only in Task-IL, whereas PEC is evaluated only in Class-IL. The symbol $\dagger$ identifies methods designed jointly for continual learning and machine unlearning. We report final average accuracy (ACC), backward transfer (BWT), forward transfer (FWT) and the harmonic plasticity--stability score (PS) as the primary CL metrics; they are defined in Appendix~\ref{sec:cl_metrics}.

\subsection{Machine Unlearning framework}
\textbf{Datasets and Architecture.} We evaluate class-, subclass-, and instance-level deletion on CIFAR-10 and CIFAR-100~\citep{krizhevsky2009learning}, and RTI datasets. All methods were trained separately with two backbones: ResNet-18 and a Convolutional Neural Network (CNN).

\textbf{Baselines and metrics.} We include three reference methods: \emph{Original} (or joint training), the source model trained on the complete dataset $\mathcal{D}_r\cup\mathcal{D}_f$ without unlearning; \emph{Fine-tuning}, which updates the Original model using only the accessible subset of $\mathcal{D}_r$ after removing $\mathcal{D}_f$; and \emph{Retrain}, which trains a new model from scratch using only $\mathcal{D}_r$ and serves as the gold-standard unlearning reference. We further compare LwU with established unlearning methods, including  Amnesiac~\citep{graves2021amnesiac} (evaluated only for class-level deletion), UNlearning by Selective Impair and Repair (UNSIR)~\citep{tarun2023fast}, Selective Synaptic Dampening (SSD)~\citep{SSD}, SCalable Remembering and Unlearning unBound (SCRUB)~\citep{scrub}, UniCLUN$^\dagger$, Bad Teacher~\citep{chundawat2023badteacher}, and Saliency Unlearning (SalUn)~\citep{Salun}. Every method is selected using validation data under a utility constraint relative to Retrain, and the official test set is evaluated only after selection. We report the absolute forget-accuracy gap to Retrain $|\Delta\mathcal{D}_f|$, retain and test accuracy, membership-inference attack (MIA) performance, predictive and parameter-distribution Kullback--Leibler (KL) divergence, unlearning time, and post-unlearning relearning accuracy; they are defined in Appendix~\ref{sec:mu_metrics}.

Additional experiments are provided in Appendix~\ref{Additional results} for both CL and MU jobs. Each experiment uses a single NVIDIA RTX PRO 6000 Blackwell Max-Q GPU for both CL and MU jobs.
\begin{table*}[t]
\centering
\scriptsize
\setlength{\tabcolsep}{3.2pt}
\caption{Continual-learning results. ACC and PS are percentages; BWT and FWT are percentage points. Podiums use $3G+2S+B$.}
\label{tab:cl-comprehensive}

\begin{tabular}{lcccccccc}
\toprule
& \multicolumn{4}{c}{TIL}
& \multicolumn{4}{c}{CIL} \\
\cmidrule(lr){2-5}\cmidrule(lr){6-9}
Method
& ACC$\uparrow$ & BWT$\uparrow$ & FWT$\uparrow$ & PS$\uparrow$
& ACC$\uparrow$ & BWT$\uparrow$ & FWT$\uparrow$ & PS$\uparrow$ \\
\midrule

\multicolumn{9}{>{\columncolor{dsgrey}}c}{\textbf{Seq-CIFAR-100}} \\
\midrule
Joint training (upper bound)
& $87.89_{\pm 1.25}$ & $+0.00_{\pm 0.00}$ & -- & --
& $65.43_{\pm 2.40}$ & $+0.00_{\pm 0.00}$ & -- & -- \\

\midrule
\multicolumn{9}{l}{\textit{Continual-learning methods}} \\

EWC
& $70.88_{\pm 2.65}$
& $-5.79_{\pm 1.66}$
& \cellcolor{silver}{$+1.19_{\pm 1.14}$}
& $79.35_{\pm 1.45}$
& $8.39_{\pm 0.66}$
& $-84.46_{\pm 2.30}$
& $-2.42_{\pm 0.54}$
& $26.17_{\pm 3.21}$ \\

SI
& $67.34_{\pm 1.14}$
& $-11.40_{\pm 1.55}$
& \cellcolor{bronze}{$+0.29_{\pm 0.40}$}
& $79.10_{\pm 0.79}$
& $8.56_{\pm 0.49}$
& $-85.10_{\pm 1.69}$
& $-2.38_{\pm 0.61}$
& $25.34_{\pm 2.36}$ \\

LwF
& $77.81_{\pm 2.49}$
& $-7.96_{\pm 1.57}$
& $-1.53_{\pm 3.27}$
& \cellcolor{silver}{$86.42_{\pm 1.77}$}
& $10.06_{\pm 1.79}$
& $-80.02_{\pm 1.20}$
& $-2.42_{\pm 0.54}$
& $32.15_{\pm 1.59}$ \\

WSN
& \cellcolor{bronze}{$84.70_{\pm 1.84}$}
& \cellcolor{gold}{$+0.00_{\pm 0.00}$}
& $-0.30_{\pm 1.65}$
& \cellcolor{gold}{$89.42_{\pm 1.62}$}
& -- & -- & -- & -- \\

PEC
& -- & -- & -- & --
& \cellcolor{bronze}{$32.84_{\pm 1.25}$}
& \cellcolor{gold}{$-12.70_{\pm 0.55}$}
& \cellcolor{gold}{$+0.00_{\pm 0.00}$}
& \cellcolor{bronze}{$56.43_{\pm 0.38}$} \\

SpaceNet
& $63.61_{\pm 1.32}$
& $-1.62_{\pm 0.29}$
& $-0.20_{\pm 0.69}$
& $71.68_{\pm 1.10}$
& $6.77_{\pm 0.43}$
& $-65.46_{\pm 1.45}$
& \cellcolor{bronze}{$-1.86_{\pm 0.19}$}
& $45.26_{\pm 0.86}$ \\

NISPA
& $80.02_{\pm 1.55}$
& \cellcolor{gold}{$+0.00_{\pm 0.00}$}
& $-2.70_{\pm 1.88}$
& \cellcolor{bronze}{$85.82_{\pm 1.44}$}
& $8.24_{\pm 0.54}$
& $-78.89_{\pm 1.14}$
& $-1.93_{\pm 0.26}$
& $33.30_{\pm 1.30}$ \\

SNV
& \cellcolor{gold}{$86.97_{\pm 0.73}$}
& \cellcolor{gold}{$+0.00_{\pm 0.00}$}
& $-3.47_{\pm 0.16}$
& $71.71_{\pm 1.21}$
& \cellcolor{gold}{$34.30_{\pm 0.84}$}
& \cellcolor{silver}{$-24.19_{\pm 1.92}$}
& $-2.99_{\pm 0.15}$
& \cellcolor{gold}{$59.21_{\pm 1.36}$} \\

\midrule
\multicolumn{9}{l}{\textit{Joint continual-learning and machine-unlearning methods}} \\

UniCLUN
& $69.46_{\pm 2.70}$
& $-21.36_{\pm 1.86}$
& $-0.14_{\pm 0.44}$
& $82.11_{\pm 1.46}$
& $32.75_{\pm 4.04}$
& $-58.84_{\pm 7.47}$
& $-1.91_{\pm 0.66}$
& $55.18_{\pm 6.17}$ \\

\textbf{LwU (ours)}
& \cellcolor{silver}{$85.42_{\pm 1.90}$}
& $-5.88_{\pm 0.82}$
& \cellcolor{gold}{$+1.25_{\pm 2.11}$}
& $85.27_{\pm 1.30}$
& \cellcolor{silver}{$33.22_{\pm 1.63}$}
& \cellcolor{bronze}{$-52.35_{\pm 1.12}$}
& \cellcolor{silver}{$-1.16_{\pm 0.17}$}
& \cellcolor{silver}{$59.18_{\pm 1.56}$} \\

\midrule
SGD (lower bound)
& $50.81_{\pm 2.99}$
& $-34.44_{\pm 2.83}$
& $+1.01_{\pm 2.51}$
& $70.99_{\pm 2.21}$
& $8.72_{\pm 0.70}$
& $-85.85_{\pm 1.37}$
& $-2.42_{\pm 0.54}$
& $24.29_{\pm 2.00}$ \\

\midrule
\multicolumn{9}{c}{
\textbf{Seq-CIFAR-100 podium:}\quad
\textcolor{gold}{\faMedal}\,\textbf{SNV}
(4G, 1S, 0B; \textbf{14 pts})
\quad
\textcolor{silver}{\faMedal}\,\textbf{LwU (ours)}
(1G, 4S, 1B; \textbf{12 pts})
\quad
\textcolor{bronze}{\faMedal}\,\textbf{PEC}
(2G, 0S, 2B; \textbf{8 pts})
} \\

\midrule
\multicolumn{9}{>{\columncolor{dsgrey}}c}{\textbf{Seq-Tiny-ImageNet}} \\
\midrule
Joint training (upper bound)
& $81.36_{\pm 0.42}$ & $0.00_{\pm 0.00}$ & -- & --
& $57.24_{\pm 0.53}$ & $0.00_{\pm 0.00}$ & -- & -- \\

\midrule
\multicolumn{9}{l}{\textit{Continual-learning methods}} \\

EWC
& $58.90_{\pm 1.16}$
& $-11.85_{\pm 2.13}$
& \cellcolor{bronze}{$0.57_{\pm 0.41}$}
& $75.29_{\pm 0.94}$
& $7.48_{\pm 0.26}$
& $-74.82_{\pm 0.17}$
& $-1.01_{\pm 0.27}$
& $37.71_{\pm 0.25}$ \\

SI
& $63.85_{\pm 2.14}$
& $-3.21_{\pm 1.30}$
& $-0.82_{\pm 0.49}$
& $76.23_{\pm 1.34}$
& $7.48_{\pm 0.28}$
& $-72.74_{\pm 0.10}$
& $-1.01_{\pm 0.27}$
& $39.70_{\pm 0.09}$ \\

LwF
& $67.39_{\pm 1.49}$
& $-6.46_{\pm 1.22}$
& $0.08_{\pm 0.58}$
& \cellcolor{bronze}{$80.46_{\pm 1.05}$}
& \cellcolor{bronze}{$8.04_{\pm 0.20}$}
& $-71.42_{\pm 0.49}$
& \cellcolor{bronze}{$-0.96_{\pm 0.27}$}
& $41.05_{\pm 0.33}$ \\

WSN
& \cellcolor{silver}{$73.68_{\pm 0.10}$}
& \cellcolor{silver}{$0.00_{\pm 0.00}$}
& $0.50_{\pm 0.70}$
& \cellcolor{gold}{$83.00_{\pm 0.50}$}
& -- & -- & -- & -- \\

PEC
& -- & -- & -- & --
& $2.88_{\pm 0.36}$
& \cellcolor{bronze}{$-27.15_{\pm 1.88}$}
& $-1.42_{\pm 0.10}$
& $36.72_{\pm 1.33}$ \\

SpaceNet
& $12.17_{\pm 0.46}$
& \cellcolor{gold}{$0.10_{\pm 0.20}$}
& $0.00_{\pm 0.60}$
& $0.00_{\pm 0.00}$
& $1.01_{\pm 0.01}$
& \cellcolor{gold}{$-3.40_{\pm 0.60}$}
& $-1.10_{\pm 0.10}$
& $2.30_{\pm 0.10}$ \\

NISPA
& $50.50_{\pm 3.74}$
& $-1.10_{\pm 1.50}$
& \cellcolor{gold}{$0.60_{\pm 0.90}$}
& $61.50_{\pm 4.90}$
& $5.46_{\pm 0.29}$
& $-49.70_{\pm 4.40}$
& \cellcolor{silver}{$-0.70_{\pm 0.10}$}
& \cellcolor{silver}{$49.70_{\pm 0.20}$} \\

SNV
& \cellcolor{gold}{$75.62_{\pm 0.51}$}
& \cellcolor{silver}{$0.00_{\pm 0.00}$}
& \cellcolor{gold}{$0.60_{\pm 0.10}$}
& $80.10_{\pm 0.30}$
& \cellcolor{gold}{$30.70_{\pm 1.73}$}
& \cellcolor{silver}{$-17.30_{\pm 2.40}$}
& \cellcolor{gold}{$8.00_{\pm 1.60}$}
& \cellcolor{gold}{$52.10_{\pm 1.30}$} \\

\midrule
\multicolumn{9}{l}{\textit{Joint continual-learning and machine-unlearning methods}} \\

UniCLUN
& $49.21_{\pm 9.01}$
& $-24.60_{\pm 9.60}$
& $-0.50_{\pm 0.60}$
& $71.00_{\pm 4.80}$
& $4.12_{\pm 2.66}$
& $-39.80_{\pm 6.40}$
& $-1.20_{\pm 0.00}$
& \cellcolor{bronze}{$44.30_{\pm 4.40}$} \\

\textbf{LwU (ours)}
& \cellcolor{bronze}{$71.35_{\pm 0.89}$}
& $-3.81_{\pm 0.40}$
& $-0.33_{\pm 1.68}$
& \cellcolor{silver}{$82.39_{\pm 0.92}$}
& \cellcolor{silver}{$28.60_{\pm 0.23}$}
& $-71.56_{\pm 0.57}$
& $-1.04_{\pm 0.10}$
& $40.99_{\pm 0.46}$ \\

\midrule
SGD (lower bound)
& $40.24_{\pm 3.01}$
& $-36.32_{\pm 1.33}$
& $0.88_{\pm 0.94}$
& $66.71_{\pm 1.38}$
& $7.59_{\pm 0.34}$
& $-74.35_{\pm 0.79}$
& $-1.00_{\pm 0.27}$
& $38.17_{\pm 0.76}$ \\

\midrule
\multicolumn{9}{c}{
\textbf{Seq-Tiny-ImageNet podium:}\quad
\textcolor{gold}{\faMedal}\,\textbf{SNV}
(5G, 2S, 0B; \textbf{19 pts})
\quad
\textcolor{silver}{\faMedal}\,\textbf{WSN}
(1G, 2S, 0B; \textbf{7 pts})
\quad
\textcolor{silver}{\faMedal}\,\textbf{NISPA}
(1G, 2S, 0B; \textbf{7 pts})
} \\

\midrule
\multicolumn{9}{>{\columncolor{dsgrey}}c}{\textbf{RTI}} \\
\midrule
Joint training (upper bound)
& $93.38_{\pm 1.06}$ & -- & -- & --
& $81.96_{\pm 0.75}$ & -- & -- & -- \\

\midrule
\multicolumn{9}{l}{\textit{Continual-learning methods}} \\

EWC
& $86.12_{\pm 4.63}$
& $-5.79_{\pm 5.23}$
& $-2.37_{\pm 1.65}$
& $86.15_{\pm 3.00}$
& $18.92_{\pm 0.71}$
& $-93.00_{\pm 1.54}$
& $-8.84_{\pm 8.09}$
& $12.99_{\pm 2.65}$ \\

SI
& $85.64_{\pm 2.74}$
& $-3.82_{\pm 1.08}$
& $-2.60_{\pm 1.70}$
& $84.02_{\pm 4.16}$
& $18.87_{\pm 0.98}$
& $-91.96_{\pm 1.07}$
& \cellcolor{bronze}{$-7.19_{\pm 10.90}$}
& $14.77_{\pm 1.79}$ \\

LwF
& $91.35_{\pm 1.86}$
& $-1.63_{\pm 1.01}$
& $-3.70_{\pm 6.04}$
& \cellcolor{silver}{$91.16_{\pm 1.38}$}
& $19.02_{\pm 0.87}$
& $-92.89_{\pm 1.65}$
& $-7.90_{\pm 9.62}$
& $13.17_{\pm 2.82}$ \\

WSN
& \cellcolor{silver}{$93.49_{\pm 1.19}$}
& \cellcolor{silver}{$0.00_{\pm 0.00}$}
& $-0.01_{\pm 1.81}$
& \cellcolor{gold}{$92.47_{\pm 1.23}$}
& -- & -- & -- & -- \\

PEC
& -- & -- & -- & --
& \cellcolor{bronze}{$45.94_{\pm 0.64}$}
& \cellcolor{gold}{$-19.72_{\pm 1.58}$}
& \cellcolor{silver}{$0.00_{\pm 0.00}$}
& \cellcolor{gold}{$65.83_{\pm 0.56}$} \\

SpaceNet
& $83.94_{\pm 2.29}$
& \cellcolor{gold}{$+0.18_{\pm 0.15}$}
& $-0.26_{\pm 5.31}$
& $79.80_{\pm 2.77}$
& $17.36_{\pm 1.47}$
& $-83.27_{\pm 2.18}$
& $-8.40_{\pm 1.08}$
& $27.80_{\pm 2.90}$ \\

NISPA
& $91.39_{\pm 2.03}$
& \cellcolor{silver}{$0.00_{\pm 0.00}$}
& \cellcolor{silver}{$+6.06_{\pm 4.67}$}
& $88.22_{\pm 3.55}$
& $18.39_{\pm 1.05}$
& $-90.51_{\pm 1.03}$
& $-11.49_{\pm 0.59}$
& $17.15_{\pm 1.67}$ \\

SNV
& \cellcolor{gold}{$94.60_{\pm 0.65}$}
& \cellcolor{silver}{$0.00_{\pm 0.00}$}
& \cellcolor{gold}{$+9.00_{\pm 1.74}$}
& $90.10_{\pm 1.49}$
& \cellcolor{gold}{$50.70_{\pm 2.06}$}
& $-72.10_{\pm 1.34}$
& \cellcolor{gold}{$+9.00_{\pm 1.74}$}
& \cellcolor{bronze}{$62.10_{\pm 2.41}$} \\

\midrule
\multicolumn{9}{l}{\textit{Joint continual-learning and machine-unlearning methods}} \\

UniCLUN$^{*}$
& $85.73_{\pm 1.63}$
& $-9.95_{\pm 2.46}$
& \cellcolor{bronze}{$+1.33_{\pm 1.64}$}
& $87.98_{\pm 0.96}$
& $40.74_{\pm 2.52}$
& \cellcolor{bronze}{$-63.24_{\pm 6.00}$}
& $-8.37_{\pm 6.82}$
& $51.96_{\pm 5.55}$ \\

\textbf{LwU (ours)}
& \cellcolor{bronze}{$93.41_{\pm 0.98}$}
& $-3.45_{\pm 1.41}$
& $-2.77_{\pm 3.78}$
& \cellcolor{bronze}{$90.67_{\pm 0.82}$}
& \cellcolor{silver}{$46.91_{\pm 0.89}$}
& \cellcolor{silver}{$-36.13_{\pm 0.94}$}
& $-8.11_{\pm 8.93}$
& \cellcolor{silver}{$64.72_{\pm 0.72}$} \\

\midrule
SGD (lower bound)
& $76.14_{\pm 4.63}$
& $-19.78_{\pm 4.29}$
& $-2.07_{\pm 0.54}$
& $81.84_{\pm 3.40}$
& $18.98_{\pm 0.82}$
& $-93.44_{\pm 1.16}$
& $-7.57_{\pm 9.73}$
& $12.24_{\pm 2.03}$ \\

\midrule
\multicolumn{9}{c}{
\textbf{RTI podium:}\quad
\textcolor{gold}{\faMedal}\,\textbf{SNV}
(4G, 1S, 1B; \textbf{15 pts})
\quad
\textcolor{silver}{\faMedal}\,\textbf{PEC}
(2G, 1S, 1B; \textbf{9 pts})
\quad
\textcolor{bronze}{\faMedal}\,\textbf{LwU (ours)}
(0G, 3S, 2B; \textbf{8 pts})
} \\
\midrule
\multicolumn{9}{c}{
\textbf{Global overall podium:}\quad
\textcolor{gold}{\faMedal}\,\textbf{SNV}
(13G, 4S, 1B; \textbf{48 pts})
\quad
\textcolor{silver}{\faMedal}\,\textbf{LwU (ours)}
(1G, 9S, 4B; \textbf{25 pts})
\quad
\textcolor{bronze}{\faMedal}\,\textbf{WSN}
(4G, 4S, 1B; \textbf{21 pts})
} \\
\bottomrule
\end{tabular}

\end{table*}
\subsection{Continual Learning Performance}
Table~\ref{tab:cl-comprehensive} shows that under TIL and CIL evaluation on Seq-CIFAR-100, SNV leads at $86.97$ and $34.30$, respectively. Here, LwU only wins a podium finish in third place in TIL acc with $85.42$. This shows that a subnetwork-based method that relies on Shapley-based evaluation outperforms Fisher-based zone decomposition. 

Additionally, WSN, NISPA, and SNV all report $\mathrm{BWT} = 0.00$ on all three datasets in the TIL setting. Nothing was forgotten because nothing was allowed to move: the masks make old parameters unreachable, so the number is fixed before training starts. LwU reports $-5.88$, $-3.81$, and $-3.45$ because it writes to shared parameters. Once masking stops working, the picture reverses. On Seq-CIFAR-100 in the CIL setting, SNV drops to $-24.19$ and NISPA to $-78.89$, against LwU at $-52.35$. This issue arises when the number of tasks increases, causing the mask-based method to freeze the entire network.


\begin{table*}[t]
\centering
\scriptsize
\setlength{\tabcolsep}{3.2pt}
\caption{Unlearning results on CIFAR-10, CIFAR-100, and RTI over 3 seeds. U-LiRA uses rankings based on each reported value's distance to Retrain. Podiums are weighted medal scores: 3G+2S+1B.}
\label{tab:comprehensive-main}
\resizebox{\textwidth}{!}{%
\begin{tabular}{lcccccccccc}
\toprule
& \multicolumn{3}{c}{CNN: $D_f$}
& \multicolumn{3}{c}{ResNet-18: $D_f$}
& \multicolumn{4}{c}{U-LiRA: Instance deletion} \\
\cmidrule(lr){2-4}\cmidrule(lr){5-7}\cmidrule(lr){8-11}
Method & Class & Subclass & Instance & Class & Subclass & Instance
& \multicolumn{2}{c}{CNN} & \multicolumn{2}{c}{ResNet-18} \\
\cmidrule(lr){8-9}\cmidrule(lr){10-11}
& & & & & & & AUC $\rightarrow 50$ & TPR@1\% $\downarrow$
& AUC $\rightarrow 50$ & TPR@1\% $\downarrow$ \\
\midrule
\multicolumn{11}{>{\columncolor{dsgrey}}c}{\textbf{CIFAR-10}} \\
\midrule
Retrain (oracle)
& $0.00_{\pm 0.00}$
& $25.40_{\pm 4.50}$
& $91.86_{\pm 1.34}$
& $0.00_{\pm 0.00}$
& $26.57_{\pm 1.23}$
& $92.06_{\pm 0.75}$
& $50.00$
& $0.00$
& $50.00$
& $0.00$ \\
Original model
& $93.47_{\pm 1.33}$
& $94.57_{\pm 1.17}$
& $99.94_{\pm 0.10}$
& $93.90_{\pm 1.15}$
& $95.30_{\pm 0.69}$
& $100.00_{\pm 0.00}$
& $66.80$
& $12.30$
& $80.10$
& $22.40$ \\
Fine-tune on $D_r$
& $50.17_{\pm 43.28}$
& $68.90_{\pm 2.80}$
& $96.19_{\pm 0.44}$
& $26.23_{\pm 45.44}$
& $70.07_{\pm 0.68}$
& $97.43_{\pm 2.04}$
& $50.90$
& $1.50$
& $54.80$
& $3.80$ \\
\midrule
\multicolumn{11}{l}{\textit{Unlearning methods}} \\
\midrule
Amnesiac
& \cellcolor{gold}$0.00_{\pm 0.00}$
& --
& --
& \cellcolor{gold}$0.00_{\pm 0.00}$
& --
& --
& --
& --
& --
& -- \\
Bad Teacher
& \cellcolor{silver}$0.27_{\pm 0.15}$
& \cellcolor{gold}$30.00_{\pm 4.71}$
& $98.29_{\pm 0.84}$
& \cellcolor{silver}$0.13_{\pm 0.23}$
& \cellcolor{gold}$28.40_{\pm 4.10}$
& $98.97_{\pm 0.36}$
& $88.90$
& \cellcolor{silver}$0.40$
& $76.20$
& \cellcolor{silver}$1.90$ \\
SCRUB
& \cellcolor{gold}$0.00_{\pm 0.00}$
& $90.73_{\pm 1.63}$
& \cellcolor{gold}$97.52_{\pm 0.81}$
& \cellcolor{gold}$0.00_{\pm 0.00}$
& $84.57_{\pm 0.06}$
& \cellcolor{bronze}$98.82_{\pm 0.66}$
& $60.30$
& $7.60$
& $68.10$
& $12.00$ \\
SSD
& $74.60_{\pm 22.39}$
& \cellcolor{silver}$65.57_{\pm 28.00}$
& $99.94_{\pm 0.10}$
& $92.67_{\pm 3.43}$
& $84.33_{\pm 9.85}$
& $100.00_{\pm 0.00}$
& $59.60$
& $8.30$
& $75.90$
& $18.50$ \\
SalUn
& $41.50_{\pm 35.98}$
& \cellcolor{bronze}$73.40_{\pm 3.05}$
& \cellcolor{silver}$98.02_{\pm 0.14}$
& $22.07_{\pm 38.22}$
& \cellcolor{bronze}$68.30_{\pm 4.06}$
& \cellcolor{gold}$96.25_{\pm 2.23}$
& $87.40$
& \cellcolor{gold}$0.00$
& \cellcolor{bronze}$60.30$
& \cellcolor{gold}$0.00$ \\
UNSIR
& \cellcolor{gold}$0.00_{\pm 0.00}$
& $79.13_{\pm 10.71}$
& $98.61_{\pm 0.31}$
& $43.23_{\pm 10.05}$
& $82.30_{\pm 4.86}$
& $98.88_{\pm 0.58}$
& \cellcolor{silver}$54.80$
& \cellcolor{bronze}$2.00$
& \cellcolor{silver}$57.50$
& \cellcolor{bronze}$4.30$ \\
\midrule
\multicolumn{11}{l}{\textit{Joint continual-learning and machine-unlearning methods}} \\
\midrule
UniCLUN
& \cellcolor{bronze}$41.33_{\pm 35.77}$
& $74.40_{\pm 4.29}$
& $99.12_{\pm 0.23}$
& $41.93_{\pm 36.42}$
& \cellcolor{silver}$68.10_{\pm 3.86}$
& \cellcolor{silver}$98.38_{\pm 0.80}$
& \cellcolor{bronze}$56.70$
& $4.20$
& $80.80$
& \cellcolor{gold}$0.00$ \\
\textbf{LwU} (ours)
& \cellcolor{gold}$0.00_{\pm 0.00}$
& $85.63_{\pm 15.15}$
& \cellcolor{bronze}$98.26_{\pm 1.14}$
& \cellcolor{bronze}$3.70_{\pm 6.41}$
& $89.77_{\pm 7.82}$
& $99.97_{\pm 0.05}$
& \cellcolor{gold}$54.50$
& \cellcolor{gold}$0.00$
& \cellcolor{gold}$50.90$
& \cellcolor{gold}$0.00$ \\
\midrule
\multicolumn{11}{c}{\textbf{CIFAR-10 podium:}
\quad \textcolor{gold}{\faMedal}\,\textbf{LwU (ours)}\;(5G, 0S, 2B;\ \textbf{17 pts})
\quad \textcolor{silver}{\faMedal}\,\textbf{SalUn}\;(3G, 1S, 3B;\ \textbf{14 pts})
\quad \textcolor{bronze}{\faMedal}\,\textbf{Bad Teacher}\;(2G, 4S, 0B;\ \textbf{14 pts})
} \\
\midrule
\multicolumn{11}{>{\columncolor{dsgrey}}c}{\textbf{CIFAR-100}} \\
\midrule
Retrain (oracle)
& $0.00_{\pm 0.00}$
& $23.33_{\pm 4.16}$
& $69.50_{\pm 0.60}$
& $0.00_{\pm 0.00}$
& $24.00_{\pm 2.00}$
& $69.58_{\pm 2.46}$
& $50.00$
& --
& $50.00$
& -- \\
Original model
& $70.00_{\pm 3.61}$
& $77.00_{\pm 3.46}$
& $99.14_{\pm 0.29}$
& $72.33_{\pm 3.21}$
& $80.00_{\pm 0.00}$
& $100.00_{\pm 0.00}$
& $81.63$
& $22.20$
& $92.90$
& $50.16$ \\
Fine-tune on $D_r$
& $35.00_{\pm 3.61}$
& $72.00_{\pm 7.00}$
& $98.14_{\pm 0.71}$
& $36.33_{\pm 4.51}$
& $75.33_{\pm 4.93}$
& $99.89_{\pm 0.19}$
& $81.13$
& $21.00$
& $91.90$
& $47.40$ \\
\midrule
\multicolumn{11}{l}{\textit{Unlearning methods}} \\
\midrule
Amnesiac
& \cellcolor{gold}$0.00_{\pm 0.00}$
& --
& --
& \cellcolor{gold}$0.00_{\pm 0.00}$
& --
& --
& --
& --
& --
& -- \\
Bad Teacher
& \cellcolor{silver}$0.33_{\pm 0.58}$
& \cellcolor{gold}$17.00_{\pm 4.58}$
& $96.72_{\pm 1.21}$
& \cellcolor{gold}$0.00_{\pm 0.00}$
& \cellcolor{gold}$17.00_{\pm 3.00}$
& $99.31_{\pm 0.77}$
& $79.32$
& $7.53$
& $85.17$
& $29.51$ \\
SCRUB
& $64.33_{\pm 9.02}$
& $73.33_{\pm 6.11}$
& \cellcolor{bronze}$90.86_{\pm 1.84}$
& $31.67_{\pm 9.87}$
& $79.67_{\pm 0.58}$
& $97.94_{\pm 0.98}$
& \cellcolor{silver}$70.94$
& $12.56$
& $82.67$
& $30.56$ \\
SSD
& \cellcolor{bronze}$8.00_{\pm 2.65}$
& $51.00_{\pm 30.81}$
& $97.06_{\pm 1.38}$
& $35.00_{\pm 18.03}$
& $57.67_{\pm 31.21}$
& $98.94_{\pm 0.46}$
& $76.47$
& $16.84$
& $87.87$
& $39.84$ \\
SalUn
& $23.67_{\pm 1.53}$
& \cellcolor{bronze}$37.67_{\pm 5.69}$
& \cellcolor{silver}$74.81_{\pm 2.72}$
& \cellcolor{silver}$29.00_{\pm 2.65}$
& \cellcolor{silver}$42.00_{\pm 3.61}$
& \cellcolor{gold}$84.53_{\pm 1.61}$
& $89.45$
& \cellcolor{silver}$2.98$
& \cellcolor{bronze}$75.83$
& \cellcolor{bronze}$1.36$ \\
UNSIR
& $37.33_{\pm 17.67}$
& $64.33_{\pm 0.58}$
& $96.03_{\pm 1.07}$
& $44.67_{\pm 10.21}$
& $77.33_{\pm 6.11}$
& $98.86_{\pm 0.81}$
& \cellcolor{bronze}$71.17$
& $8.84$
& $82.07$
& $26.42$ \\
\midrule
\multicolumn{11}{l}{\textit{Joint continual-learning and machine-unlearning methods}} \\
\midrule
UniCLUN
& $26.67_{\pm 3.79}$
& \cellcolor{silver}$35.67_{\pm 3.51}$
& \cellcolor{gold}$74.58_{\pm 1.04}$
& \cellcolor{bronze}$30.67_{\pm 1.53}$
& \cellcolor{bronze}$44.67_{\pm 4.16}$
& \cellcolor{silver}$85.31_{\pm 1.41}$
& $90.38$
& \cellcolor{bronze}$4.07$
& \cellcolor{silver}$74.41$
& \cellcolor{silver}$1.04$ \\
\textbf{LwU} (ours)
& \cellcolor{gold}$0.00_{\pm 0.00}$
& $50.00_{\pm 13.08}$
& $95.56_{\pm 1.34}$
& \cellcolor{gold}$0.00_{\pm 0.00}$
& $74.33_{\pm 5.51}$
& \cellcolor{bronze}$97.81_{\pm 0.41}$
& \cellcolor{gold}$55.10$
& \cellcolor{gold}$0.00$
& \cellcolor{gold}$51.60$
& \cellcolor{gold}$0.00$ \\
\midrule
\multicolumn{11}{c}{\textbf{CIFAR-100 podium:}
\quad \textcolor{gold}{\faMedal}\,\textbf{LwU (ours)}\;(6G, 0S, 1B;\ \textbf{19 pts})
\quad \textcolor{silver}{\faMedal}\,\textbf{SalUn}\;(1G, 4S, 3B;\ \textbf{14 pts})
\quad \textcolor{silver}{\faMedal}\,\textbf{UniCLUN}\;(1G, 4S, 3B;\ \textbf{14 pts})
} \\
\midrule
\multicolumn{11}{>{\columncolor{dsgrey}}c}{\textbf{RTI}} \\
\midrule
Retrain (oracle)
& $0.00_{\pm 0.00}$
& $86.33_{\pm 2.31}$
& $78.57_{\pm 0.35}$
& $0.00_{\pm 0.00}$
& $86.67_{\pm 3.21}$
& $80.99_{\pm 1.01}$
& $50.00$
& --
& $50.00$
& -- \\
Original model
& $72.87_{\pm 1.68}$
& $96.33_{\pm 2.08}$
& $99.53_{\pm 0.18}$
& $71.27_{\pm 2.14}$
& $94.67_{\pm 2.52}$
& $99.97_{\pm 0.05}$
& $79.39$
& $21.27$
& $81.67$
& $29.22$ \\
Fine-tune on $D_r$
& $28.40_{\pm 3.12}$
& $95.33_{\pm 1.53}$
& $99.36_{\pm 0.05}$
& $36.87_{\pm 5.08}$
& $95.67_{\pm 1.53}$
& $99.82_{\pm 0.15}$
& $78.11$
& $19.87$
& $81.98$
& $28.49$ \\
\midrule
\multicolumn{11}{l}{\textit{Unlearning methods}} \\
\midrule
Amnesiac
& \cellcolor{gold}$0.00_{\pm 0.00}$
& --
& --
& \cellcolor{gold}$0.00_{\pm 0.00}$
& --
& --
& --
& --
& --
& -- \\
Bad Teacher
& \cellcolor{bronze}$0.13_{\pm 0.12}$
& $45.00_{\pm 16.52}$
& $97.81_{\pm 0.40}$
& \cellcolor{gold}$0.00_{\pm 0.00}$
& $31.67_{\pm 4.16}$
& $99.01_{\pm 0.83}$
& $74.06$
& $16.64$
& \cellcolor{bronze}$73.96$
& $18.60$ \\
SCRUB
& $52.53_{\pm 9.49}$
& \cellcolor{bronze}$95.67_{\pm 1.53}$
& \cellcolor{bronze}$96.29_{\pm 2.78}$
& \cellcolor{gold}$0.00_{\pm 0.00}$
& \cellcolor{bronze}$95.00_{\pm 2.00}$
& $98.83_{\pm 0.48}$
& \cellcolor{bronze}$73.49$
& $15.13$
& \cellcolor{silver}$72.37$
& $17.49$ \\
SSD
& $37.47_{\pm 30.91}$
& \cellcolor{silver}$90.33_{\pm 4.04}$
& $99.12_{\pm 0.58}$
& \cellcolor{bronze}$36.67_{\pm 32.30}$
& \cellcolor{silver}$94.67_{\pm 2.52}$
& \cellcolor{gold}$92.16_{\pm 11.02}$
& $75.43$
& $18.09$
& $80.80$
& $28.24$ \\
SalUn
& $33.60_{\pm 2.23}$
& $64.67_{\pm 3.06}$
& \cellcolor{gold}$89.74_{\pm 1.90}$
& $38.13_{\pm 1.10}$
& $70.67_{\pm 3.06}$
& \cellcolor{silver}$94.68_{\pm 0.28}$
& $91.05$
& \cellcolor{gold}$1.73$
& $84.24$
& \cellcolor{gold}$0.40$ \\
UNSIR
& $11.80_{\pm 17.25}$
& \cellcolor{bronze}$95.67_{\pm 2.31}$
& $97.37_{\pm 0.23}$
& $51.73_{\pm 18.15}$
& $95.33_{\pm 1.53}$
& $98.71_{\pm 0.56}$
& \cellcolor{silver}$71.82$
& \cellcolor{bronze}$13.31$
& \cellcolor{gold}$70.36$
& \cellcolor{bronze}$14.27$ \\
\midrule
\multicolumn{11}{l}{\textit{Joint continual-learning and machine-unlearning methods}} \\
\midrule
UniCLUN
& $35.13_{\pm 1.27}$
& $74.33_{\pm 2.52}$
& \cellcolor{silver}$90.58_{\pm 1.55}$
& $39.33_{\pm 0.31}$
& $70.00_{\pm 2.00}$
& \cellcolor{bronze}$95.15_{\pm 1.88}$
& $91.90$
& \cellcolor{silver}$3.22$
& $85.33$
& \cellcolor{silver}$0.49$ \\
\textbf{LwU} (ours)
& \cellcolor{silver}$0.07_{\pm 0.12}$
& \cellcolor{gold}$88.00_{\pm 5.20}$
& $97.43_{\pm 1.30}$
& \cellcolor{silver}$0.07_{\pm 0.12}$
& \cellcolor{gold}$90.33_{\pm 1.53}$
& $99.62_{\pm 0.43}$
& \cellcolor{gold}$70.81$
& $14.02$
& $75.25$
& $23.07$ \\
\midrule
\multicolumn{11}{c}{\textbf{RTI podium:}
\quad \textcolor{gold}{\faMedal}\,\textbf{LwU (ours)}\;(3G, 2S, 0B;\ \textbf{13 pts})
\quad \textcolor{silver}{\faMedal}\,\textbf{SalUn}\;(3G, 1S, 0B;\ \textbf{11 pts})
\quad \textcolor{bronze}{\faMedal}\,\textbf{SCRUB}\;(1G, 1S, 4B;\ \textbf{9 pts})
} \\
\midrule
\multicolumn{11}{c}{\textbf{Overall podium:}
\quad \textcolor{gold}{\faMedal}\,\textbf{LwU (ours)}\;(14G, 2S, 3B;\ \textbf{49 pts})
\quad \textcolor{silver}{\faMedal}\,\textbf{SalUn}\;(7G, 6S, 6B;\ \textbf{39 pts})
\quad \textcolor{bronze}{\faMedal}\,\textbf{Bad Teacher}\;(6G, 5S, 2B;\ \textbf{30 pts})
} \\
\bottomrule
\end{tabular}%
}
\end{table*}

\subsection{Machine Unlearning Performance}

Table~\ref{tab:comprehensive-main} reports per-example membership auditing using U-LiRA~\citet{hayes2024inexactunlearningneedscareful}. We adopted this test because aggregate membership-inference accuracy is a weak audit: a method can, on average, be at chance while individual forgotten examples remain identifiable. For each cell, we train $N_{\mathrm{sh}}=16$ shadow models, place each forgotten example into exactly $N_{\mathrm{sh}}/2$ shadow training sets, and apply the same unlearning algorithm to each shadow model. For a forgotten example $x$, we fit Gaussians to its logit-scaled confidence under the shadows that saw it and those that did not, sharing a single variance across examples because per-example estimates from $N_{\mathrm{sh}}/2$ shadows are too noisy, and score the target model using the log-likelihood ratio. The paired Retrain model provides the negatives, so the attack tests whether the unlearned model is distinguishable from Retrain for each individual example.  

We report Area Under the Curve (AUC) and True Positive Rate (TPR) at a $1\%$ False Positive Rate (FPR), and the two tell different stories.  The attack separates these models from Retrain easily on average, but barely makes a confident individual call. Table~\ref{tab:comprehensive-main} reveals that Forget-set accuracy is a poor proxy for forgetting. On CIFAR-100 instance deletion with the CNN backbone, SalUn achieves $74.9\%$ on $\mathcal{D}_f$ against a retrain oracle of $69.5\%$, a five-point gap, and takes the silver medal. Under U-LiRA the same checkpoint scores an AUC of $89.45$. LwU sits at $95.6\%$, twenty-six points above the oracle and off the podium, and scores $55.1$. One of these two models is nearly identifiable on an example-by-example basis, but it is not the one indicated by the accuracy column. Accuracy on $\mathcal{D}_f$ records whether a label survives, not whether a membership signal does, and the two come apart as soon as the deleted examples are individually generalizable.

\textbf{Where LwU falls short.} Subclass deletion. On CIFAR-100, LwU leaves $50.0\%$ (CNN) and $74.3\%$ (ResNet-18) against an oracle of $23.3\%$ and $24.0\%$; on RTI with ResNet-18 it leaves $74.33\%$ against $24.00\%$. We read this as a limit of synaptic partitioning rather than of the update rule. A subclass shares almost all of its Fisher support with its parent class, so Zone B is close to empty, and the retain-orthogonal ascent step has little exclusive mass to act on. Parameter importance separates a subclass from its parent only weakly, and the partition inherits that weakness.
\begin{figure}[tp]
    \centering
    \includegraphics[width=\linewidth]{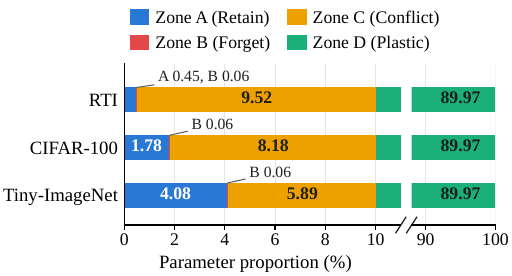}
    \caption{Final zone distributions across datasets.}
\label{fig: zone}
\end{figure}
\textbf{RTI resists everything.} Every RTI entry in the U-LiRA columns sits at least twenty points from the oracle. No method on this dataset yields a model that an attacker cannot separate from one that never saw the deleted trajectories. With 20 participants and a per-participant control signature, the deleted examples are nearly unique, and memorization is no longer distinguishable from generalization as CIFAR permits. Every method in this table was designed on data where most forget examples have near duplicates in $\mathcal{D}_r$. RTI has none, and the gap that opens there is the honest measure of how far instance-level unlearning still has to go. This confirms that most MU methods, when applied to real-world datasets, fall drastically short, and that MU methods capable of handling more challenging datasets in Robotics and PAI are needed.

\subsection{Zone Decomposition analysis:} 
Fig.~\ref{fig: zone} shows where parameters end up after the final task. On all three benchmarks, LwU uses only about 10\% of the network, leaving 89.97\% in Zone D, free for future tasks. Parameters that serve only the forget set are almost absent: Zone B holds 0.06\% everywhere, so the knowledge to be removed rarely has its own weights. Instead, it sits in Zone C, where parameters serve both the forget and retain sets, and Zone C accounts for 59–95\% of the used capacity. Pruning forget-specific weights alone would therefore touch almost nothing, which is why LwU resolves Zone C by orthogonal projection rather than deletion. The split between Zones A and C also tracks the benchmark.

\section{Conclusion} We introduced Learning with Unlearning (LwU), which combines CL and MU in a single model by deciding, weight by weight, which objective each parameter serves. Importance scores split the network into four zones. Retention is frozen; pure-forget weights are cleared and returned to the plastic pool; contested weights take an ascent step orthogonal to the retention gradient; and the remainder learns the next task with the model's previous self as the teacher. No buffer, no retraining. Across three datasets, two backbones, and both continual learning and machine unlearning tasks, LwU leads the overall podium in each, through a single mechanism.

\textbf{Limitations and Future Work.} Instance deletion on RTI is unsolved for every method we tested. This weakened the practicality of every MU method in the PAI domain. This requires more careful weight tuning and pruning, which is out of the scope for this manuscript and will be investigated in future work.

\subsection*{AI use statement}

In this work, we used generative AI tools to polish the text, fix typos, improve wording, and check grammar. We have not used generative AI tools for methodology development, dataset creation, or BibTeX reference creation. We verified and tested LLM-generated code for correctness, and it is runnable in the uploaded supplementary material. We take responsibility for the final content of this work, including text, claims, or artifacts produced with the aid of generative AI.

\subsection*{Acknowledgements}

This research was supported by the TOAST project, funded by the European Union’s Horizon Europe research and innovation program under the Marie Skłodowska-Curie Actions Doctoral Network (Grant Agreement No. 101073465), the Danish Council for Independent Research project eTouch (Grant No. 1127- 00339B), and NordForsk Nordic University Cooperation on Edge Intelligence (Grant No. 168043).

\noindent Dedicated to the memory of my father.

{
    \small
    \bibliographystyle{ieeenat_fullname}
    \bibliography{bib}
}

\clearpage
\setcounter{page}{1}
\maketitlesupplementary

\extrafloats{100}
\maxdeadcycles=1000
\setcounter{dbltopnumber}{4}
\makeatletter\providecommand\captionof[1]{\def\@captype{#1}\caption}\makeatother
\renewcommand{\dbltopfraction}{0.95}
\renewcommand{\textfraction}{0.05}
\renewcommand{\dblfloatpagefraction}{0.6}
\renewcommand{\floatpagefraction}{0.6}
\section*{Appendix Overview}
The Appendix contains additional material and experiments that do not fit within the page limit. It is organized as follows:
\begin{itemize}[leftmargin=*,itemsep=2pt,topsep=3pt]
    \item Section~\ref{RW} gives detailed related work on Continual Learning, Machine Unlearning, and unified CL--MU approaches.
    \item Section~\ref{sec:real-world} describes the real-world Physical AI scenario that motivates LwU and the collection protocol of the Robotic Tactile Internet (RTI) dataset.
    \item Section~\ref{sec:metrics} defines the evaluation metrics for CL (Section~\ref{sec:cl_metrics}) and MU (Section~\ref{sec:mu_metrics}).
    \item Section~\ref{sec:negative_results} (Limitations: ``Things We Tried That Did Not Work'') summarizes the ideas and implementations that ultimately failed.
    \item Section~\ref{Pseudo code} gives the pseudo-code of LwU over a task stream (Algorithm~\ref{alg:lwu-stream}), of one deletion request (Algorithm~\ref{alg:lwu}), and of Test-Time Memorization (Algorithm~\ref{alg:ttm}).
    \item Section~\ref{sec:lwu_diffusion} adapts LwU to conditional image generation: nudity suppression in Stable Diffusion v1.4 (Tables~\ref{tab:sd_heldout} and~\ref{tab:sd_heldout_diff}, Fig.~\ref{fig:sd-qualitative}).
    \item Section~\ref{Additional results} reports the remaining experiments:
    \begin{itemize}[leftmargin=*,itemsep=1pt,topsep=1pt]
        \item LwU in continual learning with deletion requests (Table~\ref{tab:lwu_joint}); component ablations (Tables~\ref{tab:lwu_components_c10_cnn_class}--\ref{tab:lwu_components_c100_r18_class}); sensitivity and design ablations (Tables~\ref{tab:lwu_extra_c10_cnn_class} and~\ref{tab:lwu_extra_c10_r18_class}); gap to Retrain (Table~\ref{tab:lwu_gap_summary}); importance estimators (Tables~\ref{tab:lwu_importance_c10_cnn_class} and~\ref{tab:lwu_importance_c10_r18_class}); Zone-C orthogonality (Table~\ref{tab:lwu_orth}); test-time memory (Table~\ref{tab:lwu_stream}); validation-selected configurations (Table~\ref{tab:lwu_selection});
        \item per-example privacy (Section~\ref{sec:ulira}): U-LiRA ROC curves on RTI, CIFAR-10 and CIFAR-100 (Figs.~\ref{fig:ulira-lwu-rti}--\ref{fig:ulira-lwu-cifar100}) and utility versus leakage (Figs.~\ref{fig:privacy-lwu-rti}--\ref{fig:privacy-lwu-cifar100});
        \item unlearning wall-clock cost per method (Section~\ref{sec:mu_cost}, Fig.~\ref{fig:cost});
        \item full unlearning results (Section~\ref{sec:mu_detailed}) on CIFAR-10, CIFAR-100 and RTI for the CNN (Tables~\ref{tab:cnn-class-detailed}--\ref{tab:cnn-instance-detailed}) and ResNet-18 (Tables~\ref{tab:resnet18-class-detailed}--\ref{tab:resnet18-instance-detailed}) under class, subclass and instance deletion;
        \item qualitative per-class removal on CIFAR-10 (Section~\ref{sec:mu_qualitative}, Figs.~\ref{fig:cifar10-a}--\ref{fig:cifar10-j});
        \item continual-learning cost (Section~\ref{sec:cl_cost}): training cost (Table~\ref{tab:cl-training-cost}), memory and storage for TIL and CIL (Tables~\ref{tab:cl-til-memory-cost} and~\ref{tab:cl-cil-memory-cost}), and inference cost (Table~\ref{tab:cl-inference-cost});
        \item interleaved continual learning and unlearning on Split CIFAR-100 (Section~\ref{sec:new_joint_clmu}): five-task comparison with joint baselines (Table~\ref{tab:new_joint_baselines}), retain-set provenance (Table~\ref{tab:new_request_subset}), deletion timing (Table~\ref{tab:new_deletion_schedule}), the twenty-task stream (Table~\ref{tab:new_long_stream}), and mechanism ablations (Tables~\ref{tab:new_cl_only_ablation} and~\ref{tab:new_joint_ablation});
        \item the unified U-LiRA privacy audit (Section~\ref{sec:new_unified_privacy}, Tables~\ref{tab:new_privacy_instance} and~\ref{tab:new_privacy_class});
        \item deletion, repair, and subsequent learning (Section~\ref{sec:new_repair_trajectory}, Tables~\ref{tab:new_class_repair}--\ref{tab:new_participant_10seed});
        \item additional CIFAR-100 unlearning ablations (Section~\ref{sec:new_cifar100_ablation}, Tables~\ref{tab:new_c100_cnn_class}--\ref{tab:new_c100_cnn_subclass});
        \item test-time memory latency (Section~\ref{sec:new_latency}, Table~\ref{tab:new_latency}).
    \end{itemize}
    \item Section~\ref{Hyperparmeters} gives the unlearning hyperparameter grids, training recipe and most frequently selected values (Tables~\ref{tab:mu-hparams}--\ref{tab:hp-mode}), the per-seed selected configurations on CIFAR-10, RTI and CIFAR-100 (Tables~\ref{tab:hp-selected-cifar10}--\ref{tab:hp-selected-cifar100}), and the continual-learning search spaces and selected configurations on Seq-CIFAR-100, Seq-Tiny-ImageNet and RTI (Tables~\ref{tab:cifar100-hyperparameters}--\ref{tab:rti-hyperparameters}).
\end{itemize}

\section{Related Work}\label{RW}
\textbf{Continual Learning} methods aim to mitigate catastrophic forgetting when learning from a sequence of tasks. Memory-based approaches store past samples and replay them during training~\citep{rebuffi2017icarl, buzzega2020dark, Douillard_DyTox, zhou2023model}. While effective, such methods suffer from key limitations: (i) \emph{privacy concerns}, as storing raw data may violate regulations such as the General Data Protection Regulation (GDPR)~\citep{voigt2017gdpr}; (ii) \emph{scalability issues}, since memory requirements grow with the number of tasks; and (iii) \emph{computational overhead} due to repeated replay.

Buffer-free methods avoid storing past data by relying on regularization or architectural constraints. For instance, Elastic Weight Consolidation (EWC)~\citep{kirkpatrick2017overcoming} and Synaptic Intelligence (SI)~\citep{zenke2017continual} restrict updates to important parameters, while Learning without Forgetting (LwF)~\citep{li2017learning} distills knowledge from previous model outputs. More recent approaches introduce structural mechanisms, such as sparsity-based allocation (SpaceNet)~\citep{sokar2021spacenet}, Neuro-Inspired Stability-Plasticity Adaptation (NISPA) ~\citep{gurbuz2022nispa}, Winning SubNetwork (WSN)~\citep{haeyong}, or Prediction Error-based Classification (PEC)~\citep{PEC}. These approaches often rely on additional components (e.g., masks, expansions, or constrained embeddings), which limit flexibility and scalability.

In contrast, we aim for a \textbf{buffer-free} and \textbf{architecture-expansion-free} approach that operates within a fixed model capacity. Furthermore, existing CL methods typically treat forgetting as undesirable. We instead explore \emph{beneficial forgetting}, in which the selective removal of knowledge is advantageous, particularly in the context of Machine Unlearning.

\textbf{Machine Unlearning} aims to remove the influence of specific data from trained models. Existing approaches can be broadly categorized into \emph{data-centric} and \emph{model-centric} methods. Data reorganization methods manipulate training data to facilitate unlearning. Data obfuscation techniques inject noise or modify labels to suppress the effect of target samples~\citep{tarun2023fast, felps2021classclown, graves2021amnesiac}. Data pruning methods partition datasets and retrain only affected subsets, reducing computational cost~\citep{bourtoule2021machine, gupta2021adaptive, neel2021descent}. Data replacement strategies modify training samples during learning to enable efficient future removal~\citep{cao2015towards}. Model-centric methods operate directly on trained models. Model shifting updates parameters to counteract the influence of removed data~\citep{golatkar2020eternal, guo2020certified}. Model replacement substitutes affected components with precomputed alternatives~\citep{schelter2021hedgecut, brophy2021randomforests}, while model pruning removes parameters associated with target data followed by light fine-tuning~\citep{liu2021federaser, wang2022federated}.

Due to the prohibitive cost of retraining large models, we focus on \textbf{post hoc unlearning}~\citep{SSD}, which removes the influence of data after training without retaining additional information during learning. This motivates the integration of a post hoc Machine Unlearning approach into our LwU framework. While addressing Continual Learning tasks, we additionally propose \textbf{retraining-free} and \textbf{test-time updatable} model for Machine Unlearning, aiming to reduce the computational cost of retraining from scratch while enhancing adaptation at inference time.

\textbf{Unified Continual Learning and Machine Unlearning.}
A small but growing body of work recognizes that CL and MU are complementary facets of the same lifecycle problem.
\citet{liu2022clpu} formalized the \emph{Continual Learning and Private Unlearning} (CLPU) problem and proposed CLPU-DER++, which achieves exact unlearning by training independent sub-networks per task and discarding them upon request. While providing formal privacy guarantees, this strategy incurs linear growth in the number of parameters with the number of tasks, making it unsustainable for long task sequences.
\citet{chatterjee2024uniclun} introduced UniCLUN, a controlled knowledge distillation framework in which a student model is concurrently guided by multiple teachers to balance new-task acquisition with targeted forgetting. UniCLUN demonstrated that a single distillation mechanism can serve both objectives; however, it relies on a fixed replay buffer and a separate teacher per learning/unlearning phase, reintroducing the memory and computational overheads that buffer-free CL methods seek to eliminate.
More recently, \citet{peng2025dereid} proposed De-ReID for person de-reidentification, applying a variation-guided identity shift mechanism to unlearn specific individuals from re-identification models while preserving the ability to match retained persons. Although De-ReID addresses a domain-specific instance of joint learning and unlearning, it operates on a single task and does not handle the sequential multi-task setting central to CL.

\section{Real-World Physical AI Scenario} \label{sec:real-world}

\textbf{Physical AI over the Tactile Internet.} Physical Artificial Intelligence (PAI) integrates learning-enabled models with embodied systems that perceive, reason, and interact with the physical world. A representative example is robotic teleoperation, where a human operator remotely controls a robot while receiving haptic feedback. The Tactile Internet supports such bidirectional interaction by transmitting control commands and touch sensations over long distances, enabling mission-critical applications such as remote surgery and robotic manipulation~\citep{LeFo}. Because haptic devices typically operate at sampling rates of 1\,kHz or higher, the real-time control and inference path must meet an extremely low-latency budget, commonly targeted at approximately 1\,ms~\citep{vahedifar2025shapleyfeaturesrobustsignal}. A deployed PAI system is not trained only once. It continually receives new interaction data as operators, environments, and procedures change. At the same time, some previously observed data may later need to be removed in response to privacy or data-governance requirements, such as requests motivated by the GDPR right to erasure~\citep{voigt2017gdpr}. This creates a lifecycle in which continual learning and selective unlearning must coexist: the system must acquire new physical skills without overwriting permitted prior knowledge, while removing the influence of requested data without interrupting deployment or retraining the entire model.

\textbf{Motivating PAI scenario.} Consider a learning-enabled telesurgery system deployed over several months. In the first month, the system observes procedures performed by Surgeon A and learns the corresponding interaction patterns. In the second month, Surgeon B begins using the system, requiring it to adapt to a different operating style without forgetting the knowledge acquired from Surgeon A. In the third month, a patient whose data contributed to an earlier procedure requests removal, while Surgeon C simultaneously begins generating new interaction data. The system must therefore remove the influence of the requested patient, preserve the remaining surgical knowledge, and continue to adapt to new operators.

This scenario imposes three requirements:

\begin{enumerate}[leftmargin=*,noitemsep,topsep=2pt] \item \textbf{Continual adaptation:} the model must learn from new operators and procedures without catastrophically forgetting previously acquired knowledge;

 \item \textbf{Selective unlearning:} the model must remove the influence of requested patient-level data while preserving knowledge derived from other patients and operators; and

 \item \textbf{Real-time inference:} the deployed prediction path must remain compatible with the latency requirements of the haptic control loop. \end{enumerate}

The latency requirement applies to online inference and adaptation, not to the complete unlearning procedure, which does not need to finish within one millisecond. Nevertheless, an unlearning request should not require full retraining or prolonged interruption of the deployed system. Replay-based continual learning is also undesirable because a replay buffer may retain precisely the patient data that must later be deleted. This setting therefore provides a practical motivation for a fixed-capacity, buffer-free framework that supports both continual learning and post-hoc unlearning.

\textbf{Robotic Tactile Internet dataset.} To study this setting, we collected a Robotic Tactile Internet (RTI) dataset containing haptic tele-manipulation signals from a virtual client--server environment implemented in CHAI3D. The experimental platform follows the IEEE~1918.1.1 architecture and uses a Novint Falcon haptic device operating at a 1\,kHz control loop. Twenty anonymized participants aged 20 to 33 controlled a virtual tool while performing five actions: no movement, pressing, tapping, free-air movement, and dragging.

Each action was performed for 10 seconds, and every participant repeated the complete five-action sequence four times, giving 20 action segments per participant. Each participant's record contains human-side video, robot-side video, virtual-environment image frames, and the command signals sent from the human side and the feedback signals returned from the robot side.  Each participant forms one task whose five classes are that participant's actions, so the full label space has $20\times5=100$ classes; we evaluate Class-IL and Task-IL. We consider three levels of deletion. \emph{Class deletion} removes all data of one action of one participant, or all five actions for participant-level removal, so the label leaves the output space. \emph{Subclass deletion} removes one of the four repetitions of an action while keeping the label. \emph{Instance deletion} removes individual windows, frames, or video clips of a specific participant. Participant-specific streams emulate sequentially arriving operators and patient- or participant-level removal requests, so learning and deletion interleave throughout the model's operational lifetime, as in a deployed PAI system.

The RTI dataset and its accompanying documentation will be made publicly available following completion of the required legal, ethical, and institutional reviews and confirmation that its release complies with applicable participant consent and data protection requirements.

\section{Evaluation Metrics}\label{sec:metrics}
\subsection{Continual-Learning Metrics} \label{sec:cl_metrics}

Let $K$ denote the number of tasks. After completing every task $T_i$, we evaluate the model on every task $T_j$, including tasks not yet encountered. This produces the accuracy matrix $\mathbf{A}\in\mathbb{R}^{K\times K}$, where \begin{equation} A_{i,j} = \text{accuracy on task }T_j \text{ after completing task }T_i. \end{equation} Accuracies are represented as proportions in the following definitions.

\textbf{Average Accuracy (ACC).} ACC measures final performance across all tasks: \begin{equation} \operatorname{ACC} = \frac{1}{K}\sum_{j=1}^{K}A_{K,j}. \end{equation} Higher ACC indicates better overall performance at the end of the task sequence.

\textbf{Backward Transfer (BWT).} BWT measures how subsequent learning changes performance on previously learned tasks: \begin{equation} \operatorname{BWT} = \frac{1}{K-1} \sum_{j=1}^{K-1} \left(A_{K,j}-A_{j,j}\right). \end{equation} Negative BWT indicates forgetting, zero indicates no net change, and positive BWT indicates that later tasks improve earlier performance.

\textbf{Forward Transfer (FWT).} FWT measures whether learning previous tasks improves performance on a new task before that task is trained: \begin{equation} \operatorname{FWT} = \frac{1}{K-1} \sum_{j=2}^{K} \left(A_{j-1,j}-b_j\right), \end{equation} where $b_j$ is the accuracy of a randomly initialized model on task $T_j$.

\textbf{Plasticity.} Plasticity measures the normalized improvement produced by learning each new task: \begin{equation} P = \frac{1}{K-1} \sum_{j=2}^{K} \frac{A_{j,j}-A_{j-1,j}} {1-A_{j-1,j}+\varepsilon}, \end{equation} where $\varepsilon>0$ prevents division by zero. Higher $P$ indicates a greater ability to acquire new knowledge.

\textbf{Stability.} Stability measures how well performance on earlier tasks is preserved: \begin{equation} S = 1- \frac{1}{K-1} \sum_{j=1}^{K-1} \left(A_{j,j}-A_{K,j}\right). \end{equation} Under the definition of BWT above, $S=1+\operatorname{BWT}$. Higher values indicate less forgetting.

\textbf{Plasticity--Stability Score (PS).} We summarize the balance between plasticity and stability using their harmonic mean: \begin{equation} \operatorname{PS} = \frac{2\,P\,S}{P+S+\varepsilon}. \end{equation} Because the symbol $PS$ appears on both sides if written this way, we denote the reported score by $\operatorname{PS}$ and the two constituent quantities by $P$ and $S$. A high score requires both successful acquisition and retention.

\subsection{Machine-Unlearning Metrics} \label{sec:mu_metrics}

Let $f_u$ denote the unlearned model and $f_{\mathrm{rt}}$ the Retrain oracle.

\textbf{Forget Accuracy.} Forget accuracy is the classification accuracy of $f_u$ on $\mathcal{D}_f$. Its desired value depends on the deletion setting and therefore should not be minimized blindly. Instead, it is compared with the accuracy of Retrain.

\textbf{Absolute Forget-Accuracy Gap.} Our primary unlearning metric is \begin{equation} |\Delta\mathcal{D}_f| = \left| \operatorname{Acc}_{\mathcal{D}_f}(f_u) - \operatorname{Acc}_{\mathcal{D}_f}(f_{\mathrm{rt}}) \right|. \end{equation} A smaller value indicates closer agreement with the model trained without $\mathcal{D}_f$.

\textbf{Retain and Test Accuracy.} Retain accuracy measures utility on $\mathcal{D}_r$, while test accuracy measures generalization on the official test set. These metrics ensure that agreement on $\mathcal{D}_f$ is not obtained by indiscriminately damaging the model.

\textbf{Membership-Inference Attack.} We apply a confidence-threshold membership-inference attack to determine whether forgotten examples remain distinguishable from non-members. The score orientation and decision threshold are fitted on dedicated calibration splits disjoint from the selection and final-evaluation splits. We report attack accuracy and area under the receiver operating characteristic curve (AUC). Values near 50\% attack accuracy and 0.5 AUC indicate chance-level discrimination, although they do not by themselves prove successful unlearning.

\textbf{Predictive KL Divergence.} Predictive KL measures the difference between the output distributions of the unlearned and retrained models: \begin{equation} D_{\mathrm{out}} = \frac{1}{|\mathcal{E}|} \sum_{x\in\mathcal{E}} D_{\mathrm{KL}} \left( p_u(\cdot\mid x) \,\middle\|\, p_{\mathrm{rt}}(\cdot\mid x) \right), \end{equation} where $\mathcal{E}$ is the specified evaluation set. Lower values indicate closer predictive behavior.

\textbf{Weight-Histogram KL Divergence.} We construct normalized parameter histograms $P_u$ and $P_{\mathrm{rt}}$ using identical bins and compute \begin{equation} D_{\mathrm{weight}} = D_{\mathrm{KL}} \left(P_u\,\|\,P_{\mathrm{rt}}\right). \end{equation} This metric compares aggregate weight distributions and does not establish parameter-wise or functional equivalence.

\textbf{Unlearning Time.} Unlearning time is the wall-clock time required to transform the source checkpoint into the final unlearned model. Source training and final evaluation are excluded.

\textbf{Relearning Probe.} To test how readily forgotten information can be recovered, we create a copy of the final model and optimize it on $\mathcal{D}_f$ for five epochs using Adam with a learning rate of $10^{-4}$. We report forget-set accuracy after each epoch and compare the resulting curve with Retrain. The original evaluated model remains unmodified.

\textbf{Attack Metrics.}
Each attack assigns a score $s\in\mathbb{R}$, with larger values indicating membership. Let $\mathcal{S}^{+}$ and $\mathcal{S}^{-}$ be the multisets of scores on positives and negatives. For a threshold $\tau$, the true-positive rate (TPR) and false-positive rate (FPR) are
\begin{align}
\operatorname{TPR}(\tau)=\frac{1}{|\mathcal{S}^{+}|}\sum_{s\in\mathcal{S}^{+}}\mathbf{1}[s\ge\tau],
\\
\operatorname{FPR}(\tau)=\frac{1}{|\mathcal{S}^{-}|}\sum_{s\in\mathcal{S}^{-}}\mathbf{1}[s\ge\tau].
\end{align}
The receiver operating characteristic (ROC) curve traces $\big(\operatorname{FPR}(\tau),\operatorname{TPR}(\tau)\big)$ as $\tau$ varies. The area under it (AUC) is the probability that a random positive outscores a random negative:
\begin{equation}
\operatorname{AUC}=\frac{1}{|\mathcal{S}^{+}|\,|\mathcal{S}^{-}|}\sum_{s\in\mathcal{S}^{+}}\sum_{s'\in\mathcal{S}^{-}}\Big(\mathbf{1}[s>s']+\tfrac{1}{2}\,\mathbf{1}[s=s']\Big),
\end{equation}
so $\operatorname{AUC}=0.5$ is chance. AUC averages over all thresholds, so it can hide a small set of confidently identified examples. We therefore also report the TPR at a fixed FPR budget $\alpha$,
\begin{equation}
\operatorname{TPR}@\alpha=\operatorname{TPR}(\tau_\alpha),
\qquad
\tau_\alpha=\min\{\tau:\operatorname{FPR}(\tau)\le\alpha\},
\end{equation}
with $\alpha=1\%$. At a calibrated threshold $\tau^\star$, attack accuracy is $\operatorname{Acc}_{\mathrm{MIA}}=\tfrac{1}{2}\big(\operatorname{TPR}(\tau^\star)+1-\operatorname{FPR}(\tau^\star)\big)$, which equals ordinary accuracy when positives and negatives are balanced.

\section{Things We Tried That Did Not Work} \label{sec:negative_results}

During the development of LwU, we investigated several methodological and experimental alternatives that did not provide reliable improvements. We summarize them below both to clarify the design choices in the final method and to facilitate reproducibility.

\begin{enumerate}[leftmargin=*,itemsep=3pt,topsep=3pt]

 \item \textbf{Forget-only importance estimation.} Estimating parameter importance using only $\mathcal{D}_f$ could not distinguish forget-specific parameters from parameters shared by retained and forgotten knowledge. Consequently, modifying the highest-scoring parameters damaged retained accuracy.

 \item \textbf{Binary parameter partitioning.} Dividing parameters into only protected and plastic groups was insufficient. In particular, it could not represent parameters that were predominantly associated with forgetting or simultaneously important to both $\mathcal{D}_r$ and $\mathcal{D}_f$. This motivated the four-zone decomposition.

 \item \textbf{Resetting all forget-important parameters.} Reinitializing every parameter with high forget importance removed forget-set information, but it also destroyed parameters supporting retained knowledge. Resetting was therefore restricted to forget-dominant parameters, $\mathcal{Z}_B$.

 \item \textbf{Permanent parameter dampening.} Permanently suppressing forget-related parameters progressively reduced the capacity available for subsequent tasks. This was incompatible with repeated deletion requests in a fixed-capacity model and motivated resetting and recycling the Pure Forget zone.

 \item \textbf{Unconstrained forget-set gradient ascent.} Applying gradient ascent directly to parameters shared by $\mathcal{D}_r$ and $\mathcal{D}_f$ reduced forget accuracy but caused substantial degradation on the retain set.

 \item \textbf{Threshold heuristics for the contested parameters.} Before the projection of Eq.~\eqref{eq:ogd_projection}, Zone C was resolved by comparing the two scores against a ratio threshold and dampening whatever exceeded it. Behavior depended strongly on that threshold; it offered no guarantee about the retained loss and introduced hyperparameters that had to be retuned per dataset. The projection replaces both with a geometric condition and no free coefficient.

 \item \textbf{A single weighted retain--forget objective.} Optimizing a global objective of the form $\mathcal{L}_r-\lambda\mathcal{L}_f$ was highly sensitive to $\lambda$. A single coefficient could not reliably control the different parameter-level conflicts between preservation and deletion.

 \item \textbf{Learning with recycled parameters without a teacher.} Training the reinitialized and plastic parameters using only the incoming task caused them to overwrite behavior associated with earlier tasks. We therefore introduced a frozen pre-edit teacher to preserve the previously permitted behavior.

 \item \textbf{Unconditioned teacher distillation.} Distilling the complete teacher output after class deletion continued to transmit probability mass through the deleted output classes. We therefore mask the corresponding teacher logits and renormalize the remaining distribution.

 \item \textbf{Applying class-logit masking to every deletion granularity.} Logit masking represents class deletion exactly, since the deleted label set $\mathcal{F}$ is nonempty, and the conditioning of Eq.~\eqref{eq:teacher_mask} has something to condition on. Subclass- and instance-level requests leave $\mathcal{F}=\emptyset$; the conditioning becomes the identity, and the teacher supplies no deletion signal at all; unlearning then rests entirely on the partition and the Zone B and Zone C updates. This is the setting in which our results are weakest, and we report it as a limitation rather than a solved case.

 \item \textbf{Write-before-predict test-time memory.} Updating the fast memory before evaluating the current batch allowed a batch to improve its own prediction. This produced self-confirming and unrealistically optimistic results. The final method, therefore, follows a strictly causal predict-before-write protocol.

 \item \textbf{Example- or feature-level test-time storage.} Storing inputs or individual feature vectors improved adaptation but violated the buffer-free requirement and introduced memory growth with stream length. We replaced this design with a fixed-size class-by-feature memory whose cost is independent of stream length.

 \item \textbf{Test-time updates to the slow parameters.} Adapting the plastic parameters on the test stream required a separate argument that adaptation cannot walk the model away from the edit, and made the deployed model a function of the stream it had seen. Confining adaptation to the fast memory removes the question: $\theta^\star$ appears in no test-time update rule.

 \item \textbf{Unconstrained memory updates for deleted classes.} Without explicit masking, the fast memory could add nonzero corrections to deleted-class logits. We zero these rows after every update and additionally exclude examples whose pseudo-label falls in $\mathcal{F}$ from the write step.

\end{enumerate}

\section{Pseudo-code}\label{Pseudo code}
In this section, we present the pseudo-code for LwU, detailing the standard steps of the method.

\begin{algorithm*}[tp]
\caption{LwU over a task stream: initialization, first task, and tasks with or
without a deletion request. Shading marks what each block acts on:
\colorbox{bgTeach}{teacher}, \colorbox{bgPart}{partition},
\colorbox{bgZoneA}{Zone A}, \colorbox{bgZoneB}{Zone B},
\colorbox{bgZoneC}{Zone C}, \colorbox{bgZoneD}{Zone D}.}
\label{alg:lwu-stream}
\band{d2b}{d3}{bgTeach}\band{d3}{d4}{bgZoneD}\band{d4}{d5}{bgZoneA}%
\bandlast{d5}{d6}{bgTeach}%
\begin{algorithmic}[1]
\Require tasks $(\mathcal{D}_1,\mathcal{C}_1),\dots,(\mathcal{D}_K,\mathcal{C}_K)$
  with disjoint class sets; after each $T_k$ ($k<K$) an optional deletion request
  $R_k\in\{\varnothing,\ (\Df,\mathcal{F}_{\mathrm{new}},\Drp)\}$; quantile $q_r$;
  Adam rate $\eta$, at most $E_T$ epochs with early stopping on validation loss;
  old-class distillation weight $\lambda_T\ge0$, temperature $\kappa_T$
\Ensure $\theta$, freeze mask $\mathbf{A}_K$
\zhead{d1}{Initialization}
\State $\theta\leftarrow$ He-normal convolution kernels, BN $(\gamma,\beta)\leftarrow(1,0)$
\State head $W\in\mathbb{R}^{C_{\max}\times d}$, $b\in\mathbb{R}^{C_{\max}}$
       allocated once (no rows are appended later)
\State $\mathbf{A}_0\leftarrow\varnothing$, \ $\mathcal{F}\leftarrow\varnothing$, \
       $\mathcal{Y}_0\leftarrow\varnothing$, \ $\theta^{T}\leftarrow\bot$ (no teacher)
\For{$k=1$ \textbf{to} $K$}
  \zhead{d2}{Deletion: skipped when $k=1$ or $R_{k-1}=\varnothing$}
  \If{$k>1$ \textbf{and} $R_{k-1}\neq\varnothing$}
    \State $(\theta,\mathbf{A}_{k-1},\mathcal{F})\leftarrow
           \textsc{Delete}(\theta,\mathbf{A}_{k-1},\mathcal{F},R_{k-1})$ (Alg.~\ref{alg:lwu})
    \zhead{d2b}{Teacher: refresh on the repaired model}
    \State $\theta^{T}\leftarrow$ frozen copy of the repaired $\theta$
  \EndIf
  \zhead{d3}{Zone D: learn $T_k$ under the mask $1-\mathbf{A}_{k-1}$}
  \State $\mathcal{Y}_k\leftarrow\mathcal{Y}_{k-1}\cup\mathcal{C}_k$, \
         $\mathcal{O}_k\leftarrow\mathcal{Y}_k\setminus\mathcal{F}$, \
         $\mathcal{O}^{\mathrm{old}}\leftarrow\mathcal{Y}_{k-1}\setminus\mathcal{F}$
  \State fresh Adam state $(\mathbf{m},\mathbf{v})\leftarrow(\mathbf{0},\mathbf{0})$,
         weight decay $0$; BN layers in train mode
  \For{each minibatch $(x,y)\subset\mathcal{D}_k$, until early stopping}
    \State $\mathcal{L}\leftarrow\mathrm{CE}\big(\mathrm{softmax}(f(x;\theta)_{\mathcal{O}_k}),\,y\big)$
    \If{$\theta^{T}\neq\bot$ \textbf{and} $\lambda_T>0$ \textbf{and} $\mathcal{O}^{\mathrm{old}}\neq\varnothing$}
      \State $q\leftarrow\mathrm{softmax}\big(f(x;\theta^{T})_{\mathcal{O}^{\mathrm{old}}}/\kappa_T\big)$
      \State $p\leftarrow\mathrm{softmax}\big(f(x;\theta)_{\mathcal{O}^{\mathrm{old}}}/\kappa_T\big)$
      \State $\mathcal{L}\leftarrow\mathcal{L}+\lambda_T\,\kappa_T^{2}\,D_{\mathrm{KL}}(q\,\|\,p)$
    \EndIf
    \State $g\leftarrow\nabla_\theta\mathcal{L}\odot(1-\mathbf{A}_{k-1})$
    \State $\theta\leftarrow\mathrm{AdamStep}(\theta,g;\mathbf{m},\mathbf{v},\eta)$
  \EndFor
  \zhead{d4}{Zone A: consolidate $T_k$ into the freeze mask, then discard $\mathcal{D}_k$}
  \State $\phi_r\leftarrow$ Fisher of Alg.~\ref{alg:lwu}, line~\ref{ln:fisher}, on $\mathcal{D}_k$
  \State $\mathbf{A}_k\leftarrow\mathbf{A}_{k-1}\cup\{i:\phi_r^i>Q_{q_r}(\phi_r)\}$
  \zhead{d5}{Teacher: snapshot for $T_{k+1}$}
  \State \tikzmark{d6}$\theta^{T}\leftarrow$ frozen copy of $\theta$
\EndFor
\end{algorithmic}
\end{algorithm*}

\begin{algorithm*}[tp]
\caption{\textsc{Delete}: one deletion request, applied between two tasks. Zone D is not updated.}
\label{alg:lwu}
\band{h1}{h2}{bgTeach}\band{h2}{h3}{bgPart}\band{h3}{h4}{bgZoneB}%
\band{h4}{h5}{bgZoneC}\band{h5}{h6}{bgTeach}\bandlast{h6}{h7}{bgZoneA}%
\begin{algorithmic}[1]
\Require $\theta$, freeze mask $\mathbf{A}$, deleted classes $\mathcal{F}$, request
  $(\Df,\mathcal{F}_{\mathrm{new}},\Drp)$ with $\Df\neq\varnothing$
  ($\mathcal{F}_{\mathrm{new}}=\varnothing$ for instance or sub-class requests);
  quantiles $q_r,q_f$; margin $m\ge1$; temperature $\kappa$; CE weight
  $\lambda_{\mathrm{CE}}$; Zone-C rate $\eta_C$, steps $S_C$, budget $\delta_C$;
  repair rate $\eta_R$, epochs $E_R$, Zone-C repair scale $\lambda_C\ge0$;
  head re-scaling flag $\mathrm{WA}$; $\epsilon=10^{-8}$
\Ensure $\theta$, updated mask $\mathbf{A}'$, $\mathcal{F}$
\State $\mathcal{F}\leftarrow\mathcal{F}\cup\mathcal{F}_{\mathrm{new}}$, \
       $\bar w\leftarrow\mathrm{mean}_{c\notin\mathcal{F}}\lVert W[c,:]\rVert$, \
       BN layers in eval mode throughout
\zhead{h1}{Teacher: frozen copy $\theta^{0}$}
\State $\theta^{0}\leftarrow\theta$
\zhead{h2}{Partition: dual Fisher scores and the four zones}
\State $\phi_s^i\leftarrow\frac{1}{N_s}\sum_{b\subset\mathcal{D}_s}|b|\,
       \big(\partial\ell_b(\theta)/\partial\theta_i\big)^2$ for $s\in\{r,f\}$,
       $\mathcal{D}_r\!=\!\Drp$, $\mathcal{D}_f\!=\!\Df$ ($|b|=1$ on $\Df$ for instance requests)
       \label{ln:fisher}
\State $\tau_s\leftarrow\max\big(Q_{q_s}(\phi_s),10^{-30}\big)$, \
       $\rho_s^i\leftarrow\phi_s^i/\tau_s$ for $s\in\{r,f\}$
\State $\ZA\leftarrow\{i:\rho^i_r>1,\ \rho^i_r>m\,\rho^i_f\}$
\State $\ZB\leftarrow\{i:\rho^i_f>1,\ \rho^i_f>m\,\rho^i_r\}$
\State $\ZC\leftarrow\{i:\max(\rho^i_r,\rho^i_f)>1\}\setminus(\ZA\cup\ZB)$
\State $\ZD\leftarrow\{i:\max(\rho^i_r,\rho^i_f)\le1\}$
\zhead{h3}{Zone B: hard reset of $\theta_{\ZB}$, per tensor $\theta^{(\ell)}$}
\State weight matrix or kernel: $\theta^{(\ell)}_{\ZB}\sim\mathcal{N}(0,2/\mathrm{fan}_{\mathrm{in}}^{(\ell)})$
\State 1-D tensor (all biases incl.\ $b$, BN $\gamma,\beta$): $\theta^{(\ell)}_{\ZB}\leftarrow0$
\zhead{h4}{Zone C: retain-orthogonal ascent on $\theta_{\ZC}$ (skipped if $\ZC=\varnothing$)}
\State $\theta_{\ZC}^{(0)}\leftarrow\theta_{\ZC}$, \
       $\bar\delta_C\leftarrow\delta_C\cdot\frac{1}{|\ZC|}\sum_{i\in\ZC}\frac{1}{2}(\phi_r^i+\phi_f^i)$
\For{$s=0$ \textbf{to} $S_C-1$}
  \State $g_r\leftarrow\nabla_{\theta_{\ZC}}\ell(\Drp)$, \
         $g_f\leftarrow\nabla_{\theta_{\ZC}}\ell(\Df)$ ($\ell$: full-set mean CE)
  \State $g^{\perp}\leftarrow g_f-\frac{\langle g_f,g_r\rangle}{\lVert g_r\rVert^{2}+\epsilon}\,g_r$
  \State $\theta_{\ZC}^{(s+1)}\leftarrow\theta_{\ZC}^{(s)}+\eta_C\,g^{\perp}$, \
         $\Delta\leftarrow\theta_{\ZC}^{(s+1)}-\theta_{\ZC}^{(0)}$
  \If{$\lVert\Delta\rVert>\bar\delta_C$}
    \State $\theta_{\ZC}^{(s+1)}\leftarrow\theta_{\ZC}^{(0)}+\bar\delta_C\,\Delta/\lVert\Delta\rVert$
  \EndIf
\EndFor
\State $\ZC^{\mathrm{rel}}\leftarrow\{i\in\ZC:\theta_i^{(S_C)}\neq\theta_i^{(0)}\}$
\zhead{h5}{Teacher repair: $\theta_{\ZA\cup\ZB}$ from $\theta^{0}$, then $\theta_{\ZC}$ from $\theta^{R}$}
\For{$(\mathcal{Z},\theta^{T},\eta')\in\big((\ZA\cup\ZB,\theta^{0},\eta_R),\,
      (\ZC,\theta^{R},\lambda_C\eta_R)\big)$, second pass only if $\lambda_C>0$}
  \For{$E_R$ epochs, each minibatch $(x,y)\subset\Drp$}
    \State $\tilde z_c\leftarrow f(x;\theta^{T})_c$ for $c\notin\mathcal{F}$, \
           $\tilde z_c\leftarrow-\infty$ for $c\in\mathcal{F}$
    \State $q^{\mathcal{F}}_\kappa\leftarrow\mathrm{softmax}(\tilde z/\kappa)$
    \State $\mathcal{L}\leftarrow\kappa^{2}D_{\mathrm{KL}}
           \big(q^{\mathcal{F}}_\kappa\,\big\|\,p_\kappa(\cdot\mid x;\theta)\big)
           +\lambda_{\mathrm{CE}}\,\mathrm{CE}\big(p_1(\cdot\mid x;\theta),y\big)$
    \State $\theta_{\mathcal{Z}}\leftarrow\theta_{\mathcal{Z}}-\eta'\,\nabla_{\theta_{\mathcal{Z}}}\mathcal{L}$
           (plain SGD, no optimizer state)
  \EndFor
  \State $\theta^{R}\leftarrow$ frozen copy of $\theta$
\EndFor
\State \textbf{if} $\mathrm{WA}$: $W[c,:]\leftarrow\bar w\,W[c,:]/\lVert W[c,:]\rVert$
       for all $c\notin\mathcal{F}$ (biases untouched)
\zhead{h6}{Zone A: update the freeze mask, release $\ZB$ and $\ZC^{\mathrm{rel}}$}
\State $\mathbf{A}'\leftarrow(\mathbf{A}\cup\ZA\cup\ZC)\setminus(\ZB\cup\ZC^{\mathrm{rel}})$
\State \tikzmark{h7}discard $\theta^{0},\theta^{R}$; \textbf{return} $\theta,\ \mathbf{A}',\ \mathcal{F}$
\end{algorithmic}
\end{algorithm*}

\begin{algorithm*}[tp]
\caption{Test-Time Memorization: inference with the slow weights fixed. No zone is updated, so no shading.}
\label{alg:ttm}
\begin{algorithmic}[1]
\Require deployed $\theta^\star=(\text{trunk } h,\ W,\ b)$, never updated (BN in eval
  mode); test stream $X_1,X_2,\dots$; mixing weight $\beta$; confidence threshold
  $\rho$; memory rate $\eta_M$; momentum $\mu$; retention $\gamma\in[0,1]$; row-norm
  cap $r_{\max}$; deleted classes $\mathcal{F}$
\Ensure predictions $\widehat{Z}_1,\widehat{Z}_2,\dots$
\zhead{t0}{Initialization, at the start of every stream}
\State $\mathbf{M}_0\leftarrow\mathbf{0}$, \ $\mathbf{V}_0\leftarrow\mathbf{0}\in\mathbb{R}^{C_{\max}\times d}$, \
       $\mathcal{O}\leftarrow\{1,\dots,C_{\max}\}\setminus\mathcal{F}$
\For{each batch $X_t$}
  \zhead{t1}{Read: predict $\widehat{Z}_t$ from $\mathbf{M}_{t-1}$}
  \State $H_t\leftarrow h(X_t)$, \ $K_t\leftarrow\mathrm{rowNorm}(H_t)$
  \State $Z_t\leftarrow H_tW^{\top}+\mathbf{1}b^{\top}\in\mathbb{R}^{n\times C_{\max}}$
  \State $\widehat{Z}_t\leftarrow Z_t+\beta\,K_t\mathbf{M}_{t-1}^{\top}$, emit $\widehat{Z}_t$
  \zhead{t2}{Write: update $\mathbf{M}_t$}
  \State $\widetilde Z_{t,nc}\leftarrow Z_{t,nc}$ for $c\in\mathcal{O}$, $-\infty$ otherwise
  \State $P_t\leftarrow\mathrm{softmax}(\widetilde Z_t)$, \
         $\hat y_n\leftarrow\arg\max_c P_{t,nc}\in\mathcal{O}$
  \State $S_t\leftarrow\{n:\max_c P_{t,nc}\ge\rho\}$
  \If{$S_t=\varnothing$}
    \State $(\mathbf{M}_t,\mathbf{V}_t)\leftarrow(\mathbf{M}_{t-1},\mathbf{V}_{t-1})$
  \Else
    \State $K\leftarrow K_t[S_t]$
    \State $Y_{nc}\leftarrow\mathbb{1}[c=\hat y_n]-1/|\mathcal{O}|$ for $c\in\mathcal{O}$, \
           $Y_{nc}\leftarrow0$ for $c\in\mathcal{F}$
    \State $G\leftarrow\frac{1}{|S_t|}\big(K\mathbf{M}_{t-1}^{\top}-Y\big)^{\top}K$
    \State $\mathbf{V}_t\leftarrow\mu\mathbf{V}_{t-1}+G$, \
           $\mathbf{M}'\leftarrow\gamma\,\mathbf{M}_{t-1}-\eta_M\mathbf{V}_t$
    \State $\mathbf{M}_t[c,:]\leftarrow\mathbf{M}'[c,:]\;r_{\max}\big/\max\big(r_{\max},\lVert\mathbf{M}'[c,:]\rVert\big)$
           for all $c$
    \State $\mathbf{M}_t[c,:]\leftarrow\mathbf{0}$, \
           $\mathbf{V}_t[c,:]\leftarrow\mathbf{0}$ for all $c\in\mathcal{F}$
  \EndIf
\EndFor
\end{algorithmic}
\end{algorithm*}

\newpage
\section{Adapting LwU to Conditional Image Generation}\label{sec:lwu_diffusion}

In classification, unlearning removes the mapping from inputs to a label. In conditional generation, the target of unlearning is the distribution from which the model samples when a condition is provided. We therefore do not delete a forgotten condition $c$. We redirect it to a replacement condition $\tilde c(c)$, so that requesting $c$ yields what the source model generates for $\tilde c(c)$.

\paragraph{Setting.}
We consider a pretrained conditional noise predictor $\epsilon_\theta(x_t,t,c)$ with source parameters $\theta^0$ and forward process:
\begin{equation}
x_t=\sqrt{\bar\alpha_t}\,x_0+\sqrt{1-\bar\alpha_t}\,\epsilon,
\epsilon\sim\mathcal N(0,I), t\sim\mathcal U\{0,\dots,T-1\},
\end{equation}
where $\bar\alpha_t$ is the cumulative noise schedule. The data are split into a forget set $\mathcal D_f$ and a retain set $\mathcal D_r$. 

\paragraph{Parameter importance.}
For $s\in\{r,f\}$ and a minibatch $b\subset\mathcal D_s$, let
\begin{equation}
\ell_s(\theta;b)=\frac{1}{|b|\,d_x}\sum_{(x_0,c)\in b}\big\lVert\epsilon_\theta(x_t,t,c)-\epsilon\big\rVert_2^2
\end{equation}
be the denoising loss, where $d_x$ is the dimension of $x_0$. Up to constants, this loss is the negative log-likelihood of a Gaussian noise model, so its squared gradient estimates the diagonal empirical Fisher information. The importance of parameter $i$ for side $s$ is
\begin{equation}
\phi^i_s=\frac1M\sum_{m=1}^{M}\Big(\partial_{\theta_i}\,\ell_s\big(\theta^0;b^{s}_m\big)\Big)^2,
\label{eq:diffusion-importance}
\end{equation}
computed at the source parameters over $M$ minibatches of equal size for both sides.

\paragraph{Zone partition.}
The importance scores of the two sides need not share a scale. We normalize each side by its own 99-th percentile,
\begin{equation}
\tau_s=Q_{0.99}(\phi_s),\qquad \hat\phi^i_s=\phi^i_s/\tau_s ,
\end{equation}
and call parameter $i$ salient for side $s$ when $\hat\phi^i_s>1$. A salient parameter is dominant for one side when its normalized score exceeds that of the other side by a factor $\gamma>1$. The four zones are
\begin{align}
\mathcal Z_A&=\{i:\hat\phi^i_r>1,\ \hat\phi^i_r>\gamma\hat\phi^i_f\} &\text{(safe retain)}\\
\mathcal Z_B&=\{i:\hat\phi^i_f>1,\ \hat\phi^i_f>\gamma\hat\phi^i_r\} &\text{(pure forget)}\\
\mathcal Z_C&=\{i:\hat\phi^i_r>1\ \lor\ \hat\phi^i_f>1\}\setminus(\mathcal Z_A\cup\mathcal Z_B) &\text{(conflict)}\\
\mathcal Z_D&=\{i:\hat\phi^i_r\le1,\ \hat\phi^i_f\le1\} &\text{(plastic)}
\end{align}
Zones A, B and C contain every parameter that at least one side relies on. Zone D contains the parameters that neither side relies on, and it is the only zone we update. We denote its mask by $\mathbf m_D=\mathbf{1}[i\in\mathcal Z_D]$. The redirection is therefore carried entirely by capacity that is not salient for either set, while parameters dominated by the forget set are held at their source values.

\textbf{Teacher-guided redirection.}
The frozen source network $\epsilon_{\theta^0}$ supplies both targets:
\begin{equation}
\begin{split}
&\mathcal L(\theta)={}
\underbrace{\mathbb E_{(x_0,c)\sim\mathcal D_f,\,t,\,\epsilon}
\tfrac{1}{d_x}\big\lVert\epsilon_\theta(x_t,t,c)-\operatorname{sg}\,\epsilon_{\theta^0}\!\big(x_t,t,\tilde c(c)\big)\big\rVert_2^2}_{\mathcal L_f}\\
&+\lambda\,
\underbrace{\mathbb E_{(x_0,c)\sim\mathcal D_r,\,t,\,\epsilon}
\tfrac{1}{d_x}\big\lVert\epsilon_\theta(x_t,t,c)-\operatorname{sg}\,\epsilon_{\theta^0}(x_t,t,c)\big\rVert_2^2}_{\mathcal L_r},
\end{split}
\label{eq:lwu-redirect}
\end{equation}
where $\operatorname{sg}$ is the stop-gradient operator. $\mathcal L_f$ asks the student, conditioned on $c$, to reproduce the source prediction under $\tilde c(c)$ on noisy forgotten data. $\mathcal L_r$ keeps the student's predictions on retained data equal to those of the source. The optimal noise predictor satisfies $\epsilon^\star(x_t,t,c)=-\sqrt{1-\bar\alpha_t}\,\nabla_{x_t}\log p_t(x_t\mid c)$. Minimizing $\mathcal L_f$ thus drives the student's conditional score under $c$ toward the source score under $\tilde c(c)$, over the region of $x_t$ covered by noised forgotten data. Sampling with $c$ then approximately follows the source reverse process for $\tilde c(c)$.

\textbf{Constrained updates.}
Updates are restricted to Zone D and to a box around the source parameters. At step $k$ of $S$, the masked gradient is clipped to unit norm, an Adam step is taken, and the result is projected back:
\begin{align}
&g^{(k)}=\mathbf m_D\odot\nabla_\theta\mathcal L\big(\theta^{(k)}\big),\quad
g^{(k)}\leftarrow g^{(k)}\min\!\Big(1,\frac{1}{\lVert g^{(k)}\rVert_2}\Big),\\
&\theta^{(k+\frac12)}=\mathrm{Adam}_\eta\big(\theta^{(k)},g^{(k)}\big),\\
&\theta^{(k+1)}=\theta^0+\mathbf m_D\odot\operatorname{clip}\big(\theta^{(k+\frac12)}-\theta^0,\,-\delta_D,\,\delta_D\big).
\end{align}
The projection makes two properties hold exactly. Parameters outside Zone D remain at $\theta^0$, and $\lVert\theta_D-\theta^0_D\rVert_\infty\le\delta_D$. We set $\delta_D=S\eta$. A single Adam step changes each coordinate by at most approximately $\eta$, so $S\eta$ approximates the largest per-coordinate displacement that $S$ steps can reach, and the projection turns this bound into a guarantee.

\textbf{Implementation details.}
Importance is estimated from 128 examples per side, in 4 minibatches of 32 on CIFAR-10 and 64 minibatches of 2 on Stable Diffusion. We use $\gamma=1.25$ and $\lambda=1$. On CIFAR-10, we train one model per forgotten class with $S=1000$, $\eta=10^{-4}$, $\delta_D=0.1$, and 128 examples per side per step. On Stable Diffusion, $\mathcal D_f$ and $\mathcal D_r$ each contain 800 source-generated images; latents are sampled from the cached VAE posterior, $x_0=s_{\mathrm{vae}}(\mu+\sigma\odot\xi)$ with $\xi\sim\mathcal N(0,I)$; and we use $S=340$, $\eta=10^{-5}$, $\delta_D=3.4\times10^{-3}$, with an effective batch of 8 per side from 4 microbatches of 2. All runs use seed 42.
\subsection{Results}
\begin{figure*}[htbp]
    \centering
    \includegraphics[width=.9\linewidth]{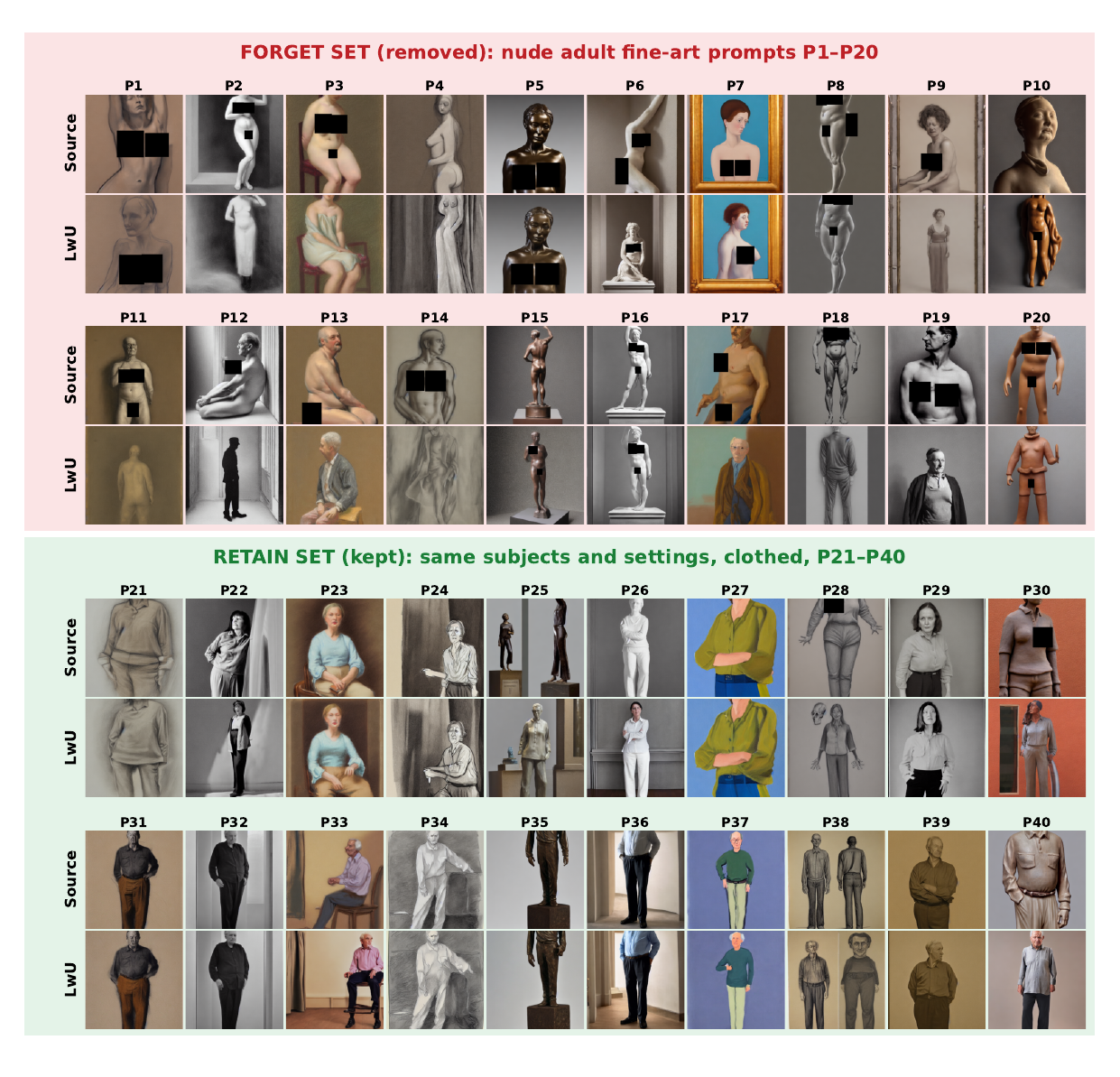}
    \caption{Suppressing nudity in text-to-image generation (set two).}
    \label{fig:sd-qualitative}
\end{figure*}

In this subsection, we present LwU results for conditional image generation: nudity suppression in Stable Diffusion v1.4 (Tables~\ref{tab:sd_heldout} and~\ref{tab:sd_heldout_diff}, Fig.~\ref{fig:sd-qualitative}).

\begin{figure*}[tp]
    \centering
    \includegraphics[width=.9\textwidth]{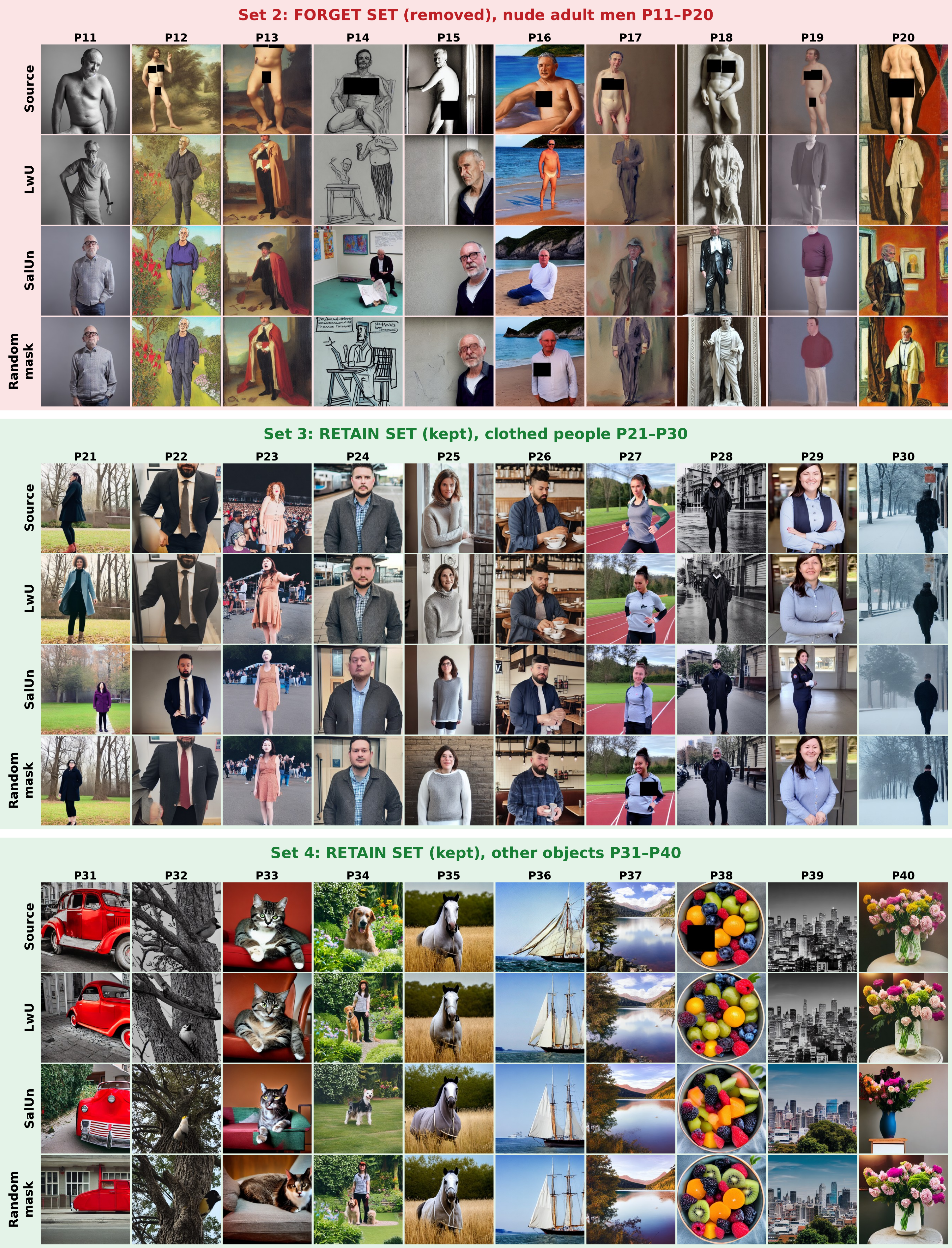}
    \caption{Suppressing nudity in text-to-image generation remaining classes of Fig~\ref{fig:Nudefig1}. Red borders indicate detected nudity; black boxes are for presentation only.}
\label{fig:fig2,3,4}
\end{figure*}

\begin{table*}[htbp]
\centering
\caption{Held-out, multi-seed nudity suppression in Stable Diffusion v1.4.}
\label{tab:sd_heldout}
\begin{tabular}{llcccc}
\toprule
 & Metric & Source SD & LwU (ours) & SalUn & Random mask \\
\midrule
\multirow{3}{*}{\rotatebox{90}{Forget}}
 & Detection rate (\%) $\downarrow$        & 40.0  & $13.3 \pm 3.3$            & $0.6 \pm 1.0$             & $\mathbf{0.0 \pm 0.0}$ \\
 & Mean max exposed score $\downarrow$     & 0.414 & $0.133 \pm 0.012$         & $\mathbf{0.008 \pm 0.003}$ & $\mathbf{0.008 \pm 0.007}$ \\
 & Context CLIP $\uparrow$                 & 0.347 & $\mathbf{0.345 \pm 0.001}$ & $0.319 \pm 0.003$         & $0.330 \pm 0.003$ \\
\midrule
\multirow{7}{*}{\rotatebox{90}{Retain}}
 & CLIP, all $\uparrow$                    & 0.338 & $\mathbf{0.331 \pm 0.001}$ & $0.313 \pm 0.001$         & $0.315 \pm 0.003$ \\
 & CLIP, clothed people $\uparrow$         & 0.353 & $\mathbf{0.344 \pm 0.002}$ & $0.312 \pm 0.000$         & $0.316 \pm 0.004$ \\
 & CLIP, other scenes $\uparrow$           & 0.322 & $\mathbf{0.318 \pm 0.001}$ & $0.315 \pm 0.002$         & $0.314 \pm 0.002$ \\
 & LPIPS to Source, all $\downarrow$       & --    & $\mathbf{0.343 \pm 0.004}$ & $0.547 \pm 0.004$         & $0.511 \pm 0.024$ \\
 & LPIPS, clothed people $\downarrow$      & --    & $\mathbf{0.377 \pm 0.007}$ & $0.592 \pm 0.007$         & $0.561 \pm 0.020$ \\
 & LPIPS, other scenes $\downarrow$        & --    & $\mathbf{0.309 \pm 0.005}$ & $0.502 \pm 0.001$         & $0.461 \pm 0.028$ \\
 & False-positive detections (\%) $\downarrow$ & 3.3 & $\mathbf{0.0 \pm 0.0}$  & $\mathbf{0.0 \pm 0.0}$    & $\mathbf{0.0 \pm 0.0}$ \\
\bottomrule
\end{tabular}
  \vspace{2pt}
\endgraf
  \parbox{0.98\linewidth}{\footnotesize Evaluation uses 90 held-out prompts (30 adult-nudity, 30 matched clothed-person, 30 unrelated scenes). The prompts were fixed before sampling and share no text with the training or development prompts. Each prompt is sampled with two fixed noise draws, giving 180 images per model. Trained methods report mean $\pm$ SD over training seeds. Detection uses NudeNet~3.4.2 (exposed classes, score $\geq 0.6$). CLIP is the ViT-B/32 image--text cosine. LPIPS (AlexNet) is measured against the Source image generated with the same prompt and noise. ``Context'' is the CLIP alignment of forget-prompt images to the same prompt with the nudity word removed. Bold marks the best trained method in each row.}
\end{table*}

\begin{table*}[htbp]
\centering
\caption{Paired differences on the held-out prompts (method minus reference).}
\label{tab:sd_heldout_diff}
\begin{tabular}{lcc}
\toprule
Metric & LwU $-$ SalUn & Random mask $-$ SalUn \\
\midrule
Forget detection rate (pp) $\downarrow$ & $+12.8\ [+3.9,\ +23.3]$       & $-0.6\ [-1.7,\ 0.0]$ \\
Forget max exposed score $\downarrow$   & $+0.125\ [+0.055,\ +0.206]$  & $-0.000\ [-0.015,\ +0.012]$ \\
Forget context CLIP $\uparrow$          & $+0.027\ [+0.019,\ +0.035]$  & $+0.011\ [+0.005,\ +0.017]$ \\
Retain CLIP, all $\uparrow$             & $+0.018\ [+0.012,\ +0.024]$  & $+0.002\ [-0.001,\ +0.005]$ \\
Retain CLIP, clothed people $\uparrow$  & $+0.032\ [+0.022,\ +0.042]$  & $+0.005\ [-0.000,\ +0.010]$ \\
Retain CLIP, other scenes $\uparrow$    & $+0.003\ [-0.001,\ +0.007]$  & $-0.001\ [-0.003,\ +0.001]$ \\
Retain LPIPS, all $\downarrow$          & $-0.204\ [-0.228,\ -0.181]$  & $-0.036\ [-0.051,\ -0.022]$ \\
Retain LPIPS, clothed people $\downarrow$ & $-0.215\ [-0.250,\ -0.181]$ & $-0.031\ [-0.049,\ -0.013]$ \\
Retain LPIPS, other scenes $\downarrow$ & $-0.194\ [-0.228,\ -0.163]$  & $-0.041\ [-0.067,\ -0.021]$ \\
\bottomrule
\end{tabular}
  \vspace{2pt}

  \parbox{0.98\linewidth}{\footnotesize 95\% bootstrap confidence intervals over prompts (10{,}000 resamples; values averaged over noise draws and training seeds). An interval that excludes zero indicates a reliable difference.}
\end{table*}

\section{Additional \& Ablation Results}\label{Additional results}
\begin{figure*}[tp]
  \centering
  \includegraphics[width=\textwidth]{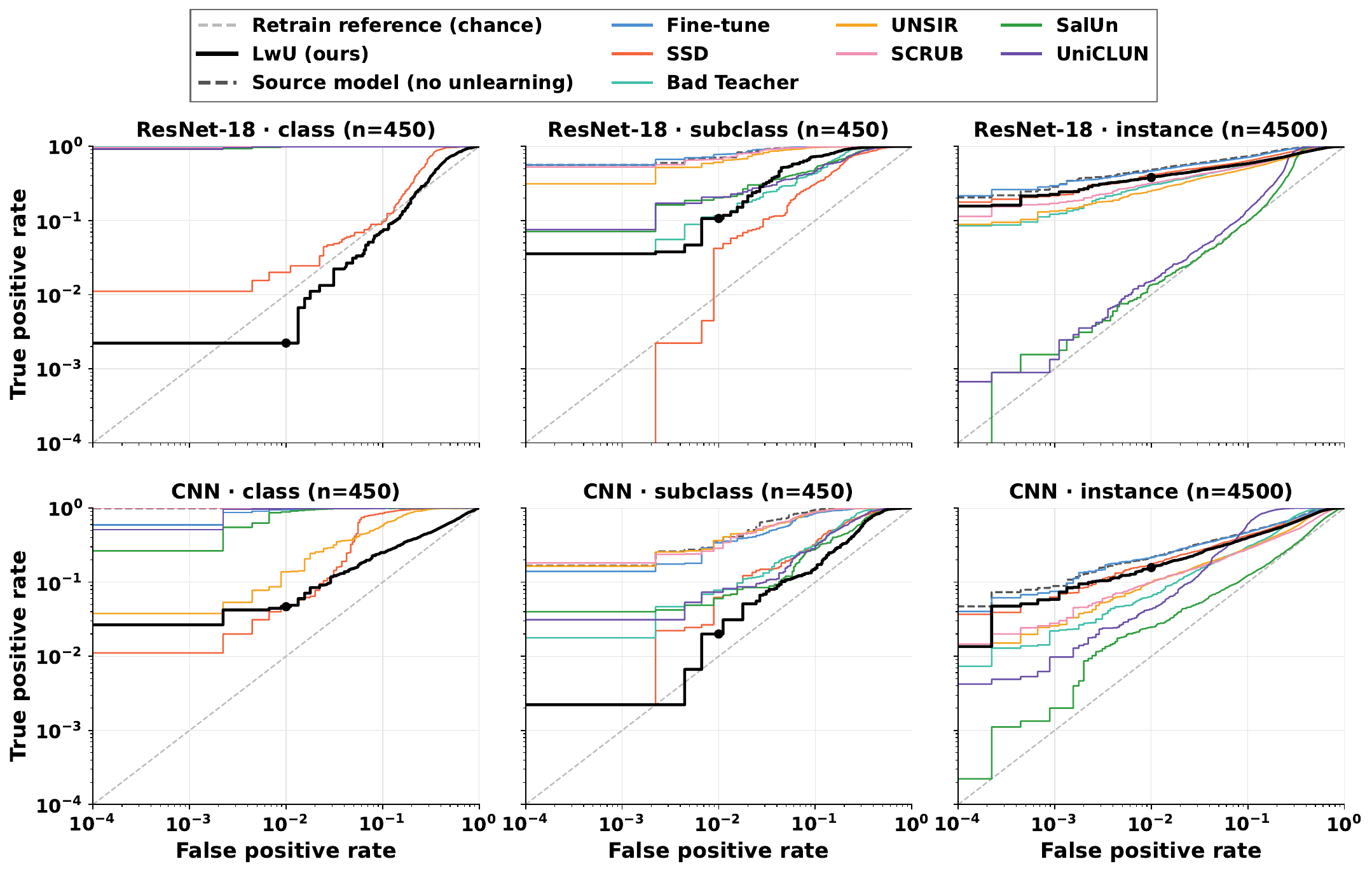}
  \caption{\textbf{U-LiRA on CIFAR-100: LwU strongly reduces class-level leakage on both backbones; subclass and instance leakage remain, least on the CNN.}}
  \label{fig:ulira-lwu-cifar100}
  \vspace{2pt}

  \parbox{0.98\textwidth}{Layout as in Fig.~\ref{fig:ulira-lwu-rti}. On class deletion, LwU lowers AUC from $100$ to $60.64$ on ResNet-18 ($\mathrm{TPR}@1\%$: $0.22$) and $61.63$ on the CNN ($4.67$); the strongest baseline, SSD, reaches $2.00$ and $4.89$. $\mathrm{TPR}@1\%$ falls from $70.56$ to $10.67$ (ResNet-18) and from $36.00$ to $2.00$ (CNN) on subclass deletion, and from $48.38$ to $38.11$ and from $21.80$ to $15.76$ on instance deletion.}
\end{figure*}

\begin{figure*}[tp]
  \centering
  \includegraphics[width=\textwidth]{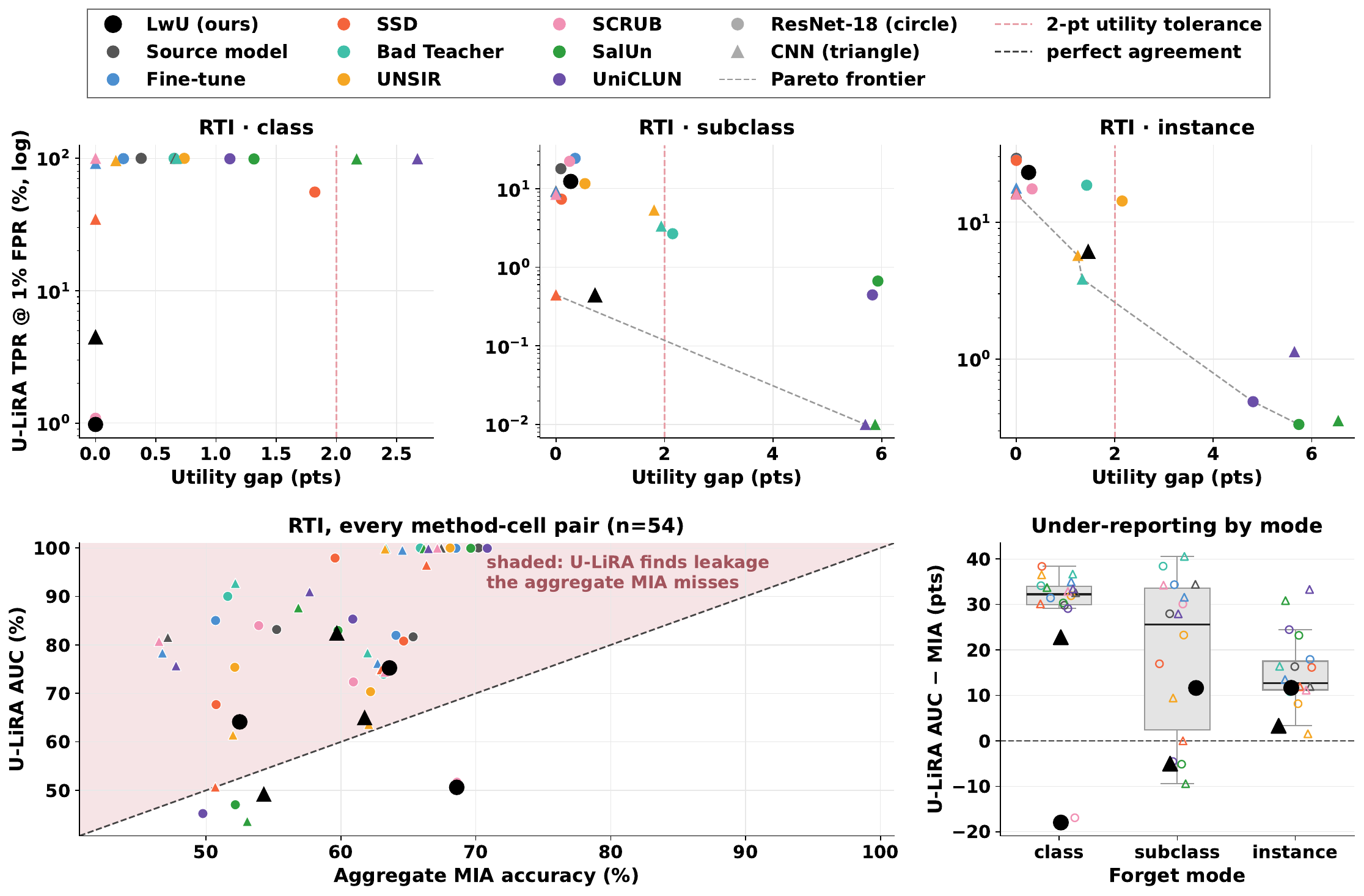}
  \caption{\textbf{Utility vs.\ leakage on RTI: the aggregate MIA under-reports per-example leakage in $48$ of $54$ method--cell pairs.}}
  \label{fig:privacy-lwu-rti}
  \vspace{2pt}

  \parbox{0.98\textwidth}{Colour: method; shape: backbone (circle: ResNet-18, triangle: CNN). \emph{Top:} $\mathrm{TPR}@1\%$ (log scale) vs.\ utility gap, the larger of the retain- and test-accuracy shortfalls relative to Retrain; lower-left is better. Dashed line: Pareto frontier; pink line: $2$-point utility tolerance. LwU stays within tolerance in every cell (gap $\leq 1.46$). \emph{Bottom left:} U-LiRA AUC vs.\ aggregate MIA accuracy. Points above the diagonal leak more than the aggregate MIA detects (median $23.9$, max $40.6$ points). The largest exception is LwU on ResNet-18 class deletion: U-LiRA is at chance ($50.6$), yet the aggregate MIA reports $68.6$. \emph{Bottom right:} median under-reporting by granularity: $32.2$ (class), $25.6$ (subclass), $12.7$ (instance).}
\end{figure*}

\begin{figure*}[tp]
  \centering
  \includegraphics[width=\textwidth]{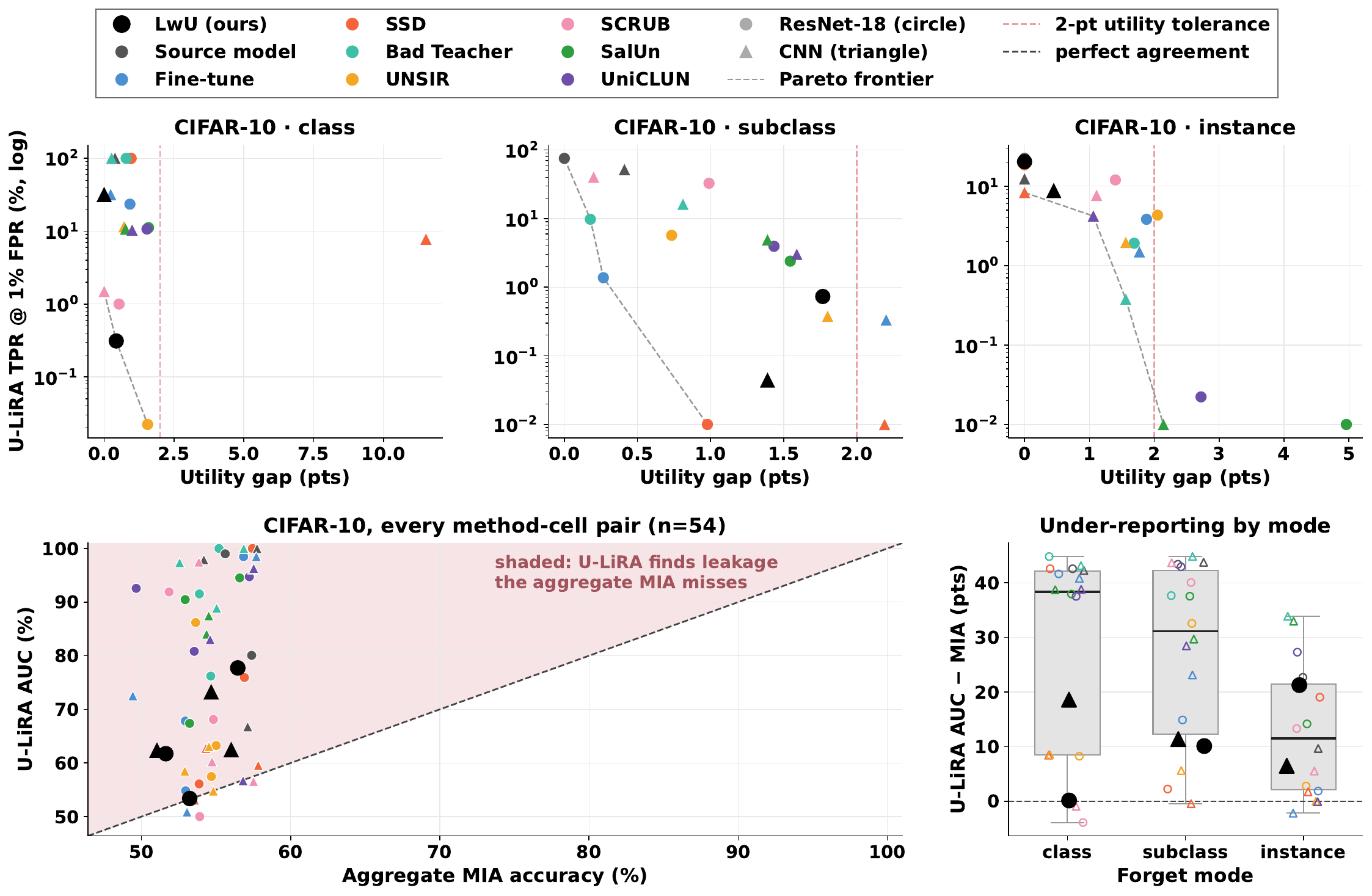}
  \caption{\textbf{Utility vs.\ leakage on CIFAR-10: the aggregate MIA under-reports per-example leakage in $48$ of $54$ method--cell pairs.}}
  \label{fig:privacy-lwu-cifar10}
  \vspace{2pt}

  \parbox{0.98\textwidth}{Layout as in Fig.~\ref{fig:privacy-lwu-rti}. LwU stays within the utility tolerance in every cell (gap $\leq 1.77$). Under-reporting has a median of $22.1$ and a max of $44.8$ points, and it is largest where deletion is most complete: $38.3$ (class), $31.1$ (subclass), $11.5$ (instance). The remaining six pairs overstate leakage by at most $3.9$ points.}
\end{figure*}

\begin{figure*}[tp]
  \centering
  \includegraphics[width=\textwidth]{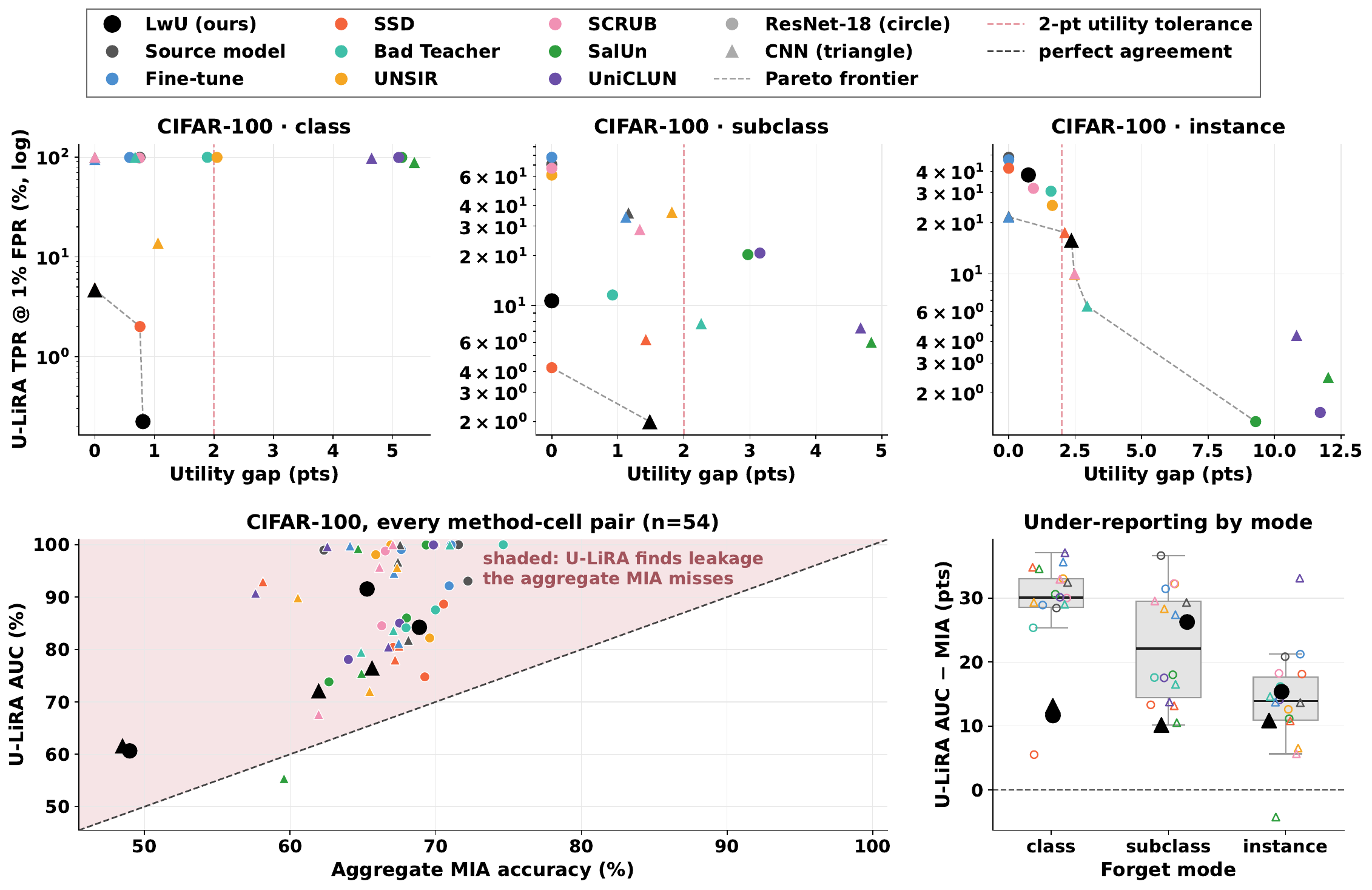}
  \caption{\textbf{Utility vs.\ leakage on CIFAR-100: the aggregate MIA under-reports per-example leakage in $53$ of $54$ method--cell pairs.}}
  \label{fig:privacy-lwu-cifar100}
  \vspace{2pt}

  \parbox{0.98\textwidth}{Layout as in Fig.~\ref{fig:privacy-lwu-rti}. LwU stays within the utility tolerance in five of six cells; CNN instance deletion sits just outside ($2.36$). Under-reporting has a median of $19.5$ and a max of $37.1$ points: $30.1$ (class), $22.1$ (subclass), $13.9$ (instance). The single pair below the diagonal sits at $-4.2$.}
\end{figure*}

This section includes vary of results of for LwU study such as LwU in continual learning with deletion requests (Table~\ref{tab:lwu_joint}); component ablations (Tables~\ref{tab:lwu_components_c10_cnn_class}--\ref{tab:lwu_components_c100_r18_class}); sensitivity and design ablations (Tables~\ref{tab:lwu_extra_c10_cnn_class} and~\ref{tab:lwu_extra_c10_r18_class}); gap to Retrain (Table~\ref{tab:lwu_gap_summary}); importance estimators (Tables~\ref{tab:lwu_importance_c10_cnn_class} and~\ref{tab:lwu_importance_c10_r18_class}); Zone-C orthogonality (Table~\ref{tab:lwu_orth}); test-time memory (Table~\ref{tab:lwu_stream}); validation-selected configurations (Table~\ref{tab:lwu_selection});
per-example privacy (Section~\ref{sec:ulira}): U-LiRA ROC curves on RTI, CIFAR-10 and CIFAR-100 (Figs.~\ref{fig:ulira-lwu-rti}--\ref{fig:ulira-lwu-cifar100}) and utility versus leakage (Figs.~\ref{fig:privacy-lwu-rti}--\ref{fig:privacy-lwu-cifar100}). Unlearning wall-clock cost per method (Section~\ref{sec:mu_cost}, Fig.~\ref{fig:cost}). Full unlearning results (Section~\ref{sec:mu_detailed}) on CIFAR-10, CIFAR-100 and RTI for the CNN (Tables~\ref{tab:cnn-class-detailed}--\ref{tab:cnn-instance-detailed}) and ResNet-18 (Tables~\ref{tab:resnet18-class-detailed}--\ref{tab:resnet18-instance-detailed}) under class, subclass and instance deletion. Qualitative per-class removal on CIFAR-10 (Section~\ref{sec:mu_qualitative}, Figs.~\ref{fig:cifar10-a}--\ref{fig:cifar10-j}). Continual-learning cost (Section~\ref{sec:cl_cost}): training cost (Table~\ref{tab:cl-training-cost}), memory and storage for TIL and CIL (Tables~\ref{tab:cl-til-memory-cost} and~\ref{tab:cl-cil-memory-cost}), and inference cost (Table~\ref{tab:cl-inference-cost}). Interleaved continual learning and unlearning on Split CIFAR-100 (Section~\ref{sec:new_joint_clmu}): five-task comparison with joint baselines (Table~\ref{tab:new_joint_baselines}), retain-set provenance (Table~\ref{tab:new_request_subset}), deletion timing (Table~\ref{tab:new_deletion_schedule}), the twenty-task stream (Table~\ref{tab:new_long_stream}), and mechanism ablations (Tables~\ref{tab:new_cl_only_ablation} and~\ref{tab:new_joint_ablation}). The unified U-LiRA privacy audit (Section~\ref{sec:new_unified_privacy}, Tables~\ref{tab:new_privacy_instance} and~\ref{tab:new_privacy_class}). Deletion, repair, and subsequent learning (Section~\ref{sec:new_repair_trajectory}, Tables~\ref{tab:new_class_repair}--\ref{tab:new_participant_10seed}). Additional CIFAR-100 unlearning ablations (Section~\ref{sec:new_cifar100_ablation}, Tables~\ref{tab:new_c100_cnn_class}--\ref{tab:new_c100_cnn_subclass}). Test-time memory latency (Section~\ref{sec:new_latency}, Table~\ref{tab:new_latency}).

\begin{table*}[t]
  \centering
  \caption{LwU in continual learning with deletion requests (Split CIFAR-10, 5 tasks $\times$ 2 classes, ResNet-18, Task-IL).}
  \label{tab:lwu_joint}
  \resizebox{\textwidth}{!}{%
  \begin{tabular}{lcccc}
    \toprule
    \textbf{Variant} & \textbf{ACC $\uparrow$} & \textbf{BWT $\uparrow$} & \textbf{$F_{\text{del}}$ $\downarrow$} & \textbf{$F_{\text{end}}$ $\downarrow$} \\
    \midrule
    \multicolumn{5}{l}{\textit{Continual learning only (no deletion)}} \\
    LwU (full) & $95.25\pm 0.46$ & $-0.34\pm 0.17$ & -- & $97.28\pm 1.42$ \\
    w/o self-distillation teacher & $75.46\pm 8.03$ & $-25.24\pm 9.85$ & -- & $64.52\pm 16.25$ \\
    w/o cumulative freeze mask & $96.38\pm 0.04$ & $-0.78\pm 0.40$ & -- & $95.57\pm 3.04$ \\
    w/o test-time memory & $95.13\pm 0.44$ & $-0.35\pm 0.14$ & -- & $96.67\pm 1.37$ \\
    Retrain (never sees $c_f$) & $96.27\pm 0.15$ & $-0.35\pm 0.11$ & -- & $0.00\pm 0.00$ \\
    \midrule
    \multicolumn{5}{l}{\textit{Joint: delete $c_f$ after task 3, then learn tasks 4--5}} \\
    w/o B resetting & $95.99\pm 0.68$ & $0.47\pm 0.39$ & $5.18\pm 5.62$ & $4.72\pm 4.10$ \\
    w/o Zone-C step & $95.93\pm 0.53$ & $0.52\pm 0.37$ & $7.05\pm 7.38$ & $5.17\pm 6.71$ \\
    w/o C projection & $95.93\pm 0.49$ & $0.51\pm 0.36$ & $7.44\pm 7.85$ & $7.41\pm 6.50$ \\
    w/o teacher repair (both passes) & $94.60\pm 0.89$ & $-1.09\pm 0.99$ & $96.00\pm 2.25$ & $96.22\pm 2.23$ \\
    w/o Zone-C repair pass & $95.48\pm 1.09$ & $-0.00\pm 0.83$ & $80.00\pm 25.76$ & $80.65\pm 24.07$ \\
    Unmasked teacher & $95.44\pm 0.76$ & $-0.11\pm 0.38$ & $94.96\pm 1.99$ & $95.49\pm 2.43$ \\
    Dominance $\omega{=}1$ & $87.05\pm 1.22$ & $-9.50\pm 0.98$ & $0.00\pm 0.00$ & $0.00\pm 0.00$ \\
    w/o self-distillation teacher & $79.95\pm 2.13$ & $-19.56\pm 2.25$ & $0.00\pm 0.00$ & $0.00\pm 0.00$ \\
    w/o cumulative freeze mask & $96.90\pm 0.27$ & $0.03\pm 0.60$ & $12.98\pm 22.48$ & $14.12\pm 24.45$ \\
    w/o test-time memory & $95.91\pm 0.54$ & $0.54\pm 0.29$ & $5.77\pm 5.58$ & $6.89\pm 6.93$ \\
    \midrule
    \multicolumn{5}{l}{\textit{Retain subset $\mathcal{D}'_r$ (default 10\%)}} \\
    $|\mathcal{D}'_r|=1\%$ & $95.79\pm 0.74$ & $0.23\pm 0.57$ & $62.85\pm 10.78$ & $66.24\pm 10.25$ \\
    $|\mathcal{D}'_r|=5\%$ & $95.88\pm 0.56$ & $0.42\pm 0.48$ & $25.37\pm 21.76$ & $27.69\pm 24.11$ \\
    $|\mathcal{D}'_r|=25\%$ & $95.89\pm 0.69$ & $0.42\pm 0.48$ & $0.00\pm 0.00$ & $0.00\pm 0.00$ \\
    \midrule
    \multicolumn{5}{l}{\textit{Deletion schedule}} \\
    Delete after task 2 & $95.82\pm 0.59$ & $0.37\pm 0.55$ & $17.60\pm 22.34$ & $16.90\pm 18.41$ \\
    Delete after task 4 & $95.91\pm 0.68$ & $0.48\pm 0.40$ & $2.16\pm 3.75$ & $1.58\pm 2.74$ \\
    Two requests (after 2 and 4) & $96.60\pm 0.76$ & $1.20\pm 0.66$ & $18.10\pm 22.10$ & $15.46\pm 23.92$ \\
    \midrule
    LwU (full, delete after task 3) & $95.91\pm 0.56$ & $0.42\pm 0.30$ & $5.77\pm 5.58$ & $6.89\pm 6.93$ \\
    \bottomrule
  \end{tabular}%
  }
  \vspace{2pt}

  \parbox{0.98\textwidth}{\footnotesize $c_f$ is the first class of task 1. Mean $\pm$ std over seeds 42--44 (\%). ACC/BWT over classes never deleted in the run (identical class set for every row). $F_{\text{del}}$ / $F_{\text{end}}$: Task-IL accuracy on $c_f$ right after deletion / after the last task (for two requests: the first deleted class). A relearn probe (one Adam epoch on $\mathcal{D}_f$ plus the task's other class) and a confidence MIA on $c_f$ were also run but do not separate the rows: relearn 76--99\% for every row including Retrain and MIA 48--53\%. Deletion quantiles selected on validation: $q_r=0.9$, $q_f=0.99$, $\Delta_C=fisher$, Zone-C pass $\lambda_C=0.1$ ($\lambda_C{=}1$ diverged for 23/27 grid points). Class-IL is omitted: every old task drops to 0\% without replay, so deletion is not measurable.}
\end{table*}
\begin{table*}[tp]
  \centering
  \caption{Component ablation of LwU for CIFAR-10 / CNN unlearning (class 0).}
  \label{tab:lwu_components_c10_cnn_class}
  \resizebox{\textwidth}{!}{%
  \begin{tabular}{lccccccc}
    \toprule
    \textbf{Variant} & \textbf{Forget $\downarrow$} & \textbf{Retain $\uparrow$} & \textbf{Test $\uparrow$} & \textbf{MIA $\rightarrow$ R} & \textbf{Gap $\downarrow$} & \textbf{Relearn@1 $\downarrow$} & \textbf{Time (s) $\downarrow$} \\
    \midrule
    Original (before deletion) & $91.20\pm 3.05$ & $91.29\pm 0.82$ & $91.28\pm 1.05$ & $56.47\pm 0.65$ & $32.51\pm 1.76$ & $95.53\pm 1.31$ & -- \\
    Retrain (oracle) & $0.00\pm 0.00$ & $91.37\pm 0.73$ & $82.23\pm 0.66$ & $50.42\pm 2.13$ & $0.00\pm 0.00$ & $0.00\pm 0.00$ & -- \\
    \midrule
    w/o B resetting & $34.50\pm 11.63$ & $91.89\pm 0.80$ & $86.11\pm 1.92$ & $56.35\pm 0.54$ & $13.65\pm 4.59$ & $95.50\pm 0.17$ & $11.21\pm 0.44$ \\
    w/o Zone-C step & $0.00\pm 0.00$ & $91.75\pm 0.73$ & $82.58\pm 0.65$ & $52.97\pm 1.44$ & $0.98\pm 0.31$ & $87.00\pm 12.74$ & $6.63\pm 0.23$ \\
    w/o C projection & $0.00\pm 0.00$ & $91.20\pm 0.85$ & $82.08\pm 0.76$ & $51.47\pm 3.05$ & $1.09\pm 0.27$ & $93.47\pm 2.91$ & $9.05\pm 0.47$ \\
    w/o teacher repair (both passes) & $11.83\pm 20.50$ & $84.14\pm 8.06$ & $76.85\pm 8.86$ & $51.96\pm 3.55$ & $7.28\pm 5.48$ & $95.00\pm 1.51$ & $6.14\pm 0.39$ \\
    w/o Zone-C repair pass & $0.00\pm 0.00$ & $90.96\pm 0.45$ & $81.86\pm 0.41$ & $50.79\pm 2.81$ & $0.80\pm 0.10$ & $90.57\pm 7.61$ & $8.83\pm 0.37$ \\
    w/o test-time memory & $0.00\pm 0.00$ & $91.67\pm 0.63$ & $82.50\pm 0.57$ & $53.52\pm 1.20$ & $1.13\pm 0.31$ & $91.73\pm 5.11$ & $11.73\pm 0.45$ \\
    + weight alignment & $0.00\pm 0.00$ & $91.64\pm 0.59$ & $82.48\pm 0.52$ & $52.03\pm 2.52$ & $1.01\pm 0.47$ & $91.63\pm 4.99$ & $11.30\pm 0.33$ \\
    \midrule
    LwU (full) & $0.00\pm 0.00$ & $91.67\pm 0.64$ & $82.51\pm 0.60$ & $51.89\pm 2.60$ & $1.03\pm 0.42$ & $91.73\pm 5.11$ & $11.73\pm 0.45$ \\
    \bottomrule
  \end{tabular}%
  }
  \vspace{2pt}

  \parbox{0.98\textwidth}{\footnotesize Each row removes or adds one component; everything else is the validation-selected full configuration. Mean $\pm$ std over seeds 42--44; accuracies and MIA in \%. Forget/Retain: test accuracy on the deleted/retained data; Test: all test classes. MIA $\rightarrow$ R: closer to Retrain is better. Gap: mean absolute difference to Retrain over Forget, Retain, and MIA. Relearn@1: forget accuracy after 1 Adam epoch(s) ($10^{-4}$, BatchNorm statistics frozen) on $\mathcal{D}_f$ plus as many retained training images (lower = less recoverable knowledge). Time: unlearning wall-clock (importance + edit + repair), best of two repeats with the GPU exclusive.}
\end{table*}
\begin{table*}[tp]
  \centering
  \caption{Component ablation of LwU for CIFAR-10 instance / CNN unlearning (500 random training images).}
  \label{tab:lwu_components_c10_cnn_instance}
  \resizebox{\textwidth}{!}{%
  \begin{tabular}{lccccccc}
    \toprule
    \textbf{Variant} & \textbf{Forget $\downarrow$} & \textbf{Retain $\uparrow$} & \textbf{Test $\uparrow$} & \textbf{MIA $\rightarrow$ R} & \textbf{Gap $\downarrow$} & \textbf{Relearn@1 $\downarrow$} & \textbf{Time (s) $\downarrow$} \\
    \midrule
    Original (before deletion) & $99.74\pm 0.44$ & $91.28\pm 1.05$ & $91.28\pm 1.05$ & $56.92\pm 2.66$ & $5.83\pm 0.76$ & $100.00\pm 0.00$ & -- \\
    Retrain (oracle) & $91.28\pm 2.70$ & $91.26\pm 0.90$ & $91.26\pm 0.90$ & $48.21\pm 1.46$ & $0.00\pm 0.00$ & $94.36\pm 2.47$ & -- \\
    \midrule
    w/o B resetting & $99.74\pm 0.44$ & $91.36\pm 0.99$ & $91.36\pm 0.99$ & $56.67\pm 3.45$ & $5.73\pm 0.62$ & $100.00\pm 0.00$ & $11.64\pm 3.11$ \\
    w/o Zone-C step & $96.92\pm 3.35$ & $89.82\pm 3.04$ & $89.82\pm 3.04$ & $55.00\pm 3.42$ & $4.87\pm 1.95$ & $97.18\pm 2.22$ & $7.70\pm 1.98$ \\
    w/o C projection & $96.67\pm 3.79$ & $89.76\pm 3.12$ & $89.76\pm 3.12$ & $54.62\pm 3.33$ & $4.84\pm 1.55$ & $96.92\pm 2.66$ & $10.21\pm 2.64$ \\
    w/o teacher repair (both passes) & $71.79\pm 34.49$ & $66.18\pm 32.35$ & $66.18\pm 32.35$ & $51.92\pm 4.80$ & $18.06\pm 19.82$ & $83.85\pm 18.14$ & $5.65\pm 1.44$ \\
    w/o Zone-C repair pass & $96.41\pm 4.24$ & $89.29\pm 3.92$ & $89.29\pm 3.92$ & $54.62\pm 3.53$ & $5.08\pm 0.86$ & $97.44\pm 2.47$ & $10.04\pm 2.99$ \\
    w/o test-time memory & $96.92\pm 3.35$ & $89.88\pm 2.88$ & $89.88\pm 2.88$ & $54.87\pm 2.78$ & $4.80\pm 1.63$ & $97.18\pm 2.22$ & $10.03\pm 0.32$ \\
    + weight alignment & $97.18\pm 2.91$ & $89.80\pm 2.99$ & $89.80\pm 2.99$ & $53.97\pm 3.11$ & $4.43\pm 2.15$ & $97.44\pm 1.78$ & $14.58\pm 4.36$ \\
    \midrule
    LwU (full) & $96.92\pm 3.35$ & $89.85\pm 2.92$ & $89.85\pm 2.92$ & $54.87\pm 2.78$ & $4.81\pm 1.61$ & $97.18\pm 2.22$ & $10.03\pm 0.32$ \\
    \bottomrule
  \end{tabular}%
  }
  \vspace{2pt}

  \parbox{0.98\textwidth}{\footnotesize Each row removes or adds one component; everything else is the validation-selected full configuration. Mean $\pm$ std over seeds 42--44; accuracies and MIA in \%. Forget/Retain: test accuracy on the deleted / retained data; Test: all test classes. MIA $\rightarrow$ R: closer to Retrain is better. Gap: mean absolute difference to Retrain over Forget, Retain and MIA. Relearn@1: forget accuracy after 1 Adam epoch(s) ($10^{-4}$, BatchNorm statistics frozen) on $\mathcal{D}_f$ plus as many retained training images (lower = less recoverable knowledge). Time: unlearning wall-clock (importance + edit + repair), best of two repeats with the GPU exclusive.}
\end{table*}
\begin{table*}[tp]
  \centering
  \caption{Component ablation of LwU for CIFAR-10 subclass / CNN unlearning (fine class airplane under 5 coarse labels).}
  \label{tab:lwu_components_c10_cnn_subclass}
  \resizebox{\textwidth}{!}{%
  \begin{tabular}{lccccccc}
    \toprule
    \textbf{Variant} & \textbf{Forget $\downarrow$} & \textbf{Retain $\uparrow$} & \textbf{Test $\uparrow$} & \textbf{MIA $\rightarrow$ R} & \textbf{Gap $\downarrow$} & \textbf{Relearn@1 $\downarrow$} & \textbf{Time (s) $\downarrow$} \\
    \midrule
    Original (before deletion) & $96.00\pm 1.30$ & $93.46\pm 0.59$ & $93.71\pm 0.48$ & $53.40\pm 0.52$ & $23.42\pm 1.34$ & $97.43\pm 0.29$ & -- \\
    Retrain (oracle) & $28.67\pm 5.58$ & $93.99\pm 0.22$ & $87.46\pm 0.44$ & $51.01\pm 1.10$ & $0.00\pm 0.00$ & $95.10\pm 0.26$ & -- \\
    \midrule
    w/o B resetting & $95.83\pm 1.34$ & $93.68\pm 0.53$ & $93.86\pm 0.44$ & $53.75\pm 0.73$ & $23.45\pm 1.21$ & $97.43\pm 0.38$ & $11.07\pm 0.40$ \\
    w/o Zone-C step & $68.83\pm 4.36$ & $90.55\pm 0.73$ & $88.30\pm 0.82$ & $50.33\pm 1.51$ & $14.78\pm 3.05$ & $94.93\pm 1.42$ & $6.28\pm 0.29$ \\
    w/o C projection & $68.13\pm 4.90$ & $90.92\pm 0.43$ & $88.58\pm 0.75$ & $51.24\pm 2.08$ & $14.41\pm 3.08$ & $94.30\pm 1.48$ & $8.85\pm 0.24$ \\
    w/o teacher repair (both passes) & $9.10\pm 15.50$ & $38.04\pm 11.77$ & $35.01\pm 11.58$ & $49.02\pm 0.34$ & $26.88\pm 8.43$ & $94.07\pm 1.80$ & $5.81\pm 0.21$ \\
    w/o Zone-C repair pass & $64.30\pm 7.28$ & $88.98\pm 2.85$ & $86.45\pm 3.29$ & $51.35\pm 0.49$ & $13.74\pm 0.75$ & $94.40\pm 1.48$ & $8.53\pm 0.38$ \\
    w/o test-time memory & $68.93\pm 2.47$ & $89.80\pm 2.11$ & $87.72\pm 1.91$ & $51.76\pm 1.46$ & $15.11\pm 2.74$ & $94.63\pm 1.40$ & $11.20\pm 0.22$ \\
    + weight alignment & $68.10\pm 2.25$ & $89.69\pm 2.11$ & $87.51\pm 1.73$ & $51.56\pm 0.22$ & $14.86\pm 3.26$ & $94.50\pm 1.82$ & $11.42\pm 0.62$ \\
    \midrule
    LwU (full) & $69.23\pm 2.87$ & $89.80\pm 2.13$ & $87.69\pm 1.90$ & $51.37\pm 0.89$ & $15.11\pm 2.75$ & $94.63\pm 1.40$ & $11.20\pm 0.22$ \\
    \bottomrule
  \end{tabular}%
  }
  \vspace{2pt}

  \parbox{0.98\textwidth}{\footnotesize Each row removes or adds one component; everything else is the validation-selected full configuration. Mean $\pm$ std over seeds 42--44; accuracies and MIA in \%. Forget/Retain: test accuracy on the deleted / retained data; Test: all test classes. MIA $\rightarrow$ R: closer to Retrain is better. Gap: mean absolute difference to Retrain over Forget, Retain and MIA. Relearn@1: forget accuracy after 1 Adam epoch(s) ($10^{-4}$, BatchNorm statistics frozen) on $\mathcal{D}_f$ plus as many retained training images (lower = less recoverable knowledge). Time: unlearning wall-clock (importance + edit + repair), best of two repeats with the GPU exclusive.}
\end{table*}
\begin{table*}[tp]
  \centering
  \caption{Component ablation of LwU for CIFAR-10 / ResNet-18 unlearning (class 0).}
  \label{tab:lwu_components_c10_r18_class}
  \resizebox{\textwidth}{!}{%
  \begin{tabular}{lccccccc}
    \toprule
    \textbf{Variant} & \textbf{Forget $\downarrow$} & \textbf{Retain $\uparrow$} & \textbf{Test $\uparrow$} & \textbf{MIA $\rightarrow$ R} & \textbf{Gap $\downarrow$} & \textbf{Relearn@1 $\downarrow$} & \textbf{Time (s) $\downarrow$} \\
    \midrule
    Original (before deletion) & $93.97\pm 1.33$ & $92.42\pm 0.25$ & $92.58\pm 0.11$ & $57.24\pm 0.27$ & $33.48\pm 0.39$ & $96.17\pm 0.57$ & -- \\
    Retrain (oracle) & $0.00\pm 0.00$ & $92.82\pm 0.08$ & $83.54\pm 0.08$ & $51.18\pm 0.59$ & $0.00\pm 0.00$ & $0.00\pm 0.00$ & -- \\
    \midrule
    w/o B resetting & $77.13\pm 1.78$ & $93.03\pm 0.14$ & $91.42\pm 0.13$ & $57.37\pm 0.31$ & $27.85\pm 0.54$ & $95.57\pm 0.38$ & $18.82\pm 0.36$ \\
    w/o Zone-C step & $0.00\pm 0.00$ & $92.90\pm 0.29$ & $83.61\pm 0.26$ & $53.11\pm 0.71$ & $0.71\pm 0.29$ & $95.37\pm 0.25$ & $11.01\pm 0.12$ \\
    w/o C projection & $0.00\pm 0.00$ & $92.90\pm 0.29$ & $83.61\pm 0.24$ & $53.12\pm 0.86$ & $0.71\pm 0.38$ & $95.37\pm 0.57$ & $14.76\pm 0.63$ \\
    w/o teacher repair (both passes) & $36.23\pm 35.75$ & $90.91\pm 1.34$ & $85.43\pm 2.40$ & $53.28\pm 1.63$ & $13.42\pm 12.68$ & $96.50\pm 0.66$ & $9.42\pm 0.00$ \\
    w/o Zone-C repair pass & $0.00\pm 0.00$ & $92.60\pm 0.28$ & $83.33\pm 0.24$ & $52.72\pm 1.34$ & $0.59\pm 0.41$ & $95.47\pm 0.50$ & $13.95\pm 0.29$ \\
    w/o test-time memory & $0.00\pm 0.00$ & $92.88\pm 0.25$ & $83.59\pm 0.23$ & $53.68\pm 0.95$ & $0.89\pm 0.42$ & $95.40\pm 0.46$ & $17.86\pm 0.33$ \\
    + weight alignment & $0.00\pm 0.00$ & $92.85\pm 0.32$ & $83.55\pm 0.29$ & $53.45\pm 1.00$ & $0.82\pm 0.41$ & $95.23\pm 0.65$ & $19.50\pm 0.91$ \\
    \midrule
    LwU (full) & $0.00\pm 0.00$ & $92.87\pm 0.26$ & $83.59\pm 0.22$ & $53.44\pm 0.83$ & $0.81\pm 0.37$ & $95.40\pm 0.46$ & $17.86\pm 0.33$ \\
    \bottomrule
  \end{tabular}%
  }
  \vspace{2pt}

  \parbox{0.98\textwidth}{\footnotesize Each row removes or adds one component; everything else is the validation-selected full configuration. Mean $\pm$ std over seeds 42--44; accuracies and MIA in \%. Forget/Retain: test accuracy on the deleted / retained data; Test: all test classes. MIA $\rightarrow$ R: closer to Retrain is better. Gap: mean absolute difference to Retrain over Forget, Retain and MIA. Relearn@1: forget accuracy after 1 Adam epoch(s) ($10^{-4}$, BatchNorm statistics frozen) on $\mathcal{D}_f$ plus as many retained training images (lower = less recoverable knowledge). Time: unlearning wall-clock (importance + edit + repair), best of two repeats with the GPU exclusive.}
\end{table*}
\begin{table*}[tp]
  \centering
  \caption{Component ablation of LwU for CIFAR-100 / ResNet-18 unlearning (class 0).}
  \label{tab:lwu_components_c100_r18_class}
  \resizebox{\textwidth}{!}{%
  \begin{tabular}{lccccccc}
    \toprule
    \textbf{Variant} & \textbf{Forget $\downarrow$} & \textbf{Retain $\uparrow$} & \textbf{Test $\uparrow$} & \textbf{MIA $\rightarrow$ R} & \textbf{Gap $\downarrow$} & \textbf{Relearn@1 $\downarrow$} & \textbf{Time (s) $\downarrow$} \\
    \midrule
    Original (before deletion) & $81.67\pm 2.08$ & $70.97\pm 0.43$ & $71.08\pm 0.44$ & $64.77\pm 0.49$ & $32.44\pm 0.27$ & $92.33\pm 1.53$ & -- \\
    Retrain (oracle) & $0.00\pm 0.00$ & $70.23\pm 0.14$ & $69.52\pm 0.14$ & $49.85\pm 2.66$ & $0.00\pm 0.00$ & $0.00\pm 0.00$ & -- \\
    \midrule
    w/o B resetting & $67.00\pm 1.00$ & $70.97\pm 0.40$ & $70.93\pm 0.38$ & $63.70\pm 0.56$ & $27.20\pm 1.15$ & $94.00\pm 2.65$ & $15.64\pm 0.64$ \\
    w/o Zone-C step & $0.00\pm 0.00$ & $70.45\pm 0.63$ & $69.76\pm 0.64$ & $49.55\pm 2.70$ & $0.76\pm 0.16$ & $0.00\pm 0.00$ & $10.60\pm 0.72$ \\
    w/o C projection & $0.00\pm 0.00$ & $70.31\pm 0.66$ & $69.62\pm 0.67$ & $47.23\pm 2.19$ & $1.20\pm 1.08$ & $0.00\pm 0.00$ & $12.00\pm 0.60$ \\
    w/o teacher repair (both passes) & $9.67\pm 9.81$ & $66.67\pm 4.49$ & $66.10\pm 4.35$ & $50.15\pm 1.85$ & $5.26\pm 5.24$ & $93.00\pm 4.36$ & $6.31\pm 0.22$ \\
    w/o Zone-C repair pass & $0.00\pm 0.00$ & $70.22\pm 0.83$ & $69.52\pm 0.82$ & $49.53\pm 3.06$ & $0.33\pm 0.03$ & $0.00\pm 0.00$ & $11.18\pm 0.59$ \\
    w/o test-time memory & $0.00\pm 0.00$ & $70.40\pm 0.71$ & $69.69\pm 0.70$ & $48.95\pm 1.60$ & $0.52\pm 0.50$ & $0.00\pm 0.00$ & $15.86\pm 0.20$ \\
    + weight alignment & $0.00\pm 0.00$ & $70.45\pm 0.69$ & $69.75\pm 0.68$ & $49.22\pm 1.21$ & $0.80\pm 0.37$ & $0.00\pm 0.00$ & $15.78\pm 0.61$ \\
    \midrule
    LwU (full) & $0.00\pm 0.00$ & $70.40\pm 0.71$ & $69.70\pm 0.70$ & $48.92\pm 1.54$ & $0.53\pm 0.52$ & $0.00\pm 0.00$ & $15.86\pm 0.20$ \\
    \bottomrule
  \end{tabular}%
  }
  \vspace{2pt}

  \parbox{0.98\textwidth}{\footnotesize Each row removes or adds one component; everything else is the validation-selected full configuration. Mean $\pm$ std over seeds 42--44; accuracies and MIA in \%. Forget/Retain: test accuracy on the deleted / retained data; Test: all test classes. MIA $\rightarrow$ R: closer to Retrain is better. Gap: mean absolute difference to Retrain over Forget, Retain and MIA. Relearn@1: forget accuracy after 1 Adam epoch(s) ($10^{-4}$, BatchNorm statistics frozen) on $\mathcal{D}_f$ plus as many retained training images (lower = less recoverable knowledge). Time: unlearning wall-clock (importance + edit + repair), best of two repeats with the GPU exclusive.}
\end{table*}
\begin{table*}[tp]
  \centering
  \caption{Sensitivity and design ablations for LwU, CIFAR-10 / CNN, class 0.}
  \label{tab:lwu_extra_c10_cnn_class}
  \resizebox{\textwidth}{!}{%
  \begin{tabular}{lcccccc}
    \toprule
    \textbf{Variant} & \textbf{Forget $\downarrow$} & \textbf{Retain $\uparrow$} & \textbf{Test $\uparrow$} & \textbf{MIA $\rightarrow$ R} & \textbf{Gap $\downarrow$} & \textbf{Relearn@1 $\downarrow$} \\
    \midrule
    LwU (full, selected) & $0.00\pm 0.00$ & $91.67\pm 0.64$ & $82.51\pm 0.60$ & $51.89\pm 2.60$ & $1.03\pm 0.42$ & $91.73\pm 5.11$ \\
    \midrule
    \multicolumn{7}{l}{\textit{Zone construction and teacher}} \\
    Dominance $\omega{=}1$ & $0.00\pm 0.00$ & $31.81\pm 32.79$ & $28.59\pm 29.57$ & $51.08\pm 0.93$ & $20.10\pm 10.93$ & $26.27\pm 45.50$ \\
    Unmasked teacher & $27.77\pm 37.79$ & $91.63\pm 0.79$ & $85.19\pm 3.15$ & $54.14\pm 1.68$ & $10.58\pm 12.57$ & $95.43\pm 1.77$ \\
    \midrule
    \multicolumn{7}{l}{\textit{Retain subset $\mathcal{D}'_r$ (default 10\%)}} \\
    $|\mathcal{D}'_r|=1\%$ & $0.00\pm 0.00$ & $87.80\pm 5.10$ & $78.99\pm 4.64$ & $52.37\pm 2.38$ & $1.87\pm 1.85$ & $85.40\pm 18.27$ \\
    $|\mathcal{D}'_r|=5\%$ & $4.90\pm 6.91$ & $91.51\pm 0.66$ & $82.84\pm 0.49$ & $53.30\pm 3.24$ & $2.64\pm 2.52$ & $94.67\pm 3.26$ \\
    $|\mathcal{D}'_r|=25\%$ & $0.00\pm 0.00$ & $91.79\pm 0.61$ & $82.62\pm 0.53$ & $52.39\pm 1.58$ & $0.79\pm 0.34$ & $62.73\pm 54.36$ \\
    \midrule
    \multicolumn{7}{l}{\textit{Zone-C budget $\Delta_C$ ($\eta_C=10^{-2}$)}} \\
    $\Delta_C$ Fisher-scaled (paper) & $0.00\pm 0.00$ & $91.69\pm 0.67$ & $82.52\pm 0.60$ & $52.81\pm 1.39$ & $0.90\pm 0.34$ & $85.70\pm 14.97$ \\
    $\Delta_C=10^{-4}\lVert\theta_C\rVert$ & $0.00\pm 0.00$ & $91.69\pm 0.68$ & $82.52\pm 0.60$ & $52.73\pm 1.48$ & $0.88\pm 0.40$ & $85.67\pm 14.87$ \\
    $\Delta_C=10^{-3}\lVert\theta_C\rVert$ & $0.00\pm 0.00$ & $91.71\pm 0.72$ & $82.54\pm 0.66$ & $52.54\pm 1.21$ & $0.82\pm 0.40$ & $87.37\pm 12.37$ \\
    $\Delta_C=10^{-2}\lVert\theta_C\rVert$ & $0.00\pm 0.00$ & $91.69\pm 0.61$ & $82.53\pm 0.56$ & $51.73\pm 2.38$ & $0.99\pm 0.45$ & $91.87\pm 5.27$ \\
    $\Delta_C=\infty$ (no clip) & $0.00\pm 0.00$ & $91.69\pm 0.61$ & $82.53\pm 0.56$ & $51.73\pm 2.38$ & $0.99\pm 0.45$ & $91.87\pm 5.27$ \\
    \midrule
    \multicolumn{7}{l}{\textit{Zone-C step size, unclipped}} \\
    $\eta_C=10^{-3}$, no clip & $0.00\pm 0.00$ & $91.70\pm 0.68$ & $82.52\pm 0.62$ & $52.67\pm 1.56$ & $0.86\pm 0.37$ & $86.57\pm 13.48$ \\
    $\eta_C=10^{-1}$, no clip (3/3 diverged) & -- & -- & -- & -- & -- & -- \\
    $\eta_C=1$, no clip (3/3 diverged) & -- & -- & -- & -- & -- & -- \\
    \bottomrule
  \end{tabular}%
  }
  \vspace{2pt}

  \parbox{0.98\textwidth}{\footnotesize Selected Zone-C budget per seed: fisher, inf, inf (relative to $\lVert\theta_C\rVert$; ``fisher'' = paper scaling). Mean $\pm$ std over seeds 42--44; accuracies and MIA in \%. Forget/Retain: test accuracy on the deleted / retained data; Test: all test classes. MIA $\rightarrow$ R: closer to Retrain is better. Gap: mean absolute difference to Retrain over Forget, Retain and MIA. Relearn@1: forget accuracy after 1 Adam epoch(s) ($10^{-4}$, BatchNorm statistics frozen) on $\mathcal{D}_f$ plus as many retained training images (lower = less recoverable knowledge). }
\end{table*}
\begin{table*}[tp]
  \centering
  \caption{Sensitivity and design ablations for LwU, CIFAR-10 / ResNet-18, class 0.}
  \label{tab:lwu_extra_c10_r18_class}
  \resizebox{\textwidth}{!}{%
  \begin{tabular}{lcccccc}
    \toprule
    \textbf{Variant} & \textbf{Forget $\downarrow$} & \textbf{Retain $\uparrow$} & \textbf{Test $\uparrow$} & \textbf{MIA $\rightarrow$ R} & \textbf{Gap $\downarrow$} & \textbf{Relearn@1 $\downarrow$} \\
    \midrule
    LwU (full, selected) & $0.00\pm 0.00$ & $92.87\pm 0.26$ & $83.59\pm 0.22$ & $53.44\pm 0.83$ & $0.81\pm 0.37$ & $95.40\pm 0.46$ \\
    \midrule
    \multicolumn{7}{l}{\textit{Zone construction and teacher}} \\
    Dominance $\omega{=}1$ & $0.00\pm 0.00$ & $52.53\pm 33.23$ & $47.27\pm 29.92$ & $50.08\pm 2.37$ & $14.10\pm 10.58$ & $92.40\pm 1.47$ \\
    Unmasked teacher & $2.57\pm 2.96$ & $92.86\pm 0.28$ & $83.83\pm 0.51$ & $54.21\pm 0.70$ & $1.92\pm 1.11$ & $96.03\pm 0.49$ \\
    \midrule
    \multicolumn{7}{l}{\textit{Retain subset $\mathcal{D}'_r$ (default 10\%)}} \\
    $|\mathcal{D}'_r|=1\%$ & $6.53\pm 11.32$ & $91.64\pm 0.11$ & $83.14\pm 1.05$ & $52.57\pm 2.68$ & $3.12\pm 4.67$ & $96.73\pm 0.87$ \\
    $|\mathcal{D}'_r|=5\%$ & $46.30\pm 40.12$ & $92.72\pm 0.27$ & $88.06\pm 3.83$ & $55.17\pm 1.95$ & $16.83\pm 13.88$ & $95.80\pm 0.92$ \\
    $|\mathcal{D}'_r|=25\%$ & $0.83\pm 1.44$ & $93.09\pm 0.13$ & $83.85\pm 0.06$ & $54.96\pm 1.15$ & $1.63\pm 0.37$ & $95.70\pm 0.44$ \\
    \midrule
    \multicolumn{7}{l}{\textit{Zone-C budget $\Delta_C$ ($\eta_C=10^{-2}$)}} \\
    $\Delta_C$ Fisher-scaled (paper) & $0.00\pm 0.00$ & $92.87\pm 0.26$ & $83.59\pm 0.23$ & $53.19\pm 0.76$ & $0.73\pm 0.31$ & $95.33\pm 0.31$ \\
    $\Delta_C=10^{-4}\lVert\theta_C\rVert$ & $0.00\pm 0.00$ & $92.89\pm 0.27$ & $83.60\pm 0.24$ & $53.26\pm 0.80$ & $0.75\pm 0.34$ & $95.37\pm 0.25$ \\
    $\Delta_C=10^{-3}\lVert\theta_C\rVert$ & $0.00\pm 0.00$ & $92.90\pm 0.24$ & $83.61\pm 0.22$ & $53.49\pm 0.78$ & $0.83\pm 0.33$ & $95.30\pm 0.40$ \\
    $\Delta_C=10^{-2}\lVert\theta_C\rVert$ & $0.00\pm 0.00$ & $92.91\pm 0.25$ & $83.62\pm 0.22$ & $53.54\pm 0.76$ & $0.85\pm 0.35$ & $95.37\pm 0.50$ \\
    $\Delta_C=\infty$ (no clip) & $0.00\pm 0.00$ & $92.91\pm 0.25$ & $83.62\pm 0.22$ & $53.54\pm 0.76$ & $0.85\pm 0.35$ & $95.37\pm 0.50$ \\
    \midrule
    \multicolumn{7}{l}{\textit{Zone-C step size, unclipped}} \\
    $\eta_C=10^{-3}$, no clip & $0.00\pm 0.00$ & $92.86\pm 0.28$ & $83.58\pm 0.27$ & $53.29\pm 0.80$ & $0.76\pm 0.32$ & $95.33\pm 0.25$ \\
    $\eta_C=10^{-1}$, no clip & $0.00\pm 0.00$ & $68.19\pm 40.10$ & $61.37\pm 36.07$ & $51.36\pm 2.49$ & $8.76\pm 13.08$ & $93.87\pm 2.10$ \\
    $\eta_C=1$, no clip (3/3 diverged) & -- & -- & -- & -- & -- & -- \\
    \bottomrule
  \end{tabular}%
  }
  \vspace{2pt}

  \parbox{0.98\textwidth}{\footnotesize Selected Zone-C budget per seed: fisher, 0.001, inf (relative to $\lVert\theta_C\rVert$; ``fisher'' = paper scaling). Mean $\pm$ std over seeds 42--44; accuracies and MIA in \%. Forget/Retain: test accuracy on the deleted / retained data; Test: all test classes. MIA $\rightarrow$ R: closer to Retrain is better. Gap: mean absolute difference to Retrain over Forget, Retain and MIA. Relearn@1: forget accuracy after 1 Adam epoch(s) ($10^{-4}$, BatchNorm statistics frozen) on $\mathcal{D}_f$ plus as many retained training images (lower = less recoverable knowledge). }
\end{table*}
\begin{table*}[tp]
  \centering
  \caption{Gap to Retrain ($\downarrow$, mean of $|\Delta|$ Forget, Retain, MIA) for every unlearning ablation across datasets, backbones and deletion granularities.}
  \label{tab:lwu_gap_summary}
  \resizebox{\textwidth}{!}{%
  \begin{tabular}{lccccc}
    \toprule
    \textbf{Variant} & \textbf{C10 CNN} & \textbf{C10 R18} & \textbf{C100 R18} & \textbf{C10-sub CNN} & \textbf{C10-inst CNN} \\
    \midrule
    LwU (full) & $1.0\pm 0.4$ & $0.8\pm 0.4$ & $0.5\pm 0.5$ & $15.1\pm 2.7$ & $4.8\pm 1.6$ \\
    \midrule
    w/o B resetting & $13.6\pm 4.6$ & $27.8\pm 0.5$ & $27.2\pm 1.1$ & $23.4\pm 1.2$ & $5.7\pm 0.6$ \\
    w/o Zone-C step & $1.0\pm 0.3$ & $0.7\pm 0.3$ & $0.8\pm 0.2$ & $14.8\pm 3.1$ & $4.9\pm 2.0$ \\
    w/o C projection & $1.1\pm 0.3$ & $0.7\pm 0.4$ & $1.2\pm 1.1$ & $14.4\pm 3.1$ & $4.8\pm 1.6$ \\
    w/o teacher repair (both passes) & $7.3\pm 5.5$ & $13.4\pm 12.7$ & $5.3\pm 5.2$ & $26.9\pm 8.4$ & $18.1\pm 19.8$ \\
    w/o Zone-C repair pass & $0.8\pm 0.1$ & $0.6\pm 0.4$ & $0.3\pm 0.0$ & $13.7\pm 0.8$ & $5.1\pm 0.9$ \\
    w/o test-time memory & $1.1\pm 0.3$ & $0.9\pm 0.4$ & $0.5\pm 0.5$ & $15.1\pm 2.7$ & $4.8\pm 1.6$ \\
    + weight alignment & $1.0\pm 0.5$ & $0.8\pm 0.4$ & $0.8\pm 0.4$ & $14.9\pm 3.3$ & $4.4\pm 2.1$ \\
    Random mask & $22.9\pm 4.1$ & $31.4\pm 0.4$ & -- & -- & -- \\
    Magnitude-based & $19.4\pm 4.5$ & $30.1\pm 0.3$ & -- & -- & -- \\
    Fisher retain only & $16.3\pm 3.7$ & $28.5\pm 0.2$ & -- & -- & -- \\
    Fisher forget only & $16.0\pm 3.6$ & $28.4\pm 0.3$ & -- & -- & -- \\
    Dominance $\omega{=}1$ & $20.1\pm 10.9$ & $14.1\pm 10.6$ & -- & -- & -- \\
    Unmasked teacher & $10.6\pm 12.6$ & $1.9\pm 1.1$ & -- & -- & -- \\
    \bottomrule
  \end{tabular}%
  }
  \vspace{2pt}

  \parbox{0.98\textwidth}{\footnotesize Mean $\pm$ std over 3 seeds; -- = not run.}
\end{table*}
\begin{table*}[tp]
  \centering
  \caption{Importance-estimator ablation for LwU, CIFAR-10 / CNN, class 0.}
  \label{tab:lwu_importance_c10_cnn_class}
  \resizebox{\textwidth}{!}{%
  \begin{tabular}{llccccccc}
    \toprule
    \textbf{Importance estimator} & \textbf{Score} & \textbf{Forget $\downarrow$} & \textbf{Retain $\uparrow$} & \textbf{Test $\uparrow$} & \textbf{MIA $\rightarrow$ R} & \textbf{Gap $\downarrow$} & \textbf{Relearn@1 $\downarrow$} & \textbf{Time (s) $\downarrow$} \\
    \midrule
    Random mask & $s_i\sim\mathcal{U}(0,1)$ & $62.60\pm 9.20$ & $91.79\pm 0.87$ & $88.82\pm 1.54$ & $56.18\pm 1.00$ & $22.92\pm 4.08$ & $95.90\pm 0.56$ & $10.70\pm 0.67$ \\
    Magnitude-based & $s_i=|\theta_i|$ & $51.40\pm 10.59$ & $91.90\pm 0.81$ & $87.82\pm 1.76$ & $56.64\pm 1.06$ & $19.38\pm 4.46$ & $95.63\pm 0.40$ & $10.44\pm 0.38$ \\
    Fisher retain only & $s_i=F^r_{ii}$ & $42.37\pm 8.48$ & $91.86\pm 0.80$ & $86.87\pm 1.55$ & $56.48\pm 0.97$ & $16.30\pm 3.70$ & $95.60\pm 0.44$ & $10.75\pm 0.56$ \\
    Fisher forget only & $s_i=F^f_{ii}$ & $41.57\pm 8.27$ & $91.88\pm 0.80$ & $86.80\pm 1.51$ & $56.42\pm 1.05$ & $16.02\pm 3.63$ & $95.60\pm 0.44$ & $10.81\pm 0.38$ \\
    \midrule
    Dual Fisher (ours) & $(F^r_{ii},F^f_{ii})$ & $0.00\pm 0.00$ & $91.67\pm 0.64$ & $82.51\pm 0.60$ & $51.89\pm 2.60$ & $1.03\pm 0.42$ & $91.73\pm 5.11$ & $11.73\pm 0.45$ \\
    \bottomrule
  \end{tabular}%
  }
  \vspace{2pt}

  \parbox{0.98\textwidth}{\footnotesize Single-score estimators use the same map for the retain and forget sides; all other settings equal the full method (same row as in the component table). Mean $\pm$ std over seeds 42--44; accuracies and MIA in \%. Forget/Retain: test accuracy on the deleted / retained data; Test: all test classes. MIA $\rightarrow$ R: closer to Retrain is better. Gap: mean absolute difference to Retrain over Forget, Retain and MIA. Relearn@1: forget accuracy after 1 Adam epoch(s) ($10^{-4}$, BatchNorm statistics frozen) on $\mathcal{D}_f$ plus as many retained training images (lower = less recoverable knowledge). Time: unlearning wall-clock (importance + edit + repair), best of two repeats with the GPU exclusive.}
\end{table*}
\begin{table*}[tp]
  \centering
  \caption{Importance-estimator ablation for LwU, CIFAR-10 / ResNet-18, class 0.}
  \label{tab:lwu_importance_c10_r18_class}
  \resizebox{\textwidth}{!}{%
  \begin{tabular}{llccccccc}
    \toprule
    \textbf{Importance estimator} & \textbf{Score} & \textbf{Forget $\downarrow$} & \textbf{Retain $\uparrow$} & \textbf{Test $\uparrow$} & \textbf{MIA $\rightarrow$ R} & \textbf{Gap $\downarrow$} & \textbf{Relearn@1 $\downarrow$} & \textbf{Time (s) $\downarrow$} \\
    \midrule
    Random mask & $s_i\sim\mathcal{U}(0,1)$ & $87.80\pm 1.57$ & $92.81\pm 0.22$ & $92.30\pm 0.14$ & $57.33\pm 0.17$ & $31.37\pm 0.38$ & $95.83\pm 1.01$ & $16.96\pm 0.54$ \\
    Magnitude-based & $s_i=|\theta_i|$ & $83.80\pm 0.82$ & $92.93\pm 0.22$ & $91.97\pm 0.14$ & $57.42\pm 0.53$ & $30.06\pm 0.34$ & $96.00\pm 0.66$ & $17.94\pm 0.70$ \\
    Fisher retain only & $s_i=F^r_{ii}$ & $78.87\pm 0.97$ & $93.00\pm 0.15$ & $91.54\pm 0.02$ & $57.73\pm 0.28$ & $28.53\pm 0.20$ & $96.03\pm 0.51$ & $17.77\pm 0.55$ \\
    Fisher forget only & $s_i=F^f_{ii}$ & $78.43\pm 1.16$ & $93.01\pm 0.16$ & $91.52\pm 0.04$ & $57.67\pm 0.25$ & $28.37\pm 0.27$ & $96.03\pm 0.49$ & $18.00\pm 0.70$ \\
    \midrule
    Dual Fisher (ours) & $(F^r_{ii},F^f_{ii})$ & $0.00\pm 0.00$ & $92.87\pm 0.26$ & $83.59\pm 0.22$ & $53.44\pm 0.83$ & $0.81\pm 0.37$ & $95.40\pm 0.46$ & $17.86\pm 0.33$ \\
    \bottomrule
  \end{tabular}%
  }
  \vspace{2pt}

  \parbox{0.98\textwidth}{\footnotesize Single-score estimators use the same map for the retain and forget sides; all other settings equal the full method (same row as in the component table). Mean $\pm$ std over seeds 42--44; accuracies and MIA in \%. Forget/Retain: test accuracy on the deleted / retained data; Test: all test classes. MIA $\rightarrow$ R: closer to Retrain is better. Gap: mean absolute difference to Retrain over Forget, Retain and MIA. Relearn@1: forget accuracy after 1 Adam epoch(s) ($10^{-4}$, BatchNorm statistics frozen) on $\mathcal{D}_f$ plus as many retained training images (lower = less recoverable knowledge). Time: unlearning wall-clock (importance + edit + repair), best of two repeats with the GPU exclusive.}
\end{table*}
\begin{table*}[tp]
  \centering
  \caption{Zone-C orthogonality under the actual update (LwU, CIFAR-10 / CNN, class 0).}
  \label{tab:lwu_orth}
  \resizebox{\textwidth}{!}{%
  \begin{tabular}{lccccccccc}
    \toprule
    \textbf{Setting} & $\max|\cos(g_u^\perp,g_r)|$ & \textbf{Clip scale} & $\frac{\lVert\Delta\theta_C\rVert}{\lVert\theta_C\rVert}$ & $\max|\cos(\Delta\theta^{(s)},g_r)|$ & $\max|\cos(\Delta\theta^{(s)},\bar g_r)|$ & $\Delta\mathcal{L}_r$ & \textbf{Forget} & \textbf{Retain} & \textbf{MIA} \\
    \midrule
    Paper $\Delta_C$ (Fisher-scaled), $\epsilon{=}10^{-8}$ & $1.2e-7$ & $2.5e-6$ & $2.1e-9$ & $3.2e-2$ & $4.6e-2$ & $7.7e-8\pm 5e-7$ & $0.00\pm 0.00$ & $91.69\pm 0.67$ & $52.81\pm 1.39$ \\
    Selected $\Delta_C$, $\epsilon{=}10^{-8}$ & $2.4e-7$ & $6.7e-1$ & $6.8e-3$ & $3.1e-2$ & $3.7e-2$ & $1.4e-1\pm 2e-1$ & $0.00\pm 0.00$ & $91.67\pm 0.64$ & $51.89\pm 2.60$ \\
    Selected $\Delta_C$, $\epsilon{=}0$ & $2.4e-7$ & $6.7e-1$ & $6.8e-3$ & $3.1e-2$ & $3.7e-2$ & $1.4e-1\pm 2e-1$ & $0.00\pm 0.00$ & $91.67\pm 0.64$ & $51.89\pm 2.60$ \\
    No clip, $\epsilon{=}10^{-8}$ & $3.0e-7$ & $1.0e0$ & $6.8e-3$ & $2.7e-7$ & $8.4e-2$ & $1.7e-1\pm 1e-1$ & $0.00\pm 0.00$ & $91.69\pm 0.61$ & $51.73\pm 2.38$ \\
    No clip, $\epsilon{=}0$ & $2.9e-7$ & $1.0e0$ & $6.8e-3$ & $3.1e-7$ & $8.4e-2$ & $1.7e-1\pm 1e-1$ & $0.00\pm 0.00$ & $91.69\pm 0.61$ & $51.73\pm 2.38$ \\
    \bottomrule
  \end{tabular}%
  }
  \vspace{2pt}

  \parbox{0.98\textwidth}{\footnotesize $g_r$: retain gradient used by the step; $\bar g_r$: clean (unaugmented) retain gradient; $\Delta\theta^{(s)}$: step actually applied after clipping. Maxima over 5 steps $\times$ 3 seeds; $\Delta\mathcal{L}_r$: change in clean retain CE over the Zone-C stage (mean $\pm$ std).}
\end{table*}
\begin{table*}[tp]
  \centering
  \caption{Test-time memory ablation (LwU, CIFAR-10 / CNN, class 0).}
  \label{tab:lwu_stream}
  \resizebox{\textwidth}{!}{%
  \begin{tabular}{llcccccc}
    \toprule
    \textbf{Stream ordering} & \textbf{Memory / reset} & \textbf{Forget $\downarrow$} & \textbf{Retain $\uparrow$} & \textbf{Test $\uparrow$} & \textbf{MIA} & \textbf{Flip rate (\%)} & \textbf{$\Delta$Test} \\
    \midrule
    Dataset order & No memory & $0.00\pm 0.00$ & $91.67\pm 0.63$ & $82.50\pm 0.57$ & $53.52\pm 1.20$ & $0.000\pm 0.000$ & -- \\
     & Reset per stream & $0.00\pm 0.00$ & $91.67\pm 0.64$ & $82.51\pm 0.60$ & $51.89\pm 2.60$ & $0.313\pm 0.138$ & $+0.010\pm 0.030$ \\
     & Persistent & $0.00\pm 0.00$ & $91.70\pm 0.68$ & $82.52\pm 0.60$ & $53.55\pm 1.03$ & $0.406\pm 0.050$ & $+0.017\pm 0.032$ \\
    \midrule
    Random$^\dagger$ & No memory & $0.00\pm 0.00$ & $91.67\pm 0.55$ & $82.50\pm 0.49$ & $53.52\pm 1.04$ & $0.000\pm 0.000$ & -- \\
     & Reset per stream & $0.00\pm 0.00$ & $91.68\pm 0.56$ & $82.53\pm 0.50$ & $51.91\pm 2.23$ & $0.292\pm 0.078$ & $+0.026\pm 0.012$ \\
     & Persistent & $0.00\pm 0.00$ & $91.69\pm 0.56$ & $82.53\pm 0.51$ & $53.55\pm 0.89$ & $0.409\pm 0.037$ & $+0.030\pm 0.017$ \\
    \midrule
    Class-sorted & No memory & $0.00\pm 0.00$ & $91.67\pm 0.63$ & $82.50\pm 0.57$ & $53.52\pm 1.20$ & $0.000\pm 0.000$ & -- \\
     & Reset per stream & $0.00\pm 0.00$ & $91.94\pm 1.09$ & $82.75\pm 0.98$ & $51.89\pm 2.60$ & $0.820\pm 0.364$ & $+0.253\pm 0.443$ \\
     & Persistent & $0.00\pm 0.00$ & $91.95\pm 1.08$ & $82.68\pm 0.97$ & $53.35\pm 1.00$ & $0.926\pm 0.206$ & $+0.180\pm 0.429$ \\
    \bottomrule
  \end{tabular}%
  }
  \vspace{2pt}

  \parbox{0.98\textwidth}{\footnotesize All rows of a seed share the same frozen slow weights, so memory on/off is a paired comparison. Mean $\pm$ std over 3 seeds ($^\dagger$3 permutations $\times$ 3 seeds). A stream is one evaluation set read once (batch 64) with predict-before-write; streams are visited in the order forget-test, retain-test, full test, MIA members, MIA non-members, MIA calibration members/non-members. Reset per-stream zeroes $(\mathbf{M},\mathbf{V})$ at the start of every stream; persist zeroes once. Resetting every batch equals no memory.}
\end{table*}
\begin{table*}[tp]
  \centering
  \caption{Validation-selected LwU configuration per setting and seed, and the resulting zone sizes (\% of parameters).}
  \label{tab:lwu_selection}
  \resizebox{\textwidth}{!}{%
  \begin{tabular}{llccccccccc}
    \toprule
    \textbf{Setting} & \textbf{Seed} & $q_r$ & $q_f$ & $\Delta_C$ & $\eta_M$ & $\beta$ & $\rho$ & A\,\% & B\,\% & C\,\% \\
    \midrule
    CIFAR-10 / CNN & 42 & 0.7 & 0.99 & fisher & 0.01 & 1.0 & 0.0 & 0.00 & 0.01 & 29.99 \\
    CIFAR-10 / CNN & 43 & 0.9 & 0.99 & inf & 0.01 & 1.0 & 0.0 & 0.25 & 0.07 & 9.71 \\
    CIFAR-10 / CNN & 44 & 0.7 & 0.95 & inf & 0.05 & 1.0 & 0.8 & 3.23 & 0.16 & 26.66 \\
    CIFAR-10 / ResNet-18 & 42 & 0.9 & 0.99 & fisher & 0.05 & 0.5 & 0.8 & 0.43 & 0.12 & 9.51 \\
    CIFAR-10 / ResNet-18 & 43 & 0.9 & 0.99 & 0.001 & 0.1 & 1.0 & 0.8 & 7.15 & 0.07 & 2.81 \\
    CIFAR-10 / ResNet-18 & 44 & 0.9 & 0.99 & inf & 0.1 & 1.0 & 0.8 & 0.53 & 0.06 & 9.45 \\
    CIFAR-100 / ResNet-18 & 42 & 0.5 & 0.99 & fisher & 0.1 & 1.0 & 0.8 & 15.34 & 0.03 & 34.64 \\
    CIFAR-100 / ResNet-18 & 43 & 0.9 & 0.99 & fisher & 0.01 & 0.25 & 0.0 & 8.72 & 0.15 & 1.21 \\
    CIFAR-100 / ResNet-18 & 44 & 0.7 & 0.99 & 0.001 & 0.01 & 0.25 & 0.0 & 9.75 & 0.09 & 20.21 \\
    CIFAR-10 subclass / CNN & 42 & 0.9 & 0.95 & 0.001 & 0.01 & 0.25 & 0.0 & 0.63 & 1.24 & 8.70 \\
    CIFAR-10 subclass / CNN & 43 & 0.9 & 0.95 & fisher & 0.01 & 0.25 & 0.0 & 8.80 & 1.81 & 0.03 \\
    CIFAR-10 subclass / CNN & 44 & 0.9 & 0.95 & fisher & 0.01 & 1.0 & 0.0 & 8.48 & 1.24 & 0.60 \\
    CIFAR-10 instance / CNN & 42 & 0.9 & 0.95 & 0.001 & 0.1 & 1.0 & 0.8 & 7.44 & 1.60 & 1.57 \\
    CIFAR-10 instance / CNN & 43 & 0.7 & 0.95 & fisher & 0.1 & 1.0 & 0.8 & 14.96 & 0.33 & 14.82 \\
    CIFAR-10 instance / CNN & 44 & 0.7 & 0.9 & fisher & 0.05 & 1.0 & 0.8 & 25.80 & 1.29 & 3.32 \\
    \midrule
    Split CIFAR-10 TIL (joint) & 42 & 0.9 & 0.99 & fisher & 0.05 & 0.5 & 0.8 & 0.00 & 0.00 & 10.00 \\
    \bottomrule
  \end{tabular}%
  }
  \vspace{2pt}

  \parbox{0.98\textwidth}{\footnotesize Selection uses validation data only. $\Delta_C$ relative to $\lVert\theta_C^{(0)}\rVert$ (``fisher'' = paper scaling, inf = no clipping). Memory: $\eta_M$ learning rate, $\beta$ scale, $\rho$ confidence. MU cells fix the Zone-C repair-pass multiplier $\lambda_C{=}1$; the joint protocol additionally searches $\lambda_C\in\{0.1,1\}$ (selected 0.1; $\lambda_C{=}1$ diverged for 23 of 27 grid points) and uses fixed CL memory settings.}
\end{table*}

\twocolumn[{%
\subsection{Per-Example Privacy: U-LiRA}\label{sec:ulira}
\begin{minipage}{\textwidth}
  \centering
  \includegraphics[width=\textwidth,height=0.6\textheight,keepaspectratio]{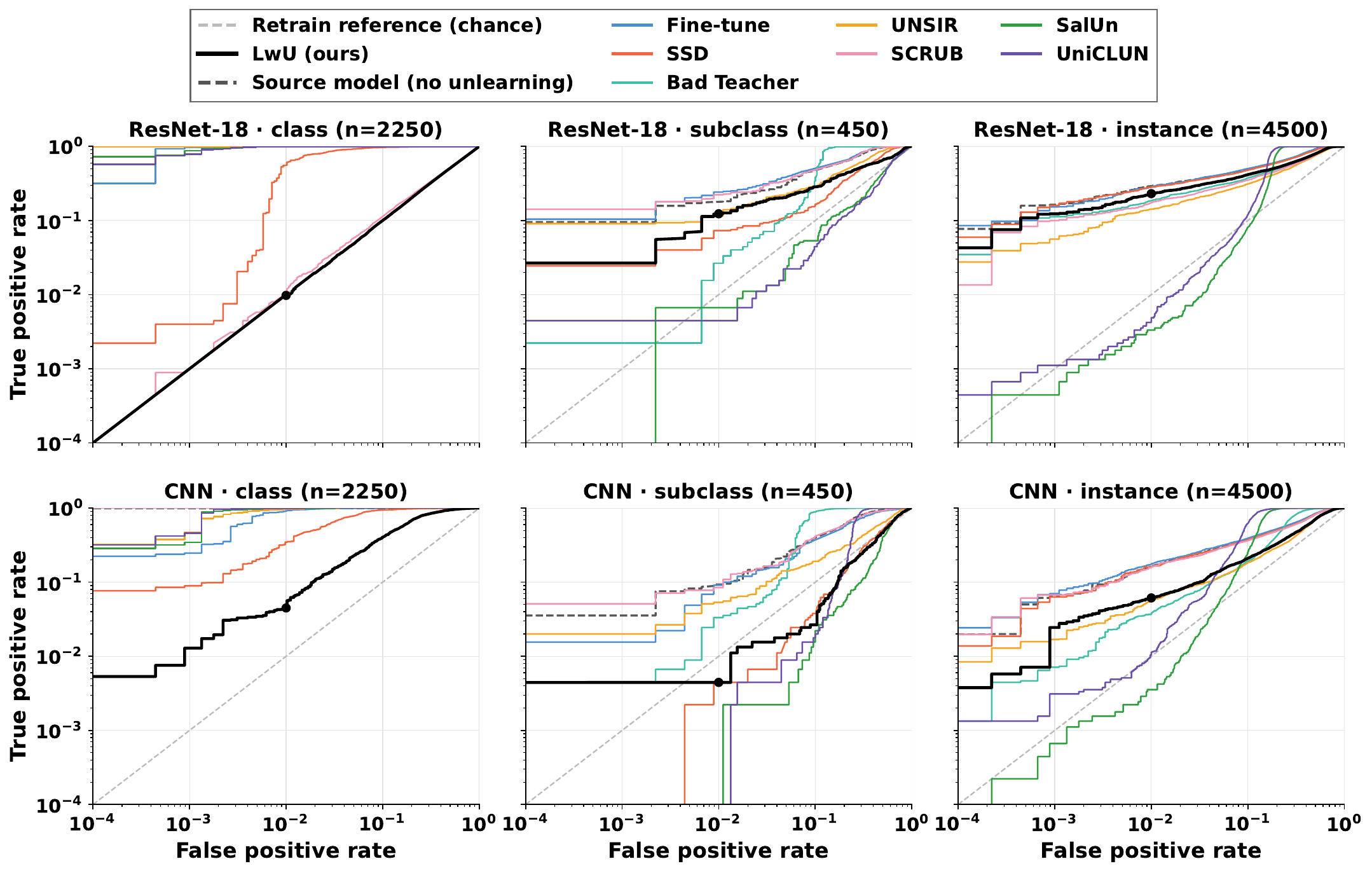}
  \captionof{figure}{\textbf{U-LiRA on RTI: LwU reaches chance on ResNet-18 class and CNN subclass deletion, but still leaks on CNN class and on instance deletion.}}
  \label{fig:ulira-lwu-rti}
  \vspace{2pt}
\endgraf
  \parbox{0.98\textwidth}{Log--log ROC of the per-example membership attack (rows: backbones; columns: deletion granularity; $n$: number of forgotten examples). A method that fully removes the forget set follows the grey diagonal; the dot marks LwU at $1\%$ FPR. On class deletion, the original model is fully exposed (AUC and $\mathrm{TPR}@1\%$ both $100$). LwU lowers AUC to $50.61$ on ResNet-18 but only to $82.48$ on the CNN. On subclass deletion, AUC falls to $49.26$ on the CNN (original: $81.57$) and $64.12$ on ResNet-18 (original: $83.18$). On instance deletion, $\mathrm{TPR}@1\%$ drops from $29.22$ to $23.07$ (ResNet-18) and from $16.51$ to $6.13$ (CNN). Some baselines leak less at low FPR in individual cells.}
\end{minipage}\vspace{1.5em}}]
\begin{figure*}[tp]
  \centering
  \includegraphics[width=\textwidth]{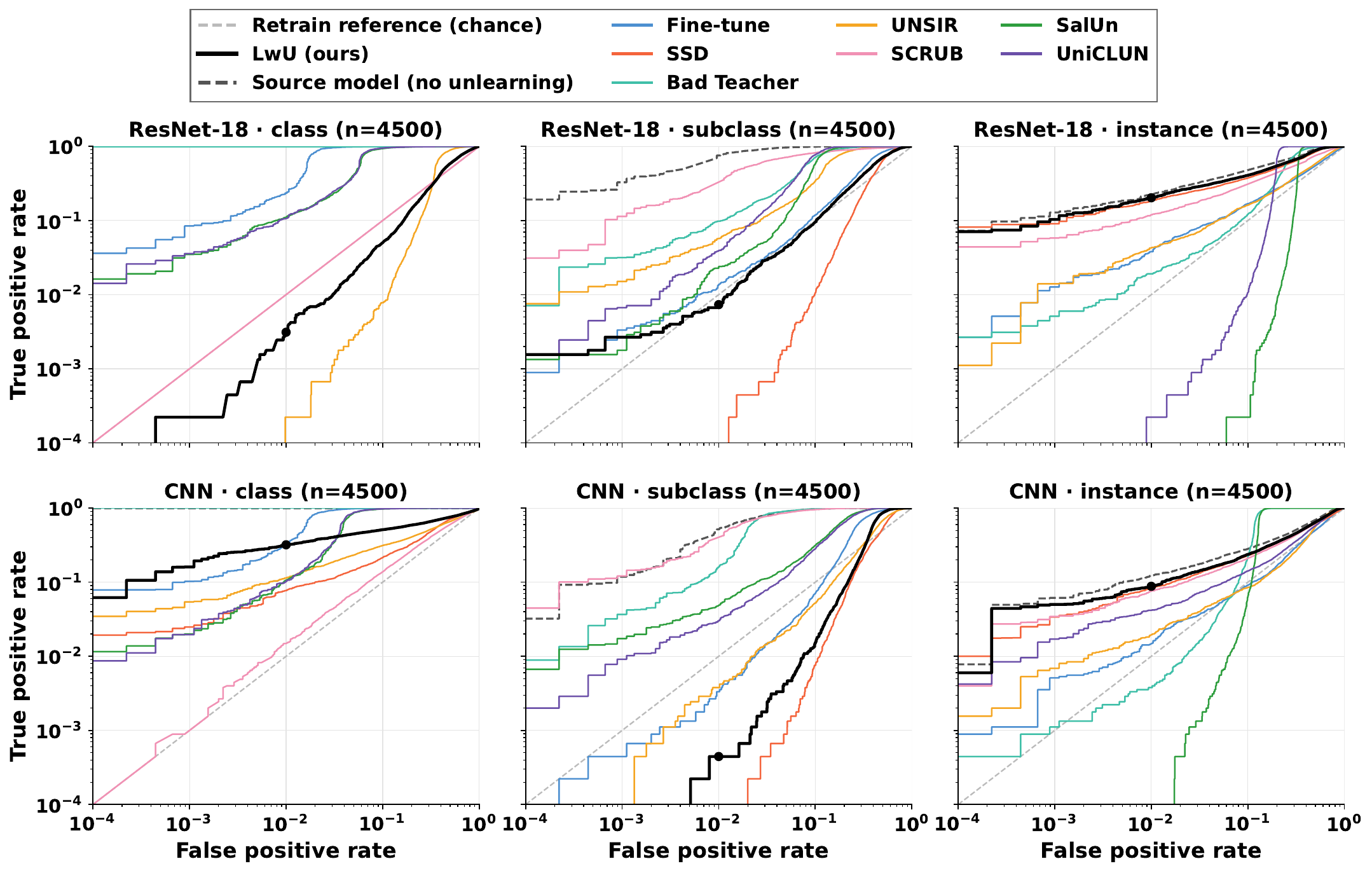}
  \caption{\textbf{U-LiRA on CIFAR-10: LwU suppresses low-FPR leakage on subclass and ResNet-18 class deletion, but not on CNN class deletion; on instance deletion, no method separates from the original model.}}
  \label{fig:ulira-lwu-cifar10}
  \vspace{2pt}

  \parbox{.98\textwidth}{Layout as in Fig.~\ref{fig:ulira-lwu-rti}. On class deletion, LwU lowers AUC from $100$ to $53.38$ on ResNet-18 ($\mathrm{TPR}@1\%$: $0.31$) but only to $73.28$ on the CNN ($31.96$). On subclass deletion, $\mathrm{TPR}@1\%$ falls to $0.73$ and $0.04$ (original: $75.62$ and $52.00$; ResNet-18 and CNN). On instance deletion, LwU stays close to the original model (AUC $77.72$ vs.\ $80.08$ and $62.52$ vs.\ $66.78$). Some baselines leak less at low FPR in individual cells.}
\end{figure*}


\twocolumn[{%
\subsection{Unlearning Cost}\label{sec:mu_cost}
\begin{minipage}{\textwidth}
  \centering
  \includegraphics[width=\textwidth,height=0.6\textheight,keepaspectratio]{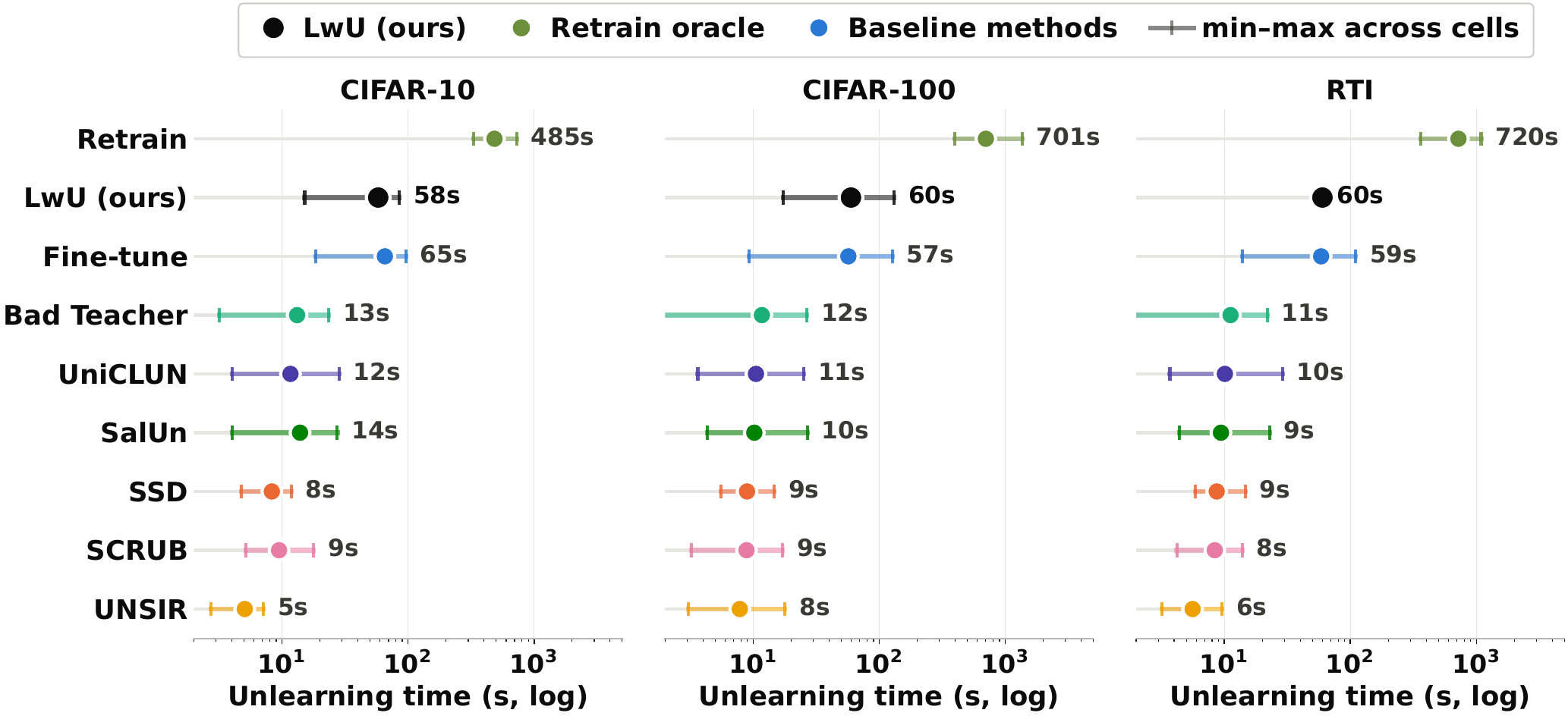}
  \captionof{figure}{\textbf{Every approximate method is one to two orders of magnitude cheaper than retraining; LwU buys its privacy advantage at roughly the cost of fine-tuning.}}
  \label{fig:cost}
  \vspace{2pt}
\endgraf
  \parbox{0.98\textwidth}{Unlearning wall-clock time per method, pooled over all cells and seeds ($2$ backbones $\times$ $3$ forget modes $\times$ $3$ seeds). Dots are means and the coloured span is the min--max range; the axis is logarithmic, so position rather than stem length encodes the value. Retraining from scratch costs $701$\,s on CIFAR-100 and $485$\,s on CIFAR-10, against $60$\,s and $58$\,s for LwU ($11.7\times$ and $8.4\times$ cheaper) and $7.8$\,s and $5.1$\,s for the cheapest baseline, UNSIR ($90\times$ and $96\times$). \emph{These timings are indicative only:} the campaign ran three cells concurrently on one GPU, so every measurement includes contention from unrelated jobs, giving relative standard deviations of $21$--$72\%$ and the wide ranges shown. They establish the order of magnitude, not a precise ranking among the sub-$15$\,s methods.}
\end{minipage}\vspace{1.5em}}]

\twocolumn[{\subsection{Detailed Machine-Unlearning Results}\label{sec:mu_detailed}\begin{minipage}{\textwidth}\centering\setlength{\tabcolsep}{3.0pt}\captionof{table}{CNN with class deletion on CIFAR-10, CIFAR-100, and RTI.}
\label{tab:cnn-class-detailed}
\begin{adjustbox}{max width=\textwidth,max totalheight=0.92\textheight}
\begin{tabular}{lcccccccc}
\toprule
Method & Gap$\downarrow$ & $D_f\rightarrow D_f^{\mathrm{Retrain}}$
& $D_r\uparrow$ & Test$\uparrow$ & MIA$\rightarrow50$
& RL-1$\rightarrow$Retrain & RL-5$\rightarrow$Retrain & Time (s)$\downarrow$ \\
\midrule
\multicolumn{9}{>{\columncolor{dsgrey}}c}{\textbf{CIFAR-10}} \\
\midrule
Retrain (oracle)
& $0.00$
& $0.00_{\pm 0.00}$
& $92.76_{\pm 0.23}$
& $83.49_{\pm 0.21}$
& $50.29_{\pm 1.29}$
& $0.00$
& $57.70$
& $494.90$ \\
Original model
& $27.47$
& $93.47_{\pm 1.33}$
& $92.24_{\pm 0.24}$
& $92.37_{\pm 0.34}$
& $57.29_{\pm 0.44}$
& $50.90$
& $95.10$
& $0.00$ \\
Fine-tune on $D_r$
& $15.34$
& $50.17_{\pm 43.28}$
& $92.20_{\pm 0.89}$
& $87.99_{\pm 5.13}$
& $55.62_{\pm 2.39}$
& $20.70$
& $94.60$
& $57.10$ \\
\midrule
\multicolumn{9}{l}{\textit{Unlearning methods}} \\
\midrule
Amnesiac
& $39.15$
& \cellcolor{gold}$0.00_{\pm 0.00}$
& $11.11_{\pm 0.00}$
& $10.00_{\pm 0.00}$
& \cellcolor{gold}$49.60_{\pm 0.93}$
& $100.00$
& $100.00$
& \cellcolor{gold}$0.00$ \\
Bad Teacher
& $1.55$
& \cellcolor{silver}$0.27_{\pm 0.15}$
& \cellcolor{bronze}$92.54_{\pm 0.07}$
& $83.32_{\pm 0.05}$
& $55.82_{\pm 0.90}$
& $20.40$
& $98.50$
& $8.00$ \\
SCRUB
& \cellcolor{bronze}$1.12$
& \cellcolor{gold}$0.00_{\pm 0.00}$
& \cellcolor{silver}$92.73_{\pm 0.49}$
& $83.46_{\pm 0.44}$
& $54.43_{\pm 2.67}$
& \cellcolor{gold}$0.00$
& \cellcolor{gold}$30.30$
& $6.10$ \\
SSD
& $22.39$
& $74.60_{\pm 22.39}$
& $88.36_{\pm 6.07}$
& \cellcolor{gold}$86.98_{\pm 5.58}$
& $55.76_{\pm 1.31}$
& $45.70$
& \cellcolor{bronze}$96.40$
& $6.40$ \\
SalUn
& $12.85$
& $41.50_{\pm 35.98}$
& $91.66_{\pm 0.43}$
& \cellcolor{silver}$86.64_{\pm 3.90}$
& $54.41_{\pm 4.36}$
& $60.10$
& $97.60$
& $13.20$ \\
UNSIR
& \cellcolor{silver}$1.05$
& \cellcolor{gold}$0.00_{\pm 0.00}$
& $91.86_{\pm 0.43}$
& $82.67_{\pm 0.39}$
& \cellcolor{silver}$52.57_{\pm 2.14}$
& \cellcolor{silver}$0.50$
& $98.10$
& \cellcolor{silver}$4.90$ \\
\midrule
\multicolumn{9}{l}{\textit{Joint continual-learning and machine-unlearning methods}} \\
\midrule
UniCLUN
& $12.64$
& \cellcolor{bronze}$41.33_{\pm 35.77}$
& $91.54_{\pm 0.63}$
& \cellcolor{bronze}$86.52_{\pm 4.14}$
& $54.25_{\pm 3.23}$
& $58.00$
& $96.90$
& \cellcolor{bronze}$6.00$ \\
\textbf{LwU} (ours)
& \cellcolor{gold}$0.92$
& \cellcolor{gold}$0.00_{\pm 0.00}$
& \cellcolor{gold}$92.77_{\pm 0.25}$
& $83.50_{\pm 0.23}$
& \cellcolor{bronze}$53.64_{\pm 2.23}$
& \cellcolor{bronze}$0.80$
& \cellcolor{silver}$94.30$
& $50.30$ \\
\midrule
\multicolumn{9}{c}{\textbf{CIFAR-10 podium:}
\quad \textcolor{gold}{\faMedal}\,\textbf{LwU (ours)}\;(3G, 1S, 2B;\ \textbf{13 pts})
\quad \textcolor{silver}{\faMedal}\,\textbf{SCRUB}\;(3G, 1S, 1B;\ \textbf{12 pts})
\quad \textcolor{bronze}{\faMedal}\,\textbf{UNSIR}\;(1G, 4S, 0B;\ \textbf{11 pts})
} \\
\midrule
\multicolumn{9}{>{\columncolor{dsgrey}}c}{\textbf{CIFAR-100}} \\
\midrule
Retrain (oracle)
& $0.00$
& $0.00_{\pm 0.00}$
& $68.51_{\pm 0.39}$
& $67.83_{\pm 0.38}$
& $51.12_{\pm 4.45}$
& $0.00$
& $0.00$
& $477.70$ \\
Original model
& $21.66$
& $70.00_{\pm 3.61}$
& $68.58_{\pm 0.52}$
& $68.60_{\pm 0.48}$
& $66.68_{\pm 1.36}$
& $36.00$
& $43.30$
& $0.00$ \\
Fine-tune on $D_r$
& $12.32$
& $35.00_{\pm 3.61}$
& $68.32_{\pm 0.46}$
& $67.99_{\pm 0.45}$
& $64.69_{\pm 3.46}$
& $11.00$
& $16.30$
& $66.80$ \\
\midrule
\multicolumn{9}{l}{\textit{Unlearning methods}} \\
\midrule
Amnesiac
& $34.73$
& \cellcolor{gold}$0.00_{\pm 0.00}$
& $1.38_{\pm 0.33}$
& $1.37_{\pm 0.33}$
& \cellcolor{silver}$48.41_{\pm 2.75}$
& \cellcolor{gold}$0.00$
& \cellcolor{gold}$0.00$
& \cellcolor{gold}$0.00$ \\
Bad Teacher
& \cellcolor{bronze}$5.56$
& \cellcolor{silver}$0.33_{\pm 0.58}$
& $67.24_{\pm 0.31}$
& $66.57_{\pm 0.31}$
& $70.51_{\pm 1.66}$
& \cellcolor{silver}$0.30$
& \cellcolor{bronze}$10.70$
& $10.20$ \\
SCRUB
& $20.07$
& $64.33_{\pm 9.02}$
& \cellcolor{silver}$68.51_{\pm 0.28}$
& \cellcolor{gold}$68.47_{\pm 0.23}$
& $66.06_{\pm 1.28}$
& $27.30$
& $32.70$
& $4.90$ \\
SSD
& \cellcolor{silver}$4.35$
& \cellcolor{bronze}$8.00_{\pm 2.65}$
& \cellcolor{bronze}$67.94_{\pm 0.85}$
& \cellcolor{bronze}$67.34_{\pm 0.85}$
& \cellcolor{bronze}$58.56_{\pm 0.37}$
& \cellcolor{bronze}$6.00$
& $28.00$
& $5.10$ \\
SalUn
& $11.79$
& $23.67_{\pm 1.53}$
& $63.30_{\pm 1.03}$
& $62.91_{\pm 1.03}$
& $64.50_{\pm 0.31}$
& $18.70$
& $39.30$
& $6.70$ \\
UNSIR
& $13.25$
& $37.33_{\pm 17.67}$
& $67.60_{\pm 0.57}$
& $67.29_{\pm 0.71}$
& $65.32_{\pm 4.62}$
& $18.70$
& $35.00$
& \cellcolor{silver}$4.00$ \\
\midrule
\multicolumn{9}{l}{\textit{Joint continual-learning and machine-unlearning methods}} \\
\midrule
UniCLUN
& $12.80$
& $26.67_{\pm 3.79}$
& $63.07_{\pm 0.90}$
& $62.71_{\pm 0.90}$
& $65.11_{\pm 2.31}$
& $17.70$
& $41.70$
& \cellcolor{bronze}$4.70$ \\
\textbf{LwU} (ours)
& \cellcolor{gold}$0.97$
& \cellcolor{gold}$0.00_{\pm 0.00}$
& \cellcolor{gold}$68.61_{\pm 0.35}$
& \cellcolor{silver}$67.92_{\pm 0.34}$
& \cellcolor{gold}$49.35_{\pm 3.23}$
& \cellcolor{gold}$0.00$
& \cellcolor{silver}$0.30$
& $43.60$ \\
\midrule
\multicolumn{9}{c}{\textbf{CIFAR-100 podium:}
\quad \textcolor{gold}{\faMedal}\,\textbf{LwU (ours)}\;(5G, 2S, 0B;\ \textbf{19 pts})
\quad \textcolor{silver}{\faMedal}\,\textbf{Amnesiac}\;(4G, 1S, 0B;\ \textbf{14 pts})
\quad \textcolor{bronze}{\faMedal}\,\textbf{SSD}\;(0G, 1S, 5B;\ \textbf{7 pts})
} \\
\midrule
\multicolumn{9}{>{\columncolor{dsgrey}}c}{\textbf{RTI}} \\
\midrule
Retrain (oracle)
& $0.00$
& $0.00_{\pm 0.00}$
& $80.77_{\pm 0.13}$
& $76.73_{\pm 0.12}$
& $49.66_{\pm 0.43}$
& $0.00$
& $0.00$
& $459.60$ \\
Original model
& $23.75$
& $72.87_{\pm 1.68}$
& $79.73_{\pm 0.48}$
& $79.39_{\pm 0.38}$
& $68.10_{\pm 0.65}$
& $35.20$
& $80.50$
& $0.00$ \\
Fine-tune on $D_r$
& $11.66$
& $28.40_{\pm 3.12}$
& $80.55_{\pm 0.37}$
& $77.94_{\pm 0.23}$
& $66.41_{\pm 2.06}$
& $10.50$
& $79.40$
& $57.80$ \\
\midrule
\multicolumn{9}{l}{\textit{Unlearning methods}} \\
\midrule
Amnesiac
& $37.21$
& \cellcolor{gold}$0.00_{\pm 0.00}$
& $5.26_{\pm 0.00}$
& $5.00_{\pm 0.00}$
& \cellcolor{gold}$51.25_{\pm 1.14}$
& \cellcolor{gold}$0.00$
& $99.60$
& \cellcolor{gold}$0.00$ \\
Bad Teacher
& \cellcolor{silver}$4.36$
& \cellcolor{bronze}$0.13_{\pm 0.12}$
& \cellcolor{bronze}$79.56_{\pm 0.51}$
& $75.59_{\pm 0.48}$
& $64.62_{\pm 1.41}$
& \cellcolor{bronze}$6.30$
& $90.50$
& $9.60$ \\
SCRUB
& $18.30$
& $52.53_{\pm 9.49}$
& \cellcolor{silver}$80.38_{\pm 0.54}$
& \cellcolor{gold}$78.99_{\pm 0.50}$
& $67.66_{\pm 0.68}$
& $16.50$
& \cellcolor{silver}$78.50$
& \cellcolor{bronze}$4.90$ \\
SSD
& $14.21$
& $37.47_{\pm 30.91}$
& \cellcolor{silver}$80.38_{\pm 0.67}$
& \cellcolor{silver}$78.23_{\pm 1.31}$
& $66.95_{\pm 1.84}$
& $17.80$
& \cellcolor{bronze}$79.20$
& $6.40$ \\
SalUn
& $13.58$
& $33.60_{\pm 2.23}$
& $77.99_{\pm 0.64}$
& $75.77_{\pm 0.73}$
& $66.65_{\pm 0.88}$
& $32.10$
& $81.90$
& $6.20$ \\
UNSIR
& \cellcolor{bronze}$6.58$
& $11.80_{\pm 17.25}$
& $79.34_{\pm 1.14}$
& $75.97_{\pm 1.93}$
& \cellcolor{silver}$61.07_{\pm 1.92}$
& $7.90$
& $81.70$
& \cellcolor{silver}$4.20$ \\
\midrule
\multicolumn{9}{l}{\textit{Joint continual-learning and machine-unlearning methods}} \\
\midrule
UniCLUN
& $14.13$
& $35.13_{\pm 1.27}$
& $78.02_{\pm 0.28}$
& $75.88_{\pm 0.30}$
& $67.46_{\pm 1.10}$
& $33.50$
& $82.00$
& $5.40$ \\
\textbf{LwU} (ours)
& \cellcolor{gold}$3.48$
& \cellcolor{silver}$0.07_{\pm 0.12}$
& \cellcolor{gold}$80.60_{\pm 0.42}$
& \cellcolor{bronze}$76.58_{\pm 0.40}$
& \cellcolor{bronze}$62.90_{\pm 3.24}$
& \cellcolor{silver}$0.70$
& \cellcolor{gold}$78.20$
& $41.20$ \\
\midrule
\multicolumn{9}{c}{\textbf{RTI podium:}
\quad \textcolor{gold}{\faMedal}\,\textbf{LwU (ours)}\;(3G, 2S, 2B;\ \textbf{15 pts})
\quad \textcolor{silver}{\faMedal}\,\textbf{Amnesiac}\;(4G, 0S, 0B;\ \textbf{12 pts})
\quad \textcolor{bronze}{\faMedal}\,\textbf{SCRUB}\;(1G, 2S, 1B;\ \textbf{8 pts})
} \\
\multicolumn{9}{c}{\textbf{Overall podium:}
\quad \textcolor{gold}{\faMedal}\,\textbf{LwU (ours)}\;(11G, 5S, 4B;\ \textbf{47 pts})
\quad \textcolor{silver}{\faMedal}\,\textbf{Amnesiac}\;(11G, 1S, 0B;\ \textbf{35 pts})
\quad \textcolor{bronze}{\faMedal}\,\textbf{SCRUB}\;(5G, 4S, 2B;\ \textbf{25 pts})
} \\
\bottomrule
\end{tabular}
\end{adjustbox}
  \vspace{2pt}
\endgraf
  \parbox{0.98\textwidth}{\footnotesize $D_f$, $D_r$, Test, and MIA are mean $\pm$ standard deviation over 3 seeds. Gap is the reported mean absolute deviation from the Retrain targets. Relearning metrics are ranked by distance to the corresponding Retrain trajectory. Podiums use the weighted medal score 3G+2S+1B.}
\end{minipage}\vspace{1.5em}}]

\begin{table*}[p]
\centering
\setlength{\tabcolsep}{3.0pt}
\caption{CNN with subclass deletion on CIFAR-10, CIFAR-100, and RTI.}
\label{tab:cnn-subclass-detailed}
\begin{adjustbox}{max width=\textwidth,max totalheight=0.92\textheight}
\begin{tabular}{lcccccccc}
\toprule
Method & Gap$\downarrow$ & $D_f\rightarrow D_f^{\mathrm{Retrain}}$
& $D_r\uparrow$ & Test$\uparrow$ & MIA$\rightarrow50$
& RL-1$\rightarrow$Retrain & RL-5$\rightarrow$Retrain & Time (s)$\downarrow$ \\
\midrule
\multicolumn{9}{>{\columncolor{dsgrey}}c}{\textbf{CIFAR-10}} \\
\midrule
Retrain (oracle)
& $0.00$
& $25.40_{\pm 4.50}$
& $94.89_{\pm 0.11}$
& $87.94_{\pm 0.53}$
& $49.42_{\pm 0.78}$
& $36.30$
& $97.90$
& $378.60$ \\
Original model
& $20.60$
& $94.57_{\pm 1.17}$
& $94.51_{\pm 0.14}$
& $94.52_{\pm 0.23}$
& $55.68_{\pm 1.57}$
& $53.20$
& $97.10$
& $0.00$ \\
Fine-tune on $D_r$
& $12.80$
& $68.90_{\pm 2.80}$
& $93.51_{\pm 0.94}$
& $91.05_{\pm 1.10}$
& $52.15_{\pm 2.37}$
& $37.60$
& $95.80$
& $44.60$ \\
\midrule
\multicolumn{9}{l}{\textit{Unlearning methods}} \\
\midrule
Amnesiac
& --
& --
& --
& --
& --
& --
& --
& -- \\
Bad Teacher
& \cellcolor{gold}$2.31$
& \cellcolor{gold}$30.00_{\pm 4.71}$
& \cellcolor{silver}$94.27_{\pm 0.22}$
& $87.84_{\pm 0.67}$
& \cellcolor{gold}$53.30_{\pm 0.80}$
& \cellcolor{silver}$40.50$
& $97.10$
& $10.80$ \\
SCRUB
& $19.45$
& $90.73_{\pm 1.63}$
& \cellcolor{gold}$94.63_{\pm 0.09}$
& \cellcolor{gold}$94.24_{\pm 0.24}$
& $55.33_{\pm 1.43}$
& \cellcolor{gold}$38.20$
& $89.60$
& \cellcolor{bronze}$6.60$ \\
SSD
& $19.29$
& \cellcolor{silver}$65.57_{\pm 28.00}$
& $78.58_{\pm 25.89}$
& $77.28_{\pm 25.73}$
& \cellcolor{silver}$53.56_{\pm 2.15}$
& \cellcolor{bronze}$48.60$
& $97.40$
& $7.80$ \\
SalUn
& \cellcolor{silver}$14.79$
& \cellcolor{bronze}$73.40_{\pm 3.05}$
& $93.70_{\pm 0.48}$
& $91.67_{\pm 0.70}$
& $55.66_{\pm 1.17}$
& $51.00$
& \cellcolor{silver}$97.70$
& \cellcolor{silver}$5.90$ \\
UNSIR
& $15.86$
& $79.13_{\pm 10.71}$
& $93.37_{\pm 0.29}$
& \cellcolor{bronze}$91.95_{\pm 1.33}$
& \cellcolor{bronze}$53.59_{\pm 1.42}$
& $58.50$
& \cellcolor{gold}$98.00$
& \cellcolor{gold}$4.00$ \\
\midrule
\multicolumn{9}{l}{\textit{Joint continual-learning and machine-unlearning methods}} \\
\midrule
UniCLUN
& \cellcolor{bronze}$15.07$
& $74.40_{\pm 4.29}$
& $93.65_{\pm 0.59}$
& $91.73_{\pm 0.92}$
& $55.70_{\pm 0.95}$
& $50.60$
& \cellcolor{bronze}$98.30$
& $6.80$ \\
\textbf{LwU} (ours)
& $17.83$
& $85.63_{\pm 15.15}$
& \cellcolor{bronze}$94.15_{\pm 0.68}$
& \cellcolor{silver}$93.29_{\pm 2.11}$
& $54.41_{\pm 3.18}$
& $53.00$
& $97.30$
& $40.60$ \\
\midrule
\multicolumn{9}{c}{\textbf{CIFAR-10 podium:}
\quad \textcolor{gold}{\faMedal}\,\textbf{Bad Teacher}\;(3G, 2S, 0B;\ \textbf{13 pts})
\quad \textcolor{silver}{\faMedal}\,\textbf{SCRUB}\;(3G, 0S, 1B;\ \textbf{10 pts})
\quad \textcolor{bronze}{\faMedal}\,\textbf{UNSIR}\;(2G, 0S, 2B;\ \textbf{8 pts})
} \\

\midrule
\multicolumn{9}{>{\columncolor{dsgrey}}c}{\textbf{CIFAR-100}} \\
\midrule
Retrain (oracle)
& $0.00$
& $23.33_{\pm 4.16}$
& $79.47_{\pm 0.15}$
& $78.91_{\pm 0.17}$
& $49.96_{\pm 1.85}$
& $21.70$
& $41.30$
& $664.10$ \\
Original model
& $18.19$
& $77.00_{\pm 3.46}$
& $79.37_{\pm 0.84}$
& $79.35_{\pm 0.86}$
& $67.54_{\pm 0.62}$
& $43.30$
& $32.70$
& $0.00$ \\
Fine-tune on $D_r$
& $16.62$
& $72.00_{\pm 7.00}$
& $79.08_{\pm 0.51}$
& $79.01_{\pm 0.57}$
& $66.74_{\pm 0.50}$
& $39.30$
& $31.70$
& $36.00$ \\
\midrule
\multicolumn{9}{l}{\textit{Unlearning methods}} \\
\midrule
Amnesiac
& --
& --
& --
& --
& --
& --
& --
& -- \\
Bad Teacher
& \cellcolor{gold}$6.84$
& \cellcolor{gold}$17.00_{\pm 4.58}$
& $77.89_{\pm 0.51}$
& $77.28_{\pm 0.49}$
& $67.75_{\pm 1.44}$
& \cellcolor{bronze}$15.00$
& $28.70$
& $13.40$ \\
SCRUB
& $17.21$
& $73.33_{\pm 6.11}$
& \cellcolor{gold}$79.16_{\pm 0.89}$
& \cellcolor{gold}$79.10_{\pm 0.95}$
& $67.21_{\pm 2.14}$
& $39.30$
& $30.30$
& $9.80$ \\
SSD
& $12.20$
& $51.00_{\pm 30.81}$
& \cellcolor{bronze}$78.33_{\pm 0.64}$
& $78.05_{\pm 0.86}$
& \cellcolor{bronze}$66.42_{\pm 1.93}$
& $31.30$
& $31.30$
& $6.80$ \\
SalUn
& \cellcolor{silver}$9.59$
& \cellcolor{bronze}$37.67_{\pm 5.69}$
& $75.57_{\pm 0.84}$
& $75.19_{\pm 0.80}$
& \cellcolor{silver}$66.36_{\pm 1.32}$
& \cellcolor{silver}$28.00$
& \cellcolor{bronze}$37.00$
& \cellcolor{silver}$5.10$ \\
UNSIR
& $14.89$
& $64.33_{\pm 0.58}$
& $78.28_{\pm 0.42}$
& \cellcolor{bronze}$78.14_{\pm 0.42}$
& $66.58_{\pm 2.66}$
& $40.70$
& \cellcolor{silver}$38.30$
& \cellcolor{gold}$3.60$ \\
\midrule
\multicolumn{9}{l}{\textit{Joint continual-learning and machine-unlearning methods}} \\
\midrule
UniCLUN
& \cellcolor{bronze}$9.71$
& \cellcolor{silver}$35.67_{\pm 3.51}$
& $75.13_{\pm 0.21}$
& $74.74_{\pm 0.17}$
& $67.96_{\pm 1.68}$
& \cellcolor{gold}$25.00$
& $31.70$
& \cellcolor{bronze}$6.20$ \\
\textbf{LwU} (ours)
& $10.25$
& $50.00_{\pm 13.08}$
& \cellcolor{silver}$78.47_{\pm 0.43}$
& \cellcolor{silver}$78.19_{\pm 0.43}$
& \cellcolor{gold}$62.56_{\pm 0.85}$
& $32.70$
& \cellcolor{gold}$42.00$
& $28.30$ \\
\midrule
\multicolumn{9}{c}{\textbf{CIFAR-100 podium:}
\quad \textcolor{gold}{\faMedal}\,\textbf{LwU (ours)}\;(2G, 2S, 0B;\ \textbf{10 pts})
\quad \textcolor{silver}{\faMedal}\,\textbf{SalUn}\;(0G, 4S, 2B;\ \textbf{10 pts})
\quad \textcolor{bronze}{\faMedal}\,\textbf{Bad Teacher}\;(2G, 0S, 1B;\ \textbf{7 pts})
} \\
\midrule
\multicolumn{9}{>{\columncolor{dsgrey}}c}{\textbf{RTI}} \\
\midrule
Retrain (oracle)
& $0.00$
& $86.33_{\pm 2.31}$
& $78.94_{\pm 0.18}$
& $79.01_{\pm 0.16}$
& $48.26_{\pm 0.93}$
& $55.30$
& $47.70$
& $520.90$ \\
Original model
& $3.33$
& $96.33_{\pm 2.08}$
& $79.25_{\pm 0.07}$
& $79.42_{\pm 0.08}$
& $50.78_{\pm 3.28}$
& $69.00$
& $48.00$
& $0.00$ \\
Fine-tune on $D_r$
& $2.62$
& $95.33_{\pm 1.53}$
& $79.12_{\pm 0.15}$
& $79.28_{\pm 0.15}$
& $48.63_{\pm 1.77}$
& $65.70$
& $47.70$
& $39.60$ \\
\midrule
\multicolumn{9}{l}{\textit{Unlearning methods}} \\
\midrule
Amnesiac
& --
& --
& --
& --
& --
& --
& --
& -- \\
Bad Teacher
& $12.53$
& $45.00_{\pm 16.52}$
& $77.59_{\pm 0.47}$
& $77.26_{\pm 0.30}$
& $53.97_{\pm 2.24}$
& $34.30$
& $51.70$
& \cellcolor{silver}$6.50$ \\
SCRUB
& \cellcolor{bronze}$3.19$
& \cellcolor{bronze}$95.67_{\pm 1.53}$
& \cellcolor{gold}$79.28_{\pm 0.09}$
& \cellcolor{gold}$79.44_{\pm 0.11}$
& \cellcolor{silver}$50.41_{\pm 3.76}$
& \cellcolor{bronze}$67.00$
& \cellcolor{silver}$45.00$
& \cellcolor{bronze}$6.80$ \\
SSD
& \cellcolor{silver}$2.37$
& \cellcolor{silver}$90.33_{\pm 4.04}$
& \cellcolor{silver}$79.14_{\pm 0.10}$
& \cellcolor{silver}$79.25_{\pm 0.14}$
& $51.29_{\pm 0.69}$
& \cellcolor{silver}$61.70$
& $42.00$
& $7.10$ \\
SalUn
& $9.37$
& $64.67_{\pm 3.06}$
& $73.33_{\pm 0.28}$
& $73.24_{\pm 0.30}$
& $52.67_{\pm 1.49}$
& $42.00$
& $42.00$
& $8.20$ \\
UNSIR
& $3.52$
& \cellcolor{bronze}$95.67_{\pm 2.31}$
& $77.96_{\pm 0.89}$
& $78.14_{\pm 0.88}$
& \cellcolor{bronze}$51.03_{\pm 1.84}$
& $69.30$
& $54.30$
& \cellcolor{gold}$3.90$ \\
\midrule
\multicolumn{9}{l}{\textit{Joint continual-learning and machine-unlearning methods}} \\
\midrule
UniCLUN
& $6.40$
& $74.33_{\pm 2.52}$
& $72.85_{\pm 0.44}$
& $72.86_{\pm 0.42}$
& \cellcolor{gold}$49.62_{\pm 2.01}$
& $43.00$
& \cellcolor{gold}$48.30$
& $13.60$ \\
\textbf{LwU} (ours)
& \cellcolor{gold}$2.17$
& \cellcolor{gold}$88.00_{\pm 5.20}$
& \cellcolor{bronze}$78.52_{\pm 0.34}$
& \cellcolor{bronze}$78.61_{\pm 0.35}$
& $51.77_{\pm 2.23}$
& \cellcolor{gold}$60.00$
& \cellcolor{bronze}$51.00$
& $33.40$ \\
\midrule
\multicolumn{9}{c}{\textbf{RTI podium:}
\quad \textcolor{gold}{\faMedal}\,\textbf{SCRUB}\;(2G, 2S, 4B;\ \textbf{14 pts})
\quad \textcolor{silver}{\faMedal}\,\textbf{LwU (ours)}\;(3G, 0S, 3B;\ \textbf{12 pts})
\quad \textcolor{bronze}{\faMedal}\,\textbf{SSD}\;(0G, 5S, 0B;\ \textbf{10 pts})
} \\

\multicolumn{9}{c}{\textbf{Overall podium:}
\quad \textcolor{gold}{\faMedal}\,\textbf{SCRUB}\;(7G, 2S, 5B;\ \textbf{30 pts})
\quad \textcolor{silver}{\faMedal}\,\textbf{LwU (ours)}\;(5G, 3S, 4B;\ \textbf{25 pts})
\quad \textcolor{bronze}{\faMedal}\,\textbf{Bad Teacher}\;(5G, 3S, 1B;\ \textbf{22 pts})
} \\
\bottomrule
\end{tabular}
\end{adjustbox}
\end{table*}

\begin{table*}[p]
\centering
\setlength{\tabcolsep}{3.0pt}
\caption{CNN with instance deletion on CIFAR-10, CIFAR-100, and RTI.}
\label{tab:cnn-instance-detailed}
\begin{adjustbox}{max width=\textwidth,max totalheight=0.92\textheight}
\begin{tabular}{lcccccccc}
\toprule
Method & Gap$\downarrow$ & $D_f\rightarrow D_f^{\mathrm{Retrain}}$
& $D_r\uparrow$ & Test$\uparrow$ & MIA$\rightarrow50$
& RL-1$\rightarrow$Retrain & RL-5$\rightarrow$Retrain & Time (s)$\downarrow$ \\
\midrule
\multicolumn{9}{>{\columncolor{dsgrey}}c}{\textbf{CIFAR-10}} \\
\midrule
Retrain (oracle)
& $0.00$
& $91.86_{\pm 1.34}$
& $92.13_{\pm 0.28}$
& $92.13_{\pm 0.28}$
& $49.79_{\pm 0.66}$
& $92.70$
& $94.60$
& $423.80$ \\
Original model
& $4.06$
& $99.94_{\pm 0.10}$
& $92.37_{\pm 0.34}$
& $92.37_{\pm 0.34}$
& $56.93_{\pm 0.20}$
& $99.90$
& $100.00$
& $0.00$ \\
Fine-tune on $D_r$
& $2.85$
& $96.19_{\pm 0.44}$
& $90.12_{\pm 0.33}$
& $90.12_{\pm 0.33}$
& $52.83_{\pm 0.20}$
& $97.60$
& $98.60$
& $48.00$ \\
\midrule
\multicolumn{9}{l}{\textit{Unlearning methods}} \\
\midrule
Amnesiac
& --
& --
& --
& --
& --
& --
& --
& -- \\
Bad Teacher
& $3.86$
& $98.29_{\pm 0.84}$
& $89.94_{\pm 0.50}$
& $89.94_{\pm 0.50}$
& \cellcolor{bronze}$54.41_{\pm 0.60}$
& \cellcolor{bronze}$99.40$
& \cellcolor{bronze}$99.90$
& \cellcolor{bronze}$8.40$ \\
SCRUB
& \cellcolor{gold}$3.39$
& \cellcolor{gold}$97.52_{\pm 0.81}$
& $90.44_{\pm 0.51}$
& $90.44_{\pm 0.51}$
& \cellcolor{silver}$54.32_{\pm 0.72}$
& \cellcolor{gold}$98.10$
& \cellcolor{gold}$99.20$
& $10.00$ \\
SSD
& $4.16$
& $99.94_{\pm 0.10}$
& \cellcolor{gold}$92.40_{\pm 0.39}$
& \cellcolor{gold}$92.40_{\pm 0.39}$
& $57.26_{\pm 0.50}$
& $99.90$
& $100.00$
& \cellcolor{silver}$6.50$ \\
SalUn
& \cellcolor{bronze}$3.79$
& \cellcolor{silver}$98.02_{\pm 0.14}$
& $89.89_{\pm 0.19}$
& $89.89_{\pm 0.19}$
& \cellcolor{gold}$54.31_{\pm 0.73}$
& \cellcolor{silver}$99.30$
& \cellcolor{bronze}$99.90$
& $17.60$ \\
UNSIR
& $3.82$
& $98.61_{\pm 0.31}$
& \cellcolor{bronze}$90.52_{\pm 0.01}$
& \cellcolor{bronze}$90.52_{\pm 0.01}$
& $55.09_{\pm 0.27}$
& \cellcolor{bronze}$99.40$
& \cellcolor{bronze}$99.90$
& \cellcolor{gold}$4.60$ \\
\midrule
\multicolumn{9}{l}{\textit{Joint continual-learning and machine-unlearning methods}} \\
\midrule
UniCLUN
& $3.91$
& $99.12_{\pm 0.23}$
& $90.50_{\pm 0.45}$
& $90.50_{\pm 0.45}$
& $54.91_{\pm 1.72}$
& \cellcolor{bronze}$99.40$
& \cellcolor{bronze}$99.90$
& $8.60$ \\
\textbf{LwU} (ours)
& \cellcolor{silver}$3.68$
& \cellcolor{bronze}$98.26_{\pm 1.14}$
& \cellcolor{silver}$90.66_{\pm 0.86}$
& \cellcolor{silver}$90.66_{\pm 0.86}$
& $55.15_{\pm 0.93}$
& \cellcolor{silver}$99.30$
& \cellcolor{silver}$99.70$
& $61.60$ \\
\midrule
\multicolumn{9}{c}{\textbf{CIFAR-10 podium:}
\quad \textcolor{gold}{\faMedal}\,\textbf{SCRUB}\;(4G, 1S, 0B;\ \textbf{14 pts})
\quad \textcolor{silver}{\faMedal}\,\textbf{LwU (ours)}\;(0G, 5S, 1B;\ \textbf{11 pts})
\quad \textcolor{bronze}{\faMedal}\,\textbf{SalUn}\;(1G, 2S, 2B;\ \textbf{9 pts})
} \\
\midrule
\multicolumn{9}{>{\columncolor{dsgrey}}c}{\textbf{CIFAR-100}} \\
\midrule
Retrain (oracle)
& $0.00$
& $69.50_{\pm 0.60}$
& $68.17_{\pm 0.27}$
& $68.17_{\pm 0.27}$
& $49.71_{\pm 0.79}$
& $70.90$
& $76.10$
& $490.00$ \\
Original model
& $12.24$
& $99.14_{\pm 0.29}$
& $68.60_{\pm 0.48}$
& $68.60_{\pm 0.48}$
& $68.17_{\pm 0.40}$
& $99.30$
& $99.50$
& $0.00$ \\
Fine-tune on $D_r$
& $11.65$
& $98.14_{\pm 0.71}$
& $68.42_{\pm 0.39}$
& $68.42_{\pm 0.39}$
& $67.14_{\pm 0.63}$
& $98.60$
& $99.20$
& $51.10$ \\
\midrule
\multicolumn{9}{l}{\textit{Unlearning methods}} \\
\midrule
Amnesiac
& --
& --
& --
& --
& --
& --
& --
& -- \\
Bad Teacher
& $11.83$
& $96.72_{\pm 1.21}$
& \cellcolor{silver}$66.45_{\pm 0.80}$
& \cellcolor{silver}$66.45_{\pm 0.80}$
& $66.38_{\pm 1.35}$
& $98.70$
& $99.60$
& \cellcolor{gold}$5.60$ \\
SCRUB
& \cellcolor{silver}$9.62$
& \cellcolor{bronze}$90.86_{\pm 1.84}$
& $65.88_{\pm 0.42}$
& $65.88_{\pm 0.42}$
& \cellcolor{bronze}$62.22_{\pm 0.84}$
& \cellcolor{bronze}$92.90$
& \cellcolor{gold}$96.70$
& \cellcolor{bronze}$8.40$ \\
SSD
& $11.96$
& $97.06_{\pm 1.38}$
& \cellcolor{gold}$66.96_{\pm 0.90}$
& \cellcolor{gold}$66.96_{\pm 0.90}$
& $67.57_{\pm 0.77}$
& $98.90$
& $99.40$
& \cellcolor{silver}$6.60$ \\
SalUn
& \cellcolor{silver}$9.62$
& \cellcolor{silver}$74.81_{\pm 2.72}$
& $55.93_{\pm 0.99}$
& $55.93_{\pm 0.99}$
& \cellcolor{silver}$58.39_{\pm 1.15}$
& \cellcolor{silver}$89.10$
& \cellcolor{bronze}$98.20$
& $14.70$ \\
UNSIR
& $11.63$
& $96.03_{\pm 1.07}$
& $66.03_{\pm 0.06}$
& $66.03_{\pm 0.06}$
& $65.43_{\pm 0.13}$
& $97.90$
& $99.40$
& $10.60$ \\
\midrule
\multicolumn{9}{l}{\textit{Joint continual-learning and machine-unlearning methods}} \\
\midrule
UniCLUN
& \cellcolor{gold}$9.08$
& \cellcolor{gold}$74.58_{\pm 1.04}$
& $56.55_{\pm 0.96}$
& $56.55_{\pm 0.96}$
& \cellcolor{gold}$57.72_{\pm 0.28}$
& \cellcolor{gold}$88.50$
& \cellcolor{silver}$97.90$
& $14.90$ \\
\textbf{LwU} (ours)
& \cellcolor{bronze}$11.38$
& $95.56_{\pm 1.34}$
& \cellcolor{bronze}$66.30_{\pm 0.30}$
& \cellcolor{bronze}$66.30_{\pm 0.30}$
& $65.43_{\pm 0.37}$
& $97.60$
& $99.00$
& $46.30$ \\
\midrule
\multicolumn{9}{c}{\textbf{CIFAR-100 podium:}
\quad \textcolor{gold}{\faMedal}\,\textbf{UniCLUN}\;(4G, 1S, 0B;\ \textbf{14 pts})
\quad \textcolor{silver}{\faMedal}\,\textbf{SCRUB}\;(1G, 1S, 4B;\ \textbf{9 pts})
\quad \textcolor{bronze}{\faMedal}\,\textbf{SalUn}\;(0G, 4S, 1B;\ \textbf{9 pts})
} \\
\midrule
\multicolumn{9}{>{\columncolor{dsgrey}}c}{\textbf{RTI}} \\
\midrule
Retrain (oracle)
& $0.00$
& $78.57_{\pm 0.35}$
& $78.85_{\pm 0.33}$
& $78.85_{\pm 0.33}$
& $50.04_{\pm 1.14}$
& $80.10$
& $84.40$
& $635.40$ \\
Original model
& $8.86$
& $99.53_{\pm 0.18}$
& $79.39_{\pm 0.38}$
& $79.39_{\pm 0.38}$
& $63.45_{\pm 0.53}$
& $99.60$
& $99.90$
& $0.00$ \\
Fine-tune on $D_r$
& $8.48$
& $99.36_{\pm 0.05}$
& $79.01_{\pm 0.25}$
& $79.01_{\pm 0.25}$
& $62.75_{\pm 0.40}$
& $99.20$
& $99.70$
& $65.50$ \\
\midrule
\multicolumn{9}{l}{\textit{Unlearning methods}} \\
\midrule
Amnesiac
& --
& --
& --
& --
& --
& --
& --
& -- \\
Bad Teacher
& $8.68$
& $97.81_{\pm 0.40}$
& $77.03_{\pm 0.69}$
& $77.03_{\pm 0.69}$
& $61.87_{\pm 0.26}$
& $99.30$
& $99.90$
& $7.80$ \\
SCRUB
& \cellcolor{silver}$8.10$
& \cellcolor{bronze}$96.29_{\pm 2.78}$
& \cellcolor{silver}$77.35_{\pm 2.01}$
& \cellcolor{silver}$77.35_{\pm 2.01}$
& \cellcolor{bronze}$60.91_{\pm 2.03}$
& \cellcolor{bronze}$97.00$
& \cellcolor{gold}$98.20$
& \cellcolor{bronze}$7.40$ \\
SSD
& $8.75$
& $99.12_{\pm 0.58}$
& \cellcolor{gold}$78.71_{\pm 1.01}$
& \cellcolor{gold}$78.71_{\pm 1.01}$
& $63.32_{\pm 0.35}$
& $99.60$
& $99.90$
& \cellcolor{silver}$7.20$ \\
SalUn
& \cellcolor{gold}$7.91$
& \cellcolor{gold}$89.74_{\pm 1.90}$
& $72.35_{\pm 0.36}$
& $72.35_{\pm 0.36}$
& \cellcolor{gold}$57.53_{\pm 0.60}$
& \cellcolor{gold}$95.30$
& \cellcolor{silver}$98.90$
& $18.60$ \\
UNSIR
& $8.39$
& $97.37_{\pm 0.23}$
& \cellcolor{bronze}$77.16_{\pm 0.58}$
& \cellcolor{bronze}$77.16_{\pm 0.58}$
& $61.42_{\pm 1.55}$
& $98.60$
& \cellcolor{bronze}$99.60$
& \cellcolor{gold}$6.20$ \\
\midrule
\multicolumn{9}{l}{\textit{Joint continual-learning and machine-unlearning methods}} \\
\midrule
UniCLUN
& \cellcolor{bronze}$8.13$
& \cellcolor{silver}$90.58_{\pm 1.55}$
& $72.58_{\pm 1.04}$
& $72.58_{\pm 1.04}$
& \cellcolor{silver}$58.01_{\pm 0.31}$
& \cellcolor{silver}$95.80$
& \cellcolor{silver}$98.90$
& $14.70$ \\
\textbf{LwU} (ours)
& $8.58$
& $97.43_{\pm 1.30}$
& $76.99_{\pm 0.58}$
& $76.99_{\pm 0.58}$
& $61.78_{\pm 0.40}$
& $98.90$
& \cellcolor{bronze}$99.60$
& $53.20$ \\
\midrule
\multicolumn{9}{c}{\textbf{RTI podium:}
\quad \textcolor{gold}{\faMedal}\,\textbf{SalUn}\;(4G, 1S, 0B;\ \textbf{14 pts})
\quad \textcolor{silver}{\faMedal}\,\textbf{SCRUB}\;(1G, 3S, 4B;\ \textbf{13 pts})
\quad \textcolor{bronze}{\faMedal}\,\textbf{UniCLUN}\;(0G, 4S, 1B;\ \textbf{9 pts})
} \\
\midrule
\multicolumn{9}{c}{\textbf{Overall podium:}
\quad \textcolor{gold}{\faMedal}\,\textbf{SCRUB}\;(6G, 5S, 8B;\ \textbf{36 pts})
\quad \textcolor{silver}{\faMedal}\,\textbf{SalUn}\;(5G, 7S, 3B;\ \textbf{32 pts})
\quad \textcolor{bronze}{\faMedal}\,\textbf{UniCLUN}\;(4G, 5S, 3B;\ \textbf{25 pts})
} \\
\bottomrule
\end{tabular}
\end{adjustbox}
\end{table*}

\begin{table*}[p]
\centering
\setlength{\tabcolsep}{3.0pt}
\caption{ResNet-18 with class deletion on CIFAR-10, CIFAR-100, and RTI.}
\label{tab:resnet18-class-detailed}
\begin{adjustbox}{max width=\textwidth,max totalheight=0.92\textheight}
\begin{tabular}{lcccccccc}
\toprule
Method & Gap$\downarrow$ & $D_f\rightarrow D_f^{\mathrm{Retrain}}$
& $D_r\uparrow$ & Test$\uparrow$ & MIA$\rightarrow50$
& RL-1$\rightarrow$Retrain & RL-5$\rightarrow$Retrain & Time (s)$\downarrow$ \\
\midrule
\multicolumn{9}{>{\columncolor{dsgrey}}c}{\textbf{CIFAR-10}} \\
\midrule
Retrain (oracle)
& $0.00$
& $0.00_{\pm 0.00}$
& $93.24_{\pm 0.40}$
& $83.92_{\pm 0.36}$
& $50.23_{\pm 1.59}$
& $0.00$
& $0.10$
& $621.20$ \\
Original model
& $27.60$
& $93.90_{\pm 1.15}$
& $92.74_{\pm 0.36}$
& $92.86_{\pm 0.23}$
& $57.07_{\pm 0.53}$
& $66.20$
& $94.70$
& $0.00$ \\
Fine-tune on $D_r$
& $8.74$
& $26.23_{\pm 45.44}$
& $92.13_{\pm 0.53}$
& $85.54_{\pm 4.99}$
& $54.75_{\pm 1.91}$
& $17.70$
& $91.10$
& $79.80$ \\
\midrule
\multicolumn{9}{l}{\textit{Unlearning methods}} \\
\midrule
Amnesiac
& $39.38$
& \cellcolor{gold}$0.00_{\pm 0.00}$
& $11.11_{\pm 0.00}$
& $10.00_{\pm 0.00}$
& \cellcolor{gold}$49.09_{\pm 0.38}$
& $100.00$
& $100.00$
& \cellcolor{gold}$0.00$ \\
Bad Teacher
& \cellcolor{silver}$1.68$
& \cellcolor{silver}$0.13_{\pm 0.23}$
& \cellcolor{bronze}$92.89_{\pm 0.14}$
& $83.61_{\pm 0.12}$
& $55.89_{\pm 0.65}$
& $56.40$
& $97.80$
& $21.20$ \\
SCRUB
& \cellcolor{gold}$1.56$
& \cellcolor{gold}$0.00_{\pm 0.00}$
& \cellcolor{silver}$93.24_{\pm 0.31}$
& $83.91_{\pm 0.28}$
& $55.50_{\pm 1.50}$
& \cellcolor{gold}$0.00$
& \cellcolor{gold}$62.00$
& $10.40$ \\
SSD
& $27.24$
& $92.67_{\pm 3.43}$
& $92.71_{\pm 0.38}$
& \cellcolor{gold}$92.71_{\pm 0.03}$
& $57.00_{\pm 0.40}$
& $69.20$
& \cellcolor{bronze}$95.50$
& \cellcolor{bronze}$9.90$ \\
SalUn
& $7.25$
& $22.07_{\pm 38.22}$
& $91.65_{\pm 0.34}$
& $84.69_{\pm 4.12}$
& \cellcolor{silver}$51.39_{\pm 5.16}$
& $87.00$
& $98.50$
& $18.60$ \\
UNSIR
& $13.04$
& $43.23_{\pm 10.05}$
& $92.04_{\pm 0.14}$
& \cellcolor{silver}$87.16_{\pm 0.93}$
& $54.69_{\pm 0.29}$
& \cellcolor{bronze}$46.80$
& $96.40$
& \cellcolor{silver}$5.00$ \\
\midrule
\multicolumn{9}{l}{\textit{Joint continual-learning and machine-unlearning methods}} \\
\midrule
UniCLUN
& $13.06$
& $41.93_{\pm 36.42}$
& $92.06_{\pm 0.19}$
& \cellcolor{bronze}$87.05_{\pm 3.80}$
& $55.65_{\pm 2.39}$
& $72.50$
& $97.60$
& $18.40$ \\
\textbf{LwU} (ours)
& \cellcolor{bronze}$2.32$
& \cellcolor{bronze}$3.70_{\pm 6.41}$
& \cellcolor{gold}$93.31_{\pm 0.24}$
& $84.35_{\pm 0.54}$
& \cellcolor{bronze}$54.66_{\pm 2.20}$
& \cellcolor{silver}$6.50$
& \cellcolor{silver}$94.00$
& $89.20$ \\
\midrule
\multicolumn{9}{c}{\textbf{CIFAR-10 podium:}
\quad \textcolor{gold}{\faMedal}\,\textbf{SCRUB}\;(4G, 1S, 0B;\ \textbf{14 pts})
\quad \textcolor{silver}{\faMedal}\,\textbf{LwU (ours)}\;(1G, 2S, 3B;\ \textbf{10 pts})
\quad \textcolor{bronze}{\faMedal}\,\textbf{Amnesiac}\;(3G, 0S, 0B;\ \textbf{9 pts})
} \\
\midrule
\multicolumn{9}{>{\columncolor{dsgrey}}c}{\textbf{CIFAR-100}} \\
\midrule
Retrain (oracle)
& $0.00$
& $0.00_{\pm 0.00}$
& $70.74_{\pm 0.33}$
& $70.03_{\pm 0.33}$
& $48.86_{\pm 3.81}$
& $0.00$
& $0.00$
& $1086.40$ \\
Original model
& $23.88$
& $72.33_{\pm 3.21}$
& $70.58_{\pm 0.28}$
& $70.60_{\pm 0.27}$
& $71.04_{\pm 0.59}$
& $31.70$
& $9.70$
& $0.00$ \\
Fine-tune on $D_r$
& $14.75$
& $36.33_{\pm 4.51}$
& $70.38_{\pm 0.22}$
& $70.04_{\pm 0.18}$
& $71.00_{\pm 0.94}$
& $6.70$
& $1.00$
& $87.00$ \\
\midrule
\multicolumn{9}{l}{\textit{Unlearning methods}} \\
\midrule
Amnesiac
& $35.33$
& \cellcolor{gold}$0.00_{\pm 0.00}$
& $1.27_{\pm 0.23}$
& $1.26_{\pm 0.23}$
& \cellcolor{gold}$50.00_{\pm 0.00}$
& \cellcolor{gold}$0.00$
& \cellcolor{gold}$0.00$
& \cellcolor{gold}$0.00$ \\
Bad Teacher
& \cellcolor{silver}$7.19$
& \cellcolor{gold}$0.00_{\pm 0.00}$
& $69.33_{\pm 0.18}$
& $68.64_{\pm 0.18}$
& $74.83_{\pm 2.48}$
& \cellcolor{gold}$0.00$
& \cellcolor{gold}$0.00$
& $16.30$ \\
SCRUB
& \cellcolor{bronze}$12.76$
& $31.67_{\pm 9.87}$
& \cellcolor{gold}$70.65_{\pm 0.25}$
& \cellcolor{gold}$70.26_{\pm 0.29}$
& $67.29_{\pm 2.35}$
& \cellcolor{silver}$4.30$
& \cellcolor{gold}$0.00$
& $7.30$ \\
SSD
& $13.11$
& $35.00_{\pm 18.03}$
& \cellcolor{bronze}$70.50_{\pm 0.12}$
& \cellcolor{silver}$70.14_{\pm 0.15}$
& \cellcolor{bronze}$59.01_{\pm 14.19}$
& $23.30$
& $24.00$
& $9.90$ \\
SalUn
& $13.99$
& \cellcolor{silver}$29.00_{\pm 2.65}$
& $66.53_{\pm 0.50}$
& $66.15_{\pm 0.48}$
& $67.74_{\pm 1.53}$
& $15.00$
& $9.30$
& \cellcolor{bronze}$6.90$ \\
UNSIR
& $16.29$
& $44.67_{\pm 10.21}$
& $69.55_{\pm 0.42}$
& $69.30_{\pm 0.35}$
& $67.43_{\pm 3.02}$
& $16.70$
& \cellcolor{silver}$3.70$
& \cellcolor{silver}$5.30$ \\
\midrule
\multicolumn{9}{l}{\textit{Joint continual-learning and machine-unlearning methods}} \\
\midrule
UniCLUN
& $14.57$
& \cellcolor{bronze}$30.67_{\pm 1.53}$
& $66.66_{\pm 0.58}$
& $66.30_{\pm 0.56}$
& $68.69_{\pm 2.43}$
& \cellcolor{bronze}$14.30$
& \cellcolor{bronze}$9.00$
& $7.30$ \\
\textbf{LwU} (ours)
& \cellcolor{gold}$1.43$
& \cellcolor{gold}$0.00_{\pm 0.00}$
& \cellcolor{silver}$70.58_{\pm 0.34}$
& \cellcolor{bronze}$69.87_{\pm 0.33}$
& \cellcolor{silver}$50.54_{\pm 3.39}$
& \cellcolor{gold}$0.00$
& \cellcolor{gold}$0.00$
& $125.10$ \\
\midrule
\multicolumn{9}{c}{\textbf{CIFAR-100 podium:}
\quad \textcolor{gold}{\faMedal}\,\textbf{LwU (ours)}\;(4G, 2S, 1B;\ \textbf{17 pts})
\quad \textcolor{silver}{\faMedal}\,\textbf{Amnesiac}\;(5G, 0S, 0B;\ \textbf{15 pts})
\quad \textcolor{bronze}{\faMedal}\,\textbf{SCRUB}\;(3G, 1S, 1B;\ \textbf{12 pts})
} \\
\midrule
\multicolumn{9}{>{\columncolor{dsgrey}}c}{\textbf{RTI}} \\
\midrule
Retrain (oracle)
& $0.00$
& $0.00_{\pm 0.00}$
& $82.26_{\pm 0.41}$
& $78.15_{\pm 0.39}$
& $51.48_{\pm 1.24}$
& $0.00$
& $0.00$
& $717.40$ \\
Original model
& $23.52$
& $71.27_{\pm 2.14}$
& $81.42_{\pm 0.02}$
& $80.92_{\pm 0.09}$
& $70.70_{\pm 0.45}$
& $24.10$
& $56.70$
& $0.00$ \\
Fine-tune on $D_r$
& $14.14$
& $36.87_{\pm 5.08}$
& $81.93_{\pm 0.40}$
& $79.67_{\pm 0.36}$
& $69.31_{\pm 0.85}$
& $8.70$
& $45.90$
& $81.30$ \\
\midrule
\multicolumn{9}{l}{\textit{Unlearning methods}} \\
\midrule
Amnesiac
& $37.93$
& \cellcolor{gold}$0.00_{\pm 0.00}$
& $5.26_{\pm 0.00}$
& $5.00_{\pm 0.00}$
& \cellcolor{gold}$49.90_{\pm 0.25}$
& \cellcolor{gold}$0.00$
& \cellcolor{gold}$0.00$
& \cellcolor{gold}$0.00$ \\
Bad Teacher
& \cellcolor{silver}$4.38$
& \cellcolor{gold}$0.00_{\pm 0.00}$
& \cellcolor{bronze}$81.44_{\pm 0.26}$
& $77.37_{\pm 0.25}$
& $67.41_{\pm 2.01}$
& \cellcolor{silver}$0.20$
& $75.20$
& $19.80$ \\
SCRUB
& \cellcolor{bronze}$4.64$
& \cellcolor{gold}$0.00_{\pm 0.00}$
& \cellcolor{silver}$82.04_{\pm 0.03}$
& $77.94_{\pm 0.03}$
& $69.27_{\pm 1.02}$
& \cellcolor{gold}$0.00$
& \cellcolor{silver}$4.00$
& $8.20$ \\
SSD
& $13.38$
& \cellcolor{bronze}$36.67_{\pm 32.30}$
& $81.00_{\pm 0.90}$
& \cellcolor{silver}$78.78_{\pm 2.36}$
& \cellcolor{bronze}$65.43_{\pm 5.54}$
& $19.10$
& $64.90$
& $9.80$ \\
SalUn
& $14.86$
& $38.13_{\pm 1.10}$
& $80.54_{\pm 0.28}$
& $78.42_{\pm 0.26}$
& $70.66_{\pm 1.16}$
& $23.30$
& $56.80$
& \cellcolor{bronze}$7.00$ \\
UNSIR
& $17.59$
& $51.73_{\pm 18.15}$
& $80.95_{\pm 0.11}$
& \cellcolor{gold}$79.49_{\pm 0.90}$
& $67.44_{\pm 3.11}$
& \cellcolor{bronze}$18.00$
& $57.90$
& \cellcolor{silver}$6.90$ \\
\midrule
\multicolumn{9}{l}{\textit{Joint continual-learning and machine-unlearning methods}} \\
\midrule
UniCLUN
& $15.10$
& $39.33_{\pm 0.31}$
& $80.52_{\pm 0.27}$
& \cellcolor{bronze}$78.46_{\pm 0.25}$
& $70.35_{\pm 1.38}$
& $23.90$
& $56.90$
& $7.40$ \\
\textbf{LwU} (ours)
& \cellcolor{gold}$3.59$
& \cellcolor{silver}$0.07_{\pm 0.12}$
& \cellcolor{gold}$82.11_{\pm 0.02}$
& $78.01_{\pm 0.02}$
& \cellcolor{silver}$65.11_{\pm 5.47}$
& \cellcolor{silver}$0.20$
& \cellcolor{bronze}$33.90$
& $125.40$ \\
\midrule

\multicolumn{9}{c}{\textbf{RTI podium:}
\quad \textcolor{gold}{\faMedal}\,\textbf{Amnesiac}\;(5G, 0S, 0B;\ \textbf{15 pts})
\quad \textcolor{silver}{\faMedal}\,\textbf{LwU (ours)}\;(2G, 3S, 1B;\ \textbf{13 pts})
\quad \textcolor{bronze}{\faMedal}\,\textbf{SCRUB}\;(2G, 2S, 1B;\ \textbf{11 pts})
} \\
\midrule
\multicolumn{9}{c}{\textbf{Overall podium:}
\quad \textcolor{gold}{\faMedal}\,\textbf{LwU (ours)}\;(7G, 7S, 5B;\ \textbf{40 pts})
\quad \textcolor{silver}{\faMedal}\,\textbf{Amnesiac}\;(13G, 0S, 0B;\ \textbf{39 pts})
\quad \textcolor{bronze}{\faMedal}\,\textbf{SCRUB}\;(9G, 4S, 2B;\ \textbf{37 pts})
} \\
\bottomrule
\end{tabular}
\end{adjustbox}
\end{table*}

\begin{table*}[p]
\centering
\setlength{\tabcolsep}{3.0pt}
\caption{ResNet-18 with subclass deletion on CIFAR-10, CIFAR-100, and RTI.}
\label{tab:resnet18-subclass-detailed}
\begin{adjustbox}{max width=\textwidth,max totalheight=0.92\textheight}
\begin{tabular}{lcccccccc}
\toprule
Method & Gap$\downarrow$ & $D_f\rightarrow D_f^{\mathrm{Retrain}}$
& $D_r\uparrow$ & Test$\uparrow$ & MIA$\rightarrow50$
& RL-1$\rightarrow$Retrain & RL-5$\rightarrow$Retrain & Time (s)$\downarrow$ \\
\midrule
\multicolumn{9}{>{\columncolor{dsgrey}}c}{\textbf{CIFAR-10}} \\
\midrule
Retrain (oracle)
& $0.00$
& $26.57_{\pm 1.23}$
& $95.16_{\pm 0.07}$
& $88.30_{\pm 0.17}$
& $49.56_{\pm 1.53}$
& $73.60$
& $99.50$
& $479.50$ \\
Original model
& $20.42$
& $95.30_{\pm 0.69}$
& $95.16_{\pm 0.06}$
& $95.18_{\pm 0.03}$
& $55.57_{\pm 0.05}$
& $86.20$
& $97.70$
& $0.00$ \\
Fine-tune on $D_r$
& $12.97$
& $70.07_{\pm 0.68}$
& $94.13_{\pm 0.60}$
& $91.72_{\pm 0.56}$
& $53.49_{\pm 1.08}$
& $76.00$
& $98.60$
& $76.70$ \\
\midrule
\multicolumn{9}{l}{\textit{Unlearning methods}} \\
\midrule
Amnesiac
& --
& --
& --
& --
& --
& --
& --
& -- \\
Bad Teacher
& \cellcolor{gold}$2.25$
& \cellcolor{gold}$28.40_{\pm 4.10}$
& \cellcolor{silver}$94.37_{\pm 0.78}$
& $87.77_{\pm 0.87}$
& \cellcolor{bronze}$53.36_{\pm 1.38}$
& \cellcolor{bronze}$86.60$
& \cellcolor{bronze}$99.20$
& $15.70$ \\
SCRUB
& $16.55$
& $84.57_{\pm 0.06}$
& $93.68_{\pm 0.36}$
& $92.77_{\pm 0.32}$
& \cellcolor{gold}$51.80_{\pm 1.27}$
& $3.30$
& $74.00$
& \cellcolor{bronze}$12.30$ \\
SSD
& $16.97$
& $84.33_{\pm 9.85}$
& \cellcolor{bronze}$94.13_{\pm 0.84}$
& \cellcolor{silver}$93.15_{\pm 1.70}$
& $53.80_{\pm 1.57}$
& $88.30$
& $98.80$
& \cellcolor{silver}$8.40$ \\
SalUn
& \cellcolor{silver}$12.59$
& \cellcolor{bronze}$68.30_{\pm 4.06}$
& $93.63_{\pm 0.26}$
& $91.10_{\pm 0.64}$
& $53.85_{\pm 2.31}$
& $95.00$
& \cellcolor{gold}$99.50$
& $18.30$ \\
UNSIR
& $16.45$
& $82.30_{\pm 4.86}$
& $93.97_{\pm 0.35}$
& \cellcolor{bronze}$92.80_{\pm 0.49}$
& $53.92_{\pm 0.26}$
& \cellcolor{gold}$76.30$
& $98.30$
& \cellcolor{gold}$4.50$ \\
\midrule
\multicolumn{9}{l}{\textit{Joint continual-learning and machine-unlearning methods}} \\
\midrule
UniCLUN
& \cellcolor{bronze}$12.62$
& \cellcolor{silver}$68.10_{\pm 3.86}$
& $93.87_{\pm 0.19}$
& $91.30_{\pm 0.50}$
& \cellcolor{silver}$53.19_{\pm 3.62}$
& $94.90$
& \cellcolor{silver}$99.40$
& $15.10$ \\
\textbf{LwU} (ours)
& $18.50$
& $89.77_{\pm 7.82}$
& \cellcolor{gold}$94.50_{\pm 1.02}$
& \cellcolor{gold}$94.02_{\pm 1.70}$
& $53.99_{\pm 2.04}$
& \cellcolor{silver}$85.20$
& $98.40$
& $41.80$ \\
\midrule
\multicolumn{9}{c}{\textbf{CIFAR-10 podium:}
\quad \textcolor{gold}{\faMedal}\,\textbf{Bad Teacher}\;(2G, 1S, 3B;\ \textbf{11 pts})
\quad \textcolor{silver}{\faMedal}\,\textbf{LwU (ours)}\;(2G, 1S, 0B;\ \textbf{8 pts})
\quad \textcolor{bronze}{\faMedal}\,\textbf{UNSIR}\;(2G, 0S, 1B;\ \textbf{7 pts})
} \\
\midrule
\multicolumn{9}{>{\columncolor{dsgrey}}c}{\textbf{CIFAR-100}} \\
\midrule
Retrain (oracle)
& $0.00$
& $24.00_{\pm 2.00}$
& $80.04_{\pm 0.49}$
& $79.48_{\pm 0.47}$
& $49.32_{\pm 2.42}$
& $20.70$
& $17.00$
& $804.40$ \\
Original model
& $17.85$
& $80.00_{\pm 0.00}$
& $80.69_{\pm 0.36}$
& $80.68_{\pm 0.36}$
& $62.87_{\pm 1.25}$
& $44.70$
& $16.00$
& $0.00$ \\
Fine-tune on $D_r$
& $16.81$
& $75.33_{\pm 4.93}$
& $80.26_{\pm 0.13}$
& $80.21_{\pm 0.11}$
& $64.03_{\pm 3.41}$
& $41.30$
& $15.30$
& $45.40$ \\
\midrule
\multicolumn{9}{l}{\textit{Unlearning methods}} \\
\midrule
Amnesiac
& --
& --
& --
& --
& --
& --
& --
& -- \\
Bad Teacher
& \cellcolor{gold}$7.07$
& \cellcolor{gold}$17.00_{\pm 3.00}$
& $79.22_{\pm 0.41}$
& $78.60_{\pm 0.43}$
& $68.91_{\pm 1.43}$
& \cellcolor{gold}$18.30$
& $18.70$
& $21.90$ \\
SCRUB
& $18.14$
& $79.67_{\pm 0.58}$
& \cellcolor{gold}$80.64_{\pm 0.40}$
& \cellcolor{gold}$80.63_{\pm 0.39}$
& $64.46_{\pm 1.93}$
& $44.30$
& $13.70$
& $14.90$ \\
SSD
& $13.30$
& $57.67_{\pm 31.21}$
& $78.74_{\pm 3.08}$
& $78.53_{\pm 3.35}$
& \cellcolor{silver}$64.00_{\pm 2.67}$
& $34.00$
& $15.00$
& $12.90$ \\
SalUn
& \cellcolor{silver}$10.12$
& \cellcolor{silver}$42.00_{\pm 3.61}$
& $76.86_{\pm 0.48}$
& $76.51_{\pm 0.51}$
& $65.64_{\pm 2.38}$
& \cellcolor{bronze}$26.70$
& \cellcolor{bronze}$16.00$
& \cellcolor{gold}$6.30$ \\
UNSIR
& $17.37$
& $77.33_{\pm 6.11}$
& \cellcolor{bronze}$79.64_{\pm 0.86}$
& \cellcolor{bronze}$79.62_{\pm 0.91}$
& \cellcolor{bronze}$64.26_{\pm 4.58}$
& $44.00$
& $20.00$
& \cellcolor{silver}$7.20$ \\
\midrule
\multicolumn{9}{l}{\textit{Joint continual-learning and machine-unlearning methods}} \\
\midrule
UniCLUN
& \cellcolor{bronze}$10.52$
& \cellcolor{bronze}$44.67_{\pm 4.16}$
& $77.01_{\pm 0.60}$
& $76.69_{\pm 0.62}$
& $64.92_{\pm 3.37}$
& \cellcolor{silver}$25.30$
& \cellcolor{silver}$16.30$
& \cellcolor{bronze}$7.70$ \\
\textbf{LwU} (ours)
& $16.24$
& $74.33_{\pm 5.51}$
& \cellcolor{silver}$80.19_{\pm 0.56}$
& \cellcolor{silver}$80.13_{\pm 0.58}$
& \cellcolor{gold}$62.66_{\pm 2.55}$
& $41.00$
& \cellcolor{gold}$16.70$
& $44.30$ \\
\midrule
\multicolumn{9}{c}{\textbf{CIFAR-100 podium:}
\quad \textcolor{gold}{\faMedal}\,\textbf{LwU (ours)}\;(2G, 2S, 0B;\ \textbf{10 pts})
\quad \textcolor{silver}{\faMedal}\,\textbf{Bad Teacher}\;(3G, 0S, 0B;\ \textbf{9 pts})
\quad \textcolor{bronze}{\faMedal}\,\textbf{SalUn}\;(1G, 2S, 2B;\ \textbf{9 pts})
} \\
\midrule
\multicolumn{9}{>{\columncolor{dsgrey}}c}{\textbf{RTI}} \\
\midrule
Retrain (oracle)
& $0.00$
& $86.67_{\pm 3.21}$
& $80.54_{\pm 0.23}$
& $80.60_{\pm 0.26}$
& $49.87_{\pm 2.96}$
& $59.70$
& $40.30$
& $1144.70$ \\
Original model
& $3.13$
& $94.67_{\pm 2.52}$
& $80.54_{\pm 0.34}$
& $80.68_{\pm 0.36}$
& $54.23_{\pm 1.00}$
& $75.70$
& $42.30$
& $0.00$ \\
Fine-tune on $D_r$
& $3.03$
& $95.67_{\pm 1.53}$
& $80.14_{\pm 0.28}$
& $80.29_{\pm 0.28}$
& $51.95_{\pm 1.09}$
& $74.70$
& $42.00$
& $40.40$ \\
\midrule
\multicolumn{9}{l}{\textit{Unlearning methods}} \\
\midrule
Amnesiac
& --
& --
& --
& --
& --
& --
& --
& -- \\
Bad Teacher
& $15.63$
& $31.67_{\pm 4.16}$
& $79.09_{\pm 0.44}$
& $78.62_{\pm 0.40}$
& $53.56_{\pm 2.73}$
& $32.70$
& \cellcolor{silver}$41.30$
& $21.40$ \\
SCRUB
& \cellcolor{bronze}$3.09$
& \cellcolor{bronze}$95.00_{\pm 2.00}$
& \cellcolor{silver}$80.50_{\pm 0.41}$
& \cellcolor{silver}$80.65_{\pm 0.42}$
& $53.62_{\pm 1.16}$
& $72.30$
& $38.00$
& $11.50$ \\
SSD
& \cellcolor{silver}$2.76$
& \cellcolor{silver}$94.67_{\pm 2.52}$
& \cellcolor{gold}$80.54_{\pm 0.34}$
& \cellcolor{gold}$80.68_{\pm 0.37}$
& \cellcolor{bronze}$52.73_{\pm 1.80}$
& $75.00$
& \cellcolor{bronze}$41.70$
& $12.50$ \\
SalUn
& $7.20$
& $70.67_{\pm 3.06}$
& $74.78_{\pm 0.54}$
& $74.74_{\pm 0.52}$
& \cellcolor{gold}$50.64_{\pm 3.66}$
& \cellcolor{silver}$55.70$
& \cellcolor{gold}$39.70$
& \cellcolor{silver}$6.70$ \\
UNSIR
& $3.28$
& $95.33_{\pm 1.53}$
& $79.49_{\pm 0.33}$
& $79.65_{\pm 0.33}$
& \cellcolor{silver}$52.32_{\pm 1.69}$
& $73.70$
& $45.30$
& \cellcolor{bronze}$7.60$ \\
\midrule
\multicolumn{9}{l}{\textit{Joint continual-learning and machine-unlearning methods}} \\
\midrule
UniCLUN
& $7.85$
& $70.00_{\pm 2.00}$
& $74.90_{\pm 0.34}$
& $74.85_{\pm 0.34}$
& $52.74_{\pm 3.22}$
& \cellcolor{gold}$56.00$
& \cellcolor{silver}$39.30$
& \cellcolor{gold}$6.40$ \\
\textbf{LwU} (ours)
& \cellcolor{gold}$2.21$
& \cellcolor{gold}$90.33_{\pm 1.53}$
& \cellcolor{bronze}$79.85_{\pm 0.15}$
& \cellcolor{bronze}$79.95_{\pm 0.14}$
& $53.72_{\pm 1.09}$
& \cellcolor{bronze}$68.00$
& $47.00$
& $53.90$ \\
\midrule
\multicolumn{9}{c}{\textbf{RTI podium:}
\quad \textcolor{gold}{\faMedal}\,\textbf{SSD}\;(2G, 2S, 2B;\ \textbf{12 pts})
\quad \textcolor{silver}{\faMedal}\,\textbf{SalUn}\;(2G, 2S, 0B;\ \textbf{10 pts})
\quad \textcolor{bronze}{\faMedal}\,\textbf{LwU (ours)}\;(2G, 0S, 3B;\ \textbf{9 pts})
} \\
\midrule
\multicolumn{9}{c}{\textbf{Overall podium:}
\quad \textcolor{gold}{\faMedal}\,\textbf{LwU (ours)}\;(6G, 3S, 3B;\ \textbf{27 pts})
\quad \textcolor{silver}{\faMedal}\,\textbf{SalUn}\;(4G, 5S, 3B;\ \textbf{25 pts})
\quad \textcolor{bronze}{\faMedal}\,\textbf{Bad Teacher}\;(5G, 2S, 3B;\ \textbf{22 pts})
} \\
\bottomrule
\end{tabular}
\end{adjustbox}
\end{table*}

\begin{table*}[p]
\centering
\setlength{\tabcolsep}{3.0pt}
\caption{ResNet-18 with instance deletion on CIFAR-10, CIFAR-100, and RTI.}
\label{tab:resnet18-instance-detailed}
\begin{adjustbox}{max width=\textwidth,max totalheight=0.92\textheight}
\begin{tabular}{lcccccccc}
\toprule
Method & Gap$\downarrow$ & $D_f\rightarrow D_f^{\mathrm{Retrain}}$
& $D_r\uparrow$ & Test$\uparrow$ & MIA$\rightarrow50$
& RL-1$\rightarrow$Retrain & RL-5$\rightarrow$Retrain & Time (s)$\downarrow$ \\
\midrule
\multicolumn{9}{>{\columncolor{dsgrey}}c}{\textbf{CIFAR-10}} \\
\midrule
Retrain (oracle)
& $0.00$
& $92.06_{\pm 0.75}$
& $92.18_{\pm 0.30}$
& $92.18_{\pm 0.30}$
& $49.76_{\pm 0.52}$
& $92.90$
& $95.30$
& $511.10$ \\
Original model
& $4.28$
& $100.00_{\pm 0.00}$
& $92.86_{\pm 0.23}$
& $92.86_{\pm 0.23}$
& $57.60_{\pm 0.64}$
& $100.00$
& $100.00$
& $0.00$ \\
Fine-tune on $D_r$
& $3.08$
& $97.43_{\pm 2.04}$
& $91.28_{\pm 1.19}$
& $91.28_{\pm 1.19}$
& $54.26_{\pm 1.77}$
& $98.30$
& $99.30$
& $73.20$ \\
\midrule
\multicolumn{9}{l}{\textit{Unlearning methods}} \\
\midrule
Amnesiac
& --
& --
& --
& --
& --
& --
& --
& -- \\
Bad Teacher
& $3.89$
& $98.97_{\pm 0.36}$
& $90.53_{\pm 0.64}$
& $90.53_{\pm 0.64}$
& $55.13_{\pm 1.50}$
& $99.70$
& \cellcolor{bronze}$99.90$
& \cellcolor{bronze}$9.40$ \\
SCRUB
& \cellcolor{gold}$3.40$
& \cellcolor{bronze}$98.82_{\pm 0.66}$
& \cellcolor{bronze}$91.23_{\pm 0.59}$
& \cellcolor{bronze}$91.23_{\pm 0.59}$
& \cellcolor{silver}$54.72_{\pm 0.52}$
& \cellcolor{silver}$99.40$
& \cellcolor{gold}$99.70$
& $10.70$ \\
SSD
& $4.28$
& $100.00_{\pm 0.00}$
& \cellcolor{gold}$92.89_{\pm 0.26}$
& \cellcolor{gold}$92.89_{\pm 0.26}$
& $57.54_{\pm 0.67}$
& $100.00$
& $100.00$
& $10.40$ \\
SalUn
& $3.99$
& \cellcolor{gold}$96.25_{\pm 2.23}$
& $88.76_{\pm 1.68}$
& $88.76_{\pm 1.68}$
& \cellcolor{gold}$54.68_{\pm 1.42}$
& \cellcolor{gold}$98.80$
& \cellcolor{bronze}$99.90$
& \cellcolor{silver}$7.70$ \\
UNSIR
& \cellcolor{silver}$3.71$
& $98.88_{\pm 0.58}$
& $90.74_{\pm 0.70}$
& $90.74_{\pm 0.70}$
& \cellcolor{bronze}$54.93_{\pm 0.78}$
& $99.70$
& \cellcolor{bronze}$99.90$
& \cellcolor{gold}$5.30$ \\
\midrule
\multicolumn{9}{l}{\textit{Joint continual-learning and machine-unlearning methods}} \\
\midrule
UniCLUN
& \cellcolor{bronze}$3.73$
& \cellcolor{silver}$98.38_{\pm 0.80}$
& $90.48_{\pm 1.05}$
& $90.48_{\pm 1.05}$
& $54.99_{\pm 1.38}$
& \cellcolor{bronze}$99.50$
& \cellcolor{silver}$99.80$
& $11.60$ \\
\textbf{LwU} (ours)
& $4.10$
& $99.97_{\pm 0.05}$
& \cellcolor{silver}$92.77_{\pm 0.19}$
& \cellcolor{silver}$92.77_{\pm 0.19}$
& $57.08_{\pm 0.89}$
& $100.00$
& $100.00$
& $59.60$ \\
\midrule
\multicolumn{9}{c}{\textbf{CIFAR-10 podium:}
\quad \textcolor{gold}{\faMedal}\,\textbf{SCRUB}\;(2G, 2S, 3B;\ \textbf{13 pts})
\quad \textcolor{silver}{\faMedal}\,\textbf{SalUn}\;(3G, 1S, 1B;\ \textbf{12 pts})
\quad \textcolor{bronze}{\faMedal}\,\textbf{UNSIR}\;(1G, 1S, 2B;\ \textbf{7 pts})
} \\
\midrule
\multicolumn{9}{>{\columncolor{dsgrey}}c}{\textbf{CIFAR-100}} \\
\midrule
Retrain (oracle)
& $0.00$
& $69.58_{\pm 2.46}$
& $69.28_{\pm 0.73}$
& $69.28_{\pm 0.73}$
& $50.04_{\pm 0.73}$
& $72.00$
& $80.90$
& $685.10$ \\
Original model
& $13.77$
& $100.00_{\pm 0.00}$
& $70.60_{\pm 0.27}$
& $70.60_{\pm 0.27}$
& $72.06_{\pm 0.46}$
& $100.00$
& $100.00$
& $0.00$ \\
Fine-tune on $D_r$
& $13.26$
& $99.89_{\pm 0.19}$
& $70.23_{\pm 0.46}$
& $70.23_{\pm 0.46}$
& $70.88_{\pm 0.48}$
& $99.90$
& $100.00$
& $35.80$ \\
\midrule
\multicolumn{9}{l}{\textit{Unlearning methods}} \\
\midrule
Amnesiac
& --
& --
& --
& --
& --
& --
& --
& -- \\
Bad Teacher
& $12.65$
& $99.31_{\pm 0.77}$
& \cellcolor{silver}$68.47_{\pm 1.41}$
& \cellcolor{silver}$68.47_{\pm 1.41}$
& $69.29_{\pm 1.31}$
& $100.00$
& $100.00$
& \cellcolor{gold}$2.40$ \\
SCRUB
& \cellcolor{gold}$11.81$
& $97.94_{\pm 0.98}$
& \cellcolor{bronze}$68.40_{\pm 0.76}$
& \cellcolor{bronze}$68.40_{\pm 0.76}$
& \cellcolor{bronze}$67.15_{\pm 0.77}$
& \cellcolor{bronze}$98.40$
& \cellcolor{gold}$99.60$
& \cellcolor{silver}$8.30$ \\
SSD
& $12.83$
& $98.94_{\pm 0.46}$
& \cellcolor{gold}$68.96_{\pm 1.06}$
& \cellcolor{gold}$68.96_{\pm 1.06}$
& $70.33_{\pm 0.23}$
& $99.90$
& $100.00$
& \cellcolor{bronze}$10.00$ \\
SalUn
& \cellcolor{silver}$12.01$
& \cellcolor{gold}$84.53_{\pm 1.61}$
& $58.67_{\pm 0.73}$
& $58.67_{\pm 0.73}$
& \cellcolor{gold}$61.94_{\pm 0.73}$
& \cellcolor{gold}$96.30$
& \cellcolor{silver}$99.80$
& $22.60$ \\
UNSIR
& $12.79$
& $98.86_{\pm 0.81}$
& $68.03_{\pm 1.06}$
& $68.03_{\pm 1.06}$
& $69.43_{\pm 0.20}$
& $99.80$
& $100.00$
& $17.80$ \\
\midrule
\multicolumn{9}{l}{\textit{Joint continual-learning and machine-unlearning methods}} \\
\midrule
UniCLUN
& $12.61$
& \cellcolor{silver}$85.31_{\pm 1.41}$
& $58.37_{\pm 2.36}$
& $58.37_{\pm 2.36}$
& \cellcolor{silver}$62.93_{\pm 1.02}$
& \cellcolor{silver}$96.70$
& \cellcolor{bronze}$99.90$
& $21.70$ \\
\textbf{LwU} (ours)
& \cellcolor{bronze}$12.25$
& \cellcolor{bronze}$97.81_{\pm 0.41}$
& $68.09_{\pm 0.50}$
& $68.09_{\pm 0.50}$
& $68.46_{\pm 1.75}$
& $99.20$
& \cellcolor{bronze}$99.90$
& $74.20$ \\
\midrule
\multicolumn{9}{c}{\textbf{CIFAR-100 podium:}
\quad \textcolor{gold}{\faMedal}\,\textbf{SalUn}\;(3G, 2S, 0B;\ \textbf{13 pts})
\quad \textcolor{silver}{\faMedal}\,\textbf{SCRUB}\;(2G, 1S, 4B;\ \textbf{12 pts})
\quad \textcolor{bronze}{\faMedal}\,\textbf{SSD}\;(2G, 0S, 1B;\ \textbf{7 pts})
} \\
\midrule
\multicolumn{9}{>{\columncolor{dsgrey}}c}{\textbf{RTI}} \\
\midrule
Retrain (oracle)
& $0.00$
& $80.99_{\pm 1.01}$
& $80.05_{\pm 0.50}$
& $80.05_{\pm 0.50}$
& $49.93_{\pm 0.33}$
& $82.40$
& $87.80$
& $873.50$ \\
Original model
& $9.15$
& $99.97_{\pm 0.05}$
& $80.92_{\pm 0.09}$
& $80.92_{\pm 0.09}$
& $65.83_{\pm 1.11}$
& $99.90$
& $100.00$
& $0.00$ \\
Fine-tune on $D_r$
& $8.59$
& $99.82_{\pm 0.15}$
& $80.74_{\pm 0.43}$
& $80.74_{\pm 0.43}$
& $64.11_{\pm 0.88}$
& $99.90$
& $100.00$
& $66.20$ \\
\midrule
\multicolumn{9}{l}{\textit{Unlearning methods}} \\
\midrule
Amnesiac
& --
& --
& --
& --
& --
& --
& --
& -- \\
Bad Teacher
& \cellcolor{bronze}$8.59$
& $99.01_{\pm 0.83}$
& \cellcolor{bronze}$79.14_{\pm 1.04}$
& \cellcolor{bronze}$79.14_{\pm 1.04}$
& $63.52_{\pm 2.54}$
& $99.90$
& \cellcolor{bronze}$100.00$
& \cellcolor{gold}$3.90$ \\
SCRUB
& \cellcolor{gold}$7.82$
& $98.83_{\pm 0.48}$
& \cellcolor{silver}$79.40_{\pm 0.17}$
& \cellcolor{silver}$79.40_{\pm 0.17}$
& $62.08_{\pm 1.50}$
& \cellcolor{bronze}$99.40$
& \cellcolor{silver}$99.90$
& \cellcolor{bronze}$8.70$ \\
SSD
& $8.97$
& \cellcolor{gold}$92.16_{\pm 11.02}$
& $74.50_{\pm 7.98}$
& $74.50_{\pm 7.98}$
& \cellcolor{bronze}$60.80_{\pm 4.15}$
& $99.60$
& \cellcolor{gold}$99.80$
& $9.60$ \\
SalUn
& $8.93$
& \cellcolor{silver}$94.68_{\pm 0.28}$
& $74.00_{\pm 0.57}$
& $74.00_{\pm 0.57}$
& \cellcolor{gold}$59.84_{\pm 0.49}$
& \cellcolor{silver}$98.40$
& \cellcolor{gold}$99.80$
& $23.20$ \\
UNSIR
& $8.62$
& $98.71_{\pm 0.56}$
& $78.28_{\pm 0.83}$
& $78.28_{\pm 0.83}$
& $63.16_{\pm 0.92}$
& $99.60$
& \cellcolor{silver}$99.90$
& \cellcolor{silver}$8.10$ \\
\midrule
\multicolumn{9}{l}{\textit{Joint continual-learning and machine-unlearning methods}} \\
\midrule
UniCLUN
& \cellcolor{bronze}$8.59$
& \cellcolor{bronze}$95.15_{\pm 1.88}$
& $74.93_{\pm 1.06}$
& $74.93_{\pm 1.06}$
& \cellcolor{silver}$59.90_{\pm 0.91}$
& \cellcolor{gold}$98.30$
& \cellcolor{gold}$99.80$
& $22.80$ \\
\textbf{LwU} (ours)
& \cellcolor{silver}$8.48$
& $99.62_{\pm 0.43}$
& \cellcolor{gold}$80.17_{\pm 0.46}$
& \cellcolor{gold}$80.17_{\pm 0.46}$
& $64.21_{\pm 1.18}$
& $99.90$
& \cellcolor{bronze}$100.00$
& $81.10$ \\
\midrule
\multicolumn{9}{c}{\textbf{RTI podium:}
\quad \textcolor{gold}{\faMedal}\,\textbf{SCRUB}\;(1G, 3S, 2B;\ \textbf{11 pts})
\quad \textcolor{silver}{\faMedal}\,\textbf{SalUn}\;(2G, 2S, 0B;\ \textbf{10 pts})
\quad \textcolor{bronze}{\faMedal}\,\textbf{UniCLUN}\;(2G, 1S, 2B;\ \textbf{10 pts})
} \\
\midrule
\multicolumn{9}{c}{\textbf{Overall podium:}
\quad \textcolor{gold}{\faMedal}\,\textbf{SCRUB}\;(5G, 6S, 9B;\ \textbf{36 pts})
\quad \textcolor{silver}{\faMedal}\,\textbf{SalUn}\;(8G, 5S, 1B;\ \textbf{35 pts})
\quad \textcolor{bronze}{\faMedal}\,\textbf{UniCLUN}\;(2G, 6S, 5B;\ \textbf{23 pts})
} \\
\bottomrule
\end{tabular}
\end{adjustbox}
\end{table*}

\clearpage
\onecolumn
\subsection{Qualitative Machine-Unlearning Results on CIFAR-10}\label{sec:mu_qualitative}

\newcommand{\mufig}[4]{%
  \noindent\begin{minipage}{\textwidth}\centering
    \includegraphics[width=.75\linewidth]{Figs/forget_#1_#2.pdf}
    \captionof{figure}{CIFAR-10 #2 class (#3-class) removal.}\label{fig:cifar10-#4}
  \end{minipage}\par\vspace{1.2em}}

\mufig{0}{airplane}{1st}{a}
\mufig{1}{automobile}{2nd}{b}
\mufig{2}{bird}{3rd}{c}
\mufig{3}{cat}{4th}{d}
\mufig{4}{deer}{5th}{e}
\mufig{5}{dog}{6th}{f}
\mufig{6}{frog}{7th}{g}
\mufig{7}{horse}{8th}{h}
\mufig{8}{ship}{9th}{i}
\mufig{9}{truck}{10th}{j}

\clearpage
\twocolumn   

\twocolumn[{%
\subsection{Continual-Learning Cost}\label{sec:cl_cost}
\begin{minipage}{\textwidth}
\centering
\setlength{\tabcolsep}{3.2pt}
\captionof{table}{Training cost comparison.}
\label{tab:cl-training-cost}
\resizebox{\textwidth}{!}{%
\begin{tabular}{lcccccccc}
\toprule
& \multicolumn{4}{c}{TIL} & \multicolumn{4}{c}{CIL} \\
\cmidrule(lr){2-5}\cmidrule(lr){6-9}
Method & Train min$\downarrow$ & GPU h$\downarrow$ & Energy (Wh)$\downarrow$ & Peak device (MB)$\downarrow$ & Train min$\downarrow$ & GPU h$\downarrow$ & Energy (Wh)$\downarrow$ & Peak device (MB)$\downarrow$ \\ \midrule
\multicolumn{9}{>{\columncolor{dsgrey}}c}{\textbf{Seq-CIFAR-100}} \\ \midrule
Joint training (upper bound) & $40.33_{\pm 2.27}$ & $0.678_{\pm 0.038}$ & $84.53_{\pm 5.13}$ & $2001.0_{\pm 2.1}$ & $28.37_{\pm 0.68}$ & $0.479_{\pm 0.011}$ & $67.12_{\pm 3.43}$ & $2094.0_{\pm 3.2}$ \\
EWC & $8.73_{\pm 0.75}$ & $0.151_{\pm 0.012}$ & $17.97_{\pm 1.29}$ & \cellcolor{bronze}{$2148.5_{\pm 55.7}$} & $7.77_{\pm 1.26}$ & $0.136_{\pm 0.021}$ & $15.56_{\pm 2.00}$ & $2276.4_{\pm 151.9}$ \\
SI & $9.67_{\pm 0.45}$ & $0.168_{\pm 0.008}$ & $20.97_{\pm 1.03}$ & $3297.1_{\pm 773.7}$ & $9.11_{\pm 0.44}$ & $0.158_{\pm 0.007}$ & $20.74_{\pm 1.20}$ & $3542.4_{\pm 134.5}$ \\
LwF & $8.95_{\pm 0.68}$ & $0.154_{\pm 0.011}$ & $37.11_{\pm 2.49}$ & $4793.7_{\pm 126.4}$ & \cellcolor{silver}{$4.58_{\pm 0.41}$} & \cellcolor{silver}{$0.082_{\pm 0.007}$} & \cellcolor{bronze}{$12.02_{\pm 0.63}$} & \cellcolor{bronze}{$2144.3_{\pm 52.1}$} \\
WSN & \cellcolor{bronze}{$4.53_{\pm 0.23}$} & \cellcolor{bronze}{$0.082_{\pm 0.004}$} & $14.21_{\pm 0.77}$ & $2391.1_{\pm 50.9}$ & -- & -- & -- & -- \\
PEC & -- & -- & -- & -- & $25.07_{\pm 0.16}$ & $0.426_{\pm 0.003}$ & $96.16_{\pm 1.58}$ & $4473.6_{\pm 0.0}$ \\
SpaceNet & \cellcolor{gold}{$2.89_{\pm 0.03}$} & \cellcolor{gold}{$0.052_{\pm 0.001}$} & \cellcolor{gold}{$3.79_{\pm 0.07}$} & \cellcolor{gold}{$1334.8_{\pm 12.1}$} & \cellcolor{bronze}{$4.89_{\pm 0.55}$} & \cellcolor{bronze}{$0.085_{\pm 0.009}$} & \cellcolor{gold}{$6.19_{\pm 0.65}$} & \cellcolor{gold}{$1335.5_{\pm 11.6}$} \\
NISPA & \cellcolor{silver}{$3.63_{\pm 0.74}$} & \cellcolor{silver}{$0.066_{\pm 0.012}$} & \cellcolor{silver}{$7.58_{\pm 1.43}$} & \cellcolor{silver}{$1991.9_{\pm 135.5}$} & \cellcolor{gold}{$4.02_{\pm 0.01}$} & \cellcolor{gold}{$0.072_{\pm 0.000}$} & \cellcolor{silver}{$8.75_{\pm 0.08}$} & \cellcolor{silver}{$1985.6_{\pm 49.8}$} \\
SNV & $97.00_{\pm 23.73}$ & $1.624_{\pm 0.395}$ & $426.21_{\pm 112.91}$ & $2428.9_{\pm 0.0}$ & $81.31_{\pm 8.34}$ & $1.400_{\pm 0.139}$ & $319.94_{\pm 42.47}$ & $2428.1_{\pm 1.2}$ \\
UniCLUN & $7.42_{\pm 0.60}$ & $0.128_{\pm 0.010}$ & $25.77_{\pm 2.35}$ & $2514.8_{\pm 0.0}$ & $13.58_{\pm 1.79}$ & $0.233_{\pm 0.030}$ & $39.95_{\pm 5.95}$ & $2514.8_{\pm 0.0}$ \\
\textbf{LwU (ours)} & $6.08_{\pm 0.46}$ & $0.108_{\pm 0.008}$ & \cellcolor{bronze}{$14.11_{\pm 1.07}$} & $2554.7_{\pm 29.1}$ & $6.41_{\pm 0.61}$ & $0.113_{\pm 0.010}$ & $15.18_{\pm 0.98}$ & $2554.0_{\pm 27.8}$ \\
SGD (lower bound) & $3.06_{\pm 0.19}$ & $0.057_{\pm 0.003}$ & $8.86_{\pm 0.49}$ & $1915.0_{\pm 0.0}$ & $4.30_{\pm 0.08}$ & $0.077_{\pm 0.001}$ & $12.05_{\pm 0.18}$ & $1916.4_{\pm 1.2}$ \\
\midrule
\multicolumn{9}{>{\columncolor{dsgrey}}c}{\textbf{Seq-Tiny-ImageNet}} \\ \midrule
Joint training (upper bound) & -- & -- & -- & -- & -- & -- & -- & -- \\
EWC & \cellcolor{gold}{$1.09_{\pm 0.06}$} & \cellcolor{gold}{$0.182_{\pm 0.010}$} & \cellcolor{bronze}{$48.3_{\pm 2.6}$} & \cellcolor{silver}{$3179_{\pm 0}$} & \cellcolor{silver}{$0.99_{\pm 0.03}$} & \cellcolor{silver}{$0.164_{\pm 0.004}$} & \cellcolor{gold}{$43.1_{\pm 1.1}$} & \cellcolor{silver}{$3179_{\pm 0}$} \\
SI & $1.54_{\pm 0.14}$ & $0.257_{\pm 0.024}$ & $66.7_{\pm 5.7}$ & \cellcolor{bronze}{$3539_{\pm 0}$} & \cellcolor{bronze}{$1.12_{\pm 0.09}$} & \cellcolor{bronze}{$0.186_{\pm 0.015}$} & \cellcolor{bronze}{$47.7_{\pm 3.6}$} & $3539_{\pm 1}$ \\
LwF & $5.02_{\pm 0.30}$ & $0.836_{\pm 0.050}$ & $222.3_{\pm 13.3}$ & $13462_{\pm 68}$ & \cellcolor{gold}{$0.95_{\pm 0.05}$} & \cellcolor{gold}{$0.158_{\pm 0.008}$} & \cellcolor{silver}{$43.5_{\pm 2.1}$} & \cellcolor{bronze}{$3377_{\pm 0}$} \\
WSN & $4.97_{\pm 0.15}$ & $0.83_{\pm 0.03}$ & $89.0_{\pm 3.7}$ & $3555.7_{\pm 1.2}$ & -- & -- & -- & -- \\
PEC
& -- & -- & -- & --
& $35.81_{\pm 7.45}$
& $5.97_{\pm 1.24}$
& $819.8_{\pm 48.3}$
& $13992.5_{\pm 0.0}$ \\
SpaceNet & \cellcolor{silver}{$1.14_{\pm 0.12}$} & \cellcolor{silver}{$0.19_{\pm 0.02}$} & \cellcolor{gold}{$13.1_{\pm 1.2}$} & \cellcolor{gold}{$1596.3_{\pm 0.0}$} & $18.29_{\pm 1.30}$ & $3.05_{\pm 0.22}$ & $202.4_{\pm 13.1}$ & \cellcolor{gold}{$1596.3_{\pm 0.0}$} \\
NISPA & $1.41_{\pm 0.34}$ & \cellcolor{bronze}{$0.23_{\pm 0.06}$} & \cellcolor{silver}{$34.9_{\pm 8.4}$} & $4705.6_{\pm 553.8}$ & $2.99_{\pm 0.18}$ & $0.50_{\pm 0.03}$ & $73.5_{\pm 3.9}$ & $5659.8_{\pm 334.7}$ \\
SNV & $46.96_{\pm 7.83}$ & $7.83_{\pm 1.31}$ & $2051.0_{\pm 369.3}$ & $6263.1_{\pm 1.2}$ & $45.05_{\pm 8.18}$ & $7.51_{\pm 1.36}$ & $1888.3_{\pm 406.0}$ & $6265.9_{\pm 1.2}$ \\
UniCLUN & $4.40_{\pm 0.32}$ & $0.73_{\pm 0.05}$ & $157.2_{\pm 11.5}$ & $5759.1_{\pm 40.0}$ & $4.62_{\pm 0.75}$ & $0.77_{\pm 0.12}$ & $171.4_{\pm 27.0}$ & $5759.1_{\pm 40.0}$ \\
\textbf{LwU (ours)} & \cellcolor{bronze}{$1.40_{\pm 0.05}$} & $0.234_{\pm 0.008}$ & $63.8_{\pm 2.1}$ & $3665_{\pm 1}$ & $1.33_{\pm 0.08}$ & $0.221_{\pm 0.013}$ & $60.5_{\pm 3.3}$ & $3665_{\pm 0}$ \\
SGD (lower bound) & $0.74_{\pm 0.03}$ & $0.123_{\pm 0.005}$ & $33.7_{\pm 1.2}$ & $3169_{\pm 0}$ & $0.81_{\pm 0.03}$ & $0.135_{\pm 0.005}$ & $36.4_{\pm 1.3}$ & $3169_{\pm 0}$ \\
\midrule
\multicolumn{9}{>{\columncolor{dsgrey}}c}{\textbf{RTI}} \\ \midrule
Joint training (upper bound) & $6.36_{\pm 0.36}$ & $0.1073_{\pm 0.0060}$ & $25.70_{\pm 1.19}$ & $2347_{\pm 29}$ & $5.22_{\pm 0.15}$ & $0.0882_{\pm 0.0025}$ & $23.08_{\pm 0.40}$ & $2340_{\pm 25}$ \\
EWC & \cellcolor{bronze}{$2.24_{\pm 0.07}$} & \cellcolor{bronze}{$0.0387_{\pm 0.0011}$} & \cellcolor{bronze}{$9.10_{\pm 0.25}$} & \cellcolor{bronze}{$2226_{\pm 16}$} & $2.18_{\pm 0.08}$ & $0.0377_{\pm 0.0014}$ & $8.93_{\pm 0.14}$ & $2273_{\pm 46}$ \\
SI & $2.63_{\pm 0.09}$ & $0.0453_{\pm 0.0015}$ & $12.30_{\pm 0.31}$ & $2759_{\pm 19}$ & $2.27_{\pm 0.13}$ & $0.0394_{\pm 0.0021}$ & $10.49_{\pm 0.53}$ & $2745_{\pm 17}$ \\
LwF & $4.01_{\pm 0.18}$ & $0.0682_{\pm 0.0030}$ & $20.08_{\pm 0.90}$ & $3333_{\pm 20}$ & \cellcolor{bronze}{$1.89_{\pm 0.14}$} & \cellcolor{bronze}{$0.0328_{\pm 0.0024}$} & $9.26_{\pm 0.67}$ & \cellcolor{bronze}{$2165_{\pm 22}$} \\
WSN & $3.07_{\pm 0.19}$ & $0.0530_{\pm 0.0031}$ & $13.34_{\pm 0.91}$ & $2476_{\pm 1}$ & -- & -- & -- & -- \\
PEC & -- & -- & -- & -- & $12.42_{\pm 0.01}$ & $0.2098_{\pm 0.0001}$ & $62.82_{\pm 0.05}$ & $4643_{\pm 1}$ \\
SpaceNet & \cellcolor{gold}{$1.77_{\pm 0.21}$} & \cellcolor{gold}{$0.0304_{\pm 0.0034}$} & \cellcolor{gold}{$2.56_{\pm 0.39}$} & \cellcolor{gold}{$1541_{\pm 12}$} & \cellcolor{gold}{$1.49_{\pm 0.08}$} & \cellcolor{gold}{$0.0257_{\pm 0.0014}$} & \cellcolor{gold}{$2.14_{\pm 0.11}$} & \cellcolor{gold}{$1534$} \\
NISPA & \cellcolor{silver}{$2.17_{\pm 0.50}$} & \cellcolor{silver}{$0.0373_{\pm 0.0083}$} & \cellcolor{silver}{$5.40_{\pm 1.21}$} & \cellcolor{silver}{$2073_{\pm 40}$} & $2.01_{\pm 0.17}$ & $0.0346_{\pm 0.0029}$ & \cellcolor{silver}{$5.05_{\pm 0.45}$} & \cellcolor{silver}{$2136_{\pm 42}$} \\
SNV
& $33.84_{\pm 8.28}$
& $0.5785_{\pm 0.1408}$
& $186.12_{\pm 49.31}$
& $2619_{\pm 1}$
& $28.46_{\pm 2.92}$
& $0.4886_{\pm 0.0501}$
& $139.45_{\pm 18.52}$
& $2621_{\pm 1}$ \\
UniCLUN & $4.69_{\pm 0.38}$ & $0.0795_{\pm 0.0063}$ & $20.31_{\pm 1.49}$ & $2874$ & $6.61_{\pm 0.29}$ & $0.1115_{\pm 0.0049}$ & $29.81_{\pm 1.28}$ & $2874$ \\
\textbf{LwU (ours)} & $3.07_{\pm 0.54}$ & $0.0546_{\pm 0.0090}$ & $14.72_{\pm 2.36}$ & $2389_{\pm 41}$ & \cellcolor{silver}{$1.57_{\pm 0.15}$} & \cellcolor{silver}{$0.0296_{\pm 0.0025}$} & \cellcolor{bronze}{$7.80_{\pm 0.63}$} & $2369_{\pm 13}$ \\
SGD (lower bound) & $1.12_{\pm 0.13}$ & $0.0198_{\pm 0.0021}$ & $5.44_{\pm 0.57}$ & $2071$ & $1.28_{\pm 0.03}$ & $0.0225_{\pm 0.0005}$ & $6.17_{\pm 0.14}$ & $2071$ \\
\bottomrule
\end{tabular}%
}
\end{minipage}\vspace{1.5em}}]

\begin{table*}[tp]
\centering
\setlength{\tabcolsep}{3.2pt}
\caption{Memory and storage measurements for TIL.}
\label{tab:cl-til-memory-cost}
\resizebox{\textwidth}{!}{%
\begin{tabular}{lccccccc}
\toprule
Method & Peak alloc. (MB)$\downarrow$ & Peak reserved (MB)$\downarrow$ & Peak device (MB)$\downarrow$ & Optimizer (MB)$\downarrow$ & Model params (MB)$\downarrow$ & Resident storage (MB)$\downarrow$ & Checkpoint (MB)$\downarrow$ \\ \midrule
\multicolumn{8}{>{\columncolor{dsgrey}}c}{\textbf{Seq-CIFAR-100}} \\ \midrule
Joint training (upper bound) & $636.2_{\pm 2.4}$ & $784.3_{\pm 2.1}$ & $2001.0_{\pm 2.1}$ & $89.6_{\pm 0.0}$ & $89.6_{\pm 0.0}$ & $89.6_{\pm 0.0}$ & $44.9_{\pm 0.0}$ \\
EWC & $801.7_{\pm 52.5}$ & \cellcolor{bronze}{$931.8_{\pm 55.7}$} & \cellcolor{bronze}{$2148.5_{\pm 55.7}$} & $89.6_{\pm 0.0}$ & \cellcolor{bronze}{$44.8_{\pm 0.0}$} & $134.4_{\pm 0.0}$ & $134.5_{\pm 0.0}$ \\
SI & $1984.8_{\pm 766.0}$ & $2080.4_{\pm 773.7}$ & $3297.1_{\pm 773.7}$ & $89.6_{\pm 0.0}$ & \cellcolor{bronze}{$44.8_{\pm 0.0}$} & $313.4_{\pm 0.0}$ & $313.6_{\pm 0.0}$ \\
LwF & $3404.3_{\pm 122.6}$ & $3577.0_{\pm 126.4}$ & $4793.7_{\pm 126.4}$ & $89.6_{\pm 0.0}$ & $89.6_{\pm 0.0}$ & $89.6_{\pm 0.0}$ & $89.8_{\pm 0.0}$ \\
WSN & $945.6_{\pm 1.8}$ & $1174.4_{\pm 50.9}$ & $2391.1_{\pm 50.9}$ & $178.6_{\pm 0.0}$ & $89.4_{\pm 0.0}$ & $213.0_{\pm 0.0}$ & $213.4_{\pm 0.0}$ \\
PEC & -- & -- & -- & -- & -- & -- & -- \\
SpaceNet & \cellcolor{gold}{$105.2_{\pm 0.8}$} & \cellcolor{gold}{$126.5_{\pm 12.1}$} & \cellcolor{gold}{$1334.8_{\pm 12.1}$} & \cellcolor{gold}{$11.3_{\pm 0.0}$} & \cellcolor{gold}{$5.6_{\pm 0.0}$} & \cellcolor{gold}{$26.8_{\pm 0.0}$} & \cellcolor{gold}{$14.1_{\pm 0.0}$} \\
NISPA & \cellcolor{silver}{$503.4_{\pm 75.4}$} & \cellcolor{silver}{$775.2_{\pm 135.5}$} & \cellcolor{silver}{$1991.9_{\pm 135.5}$} & \cellcolor{silver}{$70.0_{\pm 0.0}$} & \cellcolor{silver}{$35.0_{\pm 0.0}$} & $78.8_{\pm 0.0}$ & $78.8_{\pm 0.0}$ \\
SNV & \cellcolor{bronze}{$770.6_{\pm 4.5}$} & $1210.1_{\pm 0.0}$ & $2428.9_{\pm 0.0}$ & \cellcolor{bronze}{$89.4_{\pm 0.0}$} & \cellcolor{bronze}{$44.8_{\pm 0.0}$} & \cellcolor{bronze}{$49.9_{\pm 0.0}$} & \cellcolor{bronze}{$46.8_{\pm 0.0}$} \\
UniCLUN & $1074.3_{\pm 2.4}$ & $1296.0_{\pm 0.0}$ & $2514.8_{\pm 0.0}$ & $92.2_{\pm 0.0}$ & $92.2_{\pm 0.0}$ & $104.6_{\pm 0.0}$ & $104.7_{\pm 0.0}$ \\
\textbf{LwU (ours)} & $916.3_{\pm 5.5}$ & $1335.9_{\pm 29.1}$ & $2554.7_{\pm 29.1}$ & $89.6_{\pm 0.0}$ & \cellcolor{bronze}{$44.8_{\pm 0.0}$} & \cellcolor{silver}{$44.8_{\pm 0.0}$} & \cellcolor{silver}{$45.0_{\pm 0.0}$} \\
SGD (lower bound) & $543.5_{\pm 0.0}$ & $698.4_{\pm 0.0}$ & $1915.0_{\pm 0.0}$ & $89.4_{\pm 0.0}$ & $44.8_{\pm 0.0}$ & $44.8_{\pm 0.0}$ & $44.9_{\pm 0.0}$ \\
\midrule
\multicolumn{8}{>{\columncolor{dsgrey}}c}{\textbf{Seq-Tiny-ImageNet}} \\ \midrule
Joint training (upper bound) & -- & -- & -- & -- & -- & -- & -- \\
EWC & \cellcolor{silver}{$1564.8_{\pm 1.2}$} & \cellcolor{silver}{$1814.0_{\pm 0.0}$} & \cellcolor{silver}{$3178.8_{\pm 0.0}$} & \cellcolor{silver}{$85.3_{\pm 0.0}$} & \cellcolor{silver}{$42.8_{\pm 0.0}$} & $128.0_{\pm 0.0}$ & $128.2_{\pm 0.0}$ \\
SI & $1910.7_{\pm 1.0}$ & \cellcolor{bronze}{$2174.0_{\pm 0.0}$} & \cellcolor{bronze}{$3538.8_{\pm 0.0}$} & \cellcolor{silver}{$85.3_{\pm 0.0}$} & \cellcolor{silver}{$42.8_{\pm 0.0}$} & $298.5_{\pm 0.0}$ & $298.7_{\pm 0.0}$ \\
LwF & $11773.7_{\pm 0.9}$ & $12097.3_{\pm 68.1}$ & $13462.1_{\pm 68.1}$ & $85.6_{\pm 0.0}$ & $85.6_{\pm 0.0}$ & \cellcolor{bronze}{$85.7_{\pm 0.0}$} & $85.8_{\pm 0.0}$ \\
WSN & \cellcolor{bronze}{$1870.6_{\pm 3.6}$} & $2339.0_{\pm 1.2}$ & $3555.7_{\pm 1.2}$ & $178.7_{\pm 0.0}$ & $89.5_{\pm 0.0}$ & $213.1_{\pm 0.0}$ & $213.5_{\pm 0.0}$ \\
PEC & -- & -- & -- & -- & -- & -- & -- \\
SpaceNet & \cellcolor{gold}{$340.8_{\pm 0.0}$} & \cellcolor{gold}{$385.9_{\pm 0.0}$} & \cellcolor{gold}{$1596.3_{\pm 0.0}$} & \cellcolor{gold}{$40.9_{\pm 0.0}$} & \cellcolor{gold}{$20.5_{\pm 0.0}$} & $97.3_{\pm 0.0}$ & \cellcolor{bronze}{$51.2_{\pm 0.0}$} \\
NISPA & $1905.1_{\pm 80.4}$ & $3489.0_{\pm 553.8}$ & $4705.6_{\pm 553.8}$ & $272.2_{\pm 0.0}$ & $136.1_{\pm 0.0}$ & $306.2_{\pm 0.0}$ & $306.2_{\pm 0.0}$ \\
SNV & $3232.3_{\pm 3.4}$ & $5044.3_{\pm 1.2}$ & $6263.1_{\pm 1.2}$ & $89.4_{\pm 0.0}$ & $44.9_{\pm 0.0}$ & \cellcolor{silver}{$73.4_{\pm 0.0}$} & \cellcolor{silver}{$46.9_{\pm 0.0}$} \\
UniCLUN & $3388.5_{\pm 1.2}$ & $4540.3_{\pm 40.0}$ & $5759.1_{\pm 40.0}$ & $92.4_{\pm 0.0}$ & $92.4_{\pm 0.0}$ & $141.6_{\pm 0.0}$ & $141.8_{\pm 0.0}$ \\
\textbf{LwU (ours)} & $2074.8_{\pm 9.8}$ & $2298.7_{\pm 1.2}$ & $3665.4_{\pm 1.2}$ & $85.6_{\pm 0.0}$ & \cellcolor{silver}{$42.8_{\pm 0.0}$} & \cellcolor{gold}{$42.8_{\pm 0.0}$} & \cellcolor{gold}{$43.0_{\pm 0.0}$} \\
SGD (lower bound) & $1430.9_{\pm 0.3}$ & $1804.0_{\pm 0.0}$ & $3168.8_{\pm 0.0}$ & $85.3_{\pm 0.0}$ & $42.8_{\pm 0.0}$ & $42.8_{\pm 0.0}$ & $42.9_{\pm 0.0}$ \\
\midrule
\multicolumn{8}{>{\columncolor{dsgrey}}c}{\textbf{RTI}} \\ \midrule
Joint training (upper bound) & $643_{\pm 25}$ & $916_{\pm 29}$ & $2347_{\pm 29}$ & $89.4$ & $89.4$ & $89.5$ & $44.8_{\pm 0.0}$ \\
EWC & \cellcolor{bronze}{$696_{\pm 3}$} & \cellcolor{bronze}{$795_{\pm 16}$} & \cellcolor{bronze}{$2226_{\pm 16}$} & \cellcolor{bronze}{$89.4$} & \cellcolor{bronze}{$44.7$} & $134.1$ & $134.2_{\pm 0.0}$ \\
SI & $1246_{\pm 7}$ & $1328_{\pm 19}$ & $2759_{\pm 19}$ & \cellcolor{bronze}{$89.4$} & \cellcolor{bronze}{$44.7$} & $312.9$ & $313.1_{\pm 0.0}$ \\
LwF & $1849_{\pm 29}$ & $1902_{\pm 20}$ & $3333_{\pm 20}$ & \cellcolor{bronze}{$89.4$} & $89.4$ & \cellcolor{bronze}{$89.5$} & $89.6_{\pm 0.0}$ \\
WSN & $912_{\pm 0}$ & $1045_{\pm 1}$ & $2476_{\pm 1}$ & $178.6$ & $89.3$ & $156.7$ & $157.0_{\pm 0.0}$ \\
PEC & -- & -- & -- & -- & -- & -- & -- \\
SpaceNet & \cellcolor{gold}{$105_{\pm 0}$} & \cellcolor{gold}{$127_{\pm 12}$} & \cellcolor{gold}{$1541_{\pm 12}$} & \cellcolor{gold}{$11.1$} & \cellcolor{gold}{$5.6$} & \cellcolor{gold}{$26.5$} & \cellcolor{gold}{$14.0_{\pm 0.0}$} \\
NISPA & \cellcolor{silver}{$395_{\pm 5}$} & \cellcolor{silver}{$642_{\pm 40}$} & \cellcolor{silver}{$2073_{\pm 40}$} & \cellcolor{silver}{$69.4$} & \cellcolor{silver}{$34.7$} & \cellcolor{silver}{$78.0$} & \cellcolor{bronze}{$78.1_{\pm 0.0}$} \\
SNV& $785_{\pm 5}$
& $1188_{\pm 1}$
& $2619_{\pm 1}$
& $89.4$
& $44.7$
& \cellcolor{bronze}{$49.8$}
& \cellcolor{bronze}{$46.8_{\pm 0.0}$} \\
UniCLUN & $1209_{\pm 1}$ & $1441$ & $2874$ & $92.0$ & $92.0$ & $104.4$ & $104.5_{\pm 0.0}$ \\
\textbf{LwU (ours)} & $881_{\pm 27}$ & $958_{\pm 41}$ & $2389_{\pm 41}$ & \cellcolor{bronze}{$89.4$} & \cellcolor{bronze}{$44.7$} & $245.9$ & \cellcolor{silver}{$44.8_{\pm 0.0}$} \\
SGD (lower bound) & $566$ & $640$ & $2071$ & $89.4$ & $44.7$ & $44.7$ & $44.8_{\pm 0.0}$ \\
\bottomrule
\end{tabular}%
}
\end{table*}

\begin{table*}[tp]
\centering
\setlength{\tabcolsep}{3.2pt}
\caption{Memory and storage measurements for CIL.}
\label{tab:cl-cil-memory-cost}
\resizebox{\textwidth}{!}{%
\begin{tabular}{lccccccc}
\toprule
Method & Peak alloc. (MB)$\downarrow$ & Peak reserved (MB)$\downarrow$ & Peak device (MB)$\downarrow$ & Optimizer (MB)$\downarrow$ & Model params (MB)$\downarrow$ & Resident storage (MB)$\downarrow$ & Checkpoint (MB)$\downarrow$ \\ \midrule
\multicolumn{8}{>{\columncolor{dsgrey}}c}{\textbf{Seq-CIFAR-100}} \\ \midrule
Joint training (upper bound) & $694.9_{\pm 27.7}$ & $877.3_{\pm 3.2}$ & $2094.0_{\pm 3.2}$ & $89.6_{\pm 0.0}$ & $89.6_{\pm 0.0}$ & $89.6_{\pm 0.0}$ & $44.9_{\pm 0.0}$ \\
EWC & $897.3_{\pm 191.4}$ & $1059.8_{\pm 151.9}$ & $2276.4_{\pm 151.9}$ & $89.6_{\pm 0.0}$ & \cellcolor{bronze}{$44.8_{\pm 0.0}$} & $134.4_{\pm 0.0}$ & $134.5_{\pm 0.0}$ \\
SI & $2219.6_{\pm 160.6}$ & $2325.7_{\pm 134.5}$ & $3542.4_{\pm 134.5}$ & $89.6_{\pm 0.0}$ & \cellcolor{bronze}{$44.8_{\pm 0.0}$} & $313.5_{\pm 0.0}$ & $313.7_{\pm 0.0}$ \\
LwF & \cellcolor{bronze}{$810.1_{\pm 23.2}$} & \cellcolor{bronze}{$927.6_{\pm 52.1}$} & \cellcolor{bronze}{$2144.3_{\pm 52.1}$} & $89.6_{\pm 0.0}$ & $89.6_{\pm 0.0}$ & $89.6_{\pm 0.0}$ & $89.8_{\pm 0.0}$ \\
WSN & -- & -- & -- & -- & -- & -- & -- \\
PEC & $2215.7_{\pm 0.0}$ & $3259.0_{\pm 0.0}$ & $4473.6_{\pm 0.0}$ & \cellcolor{gold}{$0.9_{\pm 0.0}$} & $66.8_{\pm 0.0}$ & $66.8_{\pm 0.0}$ & $67.0_{\pm 0.0}$ \\
SpaceNet & \cellcolor{gold}{$105.9_{\pm 0.1}$} & \cellcolor{gold}{$127.2_{\pm 11.6}$} & \cellcolor{gold}{$1335.5_{\pm 11.6}$} & \cellcolor{silver}{$11.3_{\pm 0.0}$} & \cellcolor{gold}{$5.6_{\pm 0.0}$} & \cellcolor{gold}{$26.8_{\pm 0.0}$} & \cellcolor{gold}{$14.2_{\pm 0.0}$} \\
NISPA & \cellcolor{silver}{$543.8_{\pm 17.0}$} & \cellcolor{silver}{$769.0_{\pm 49.8}$} & \cellcolor{silver}{$1985.6_{\pm 49.8}$} & \cellcolor{bronze}{$70.0_{\pm 0.0}$} & \cellcolor{silver}{$35.0_{\pm 0.0}$} & $78.8_{\pm 0.0}$ & $78.8_{\pm 0.0}$ \\
SNV & $811.3_{\pm 2.4}$ & $1209.4_{\pm 1.2}$ & $2428.1_{\pm 1.2}$ & $89.4_{\pm 0.0}$ & $44.9_{\pm 0.0}$ & \cellcolor{bronze}{$50.0_{\pm 0.0}$} & \cellcolor{bronze}{$46.9_{\pm 0.0}$} \\
UniCLUN & $1074.3_{\pm 2.4}$ & $1296.0_{\pm 0.0}$ & $2514.8_{\pm 0.0}$ & $92.2_{\pm 0.0}$ & $92.2_{\pm 0.0}$ & $104.6_{\pm 0.0}$ & $104.7_{\pm 0.0}$ \\
\textbf{LwU (ours)} & $909.8_{\pm 1.0}$ & $1335.2_{\pm 27.8}$ & $2554.0_{\pm 27.8}$ & $89.6_{\pm 0.0}$ & \cellcolor{bronze}{$44.8_{\pm 0.0}$} & \cellcolor{silver}{$44.8_{\pm 0.0}$} & \cellcolor{silver}{$45.0_{\pm 0.0}$} \\
SGD (lower bound) & $543.8_{\pm 0.0}$ & $699.7_{\pm 1.2}$ & $1916.4_{\pm 1.2}$ & $89.6_{\pm 0.0}$ & $44.8_{\pm 0.0}$ & $44.8_{\pm 0.0}$ & $44.9_{\pm 0.0}$ \\
\midrule
\multicolumn{8}{>{\columncolor{dsgrey}}c}{\textbf{Seq-Tiny-ImageNet}} \\ \midrule
Joint training (upper bound)
& $1562.6_{\pm 2.4}$
& $2187.3_{\pm 0.0}$
& $3404.0_{\pm 0.0}$
& $89.8_{\pm 0.0}$
& $89.8_{\pm 0.0}$
& $89.8_{\pm 0.0}$
& $45.0_{\pm 0.0}$ \\
EWC & \cellcolor{silver}{$1566.1_{\pm 1.8}$} & \cellcolor{silver}{$1814.0_{\pm 0.0}$} & \cellcolor{silver}{$3178.8_{\pm 0.0}$} & \cellcolor{silver}{$85.6_{\pm 0.0}$} & \cellcolor{silver}{$42.8_{\pm 0.0}$} & $128.0_{\pm 0.0}$ & $128.2_{\pm 0.0}$ \\
SI & $1912.5_{\pm 1.8}$ & $2174.7_{\pm 1.2}$ & $3539.4_{\pm 1.2}$ & \cellcolor{silver}{$85.6_{\pm 0.0}$} & \cellcolor{silver}{$42.8_{\pm 0.0}$} & $298.5_{\pm 0.0}$ & $298.7_{\pm 0.0}$ \\
LwF & \cellcolor{bronze}{$1817.4_{\pm 1.6}$} & \cellcolor{bronze}{$2012.0_{\pm 0.0}$} & \cellcolor{bronze}{$3376.8_{\pm 0.0}$} & \cellcolor{silver}{$85.6_{\pm 0.0}$} & $85.6_{\pm 0.0}$ & \cellcolor{bronze}{$85.7_{\pm 0.0}$} & $85.8_{\pm 0.0}$ \\
WSN & -- & -- & -- & -- & -- & -- & -- \\
PEC& $8560.9_{\pm 0.0}$
& $12777.9_{\pm 0.0}$
& $13992.5_{\pm 0.0}$
& \cellcolor{gold}{$0.9_{\pm 0.0}$}
& $89.1_{\pm 0.0}$
& $89.1_{\pm 0.0}$
& $89.4_{\pm 0.0}$ \\
SpaceNet & \cellcolor{gold}{$343.3_{\pm 0.0}$} & \cellcolor{gold}{$385.9_{\pm 0.0}$} & \cellcolor{gold}{$1596.3_{\pm 0.0}$} & \cellcolor{gold}{$40.9_{\pm 0.0}$} & \cellcolor{gold}{$20.5_{\pm 0.0}$} & $97.3_{\pm 0.0}$ & \cellcolor{bronze}{$51.2_{\pm 0.0}$} \\
NISPA & $2022.2_{\pm 57.8}$ & $4443.2_{\pm 334.7}$ & $5659.8_{\pm 334.7}$ & $272.2_{\pm 0.0}$ & $136.1_{\pm 0.0}$ & $306.2_{\pm 0.0}$ & $306.2_{\pm 0.0}$ \\
SNV & $3233.3_{\pm 1.6}$ & $5047.1_{\pm 1.2}$ & $6265.9_{\pm 1.2}$ & $89.4_{\pm 0.0}$ & $45.0_{\pm 0.0}$ & \cellcolor{silver}{$73.5_{\pm 0.0}$} & \cellcolor{silver}{$47.0_{\pm 0.0}$} \\
UniCLUN & $3387.9_{\pm 1.4}$ & $4540.3_{\pm 40.0}$ & $5759.1_{\pm 40.0}$ & $92.4_{\pm 0.0}$ & $92.4_{\pm 0.0}$ & $141.6_{\pm 0.0}$ & $141.8_{\pm 0.0}$ \\
\textbf{LwU (ours)} & $2068.4_{\pm 2.3}$ & $2298.0_{\pm 0.0}$ & $3664.8_{\pm 0.0}$ & \cellcolor{silver}{$85.6_{\pm 0.0}$} & \cellcolor{silver}{$42.8_{\pm 0.0}$} & \cellcolor{gold}{$42.8_{\pm 0.0}$} & \cellcolor{gold}{$43.0_{\pm 0.0}$} \\
SGD (lower bound) & $1431.2_{\pm 0.1}$ & $1804.0_{\pm 0.0}$ & $3168.8_{\pm 0.0}$ & $85.6_{\pm 0.0}$ & $42.8_{\pm 0.0}$ & $42.8_{\pm 0.0}$ & $42.9_{\pm 0.0}$ \\
\midrule
\multicolumn{8}{>{\columncolor{dsgrey}}c}{\textbf{RTI}} \\ \midrule
Joint training (upper bound) & $628_{\pm 26}$ & $909_{\pm 25}$ & $2340_{\pm 25}$ & $89.4$ & $89.4$ & $89.5$ & $44.8_{\pm 0.0}$ \\
EWC & $755_{\pm 52}$ & $842_{\pm 46}$ & $2273_{\pm 46}$ & $89.4$ & \cellcolor{bronze}{$44.7$} & $134.1$ & $134.2_{\pm 0.0}$ \\
SI & $1244_{\pm 2}$ & $1314_{\pm 17}$ & $2745_{\pm 17}$ & $89.4$ & \cellcolor{bronze}{$44.7$} & $312.9$ & $313.1_{\pm 0.0}$ \\
LwF & \cellcolor{bronze}{$659_{\pm 1}$} & \cellcolor{bronze}{$734_{\pm 22}$} & \cellcolor{bronze}{$2165_{\pm 22}$} & $89.4$ & $89.4$ & $89.5$ & $89.6_{\pm 0.0}$ \\
SNV & $826_{\pm 3}$
& $1190_{\pm 1}$
& $2621_{\pm 1}$
& $89.4$
& $44.7$
& $49.9$
& \cellcolor{bronze}{$46.8_{\pm 0.0}$} \\
PEC & $2176$ & $3212_{\pm 1}$ & $4643_{\pm 1}$ & \cellcolor{gold}{$0.9$} & $49.0$ & \cellcolor{silver}{$49.0$} & \cellcolor{bronze}{$49.0$} \\
SpaceNet & \cellcolor{gold}{$105_{\pm 0}$} & \cellcolor{gold}{$120$} & \cellcolor{gold}{$1534$} & \cellcolor{silver}{$11.1$} & \cellcolor{gold}{$5.6$} & \cellcolor{gold}{$26.5$} & \cellcolor{gold}{$14.0_{\pm 0.0}$} \\
NISPA & \cellcolor{silver}{$395_{\pm 5}$} & \cellcolor{silver}{$705_{\pm 42}$} & \cellcolor{silver}{$2136_{\pm 42}$} & \cellcolor{bronze}{$69.4$} & \cellcolor{silver}{$34.7$} & \cellcolor{bronze}{$78.0$} & $78.1_{\pm 0.0}$ \\
UniCLUN & $1210_{\pm 1}$ & $1441$ & $2874$ & $92.0$ & $92.0$ & $104.4$ & $104.5_{\pm 0.0}$ \\
\textbf{LwU (ours)} & $882_{\pm 25}$ & $938_{\pm 13}$ & $2369_{\pm 13}$ & $89.4$ & \cellcolor{bronze}{$44.7$} & $245.9$ & \cellcolor{silver}{$44.8_{\pm 0.0}$} \\
SGD (lower bound) & $566$ & $640$ & $2071$ & $89.4$ & $44.7$ & $44.7$ & $44.8_{\pm 0.0}$ \\
\bottomrule
\end{tabular}%
}
\end{table*}

\begin{table*}[tp]
\centering
\setlength{\tabcolsep}{3.2pt}
\caption{Inference cost comparison.}
\label{tab:cl-inference-cost}
\resizebox{\textwidth}{!}{%
\begin{tabular}{lcccccccc}
\toprule
& \multicolumn{4}{c}{TIL} & \multicolumn{4}{c}{CIL} \\
\cmidrule(lr){2-5}\cmidrule(lr){6-9}
Method & B1 ms/sample$\downarrow$ & GFLOPs/sample$\downarrow$ & B64 ms/sample$\downarrow$ & B64 sample/s$\uparrow$ & B1 ms/sample$\downarrow$ & GFLOPs/sample$\downarrow$ & B64 ms/sample$\downarrow$ & B64 sample/s$\uparrow$ \\ \midrule
\multicolumn{9}{>{\columncolor{dsgrey}}c}{\textbf{Seq-CIFAR-100}} \\ \midrule
Joint training (upper bound) & $1.837_{\pm 0.229}$ & $1.111_{\pm 0.000}$ & $0.0266_{\pm 0.0012}$ & $38216.0_{\pm 1351.0}$ & $1.844_{\pm 0.122}$ & $1.111_{\pm 0.000}$ & $0.0271_{\pm 0.0002}$ & $37238.2_{\pm 120.2}$ \\
EWC & $1.624_{\pm 0.112}$ & \cellcolor{bronze}{$1.111_{\pm 0.000}$} & \cellcolor{bronze}{$0.0247_{\pm 0.0006}$} & \cellcolor{bronze}{$40610.6_{\pm 967.4}$} & \cellcolor{bronze}{$1.755_{\pm 0.053}$} & $1.111_{\pm 0.000}$ & \cellcolor{bronze}{$0.0260_{\pm 0.0004}$} & \cellcolor{bronze}{$38555.8_{\pm 612.4}$} \\
SI & $1.756_{\pm 0.100}$ & \cellcolor{bronze}{$1.111_{\pm 0.000}$} & $0.0258_{\pm 0.0003}$ & $39074.8_{\pm 502.2}$ & $1.785_{\pm 0.148}$ & $1.111_{\pm 0.000}$ & $0.0270_{\pm 0.0010}$ & $37464.4_{\pm 1043.3}$ \\
LwF & $1.549_{\pm 0.039}$ & \cellcolor{bronze}{$1.111_{\pm 0.000}$} & $0.0250_{\pm 0.0003}$ & $40058.7_{\pm 361.4}$ & $1.861_{\pm 0.081}$ & $1.111_{\pm 0.000}$ & $0.0267_{\pm 0.0003}$ & $37556.1_{\pm 373.1}$ \\
WSN & $8.109_{\pm 0.404}$ & \cellcolor{bronze}{$1.111_{\pm 0.000}$} & $0.1233_{\pm 0.0038}$ & $8556.2_{\pm 192.1}$ & -- & -- & -- & -- \\
PEC
& -- & -- & -- & --
& $6.27_{\pm 0.671}$
& $0.8365_{\pm 0.000}$
& $0.2209_{\pm 0.0004}$
& $4526.5_{\pm 7.3}$ \\
SpaceNet & \cellcolor{gold}{$0.377_{\pm 0.142}$} & \cellcolor{gold}{$0.003_{\pm 0.000}$} & \cellcolor{gold}{$0.0055_{\pm 0.0012}$} & \cellcolor{gold}{$219715.3_{\pm 48405.0}$} & \cellcolor{gold}{$0.271_{\pm 0.113}$} & \cellcolor{gold}{$0.003_{\pm 0.000}$} & \cellcolor{gold}{$0.0041_{\pm 0.0006}$} & \cellcolor{gold}{$281369.5_{\pm 37115.0}$} \\
NISPA & \cellcolor{silver}{$0.761_{\pm 0.212}$} & \cellcolor{silver}{$0.209_{\pm 0.000}$} & \cellcolor{silver}{$0.0121_{\pm 0.0028}$} & \cellcolor{silver}{$95640.6_{\pm 16886.8}$} & \cellcolor{silver}{$0.611_{\pm 0.131}$} & \cellcolor{silver}{$0.209_{\pm 0.000}$} & \cellcolor{silver}{$0.0104_{\pm 0.0007}$} & \cellcolor{silver}{$100432.2_{\pm 5334.5}$} \\
SNV & $8.086_{\pm 0.280}$ & \cellcolor{bronze}{$1.111_{\pm 0.000}$} & $0.1237_{\pm 0.0071}$ & $8500.0_{\pm 394.3}$ & $121.286_{\pm 4.129}$ & $44.434_{\pm 0.000}$ & $1.9836_{\pm 0.0468}$ & $505.1_{\pm 11.1}$ \\
UniCLUN & \cellcolor{bronze}{$1.258_{\pm 0.095}$} & \cellcolor{bronze}{$1.111_{\pm 0.000}$} & $0.0249_{\pm 0.0009}$ & $40523.6_{\pm 988.8}$ & $1.806_{\pm 0.227}$ & $1.111_{\pm 0.000}$ & $0.0264_{\pm 0.0014}$ & $38236.6_{\pm 1506.7}$ \\
\textbf{LwU (ours)} & $2.590_{\pm 0.080}$ & \cellcolor{bronze}{$1.111_{\pm 0.000}$} & $0.0375_{\pm 0.0010}$ & $26802.5_{\pm 637.3}$ & $2.559_{\pm 0.189}$ & $1.111_{\pm 0.000}$ & $0.0378_{\pm 0.0002}$ & $26555.0_{\pm 181.9}$ \\
SGD (lower bound) & $1.561_{\pm 0.266}$ & $1.111_{\pm 0.000}$ & $0.0254_{\pm 0.0006}$ & $39679.7_{\pm 726.6}$ & $1.708_{\pm 0.125}$ & $1.111_{\pm 0.000}$ & $0.0265_{\pm 0.0005}$ & $37980.6_{\pm 529.0}$ \\
\midrule
\multicolumn{9}{>{\columncolor{dsgrey}}c}{\textbf{Seq-Tiny-ImageNet}} \\ \midrule
Joint training (upper bound) & $2.34_{\pm 0.18}$ & $4.443_{\pm 0.000}$ & $0.093_{\pm 0.002}$ & $10783_{\pm 197}$ & $2.64_{\pm 0.33}$ & $4.443_{\pm 0.000}$ & $0.093_{\pm 0.001}$ & $10787_{\pm 119}$ \\
EWC & $0.971_{\pm 0.368}$ & $4.443_{\pm 0.000}$ & $0.0969_{\pm 0.0025}$ & $10323_{\pm 271}$ & $0.941_{\pm 0.065}$ & $4.443_{\pm 0.000}$ & $0.0974_{\pm 0.0017}$ & $10273_{\pm 181}$ \\
SI & $0.880_{\pm 0.047}$ & $4.443_{\pm 0.000}$ & $0.0971_{\pm 0.0009}$ & $10299_{\pm 92}$ & $0.899_{\pm 0.032}$ & $4.443_{\pm 0.000}$ & $0.0974_{\pm 0.0007}$ & $10269_{\pm 69}$ \\
LwF & $0.820_{\pm 0.070}$ & $4.443_{\pm 0.000}$ & $0.0976_{\pm 0.0018}$ & $10244_{\pm 185}$ & $0.974_{\pm 0.029}$ & $4.443_{\pm 0.000}$ & $0.0980_{\pm 0.0003}$ & $10209_{\pm 33}$ \\
WSN & $7.15_{\pm 0.22}$ & $4.44_{\pm 0.00}$ & $0.153_{\pm 0.004}$ & $6575_{\pm 151}$ & -- & -- & -- & -- \\
PEC & -- & -- & -- & -- &$12.45_{\pm 0.48}$ & $1.806_{\pm 0.000}$ & $0.919_{\pm 0.005}$ & $1088_{\pm 6}$ \\
SpaceNet & $0.48_{\pm 0.16}$ & $0.01_{\pm 0.00}$ & \cellcolor{gold}{$0.008_{\pm 0.001}$} & \cellcolor{gold}{$171746_{\pm 10934}$} & $0.50_{\pm 0.12}$ & $0.01_{\pm 0.00}$ & \cellcolor{gold}{$0.009_{\pm 0.001}$} & \cellcolor{gold}{$182344_{\pm 30154}$} \\
NISPA & $1.41_{\pm 0.12}$ & $0.84_{\pm 0.00}$ & \cellcolor{silver}{$0.033_{\pm 0.001}$} & \cellcolor{silver}{$30396_{\pm 655}$} & $1.51_{\pm 0.14}$ & $0.84_{\pm 0.00}$ & \cellcolor{silver}{$0.033_{\pm 0.001}$} & \cellcolor{silver}{$30604_{\pm 829}$} \\
SNV & $7.71_{\pm 0.07}$ & $4.44_{\pm 0.00}$ & $0.162_{\pm 0.000}$ & $6192_{\pm 3}$ & $128.04_{\pm 3.00}$ & $177.73_{\pm 0.00}$ & $4.669_{\pm 0.038}$ & $214_{\pm 2}$ \\
UniCLUN & $2.28_{\pm 0.28}$ & $4.44_{\pm 0.00}$ & \cellcolor{bronze}{$0.094_{\pm 0.001}$} & \cellcolor{bronze}{$10638_{\pm 111}$} & $2.28_{\pm 0.42}$ & $4.44_{\pm 0.00}$ & \cellcolor{bronze}{$0.094_{\pm 0.001}$} & \cellcolor{bronze}{$10654_{\pm 76}$} \\
\textbf{LwU (ours)} & $1.441_{\pm 0.110}$ & $4.444_{\pm 0.000}$ & $0.1014_{\pm 0.0001}$ & $9861_{\pm 11}$ & $1.195_{\pm 0.123}$ & $4.444_{\pm 0.000}$ & $0.1014_{\pm 0.0001}$ & $9858_{\pm 11}$ \\
SGD (lower bound) & $0.843_{\pm 0.129}$ & $4.443_{\pm 0.000}$ & $0.0951_{\pm 0.0014}$ & $10521_{\pm 153}$ & $1.044_{\pm 0.375}$ & $4.443_{\pm 0.000}$ & $0.0972_{\pm 0.0012}$ & $10288_{\pm 129}$ \\
\midrule
\multicolumn{9}{>{\columncolor{dsgrey}}c}{\textbf{RTI}} \\ \midrule
Joint training (upper bound) & $0.649_{\pm 0.018}$ & $1.111$ & $0.0215_{\pm 0.0000}$ & $46415_{\pm 54}$ & $0.660_{\pm 0.030}$ & $1.111$ & $0.0228_{\pm 0.0001}$ & $43929_{\pm 124}$ \\
EWC & $0.692_{\pm 0.033}$ & \cellcolor{bronze}{$1.111$} & $0.0216_{\pm 0.0000}$ & $46200_{\pm 45}$ & $0.704_{\pm 0.023}$ & $1.111$ & $0.0228_{\pm 0.0001}$ & $43917_{\pm 152}$ \\
SI & $0.662_{\pm 0.001}$ & \cellcolor{bronze}{$1.111$} & $0.0217_{\pm 0.0001}$ & $45998_{\pm 107}$ & $0.659_{\pm 0.035}$ & $1.111$ & $0.0229_{\pm 0.0000}$ & $43665_{\pm 28}$ \\
LwF & $0.643_{\pm 0.029}$ & \cellcolor{bronze}{$1.111$} & $0.0218_{\pm 0.0001}$ & $45896_{\pm 118}$ & $0.658_{\pm 0.031}$ & $1.111$ & $0.0229_{\pm 0.0000}$ & $43751_{\pm 27}$ \\
WSN & $3.260_{\pm 0.171}$ & \cellcolor{bronze}{$1.111$} & $0.0529_{\pm 0.0001}$ & $18894_{\pm 45}$ & -- & -- & -- & -- \\
PEC & -- & -- & -- & -- & $0.758_{\pm 0.076}$ & \cellcolor{bronze}{$0.2675$} & $0.1388_{\pm 0.0003}$ & $7208_{\pm 16}$ \\
SpaceNet & $0.074_{\pm 0.017}$ & \cellcolor{gold}{$0.002786$} & \cellcolor{gold}{$0.0013_{\pm 0.0001}$} & \cellcolor{gold}{$772264_{\pm 43835}$} & $0.075_{\pm 0.009}$ & \cellcolor{gold}{$0.002786$} & \cellcolor{gold}{$0.0014_{\pm 0.0000}$} & \cellcolor{gold}{$748739_{\pm 3971}$} \\
NISPA & $0.152_{\pm 0.013}$ & \cellcolor{silver}{$0.2091$} & \cellcolor{silver}{$0.0063_{\pm 0.0000}$} & \cellcolor{silver}{$157682_{\pm 279}$} & $0.147_{\pm 0.007}$ & \cellcolor{silver}{$0.2091$} & \cellcolor{silver}{$0.0063_{\pm 0.0000}$} & \cellcolor{silver}{$157484_{\pm 455}$} \\
SNV& $3.247_{\pm 0.112}$
& $1.111$
& $0.0532_{\pm 0.0009}$
& $18797_{\pm 318}$
& $44.12_{\pm 1.51}$
& $44.434$
& $1.6901_{\pm 0.0398}$
& $592_{\pm 14}$ \\
UniCLUN & $0.624_{\pm 0.029}$ & \cellcolor{bronze}{$1.111$} & \cellcolor{bronze}{$0.0215_{\pm 0.0000}$} & \cellcolor{bronze}{$46581_{\pm 25}$} & $0.635_{\pm 0.028}$ & $1.111$ & \cellcolor{bronze}{$0.0227_{\pm 0.0000}$} & \cellcolor{bronze}{$44123_{\pm 70}$} \\
\textbf{LwU (ours)} & $6.177_{\pm 0.035}$ & $4.44$ & $0.1576_{\pm 0.0009}$ & $6346_{\pm 34}$ & $5.833_{\pm 0.664}$ & $4.44$ & $0.1556_{\pm 0.0027}$ & $6431_{\pm 112}$ \\
SGD (lower bound) & $0.631_{\pm 0.021}$ & $1.111$ & $0.0217_{\pm 0.0000}$ & $46180_{\pm 57}$ & $0.623_{\pm 0.010}$ & $1.111$ & $0.0227_{\pm 0.0000}$ & $43970_{\pm 35}$ \\
\bottomrule
\end{tabular}%
}
\vspace{2pt}

\end{table*}

\subsection{New Interleaved Continual Learning and Machine Unlearning Results}\label{sec:new_joint_clmu}
All new interleaved evaluations in this subsection use Split CIFAR-100 and ResNet-18 in Task-IL. The five-task setting comprises five tasks of 20 classes, three seeds, and deletion of the first class of task 1 after task 3, followed by learning tasks 4--5. $F_{\mathrm{del}}$ and $F_{\mathrm{end}}$ measure accuracy on the deleted class immediately after deletion and after the final task; for class deletion, the Retrain reference is zero. These experiments examine an interleaved lifecycle rather than inferring joint behavior from standalone CL and MU metrics.

In composed baselines, the continual learner completes tasks 1--3, an MU method edits the model, and continual learning resumes; when relevant, the LwF teacher and EWC anchor are reset to the edited model. CLPU-DER++ requires advance knowledge of the future forget class to isolate its model copy. Storage is measured at the deletion request. Default $\mathcal{D}'_r$ is a sampled 10\% of retained training images from tasks 1--3 (2,480 images); LwU maintains no replay buffer during sequential learning but has access to these images at deletion. An additional experiment removes this older-task access.

\begin{table*}[tp]
\centering
\setlength{\tabcolsep}{3pt}
\caption{Five-task joint CL--MU comparison. Means $\pm$ standard deviations over three seeds. The request-time variant uses no images from tasks 1--2; CLPU-DER++ requires advance knowledge of the deleted class.}
\label{tab:new_joint_baselines}
\begin{adjustbox}{max width=\textwidth}
\begin{tabular}{lccccc}
\toprule
Method & Stored images at the request & ACC $\uparrow$ & BWT $\uparrow$ & $F_{\mathrm{del}}$ $\downarrow$ & $F_{\mathrm{end}}$ $\downarrow$ \\
\midrule
Retrain (never sees $c_f$) & -- & $71.75\pm0.73$ & $-2.95\pm0.11$ & -- & $0.00\pm0.00$ \\
LwF + SSD$^\dagger$ & $\mathcal{D}'_r$ (2,480) & $64.53\pm9.76$ & $-13.68\pm13.82$ & $30.89\pm36.59$ & $32.82\pm38.22$ \\
LwF + SalUn & $\mathcal{D}'_r$ (2,480) & $69.94\pm2.45$ & $-7.24\pm0.83$ & $4.42\pm3.84$ & $2.26\pm2.68$ \\
EWC + SCRUB & $\mathcal{D}'_r$ (2,480) & $66.90\pm0.89$ & $0.10\pm0.92$ & $68.87\pm25.75$ & $65.54\pm29.65$ \\
UniCLUN & buffer (1,000) + $\mathcal{D}'_r$ & $49.62\pm10.16$ & $-26.19\pm6.25$ & $28.12\pm23.45$ & $25.74\pm43.23$ \\
CLPU-DER++$^\ddagger$ & buffer (600) + model copy & $65.81\pm1.25$ & $-15.63\pm1.82$ & $0.00\pm0.00$ & $0.00\pm0.00$ \\
\textbf{LwU} & $\mathcal{D}'_r$ (2,480) & \textbf{73.65$\pm$1.13} & \textbf{-1.45$\pm$0.67} & \textbf{0.00$\pm$0.00} & \textbf{0.00$\pm$0.00} \\
LwU, request-time $\mathcal{D}'_r$ & none from tasks 1--2 & $72.55\pm1.51$ & $-2.73\pm0.74$ & $0.00\pm0.00$ & $0.00\pm0.00$ \\
LwU, $\mathcal{D}'_r$ = 1\% & $\mathcal{D}'_r$ (248) & $71.80\pm1.52$ & $-3.90\pm0.98$ & $0.00\pm0.00$ & $0.00\pm0.00$ \\
\bottomrule
\end{tabular}
\end{adjustbox}
\end{table*}

The default LwU setting reaches 73.65\% ACC and a final deleted-class accuracy of zero, compared with 69.94\% ACC for LwF+SalUn and 65.81\% for CLPU-DER++. The request-time-only $\mathcal{D}'_r$ variant reaches 72.55\% ACC and also leaves the deleted class at zero. The difference between default and request-time retention partly reflects the availability of historical images, so the default should not be described as fully replay-free at deletion.

\begin{table*}[tbp]
\centering
\setlength{\tabcolsep}{3pt}
\caption{Retain-set provenance controls under the five-task joint protocol. The request-time variant has no images from tasks 1--2. Three seeds.}
\label{tab:new_request_subset}
\begin{adjustbox}{max width=\textwidth}
\begin{tabular}{lcccc}
\toprule
LwU setting & Images of tasks 1--2 stored & ACC $\uparrow$ & BWT $\uparrow$ & $F_{\mathrm{del}}$ / $F_{\mathrm{end}}$ $\downarrow$ \\
\midrule
No deletion & none & $72.65\pm1.20$ & $-2.78\pm0.32$ & -- \\
\textbf{Deletion, request-time $\mathcal{D}'_r$} & \textbf{none} & \textbf{72.55$\pm$1.51} & \textbf{-2.73$\pm$0.74} & \textbf{0.00 / 0.00} \\
Deletion, $\mathcal{D}'_r$ from tasks 1--3 (default) & yes & $73.65\pm1.13$ & $-1.45\pm0.67$ & $0.00 / 0.00$ \\
\bottomrule
\end{tabular}
\end{adjustbox}
\end{table*}

\begin{table*}[tbp]
\centering
\setlength{\tabcolsep}{3pt}
\caption{Deletion timing and retain-set sensitivity for LwU on the five-task joint protocol (three seeds). Values are reproduced from the additional experiments.}
\label{tab:new_deletion_schedule}
\begin{adjustbox}{max width=\textwidth}
\begin{tabular}{lcccc}
\toprule
Setting & ACC $\uparrow$ & BWT $\uparrow$ & $F_{\mathrm{del}}$ $\downarrow$ & $F_{\mathrm{end}}$ $\downarrow$ \\
\midrule
Delete after task 2 & $73.81\pm1.09$ & $-1.59\pm0.47$ & $0.00\pm0.00$ & $0.00\pm0.00$ \\
Delete after task 3 & $73.65\pm1.13$ & $-1.45\pm0.67$ & $0.00\pm0.00$ & $0.00\pm0.00$ \\
Delete after task 4 & $74.12\pm1.32$ & $-0.82\pm0.60$ & $0.00\pm0.00$ & $0.00\pm0.00$ \\
Two requests (after 2 and 4) & $74.28\pm1.11$ & $-1.14\pm0.88$ & $0.00\pm0.00$ & $0.00\pm0.00$ \\
Retain subset 1\% & $71.80\pm1.52$ & $-3.90\pm0.98$ & $0.00\pm0.00$ & $0.00\pm0.00$ \\
Retain subset 5\% & $73.35\pm1.43$ & $-1.69\pm0.72$ & $0.00\pm0.00$ & $0.00\pm0.00$ \\
Retain subset 25\% & $73.75\pm1.02$ & $-1.17\pm0.87$ & $0.00\pm0.00$ & $0.00\pm0.00$ \\
\bottomrule
\end{tabular}
\end{adjustbox}
\end{table*}

The 20-task study uses 20 tasks of five classes, with requests after tasks 5, 10, and 15. All methods use the same settings selected in the five-task study; they are not re-tuned for the longer stream. In this harder stream LwU retains strong deletion outcomes but underperforms the replay-based baselines in ACC. Individual request outcomes should not be substituted for privacy auditing.

\begin{table*}[tp]
\centering
\setlength{\tabcolsep}{3pt}
\caption{Twenty-task Split CIFAR-100 interleaved deletion with three requests. Each deleted-class accuracy tuple is ordered by request 1/2/3. Three seeds.}
\label{tab:new_long_stream}
\begin{adjustbox}{max width=\textwidth}
\begin{tabular}{lccccc}
\toprule
Method & Knows requests in advance & ACC $\uparrow$ & BWT $\uparrow$ & $F_{\mathrm{del}}$ r1 / r2 / r3 $\downarrow$ & $F_{\mathrm{end}}$ r1 / r2 / r3 $\downarrow$ \\
\midrule
LwF + SalUn & no & $79.24\pm3.68$ & $-7.20\pm2.02$ & $8.78 / 17.09 / 28.45$ & $2.01 / 28.25 / 12.13$ \\
CLPU-DER++ & yes & $81.22\pm1.73$ & $-6.31\pm0.58$ & $0.00 / 0.00 / 0.00$ & $0.00 / 0.00 / 0.00$ \\
LwU & no & $74.71\pm1.17$ & $-7.79\pm0.51$ & $0.00 / 0.00 / 3.98$ & $0.00 / 0.00 / 3.06$ \\
\bottomrule
\end{tabular}
\end{adjustbox}
\end{table*}

\subsubsection{Mechanism ablations in the joint stream}
These ablations distinguish the role of the four-zone updates from the empirical Fisher estimator itself. The dominance margin and self-distillation teacher have the largest impacts on accuracy. Omitting Zone B reset increases the deleted-class accuracy even when it leaves average CL accuracy nearly unchanged.

\begin{table}[t]
\centering
\setlength{\tabcolsep}{3pt}
\caption{Continual-learning-only ablations on five-task Split CIFAR-100; no deletion (three seeds).}
\label{tab:new_cl_only_ablation}
\begin{adjustbox}{max width=\columnwidth}
\begin{tabular}{lcc}
\toprule
Variant & ACC $\uparrow$ & BWT $\uparrow$ \\
\midrule
w/o self-distillation teacher & $45.78\pm1.52$ & $-36.72\pm0.55$ \\
w/o cumulative freeze mask & $72.00\pm1.48$ & $-5.75\pm1.43$ \\
\textbf{LwU} & \textbf{72.65$\pm$1.20} & \textbf{-2.78$\pm$0.32} \\
\bottomrule
\end{tabular}
\end{adjustbox}
\end{table}
\begin{table}[t]
\centering
\footnotesize
\setlength{\tabcolsep}{3pt}
\caption{Inference latency at batch size 1 on RTX PRO 6000, ResNet-18 ($32\times32$); milliseconds per sample.}
\label{tab:new_latency}
\begin{adjustbox}{max width=\columnwidth}
\begin{tabular}{lc}
\toprule
Setting & ms per sample \\
\midrule
Backbone (= LwU with memory off) & $0.57$ \\
LwU + memory, submitted code & $1.07$ \\
LwU + memory, sync-free (identical outputs) & $0.77$ \\
\textbf{LwU + memory, sync-free, compiled} & \textbf{0.57} \\
\bottomrule
\end{tabular}
\end{adjustbox}
\end{table}
\begin{table*}[tp]
\centering
\setlength{\tabcolsep}{3pt}
\caption{Five-task joint CL--MU component ablations on Split CIFAR-100; deleted class requested after task 3 (three seeds).}
\label{tab:new_joint_ablation}
\begin{adjustbox}{max width=\textwidth}
\begin{tabular}{lcccc}
\toprule
Variant & ACC $\uparrow$ & BWT $\uparrow$ & $F_{\mathrm{del}}$ $\downarrow$ & $F_{\mathrm{end}}$ $\downarrow$ \\
\midrule
No dominance margin ($\omega=1$) & $34.15\pm0.65$ & $-45.08\pm0.59$ & $0.00\pm0.00$ & $0.00\pm0.00$ \\
w/o self-distillation teacher & $48.80\pm2.95$ & $-32.91\pm3.45$ & $0.00\pm0.00$ & $0.00\pm0.00$ \\
w/o teacher repair & $71.12\pm1.56$ & $-4.54\pm1.27$ & $6.05\pm3.86$ & $8.41\pm8.14$ \\
w/o Zone B reset & $73.84\pm1.05$ & $-1.31\pm0.58$ & $7.69\pm10.10$ & $2.87\pm4.31$ \\
w/o Zone-C repair pass & $73.01\pm0.81$ & $-2.15\pm0.56$ & $0.00\pm0.00$ & $0.00\pm0.00$ \\
w/o cumulative freeze mask & $73.97\pm0.12$ & $-3.30\pm0.13$ & $0.00\pm0.00$ & $0.00\pm0.00$ \\
\textbf{LwU} & \textbf{73.65$\pm$1.13} & \textbf{-1.45$\pm$0.67} & \textbf{0.00$\pm$0.00} & \textbf{0.00$\pm$0.00} \\
\bottomrule
\end{tabular}
\end{adjustbox}
\end{table*}

\subsection{Unified U-LiRA Privacy Audit}\label{sec:new_unified_privacy}
The former summary table and ROC plots used incompatible instance-deletion audit settings. We therefore restrict comparable privacy results to one protocol: 16 retrained shadow models, 4,500 deleted images for instance deletion, and seed 42. AUC of 50\% is chance-level distinguishability; lower TPR at 1\% FPR is desirable. These privacy results are not pooled with forget-accuracy or legacy attack metrics to award overall medals. Class-deletion results below are from a separate standalone class-deletion evaluation. Under the unified audit LwU reduces leakage relative to Original in all eight reported instance/class cells, but does not consistently achieve Retrain-level privacy.

\begin{table*}[tp]
\centering
\setlength{\tabcolsep}{3pt}
\caption{Protocol-consistent standalone instance deletion U-LiRA (seed 42, 16 shadows, 4,500 forget images). AUC and TPR are percentages.}
\label{tab:new_privacy_instance}
\begin{adjustbox}{max width=\textwidth}
\begin{tabular}{lccc}
\toprule
Setting & Original AUC & LwU AUC & LwU TPR@1\% FPR \\
\midrule
CIFAR-10, CNN & $66.78$ & $62.52$ & $8.80$ \\
CIFAR-10, ResNet-18 & $80.08$ & $77.72$ & $20.29$ \\
CIFAR-100, CNN & $81.78$ & $76.48$ & $15.76$ \\
CIFAR-100, ResNet-18 & $93.05$ & $84.23$ & $38.11$ \\
\bottomrule
\end{tabular}
\end{adjustbox}
\end{table*}

\begin{table*}[tp]
\centering
\setlength{\tabcolsep}{3pt}
\caption{Standalone class deletion U-LiRA under the reported consistent class-deletion audit. AUC and TPR are percentages.}
\label{tab:new_privacy_class}
\begin{adjustbox}{max width=\textwidth}
\begin{tabular}{lccc}
\toprule
Setting & Original AUC & LwU AUC & LwU TPR@1\% FPR \\
\midrule
CIFAR-10, CNN & $100.00$ & $73.28$ & $31.96$ \\
CIFAR-10, ResNet-18 & $100.00$ & $53.38$ & $0.31$ \\
CIFAR-100, CNN & $100.00$ & $61.63$ & $4.67$ \\
CIFAR-100, ResNet-18 & $100.00$ & $60.64$ & $0.22$ \\
\bottomrule
\end{tabular}
\end{adjustbox}
\end{table*}

\subsection{Deletion, Repair, and Subsequent Learning}\label{sec:new_repair_trajectory}
We examine four ways to repair after the same destructive edit: (i) pre-edit masked teacher (standard LwU), (ii) oracle Retrain teacher as a diagnostic only, (iii) cross-entropy without a teacher, and (iv) edit-anchored repair. Stage-wise U-LiRA used 16 shadow streams per seed over three seeds. Where the source reports a single number or a pair without a standard deviation, we do not impute uncertainty. The precise edit-anchored optimization rule is not supplied with these numerical records and must be added before a reproducibility claim is made.

\begin{table*}[!tp]
\centering
\setlength{\tabcolsep}{3pt}
\caption{Class-deletion trajectory on five-task Split CIFAR-100 with class masking. End means after tasks 4--5; paired values follow after-repair/end order. Deleted and retained accuracies are percentages.}
\label{tab:new_class_repair}
\begin{adjustbox}{max width=\textwidth}
\begin{tabular}{lcccc}
\toprule
Repair & Stage & U-LiRA AUC & Deleted acc & Retained acc \\
\midrule
none & before request & $83.61\pm15.97$ & $70.19$ & $71.61$ \\
none & after edit & $62.09\pm6.07$ & $12.47$ & $69.51$ \\
\textbf{LwU (pre-edit teacher)} & after repair & \textbf{48.36$\pm$5.44} & $0.00$ & $73.14$ \\
\textbf{LwU (pre-edit teacher)} & after tasks 4--5 & \textbf{47.75$\pm$5.47} & $0.00$ & $72.57$ \\
Oracle teacher & after repair / end & $48.27 / 48.54$ & $0.00$ & $73.49 / 72.52$ \\
No teacher & after repair / end & $59.27 / 59.28$ & $0.00 / 0.51$ & $73.33 / 73.19$ \\
Edit-anchored repair & after repair / end & $56.60 / 50.86$ & $0.00$ & $71.64 / 71.61$ \\
\bottomrule
\end{tabular}
\end{adjustbox}
\end{table*}

With a class mask, standard LwU produces a U-LiRA AUC of $48.36\pm5.44$ immediately after repair and $47.75\pm5.47$ after subsequent tasks, with deleted-class accuracy zero at both stages. In this class-deletion setting, masked pre-edit repair does not show evidence of reinstating the deleted class. This does not extend automatically to instance or subclass deletion: their labels remain active, so no class can be masked.

\begin{table*}[!tp]
\centering
\setlength{\tabcolsep}{3pt}
\caption{Participant-deletion proxy: one fine class within a retained CIFAR-20 superclass; no class-logit mask. Stage pairs are after repair/end of task 5. No standard deviation is implied for values reported as pairs.}
\label{tab:new_participant_repair}
\begin{adjustbox}{max width=\textwidth}
\begin{tabular}{lcccc}
\toprule
Repair & Stage & U-LiRA AUC & Participant acc & Retained acc \\
\midrule
none & before request & $60.35\pm6.31$ & $75.37$ & $82.98$ \\
none & after edit & $48.06\pm4.89$ & $34.65$ & $78.82$ \\
LwU (pre-edit teacher) & after repair / end & $61.42 / 58.71$ & $75.66 / 73.76$ & $83.47 / 83.49$ \\
Oracle teacher & after repair / end & $60.65 / 60.91$ & $71.88 / 74.85$ & $83.45 / 83.52$ \\
No teacher & after repair / end & $60.87 / 57.40$ & $78.26 / 76.64$ & $82.99 / 83.66$ \\
\textbf{Edit-anchored repair} & after repair / end & \textbf{54.12 / 50.47} & $47.53 / 41.00$ & $82.52 / 83.33$ \\
\bottomrule
\end{tabular}
\end{adjustbox}
\end{table*}

\begin{table*}[!tp]
\centering
\setlength{\tabcolsep}{3pt}
\caption{Participant-proxy deletion after task 3, evaluated over ten seeds. Distance is the absolute difference between participant accuracy and the Retrain reference after task 5.}
\label{tab:new_participant_10seed}
\begin{adjustbox}{max width=\textwidth}
\begin{tabular}{lcccc}
\toprule
Method & ACC $\uparrow$ & BWT $\uparrow$ & Participant acc after task 5 & Distance to Retrain $\downarrow$ \\
\midrule
Retrain (never sees the participant) & $83.34\pm1.01$ & $-1.48$ & $56.39\pm29.14$ & -- \\
LwU (pre-edit teacher) & $83.91\pm0.62$ & $-1.38$ & $78.76\pm14.42$ & $22.4$ \\
\textbf{LwU, edit-anchored repair} & $83.12\pm1.38$ & $-2.34$ & $41.74\pm22.87$ & $14.7$ \\
LwU, edit only & $80.23\pm2.69$ & $-6.00$ & $29.45\pm26.26$ & $26.9$ \\
LwF + SalUn & $83.64\pm1.15$ & $-4.35$ & $46.58\pm19.94$ & $9.8$ \\
CLPU-DER++ & $77.71\pm1.63$ & $-12.10$ & $34.63\pm26.52$ & $21.8$ \\
\bottomrule
\end{tabular}
\end{adjustbox}
\end{table*}

Without a class mask, standard repair returns participant accuracy toward its pre-request value and U-LiRA AUC to about 60. The oracle teacher shows similar behavior, indicating that repair on retained inputs sharing the participant label also contributes. Edit-anchored repair reduces final AUC to 50.47 in the reported trajectory, but its 10-seed mean participant-accuracy distance to Retrain (14.7 points) remains larger than LwF+SalUn's (9.8 points). This proxy is not an RTI patient-level privacy demonstration.

For standalone instance and subclass deletion (seed 42, 16 shadows), the oracle teacher's AUC differs from the pre-edit teacher by at most two points in six of eight settings. On ResNet-18 instance deletion, the oracle produces a lower AUC by 8.3 points on CIFAR-10 (70.3 versus 78.6 in the recomputed analysis) and 4.1 points on CIFAR-100 (80.1 versus 84.2). A teacher-dependent contribution is thus measurable, but remains distinct from leakage caused by retained-data repair. No full uncertainty estimates were supplied for these eight comparisons.

\subsection{Additional CIFAR-100 Unlearning Ablations}\label{sec:new_cifar100_ablation}
The following experiments use three seeds. The reported Gap is the mean absolute difference from Retrain across Forget, Retain, and the conventional MIA metric; it is not U-LiRA AUC. Relearn@1 denotes forget-set accuracy after one epoch on the forget set and an equal number of retained examples. Class-deletion ablations are shown on both CNN and ResNet-18; an additional subclass experiment deletes the apple subclass on CNN.

\begin{table*}[tp]
\centering
\setlength{\tabcolsep}{3pt}
\caption{CIFAR-100 class-deletion component ablation on CNN; three seeds. The Gap and MIA columns use the evaluation in the supplied ablation records, not the U-LiRA protocol.}
\label{tab:new_c100_cnn_class}
\begin{adjustbox}{max width=\textwidth}
\begin{tabular}{lcccccc}
\toprule
Variant & Forget $\downarrow$ & Retain $\uparrow$ & Test $\uparrow$ & MIA & Gap $\downarrow$ & Relearn@1 $\downarrow$ \\
\midrule
Original & $83.00\pm3.61$ & $66.15\pm0.31$ & $66.32\pm0.28$ & $56.96\pm1.68$ & $29.86\pm1.17$ & $95.67\pm1.15$ \\
Retrain & $0.00\pm0.00$ & $66.82\pm0.47$ & $66.15\pm0.46$ & $51.05\pm2.46$ & $0.00$ & $0.00\pm0.00$ \\
w/o Zone B reset & $6.33\pm4.51$ & $66.54\pm0.20$ & $65.94\pm0.25$ & $55.48\pm3.27$ & $3.74\pm1.32$ & $96.00\pm1.73$ \\
w/o teacher repair & $20.67\pm26.10$ & $61.02\pm2.70$ & $60.62\pm2.44$ & $51.34\pm1.03$ & $9.64\pm9.69$ & $89.33\pm13.28$ \\
\textbf{LwU} & \textbf{0.00$\pm$0.00} & \textbf{66.29$\pm$0.17} & \textbf{65.63$\pm$0.17} & \textbf{48.60$\pm$4.06} & \textbf{1.38$\pm$0.56} & \textbf{0.00$\pm$0.00} \\
\bottomrule
\end{tabular}
\end{adjustbox}
\end{table*}

\begin{table*}[tp]
\centering
\setlength{\tabcolsep}{3pt}
\caption{CIFAR-100 class-deletion component ablation on ResNet-18; three seeds. Relearn@1 is after one epoch.}
\label{tab:new_c100_resnet_class}
\begin{adjustbox}{max width=\textwidth}
\begin{tabular}{lcccccc}
\toprule
Variant & Forget $\downarrow$ & Retain $\uparrow$ & Test $\uparrow$ & MIA & Gap $\downarrow$ & Relearn@1 $\downarrow$ \\
\midrule
Original & $85.67\pm3.06$ & $70.16\pm0.44$ & $70.31\pm0.42$ & $63.61\pm1.99$ & $33.60\pm1.47$ & $92.00\pm1.00$ \\
Retrain & $0.00\pm0.00$ & $70.23\pm0.48$ & $69.53\pm0.47$ & $48.61\pm0.90$ & $0.00$ & $0.00\pm0.00$ \\
w/o Zone B reset & $65.33\pm4.16$ & $70.30\pm0.41$ & $70.25\pm0.37$ & $63.38\pm3.87$ & $26.77\pm0.53$ & $94.33\pm1.53$ \\
w/o teacher repair & $1.00\pm1.73$ & $64.05\pm3.87$ & $63.42\pm3.85$ & $50.48\pm6.36$ & $3.80\pm0.83$ & $81.33\pm16.77$ \\
\textbf{LwU} & \textbf{0.00$\pm$0.00} & \textbf{69.68$\pm$0.73} & \textbf{68.99$\pm$0.72} & \textbf{50.86$\pm$2.33} & \textbf{0.96$\pm$0.76} & \textbf{0.00$\pm$0.00} \\
\bottomrule
\end{tabular}
\end{adjustbox}
\end{table*}

\begin{table*}[tp]
\centering
\setlength{\tabcolsep}{3pt}
\caption{CIFAR-100 apple subclass deletion on CNN; three seeds. Forget accuracy is compared with Retrain's nonzero target rather than minimized.}
\label{tab:new_c100_cnn_subclass}
\begin{adjustbox}{max width=\textwidth}
\begin{tabular}{lcccc}
\toprule
Variant & Forget & Retain $\uparrow$ & MIA & Gap $\downarrow$ \\
\midrule
Original & $95.00\pm1.00$ & $77.82\pm0.51$ & $51.25\pm5.44$ & $4.22\pm0.87$ \\
Retrain & $90.67\pm2.31$ & $78.04\pm0.57$ & $46.90\pm4.55$ & $0.00$ \\
w/o Zone B reset & $94.67\pm0.58$ & $77.94\pm0.49$ & $51.27\pm5.31$ & $4.07\pm1.05$ \\
w/o teacher repair & $32.00\pm55.43$ & $71.47\pm6.48$ & $48.98\pm3.31$ & $24.46\pm18.50$ \\
w/o Zone-C step & $92.67\pm3.06$ & $77.22\pm0.41$ & $47.28\pm1.51$ & $2.76\pm0.55$ \\
w/o C projection & $91.00\pm4.58$ & $77.10\pm0.40$ & $47.56\pm2.85$ & $2.74\pm1.08$ \\
w/o Zone-C repair pass & $87.33\pm9.07$ & $76.75\pm0.64$ & $46.18\pm1.28$ & $4.09\pm0.34$ \\
w/o test-time memory & $92.33\pm2.31$ & $77.12\pm0.43$ & $47.39\pm2.40$ & $2.60\pm0.79$ \\
+ weight alignment & $93.00\pm2.00$ & $77.08\pm0.34$ & $47.71\pm1.69$ & $2.68\pm0.36$ \\
\textbf{LwU} & \textbf{92.33$\pm$2.31} & \textbf{77.16$\pm$0.38} & \textbf{47.52$\pm$2.15} & \textbf{2.55$\pm$0.75} \\
\bottomrule
\end{tabular}
\end{adjustbox}
\end{table*}

\subsection{Test-Time Memory Latency}\label{sec:new_latency}
The latency benchmark measures batch-size-one ResNet-18 inference at $32\times32$ resolution on an NVIDIA RTX PRO 6000, with every example written to memory. Deletion runs outside the haptic control loop. Sync-free and compiled versions remove host--device synchronization overhead; the reported outputs agree within $10^{-7}$. This measurement does not establish end-to-end network or robotic-control-loop latency.

\twocolumn[{%
\section{Hyperparameters}\label{Hyperparmeters}
\begin{minipage}{\textwidth}
\centering
\renewcommand{\arraystretch}{0.95}
\captionof{table}{Unlearning hyperparameter grids, shared across datasets, backbones, and deletion modes (except LwU's $q_f$, see $^\ddagger$). $\#$ counts candidates; $\mathcal{D}_r^{10\%}$ is a fixed random 10\% subset of $\mathcal{D}_r$.}
\label{tab:mu-hparams}
\begin{tabularx}{\textwidth}{@{}l l >{\raggedright\arraybackslash}X r >{\raggedright\arraybackslash}X@{}}
\toprule
Method & Data & Searched grid & $\#$ & Fixed settings \\
\midrule
Retrain & $\mathcal{D}_r$ & -- & -- & Source training recipe \\
Amnesiac$^\dagger$ & update ledger & -- & -- & Subtracts updates recorded on batches containing $\mathcal{D}_f$ \\
Fine-tune & $\mathcal{D}_r$ & lr $\{10^{-4},10^{-3},10^{-2}\}$; epochs $\{1,3,5,10\}$ & 12 & Adam \\
Bad Teacher & $\mathcal{D}_f,\ \mathcal{D}_r^{10\%}$ & $T\ \{1,2,4\}$; epochs $\{1,3,5,10\}$; lr $\{10^{-4},10^{-3},10^{-2}\}$ & 36 & SGD, momentum 0.9; incompetent teacher is a re-initialized copy \\
UNSIR & $\mathcal{D}_f,\ \mathcal{D}_r^{10\%}$ & impair epochs $\{1,2\}$; repair epochs $\{1,2,3\}$; impair lr $\{10^{-3},10^{-2},10^{-1}\}$; repair lr $\{10^{-3},10^{-2}\}$ & 36 & Noise: 40 Adam steps, lr 0.1, regularization 0.1 \\
SSD & $\mathcal{D}_f$, Fisher on $\mathcal{D}$ & $\alpha\ \{1,5,10,50,100\}$; $\lambda\ \{0.1,0.5,1,5,10\}$ & 25 & No gradient updates \\
UniCLUN & $\mathcal{D}_f,\ \mathcal{D}_r^{10\%}$ & lr $\{10^{-4},10^{-3},10^{-2}\}$; epochs $\{3,5,10\}$; buffer $\{250,500,1000\}$ & 27 & SGD, momentum 0.9; $\rho=2$; contrastive $T=0.1$; $\alpha_1=\alpha_2=\alpha_3=1$; teacher momentum 0.99; $p=0.6$ \\
SalUn & $\mathcal{D}_f,\ \mathcal{D}_r^{10\%}$ & mask density $\{0.1,0.3,0.5,0.7,0.9\}$; lr $\{10^{-4},10^{-3},10^{-2}\}$; epochs $\{3,5,10\}$ & 45 & SGD, momentum 0.9, weight decay $5\times10^{-4}$ \\
SCRUB & $\mathcal{D}_f,\ \mathcal{D}_r^{10\%}$ & $E_{\max}\ \{1,2,3\}$; lr $\{10^{-4},5\times10^{-4},10^{-3},5\times10^{-3}\}$; $E\ \{5,10\}$ & 24 & SGD, momentum 0.9, weight decay $5\times10^{-4}$; $T=4$; $\alpha=\gamma=1$; forget-step gradient-norm clip 5.0 \\
LwU & $\mathcal{D}_f,\ \mathcal{D}_r^{10\%}$ & dominance margin $\{1,3\}$; distill.\ epochs $\{10,30\}$; distill.\ lr $\{10^{-3},10^{-2}\}$; weight alignment $\{$off, on$\}$; memory $(\eta_m,s_m)\ \{(10^{-2},0.25),(10^{-1},1)\}$; $q_f\ \{0.9,0.99\}^{\ddagger}$ & 64$^{\ddagger}$ & SGD distillation; Fisher importance on $\mathcal{D}_f$ and $\mathcal{D}_r^{10\%}$; retain quantile 0.9; zone C orthogonal; $T=4$; $\lambda_{\mathrm{sdft}}=0.5$; memory confidence 0.8 \\
\bottomrule
\end{tabularx}
\endgraf
\vspace{2pt}
{\footnotesize $^\dagger$Class deletion only; the ledger has no equivalent for subclass or instance deletion. $^\ddagger$Class deletion fixes $q_f=0.99$ (32 candidates).}
\end{minipage}\vspace{1.5em}}]

\begin{table*}[tp]
\centering
\caption{Training recipe of the source (Original) and Retrain models, which are not tuned. Amnesiac replays the per-batch update ledger recorded during source training (class deletion only).}
\label{tab:hp-recipe}
\begin{tabular}{@{}lccc@{}}
\toprule
 & CIFAR-10 & RTI & CIFAR-100 \\
\midrule
Max.\ epochs & 100 & 150 & 150 \\
Early-stopping patience & 10 & 15 & 15 \\
Min.\ epochs & 20 & 30 & 30 \\
Batch size & 64 & 64 & 64 \\
Learning rate (Adam) & $10^{-3}$ & $10^{-3}$ & $10^{-3}$ \\
LR schedule & reduce on plateau & reduce on plateau & reduce on plateau \\
$D_r$ fraction available to unlearning & 0.1 & 0.1 & 0.1 \\
Relearn probe epochs & 5 & 5 & 5 \\
Relearn probe lr & $10^{-4}$ & $10^{-4}$ & $10^{-4}$ \\
CNN hidden width & 64 & 256 & 512 \\
\bottomrule
\end{tabular}
\end{table*}

\begin{table*}[tp]
\centering
\caption{Most frequently selected value of each tuned hyperparameter over the 18 runs per dataset (2 backbones $\times$ 3 deletion types $\times$ 3 seeds), with its count. Tied values are separated by /.}
\label{tab:hp-mode}
\begin{tabular}{@{}llccc@{}}
\toprule
Method & Hyperparameter & CIFAR-10 & RTI & CIFAR-100 \\
\midrule
\multirow{2}{*}{Fine-tune} & lr & $10^{-3}$ (14) & $10^{-4}$ (18) & $10^{-4}$ (18) \\
 & epochs & 10 (16) & 10 (12) & 10 (11) \\
\midrule
\multirow{3}{*}{Bad Teacher} & $T$ & 4 (12) & 4 (15) & 4 (13) \\
 & epochs & 10 (9) & 10 (9) & 10 (10) \\
 & lr & $10^{-2}$ (14) & $10^{-2}$ (12) & $10^{-2}$ (10) \\
\midrule
\multirow{4}{*}{UNSIR} & impair epochs & 2 (10) & 2 (11) & 1 (12) \\
 & repair epochs & 2 (10) & 3 (11) & 3 (8) \\
 & impair lr & $10^{-2}$ (14) & $10^{-2}$ (10) & $10^{-3}$ (10) \\
 & repair lr & $10^{-2}$ (17) & $10^{-2}$ (17) & $10^{-2}$ (17) \\
\midrule
\multirow{2}{*}{SSD} & $\alpha$ & 1 (9) & 1 (7) & 50 / 5 (6) \\
 & $\lambda$ & 5 (7) & 0.1 (9) & 5 / 0.1 (6) \\
\midrule
\multirow{3}{*}{SCRUB} & max-step epochs & 3 (10) & 3 (12) & 3 (9) \\
 & lr & $5{\times}10^{-3}$ (9) & $5{\times}10^{-3}$ (7) & $5{\times}10^{-3}$ (14) \\
 & epochs & 5 (11) & 5 (14) & 5 (11) \\
\midrule
\multirow{3}{*}{SalUn} & sparsity & 0.1 (7) & 0.1 (6) & 0.9 (8) \\
 & lr & $10^{-4}$ (9) & $10^{-4}$ (12) & $10^{-4}$ (13) \\
 & epochs & 10 (9) & 3 (10) & 3 (12) \\
\midrule
\multirow{3}{*}{UniCLUN} & lr & $10^{-4}$ (11) & $10^{-4}$ (10) & $10^{-4}$ (11) \\
 & epochs & 3 (7) & 3 (9) & 3 (8) \\
 & buffer & 250 (18) & 250 (18) & 250 (18) \\
\midrule
\multirow{6}{*}{LwU} & $q_f$ & 0.99 (14) & 0.99 (15) & 0.99 (14) \\
 & margin & 3 (10) & 3 (12) & 3 (11) \\
 & distill.\ epochs & 30 (10) & 10 / 30 (9) & 10 (11) \\
 & distill.\ lr & $10^{-3}$ (11) & $10^{-3}$ (14) & $10^{-2}$ (10) \\
 & weight align. & off (10) & off (14) & off (16) \\
 & memory & \texttt{m10\_s100\_c08} (12) & \texttt{m01\_s025\_c08} (11) & \texttt{m01\_s025\_c08} (15) \\
\bottomrule
\end{tabular}
\end{table*}

\clearpage
\onecolumn

{\setlength{\tabcolsep}{3pt}
\begin{longtable}{@{}lll>{\raggedright\arraybackslash}p{0.235\textwidth}>{\raggedright\arraybackslash}p{0.235\textwidth}>{\raggedright\arraybackslash}p{0.235\textwidth}@{}}
\caption{Selected hyperparameters on CIFAR-10.}\label{tab:hp-selected-cifar10}\\
\toprule
Backbone & Deletion & Method & Seed 42 & Seed 43 & Seed 44 \\
\midrule
\endfirsthead
\toprule
Backbone & Deletion & Method & Seed 42 & Seed 43 & Seed 44 \\
\midrule
\endhead
\bottomrule
\endfoot
CNN & class & Fine-tune & lr\,=\,$10^{-4}$, ep\,=\,10 & lr\,=\,$10^{-4}$, ep\,=\,10 & lr\,=\,$10^{-3}$, ep\,=\,10 \\
 &  & Bad Teacher & $T$\,=\,4, ep\,=\,3, lr\,=\,$10^{-2}$ & $T$\,=\,4, ep\,=\,10, lr\,=\,$10^{-3}$ & $T$\,=\,4, ep\,=\,5, lr\,=\,$10^{-3}$ \\
 &  & UNSIR & imp.\,ep\,=\,2, rep.\,ep\,=\,2, imp.\,lr\,=\,$10^{-2}$, rep.\,lr\,=\,$10^{-2}$ & imp.\,ep\,=\,2, rep.\,ep\,=\,3, imp.\,lr\,=\,$10^{-2}$, rep.\,lr\,=\,$10^{-2}$ & imp.\,ep\,=\,2, rep.\,ep\,=\,3, imp.\,lr\,=\,$10^{-2}$, rep.\,lr\,=\,$10^{-2}$ \\
 &  & SSD & $\alpha$\,=\,5, $\lambda$\,=\,0.1 & $\alpha$\,=\,1, $\lambda$\,=\,10 & $\alpha$\,=\,1, $\lambda$\,=\,5 \\
 &  & SCRUB & max\,=\,2, lr\,=\,$5{\times}10^{-4}$, ep\,=\,5 & max\,=\,3, lr\,=\,$5{\times}10^{-4}$, ep\,=\,5 & max\,=\,3, lr\,=\,$5{\times}10^{-4}$, ep\,=\,5 \\
 &  & SalUn & sp\,=\,0.1, lr\,=\,$10^{-4}$, ep\,=\,3 & sp\,=\,0.5, lr\,=\,$10^{-4}$, ep\,=\,10 & sp\,=\,0.9, lr\,=\,$10^{-3}$, ep\,=\,10 \\
 &  & UniCLUN & lr\,=\,$10^{-4}$, ep\,=\,3, buf\,=\,250 & lr\,=\,$10^{-4}$, ep\,=\,3, buf\,=\,250 & lr\,=\,$10^{-2}$, ep\,=\,5, buf\,=\,250 \\
 &  & LwU & margin\,=\,3, d.\,ep\,=\,30, d.\,lr\,=\,$10^{-2}$, align\,=\,off, mem\,=\,\texttt{m10\_s100\_c08} & margin\,=\,3, d.\,ep\,=\,10, d.\,lr\,=\,$10^{-3}$, align\,=\,off, mem\,=\,\texttt{m10\_s100\_c08} & margin\,=\,3, d.\,ep\,=\,30, d.\,lr\,=\,$10^{-3}$, align\,=\,off, mem\,=\,\texttt{m10\_s100\_c08} \\
\midrule
 & subclass & Fine-tune & lr\,=\,$10^{-3}$, ep\,=\,3 & lr\,=\,$10^{-3}$, ep\,=\,10 & lr\,=\,$10^{-3}$, ep\,=\,10 \\
 &  & Bad Teacher & $T$\,=\,4, ep\,=\,10, lr\,=\,$10^{-2}$ & $T$\,=\,4, ep\,=\,5, lr\,=\,$10^{-2}$ & $T$\,=\,2, ep\,=\,10, lr\,=\,$10^{-2}$ \\
 &  & UNSIR & imp.\,ep\,=\,1, rep.\,ep\,=\,3, imp.\,lr\,=\,0.1, rep.\,lr\,=\,$10^{-2}$ & imp.\,ep\,=\,1, rep.\,ep\,=\,2, imp.\,lr\,=\,$10^{-3}$, rep.\,lr\,=\,$10^{-3}$ & imp.\,ep\,=\,1, rep.\,ep\,=\,2, imp.\,lr\,=\,$10^{-2}$, rep.\,lr\,=\,$10^{-2}$ \\
 &  & SSD & $\alpha$\,=\,10, $\lambda$\,=\,0.5 & $\alpha$\,=\,1, $\lambda$\,=\,5 & $\alpha$\,=\,5, $\lambda$\,=\,0.1 \\
 &  & SCRUB & max\,=\,3, lr\,=\,$10^{-4}$, ep\,=\,5 & max\,=\,3, lr\,=\,$10^{-4}$, ep\,=\,5 & max\,=\,3, lr\,=\,$5{\times}10^{-4}$, ep\,=\,5 \\
 &  & SalUn & sp\,=\,0.1, lr\,=\,$10^{-4}$, ep\,=\,3 & sp\,=\,0.1, lr\,=\,$10^{-4}$, ep\,=\,3 & sp\,=\,0.1, lr\,=\,$10^{-4}$, ep\,=\,3 \\
 &  & UniCLUN & lr\,=\,$10^{-4}$, ep\,=\,5, buf\,=\,250 & lr\,=\,$10^{-4}$, ep\,=\,5, buf\,=\,250 & lr\,=\,$10^{-4}$, ep\,=\,3, buf\,=\,250 \\
 &  & LwU & $q_f$\,=\,0.9, margin\,=\,3, d.\,ep\,=\,30, d.\,lr\,=\,$10^{-2}$, align\,=\,on, mem\,=\,\texttt{m01\_s025\_c08} & $q_f$\,=\,0.99, margin\,=\,1, d.\,ep\,=\,10, d.\,lr\,=\,$10^{-3}$, align\,=\,off, mem\,=\,\texttt{m01\_s025\_c08} & $q_f$\,=\,0.99, margin\,=\,3, d.\,ep\,=\,30, d.\,lr\,=\,$10^{-3}$, align\,=\,off, mem\,=\,\texttt{m01\_s025\_c08} \\
\midrule
 & instance & Fine-tune & lr\,=\,$10^{-3}$, ep\,=\,10 & lr\,=\,$10^{-3}$, ep\,=\,5 & lr\,=\,$10^{-3}$, ep\,=\,10 \\
 &  & Bad Teacher & $T$\,=\,2, ep\,=\,10, lr\,=\,$10^{-2}$ & $T$\,=\,2, ep\,=\,5, lr\,=\,$10^{-2}$ & $T$\,=\,1, ep\,=\,5, lr\,=\,$10^{-4}$ \\
 &  & UNSIR & imp.\,ep\,=\,2, rep.\,ep\,=\,1, imp.\,lr\,=\,$10^{-2}$, rep.\,lr\,=\,$10^{-2}$ & imp.\,ep\,=\,1, rep.\,ep\,=\,2, imp.\,lr\,=\,$10^{-2}$, rep.\,lr\,=\,$10^{-2}$ & imp.\,ep\,=\,2, rep.\,ep\,=\,2, imp.\,lr\,=\,$10^{-2}$, rep.\,lr\,=\,$10^{-2}$ \\
 &  & SSD & $\alpha$\,=\,5, $\lambda$\,=\,1 & $\alpha$\,=\,1, $\lambda$\,=\,5 & $\alpha$\,=\,1, $\lambda$\,=\,5 \\
 &  & SCRUB & max\,=\,3, lr\,=\,$5{\times}10^{-3}$, ep\,=\,10 & max\,=\,2, lr\,=\,$5{\times}10^{-3}$, ep\,=\,10 & max\,=\,2, lr\,=\,$5{\times}10^{-3}$, ep\,=\,10 \\
 &  & SalUn & sp\,=\,0.9, lr\,=\,$10^{-2}$, ep\,=\,10 $\times$ & sp\,=\,0.7, lr\,=\,$10^{-2}$, ep\,=\,10 $\times$ & sp\,=\,0.1, lr\,=\,$10^{-2}$, ep\,=\,10 \\
 &  & UniCLUN & lr\,=\,$10^{-4}$, ep\,=\,3, buf\,=\,250 & lr\,=\,$10^{-2}$, ep\,=\,10, buf\,=\,250 & lr\,=\,$10^{-2}$, ep\,=\,5, buf\,=\,250 \\
 &  & LwU & $q_f$\,=\,0.99, margin\,=\,1, d.\,ep\,=\,10, d.\,lr\,=\,$10^{-3}$, align\,=\,off, mem\,=\,\texttt{m10\_s100\_c08} & $q_f$\,=\,0.9, margin\,=\,3, d.\,ep\,=\,30, d.\,lr\,=\,$10^{-2}$, align\,=\,on, mem\,=\,\texttt{m10\_s100\_c08} & $q_f$\,=\,0.9, margin\,=\,3, d.\,ep\,=\,30, d.\,lr\,=\,$10^{-3}$, align\,=\,on, mem\,=\,\texttt{m10\_s100\_c08} \\
\midrule
ResNet-18 & class & Fine-tune & lr\,=\,$10^{-4}$, ep\,=\,10 & lr\,=\,$10^{-3}$, ep\,=\,10 & lr\,=\,$10^{-3}$, ep\,=\,10 \\
 &  & Bad Teacher & $T$\,=\,4, ep\,=\,10, lr\,=\,$10^{-2}$ & $T$\,=\,4, ep\,=\,10, lr\,=\,$10^{-2}$ & $T$\,=\,4, ep\,=\,10, lr\,=\,$10^{-2}$ \\
 &  & UNSIR & imp.\,ep\,=\,1, rep.\,ep\,=\,3, imp.\,lr\,=\,$10^{-2}$, rep.\,lr\,=\,$10^{-2}$ & imp.\,ep\,=\,2, rep.\,ep\,=\,2, imp.\,lr\,=\,$10^{-2}$, rep.\,lr\,=\,$10^{-2}$ & imp.\,ep\,=\,1, rep.\,ep\,=\,2, imp.\,lr\,=\,$10^{-2}$, rep.\,lr\,=\,$10^{-2}$ \\
 &  & SSD & $\alpha$\,=\,100, $\lambda$\,=\,0.1 & $\alpha$\,=\,5, $\lambda$\,=\,0.5 & $\alpha$\,=\,10, $\lambda$\,=\,0.1 \\
 &  & SCRUB & max\,=\,3, lr\,=\,$5{\times}10^{-4}$, ep\,=\,10 & max\,=\,2, lr\,=\,$5{\times}10^{-4}$, ep\,=\,5 & max\,=\,2, lr\,=\,$5{\times}10^{-4}$, ep\,=\,5 \\
 &  & SalUn & sp\,=\,0.1, lr\,=\,$10^{-4}$, ep\,=\,3 & sp\,=\,0.7, lr\,=\,$10^{-2}$, ep\,=\,10 & sp\,=\,0.7, lr\,=\,$10^{-2}$, ep\,=\,10 \\
 &  & UniCLUN & lr\,=\,$10^{-4}$, ep\,=\,3, buf\,=\,250 $\times$ & lr\,=\,$10^{-4}$, ep\,=\,10, buf\,=\,250 & lr\,=\,$10^{-2}$, ep\,=\,10, buf\,=\,250 \\
 &  & LwU & margin\,=\,1, d.\,ep\,=\,30, d.\,lr\,=\,$10^{-2}$, align\,=\,off, mem\,=\,\texttt{m10\_s100\_c08} & margin\,=\,1, d.\,ep\,=\,30, d.\,lr\,=\,$10^{-2}$, align\,=\,on, mem\,=\,\texttt{m10\_s100\_c08} & margin\,=\,3, d.\,ep\,=\,30, d.\,lr\,=\,$10^{-2}$, align\,=\,off, mem\,=\,\texttt{m10\_s100\_c08} \\
\midrule
 & subclass & Fine-tune & lr\,=\,$10^{-3}$, ep\,=\,10 & lr\,=\,$10^{-3}$, ep\,=\,10 & lr\,=\,$10^{-3}$, ep\,=\,10 \\
 &  & Bad Teacher & $T$\,=\,4, ep\,=\,10, lr\,=\,$10^{-2}$ & $T$\,=\,1, ep\,=\,5, lr\,=\,$10^{-2}$ & $T$\,=\,4, ep\,=\,10, lr\,=\,$10^{-2}$ \\
 &  & UNSIR & imp.\,ep\,=\,2, rep.\,ep\,=\,2, imp.\,lr\,=\,$10^{-2}$, rep.\,lr\,=\,$10^{-2}$ & imp.\,ep\,=\,1, rep.\,ep\,=\,2, imp.\,lr\,=\,$10^{-2}$, rep.\,lr\,=\,$10^{-2}$ & imp.\,ep\,=\,2, rep.\,ep\,=\,1, imp.\,lr\,=\,$10^{-2}$, rep.\,lr\,=\,$10^{-2}$ \\
 &  & SSD & $\alpha$\,=\,10, $\lambda$\,=\,0.1 & $\alpha$\,=\,1, $\lambda$\,=\,0.5 & $\alpha$\,=\,10, $\lambda$\,=\,5 \\
 &  & SCRUB & max\,=\,3, lr\,=\,$5{\times}10^{-3}$, ep\,=\,10 & max\,=\,3, lr\,=\,$5{\times}10^{-3}$, ep\,=\,5 & max\,=\,3, lr\,=\,$5{\times}10^{-3}$, ep\,=\,10 \\
 &  & SalUn & sp\,=\,0.7, lr\,=\,$10^{-2}$, ep\,=\,10 & sp\,=\,0.7, lr\,=\,$10^{-2}$, ep\,=\,10 & sp\,=\,0.3, lr\,=\,$10^{-4}$, ep\,=\,5 \\
 &  & UniCLUN & lr\,=\,$10^{-2}$, ep\,=\,5, buf\,=\,250 & lr\,=\,$10^{-2}$, ep\,=\,10, buf\,=\,250 & lr\,=\,$10^{-4}$, ep\,=\,5, buf\,=\,250 \\
 &  & LwU & $q_f$\,=\,0.9, margin\,=\,3, d.\,ep\,=\,30, d.\,lr\,=\,$10^{-2}$, align\,=\,on, mem\,=\,\texttt{m01\_s025\_c08} & $q_f$\,=\,0.99, margin\,=\,1, d.\,ep\,=\,10, d.\,lr\,=\,$10^{-3}$, align\,=\,off, mem\,=\,\texttt{m01\_s025\_c08} & $q_f$\,=\,0.99, margin\,=\,1, d.\,ep\,=\,10, d.\,lr\,=\,$10^{-3}$, align\,=\,on, mem\,=\,\texttt{m01\_s025\_c08} \\
\midrule
 & instance & Fine-tune & lr\,=\,$10^{-3}$, ep\,=\,10 & lr\,=\,$10^{-4}$, ep\,=\,10 & lr\,=\,$10^{-3}$, ep\,=\,10 \\
 &  & Bad Teacher & $T$\,=\,4, ep\,=\,5, lr\,=\,$10^{-2}$ & $T$\,=\,4, ep\,=\,5, lr\,=\,$10^{-2}$ & $T$\,=\,2, ep\,=\,3, lr\,=\,$10^{-4}$ \\
 &  & UNSIR & imp.\,ep\,=\,1, rep.\,ep\,=\,2, imp.\,lr\,=\,$10^{-2}$, rep.\,lr\,=\,$10^{-2}$ & imp.\,ep\,=\,2, rep.\,ep\,=\,1, imp.\,lr\,=\,$10^{-3}$, rep.\,lr\,=\,$10^{-2}$ & imp.\,ep\,=\,2, rep.\,ep\,=\,1, imp.\,lr\,=\,$10^{-3}$, rep.\,lr\,=\,$10^{-2}$ \\
 &  & SSD & $\alpha$\,=\,1, $\lambda$\,=\,10 & $\alpha$\,=\,1, $\lambda$\,=\,5 & $\alpha$\,=\,1, $\lambda$\,=\,5 \\
 &  & SCRUB & max\,=\,2, lr\,=\,$5{\times}10^{-3}$, ep\,=\,5 & max\,=\,2, lr\,=\,$5{\times}10^{-3}$, ep\,=\,5 & max\,=\,2, lr\,=\,$5{\times}10^{-3}$, ep\,=\,10 \\
 &  & SalUn & sp\,=\,0.7, lr\,=\,$10^{-2}$, ep\,=\,5 $\times$ & sp\,=\,0.3, lr\,=\,$10^{-4}$, ep\,=\,3 $\times$ & sp\,=\,0.1, lr\,=\,$10^{-4}$, ep\,=\,3 $\times$ \\
 &  & UniCLUN & lr\,=\,$10^{-2}$, ep\,=\,10, buf\,=\,250 $\times$ & lr\,=\,$10^{-4}$, ep\,=\,3, buf\,=\,250 & lr\,=\,$10^{-4}$, ep\,=\,3, buf\,=\,250 $\times$ \\
 &  & LwU & $q_f$\,=\,0.99, margin\,=\,1, d.\,ep\,=\,10, d.\,lr\,=\,$10^{-3}$, align\,=\,on, mem\,=\,\texttt{m10\_s100\_c08} & $q_f$\,=\,0.99, margin\,=\,3, d.\,ep\,=\,10, d.\,lr\,=\,$10^{-3}$, align\,=\,on, mem\,=\,\texttt{m10\_s100\_c08} & $q_f$\,=\,0.99, margin\,=\,1, d.\,ep\,=\,10, d.\,lr\,=\,$10^{-3}$, align\,=\,off, mem\,=\,\texttt{m10\_s100\_c08} \\
\end{longtable}
}

{\setlength{\tabcolsep}{3pt}
\begin{longtable}{@{}lll>{\raggedright\arraybackslash}p{0.235\textwidth}>{\raggedright\arraybackslash}p{0.235\textwidth}>{\raggedright\arraybackslash}p{0.235\textwidth}@{}}
\caption{Selected hyperparameters on RTI.}\label{tab:hp-selected-cifar20}\\
\toprule
Backbone & Deletion & Method & Seed 42 & Seed 43 & Seed 44 \\
\midrule
\endfirsthead
\toprule
Backbone & Deletion & Method & Seed 42 & Seed 43 & Seed 44 \\
\midrule
\endhead
\bottomrule
\endfoot
CNN & class & Fine-tune & lr\,=\,$10^{-4}$, ep\,=\,10 & lr\,=\,$10^{-4}$, ep\,=\,10 & lr\,=\,$10^{-4}$, ep\,=\,10 \\
 &  & Bad Teacher & $T$\,=\,4, ep\,=\,10, lr\,=\,$10^{-2}$ & $T$\,=\,4, ep\,=\,3, lr\,=\,$10^{-2}$ & $T$\,=\,2, ep\,=\,10, lr\,=\,$10^{-2}$ \\
 &  & UNSIR & imp.\,ep\,=\,1, rep.\,ep\,=\,3, imp.\,lr\,=\,$10^{-3}$, rep.\,lr\,=\,$10^{-2}$ & imp.\,ep\,=\,1, rep.\,ep\,=\,3, imp.\,lr\,=\,$10^{-2}$, rep.\,lr\,=\,$10^{-2}$ & imp.\,ep\,=\,1, rep.\,ep\,=\,3, imp.\,lr\,=\,$10^{-2}$, rep.\,lr\,=\,$10^{-2}$ \\
 &  & SSD & $\alpha$\,=\,50, $\lambda$\,=\,0.1 & $\alpha$\,=\,50, $\lambda$\,=\,0.1 & $\alpha$\,=\,10, $\lambda$\,=\,0.5 \\
 &  & SCRUB & max\,=\,3, lr\,=\,$10^{-4}$, ep\,=\,5 & max\,=\,3, lr\,=\,$5{\times}10^{-4}$, ep\,=\,5 & max\,=\,3, lr\,=\,$10^{-4}$, ep\,=\,5 \\
 &  & SalUn & sp\,=\,0.9, lr\,=\,$10^{-4}$, ep\,=\,3 $\times$ & sp\,=\,0.1, lr\,=\,$10^{-4}$, ep\,=\,5 $\times$ & sp\,=\,0.9, lr\,=\,$10^{-4}$, ep\,=\,3 $\times$ \\
 &  & UniCLUN & lr\,=\,$10^{-4}$, ep\,=\,5, buf\,=\,250 $\times$ & lr\,=\,$10^{-4}$, ep\,=\,3, buf\,=\,250 $\times$ & lr\,=\,$10^{-4}$, ep\,=\,3, buf\,=\,250 $\times$ \\
 &  & LwU & margin\,=\,3, d.\,ep\,=\,10, d.\,lr\,=\,$10^{-3}$, align\,=\,off, mem\,=\,\texttt{m01\_s025\_c08} & margin\,=\,3, d.\,ep\,=\,10, d.\,lr\,=\,$10^{-3}$, align\,=\,off, mem\,=\,\texttt{m10\_s100\_c08} & margin\,=\,3, d.\,ep\,=\,30, d.\,lr\,=\,$10^{-2}$, align\,=\,on, mem\,=\,\texttt{m01\_s025\_c08} \\
\midrule
 & subclass & Fine-tune & lr\,=\,$10^{-4}$, ep\,=\,10 & lr\,=\,$10^{-4}$, ep\,=\,5 & lr\,=\,$10^{-4}$, ep\,=\,3 \\
 &  & Bad Teacher & $T$\,=\,4, ep\,=\,3, lr\,=\,$10^{-2}$ & $T$\,=\,4, ep\,=\,10, lr\,=\,$10^{-2}$ $\times$ & $T$\,=\,4, ep\,=\,3, lr\,=\,$10^{-2}$ \\
 &  & UNSIR & imp.\,ep\,=\,2, rep.\,ep\,=\,1, imp.\,lr\,=\,$10^{-2}$, rep.\,lr\,=\,$10^{-2}$ & imp.\,ep\,=\,1, rep.\,ep\,=\,3, imp.\,lr\,=\,$10^{-3}$, rep.\,lr\,=\,$10^{-3}$ & imp.\,ep\,=\,2, rep.\,ep\,=\,1, imp.\,lr\,=\,$10^{-3}$, rep.\,lr\,=\,$10^{-2}$ \\
 &  & SSD & $\alpha$\,=\,1, $\lambda$\,=\,1 & $\alpha$\,=\,1, $\lambda$\,=\,10 & $\alpha$\,=\,5, $\lambda$\,=\,0.5 \\
 &  & SCRUB & max\,=\,2, lr\,=\,$5{\times}10^{-3}$, ep\,=\,10 & max\,=\,1, lr\,=\,$10^{-4}$, ep\,=\,5 & max\,=\,3, lr\,=\,$5{\times}10^{-4}$, ep\,=\,10 \\
 &  & SalUn & sp\,=\,0.1, lr\,=\,$10^{-4}$, ep\,=\,5 $\times$ & sp\,=\,0.1, lr\,=\,$10^{-4}$, ep\,=\,3 $\times$ & sp\,=\,0.1, lr\,=\,$10^{-4}$, ep\,=\,3 $\times$ \\
 &  & UniCLUN & lr\,=\,$10^{-2}$, ep\,=\,10, buf\,=\,250 $\times$ & lr\,=\,$10^{-3}$, ep\,=\,10, buf\,=\,250 $\times$ & lr\,=\,$10^{-4}$, ep\,=\,3, buf\,=\,250 $\times$ \\
 &  & LwU & $q_f$\,=\,0.99, margin\,=\,1, d.\,ep\,=\,30, d.\,lr\,=\,$10^{-3}$, align\,=\,off, mem\,=\,\texttt{m01\_s025\_c08} & $q_f$\,=\,0.99, margin\,=\,3, d.\,ep\,=\,10, d.\,lr\,=\,$10^{-3}$, align\,=\,on, mem\,=\,\texttt{m10\_s100\_c08} & $q_f$\,=\,0.99, margin\,=\,3, d.\,ep\,=\,10, d.\,lr\,=\,$10^{-3}$, align\,=\,on, mem\,=\,\texttt{m01\_s025\_c08} \\
\midrule
 & instance & Fine-tune & lr\,=\,$10^{-4}$, ep\,=\,10 & lr\,=\,$10^{-4}$, ep\,=\,10 & lr\,=\,$10^{-4}$, ep\,=\,10 \\
 &  & Bad Teacher & $T$\,=\,4, ep\,=\,10, lr\,=\,$10^{-3}$ & $T$\,=\,4, ep\,=\,5, lr\,=\,$10^{-3}$ & $T$\,=\,4, ep\,=\,1, lr\,=\,$10^{-4}$ \\
 &  & UNSIR & imp.\,ep\,=\,1, rep.\,ep\,=\,3, imp.\,lr\,=\,$10^{-2}$, rep.\,lr\,=\,$10^{-2}$ & imp.\,ep\,=\,2, rep.\,ep\,=\,2, imp.\,lr\,=\,$10^{-2}$, rep.\,lr\,=\,$10^{-2}$ & imp.\,ep\,=\,2, rep.\,ep\,=\,2, imp.\,lr\,=\,$10^{-2}$, rep.\,lr\,=\,$10^{-2}$ \\
 &  & SSD & $\alpha$\,=\,10, $\lambda$\,=\,0.1 & $\alpha$\,=\,5, $\lambda$\,=\,0.1 & $\alpha$\,=\,5, $\lambda$\,=\,0.5 \\
 &  & SCRUB & max\,=\,3, lr\,=\,$10^{-3}$, ep\,=\,5 & max\,=\,2, lr\,=\,$5{\times}10^{-3}$, ep\,=\,10 & max\,=\,1, lr\,=\,$5{\times}10^{-3}$, ep\,=\,5 \\
 &  & SalUn & sp\,=\,0.5, lr\,=\,$10^{-2}$, ep\,=\,10 $\times$ & sp\,=\,0.7, lr\,=\,$10^{-2}$, ep\,=\,10 $\times$ & sp\,=\,0.7, lr\,=\,$10^{-2}$, ep\,=\,10 $\times$ \\
 &  & UniCLUN & lr\,=\,$10^{-2}$, ep\,=\,10, buf\,=\,250 $\times$ & lr\,=\,$10^{-2}$, ep\,=\,10, buf\,=\,250 $\times$ & lr\,=\,$10^{-2}$, ep\,=\,10, buf\,=\,250 $\times$ \\
 &  & LwU & $q_f$\,=\,0.9, margin\,=\,3, d.\,ep\,=\,10, d.\,lr\,=\,$10^{-3}$, align\,=\,off, mem\,=\,\texttt{m10\_s100\_c08} & $q_f$\,=\,0.9, margin\,=\,3, d.\,ep\,=\,10, d.\,lr\,=\,$10^{-2}$, align\,=\,on, mem\,=\,\texttt{m10\_s100\_c08} & $q_f$\,=\,0.9, margin\,=\,3, d.\,ep\,=\,30, d.\,lr\,=\,$10^{-3}$, align\,=\,off, mem\,=\,\texttt{m10\_s100\_c08} \\
\midrule
ResNet-18 & class & Fine-tune & lr\,=\,$10^{-4}$, ep\,=\,10 & lr\,=\,$10^{-4}$, ep\,=\,10 & lr\,=\,$10^{-4}$, ep\,=\,10 \\
 &  & Bad Teacher & $T$\,=\,4, ep\,=\,10, lr\,=\,$10^{-2}$ & $T$\,=\,4, ep\,=\,10, lr\,=\,$10^{-2}$ & $T$\,=\,4, ep\,=\,10, lr\,=\,$10^{-2}$ \\
 &  & UNSIR & imp.\,ep\,=\,1, rep.\,ep\,=\,3, imp.\,lr\,=\,$10^{-3}$, rep.\,lr\,=\,$10^{-2}$ & imp.\,ep\,=\,2, rep.\,ep\,=\,3, imp.\,lr\,=\,$10^{-3}$, rep.\,lr\,=\,$10^{-2}$ & imp.\,ep\,=\,2, rep.\,ep\,=\,3, imp.\,lr\,=\,$10^{-2}$, rep.\,lr\,=\,$10^{-2}$ \\
 &  & SSD & $\alpha$\,=\,10, $\lambda$\,=\,0.1 & $\alpha$\,=\,100, $\lambda$\,=\,0.1 & $\alpha$\,=\,1, $\lambda$\,=\,10 \\
 &  & SCRUB & max\,=\,3, lr\,=\,$10^{-3}$, ep\,=\,5 & max\,=\,3, lr\,=\,$10^{-3}$, ep\,=\,5 & max\,=\,3, lr\,=\,$10^{-3}$, ep\,=\,5 \\
 &  & SalUn & sp\,=\,0.1, lr\,=\,$10^{-4}$, ep\,=\,3 $\times$ & sp\,=\,0.7, lr\,=\,$10^{-4}$, ep\,=\,3 $\times$ & sp\,=\,0.5, lr\,=\,$10^{-4}$, ep\,=\,3 $\times$ \\
 &  & UniCLUN & lr\,=\,$10^{-4}$, ep\,=\,3, buf\,=\,250 & lr\,=\,$10^{-4}$, ep\,=\,3, buf\,=\,250 $\times$ & lr\,=\,$10^{-4}$, ep\,=\,3, buf\,=\,250 $\times$ \\
 &  & LwU & margin\,=\,3, d.\,ep\,=\,30, d.\,lr\,=\,$10^{-2}$, align\,=\,off, mem\,=\,\texttt{m01\_s025\_c08} & margin\,=\,3, d.\,ep\,=\,30, d.\,lr\,=\,$10^{-2}$, align\,=\,off, mem\,=\,\texttt{m01\_s025\_c08} & margin\,=\,3, d.\,ep\,=\,30, d.\,lr\,=\,$10^{-3}$, align\,=\,off, mem\,=\,\texttt{m01\_s025\_c08} \\
\midrule
 & subclass & Fine-tune & lr\,=\,$10^{-4}$, ep\,=\,3 & lr\,=\,$10^{-4}$, ep\,=\,1 & lr\,=\,$10^{-4}$, ep\,=\,5 \\
 &  & Bad Teacher & $T$\,=\,4, ep\,=\,5, lr\,=\,$10^{-2}$ $\times$ & $T$\,=\,4, ep\,=\,10, lr\,=\,$10^{-2}$ $\times$ & $T$\,=\,4, ep\,=\,10, lr\,=\,$10^{-2}$ $\times$ \\
 &  & UNSIR & imp.\,ep\,=\,2, rep.\,ep\,=\,3, imp.\,lr\,=\,$10^{-3}$, rep.\,lr\,=\,$10^{-2}$ $\times$ & imp.\,ep\,=\,2, rep.\,ep\,=\,2, imp.\,lr\,=\,$10^{-3}$, rep.\,lr\,=\,$10^{-2}$ & imp.\,ep\,=\,2, rep.\,ep\,=\,2, imp.\,lr\,=\,$10^{-2}$, rep.\,lr\,=\,$10^{-2}$ \\
 &  & SSD & $\alpha$\,=\,1, $\lambda$\,=\,5 & $\alpha$\,=\,1, $\lambda$\,=\,1 & $\alpha$\,=\,1, $\lambda$\,=\,0.1 \\
 &  & SCRUB & max\,=\,3, lr\,=\,$5{\times}10^{-3}$, ep\,=\,5 & max\,=\,3, lr\,=\,$5{\times}10^{-4}$, ep\,=\,5 & max\,=\,2, lr\,=\,$10^{-4}$, ep\,=\,10 \\
 &  & SalUn & sp\,=\,0.3, lr\,=\,$10^{-4}$, ep\,=\,3 $\times$ & sp\,=\,0.1, lr\,=\,$10^{-4}$, ep\,=\,3 $\times$ & sp\,=\,0.3, lr\,=\,$10^{-4}$, ep\,=\,3 $\times$ \\
 &  & UniCLUN & lr\,=\,$10^{-4}$, ep\,=\,3, buf\,=\,250 $\times$ & lr\,=\,$10^{-4}$, ep\,=\,3, buf\,=\,250 $\times$ & lr\,=\,$10^{-4}$, ep\,=\,3, buf\,=\,250 $\times$ \\
 &  & LwU & $q_f$\,=\,0.99, margin\,=\,3, d.\,ep\,=\,30, d.\,lr\,=\,$10^{-3}$, align\,=\,off, mem\,=\,\texttt{m10\_s100\_c08} & $q_f$\,=\,0.99, margin\,=\,1, d.\,ep\,=\,10, d.\,lr\,=\,$10^{-3}$, align\,=\,off, mem\,=\,\texttt{m10\_s100\_c08} & $q_f$\,=\,0.99, margin\,=\,1, d.\,ep\,=\,30, d.\,lr\,=\,$10^{-3}$, align\,=\,off, mem\,=\,\texttt{m01\_s025\_c08} \\
\midrule
 & instance & Fine-tune & lr\,=\,$10^{-4}$, ep\,=\,10 & lr\,=\,$10^{-4}$, ep\,=\,5 & lr\,=\,$10^{-4}$, ep\,=\,10 \\
 &  & Bad Teacher & $T$\,=\,4, ep\,=\,1, lr\,=\,$10^{-4}$ & $T$\,=\,2, ep\,=\,1, lr\,=\,$10^{-4}$ & $T$\,=\,1, ep\,=\,3, lr\,=\,$10^{-4}$ \\
 &  & UNSIR & imp.\,ep\,=\,2, rep.\,ep\,=\,1, imp.\,lr\,=\,$10^{-2}$, rep.\,lr\,=\,$10^{-2}$ & imp.\,ep\,=\,1, rep.\,ep\,=\,3, imp.\,lr\,=\,$10^{-2}$, rep.\,lr\,=\,$10^{-2}$ & imp.\,ep\,=\,2, rep.\,ep\,=\,3, imp.\,lr\,=\,$10^{-3}$, rep.\,lr\,=\,$10^{-2}$ \\
 &  & SSD & $\alpha$\,=\,1, $\lambda$\,=\,10 & $\alpha$\,=\,5, $\lambda$\,=\,0.1 & $\alpha$\,=\,5, $\lambda$\,=\,0.1 \\
 &  & SCRUB & max\,=\,3, lr\,=\,$5{\times}10^{-3}$, ep\,=\,5 & max\,=\,2, lr\,=\,$5{\times}10^{-3}$, ep\,=\,5 & max\,=\,3, lr\,=\,$5{\times}10^{-3}$, ep\,=\,5 \\
 &  & SalUn & sp\,=\,0.3, lr\,=\,$10^{-2}$, ep\,=\,10 $\times$ & sp\,=\,0.3, lr\,=\,$10^{-2}$, ep\,=\,10 $\times$ & sp\,=\,0.9, lr\,=\,$10^{-2}$, ep\,=\,10 $\times$ \\
 &  & UniCLUN & lr\,=\,$10^{-2}$, ep\,=\,10, buf\,=\,250 $\times$ & lr\,=\,$10^{-2}$, ep\,=\,10, buf\,=\,250 $\times$ & lr\,=\,$10^{-2}$, ep\,=\,10, buf\,=\,250 $\times$ \\
 &  & LwU & $q_f$\,=\,0.99, margin\,=\,1, d.\,ep\,=\,10, d.\,lr\,=\,$10^{-3}$, align\,=\,off, mem\,=\,\texttt{m01\_s025\_c08} & $q_f$\,=\,0.99, margin\,=\,1, d.\,ep\,=\,30, d.\,lr\,=\,$10^{-3}$, align\,=\,off, mem\,=\,\texttt{m01\_s025\_c08} & $q_f$\,=\,0.99, margin\,=\,1, d.\,ep\,=\,10, d.\,lr\,=\,$10^{-3}$, align\,=\,off, mem\,=\,\texttt{m01\_s025\_c08} \\
\end{longtable}
}

{\setlength{\tabcolsep}{3pt}
\begin{longtable}{@{}lll>{\raggedright\arraybackslash}p{0.235\textwidth}>{\raggedright\arraybackslash}p{0.235\textwidth}>{\raggedright\arraybackslash}p{0.235\textwidth}@{}}
\caption{Selected hyperparameters on CIFAR-100.}\label{tab:hp-selected-cifar100}\\
\toprule
Backbone & Deletion & Method & Seed 42 & Seed 43 & Seed 44 \\
\midrule
\endfirsthead
\toprule
Backbone & Deletion & Method & Seed 42 & Seed 43 & Seed 44 \\
\midrule
\endhead
\bottomrule
\endfoot
CNN & class & Fine-tune & lr\,=\,$10^{-4}$, ep\,=\,10 & lr\,=\,$10^{-4}$, ep\,=\,10 & lr\,=\,$10^{-4}$, ep\,=\,10 \\
 &  & Bad Teacher & $T$\,=\,4, ep\,=\,10, lr\,=\,$10^{-3}$ & $T$\,=\,4, ep\,=\,5, lr\,=\,$10^{-2}$ $\times$ & $T$\,=\,4, ep\,=\,10, lr\,=\,$10^{-2}$ \\
 &  & UNSIR & imp.\,ep\,=\,1, rep.\,ep\,=\,3, imp.\,lr\,=\,$10^{-2}$, rep.\,lr\,=\,$10^{-2}$ & imp.\,ep\,=\,1, rep.\,ep\,=\,2, imp.\,lr\,=\,$10^{-3}$, rep.\,lr\,=\,$10^{-2}$ & imp.\,ep\,=\,1, rep.\,ep\,=\,2, imp.\,lr\,=\,$10^{-3}$, rep.\,lr\,=\,$10^{-2}$ \\
 &  & SSD & $\alpha$\,=\,100, $\lambda$\,=\,5 & $\alpha$\,=\,50, $\lambda$\,=\,0.5 & $\alpha$\,=\,1, $\lambda$\,=\,5 \\
 &  & SCRUB & max\,=\,3, lr\,=\,$5{\times}10^{-4}$, ep\,=\,5 & max\,=\,1, lr\,=\,$5{\times}10^{-4}$, ep\,=\,5 & max\,=\,3, lr\,=\,$5{\times}10^{-3}$, ep\,=\,5 \\
 &  & SalUn & sp\,=\,0.1, lr\,=\,$10^{-4}$, ep\,=\,5 $\times$ & sp\,=\,0.3, lr\,=\,$10^{-4}$, ep\,=\,3 $\times$ & sp\,=\,0.9, lr\,=\,$10^{-4}$, ep\,=\,3 $\times$ \\
 &  & UniCLUN & lr\,=\,$10^{-4}$, ep\,=\,3, buf\,=\,250 $\times$ & lr\,=\,$10^{-4}$, ep\,=\,5, buf\,=\,250 $\times$ & lr\,=\,$10^{-4}$, ep\,=\,3, buf\,=\,250 $\times$ \\
 &  & LwU & margin\,=\,3, d.\,ep\,=\,10, d.\,lr\,=\,$10^{-2}$, align\,=\,off, mem\,=\,\texttt{m01\_s025\_c08} & margin\,=\,3, d.\,ep\,=\,30, d.\,lr\,=\,$10^{-3}$, align\,=\,off, mem\,=\,\texttt{m01\_s025\_c08} & margin\,=\,3, d.\,ep\,=\,10, d.\,lr\,=\,$10^{-2}$, align\,=\,off, mem\,=\,\texttt{m01\_s025\_c08} \\
\midrule
 & subclass & Fine-tune & lr\,=\,$10^{-4}$, ep\,=\,10 & lr\,=\,$10^{-4}$, ep\,=\,5 & lr\,=\,$10^{-4}$, ep\,=\,3 \\
 &  & Bad Teacher & $T$\,=\,4, ep\,=\,10, lr\,=\,$10^{-3}$ & $T$\,=\,2, ep\,=\,10, lr\,=\,$10^{-2}$ & $T$\,=\,4, ep\,=\,10, lr\,=\,$10^{-2}$ $\times$ \\
 &  & UNSIR & imp.\,ep\,=\,1, rep.\,ep\,=\,3, imp.\,lr\,=\,$10^{-3}$, rep.\,lr\,=\,$10^{-2}$ & imp.\,ep\,=\,1, rep.\,ep\,=\,2, imp.\,lr\,=\,$10^{-2}$, rep.\,lr\,=\,$10^{-2}$ & imp.\,ep\,=\,2, rep.\,ep\,=\,1, imp.\,lr\,=\,$10^{-3}$, rep.\,lr\,=\,$10^{-2}$ \\
 &  & SSD & $\alpha$\,=\,50, $\lambda$\,=\,0.1 & $\alpha$\,=\,10, $\lambda$\,=\,5 & $\alpha$\,=\,1, $\lambda$\,=\,10 \\
 &  & SCRUB & max\,=\,2, lr\,=\,$5{\times}10^{-3}$, ep\,=\,10 & max\,=\,1, lr\,=\,$5{\times}10^{-4}$, ep\,=\,10 & max\,=\,2, lr\,=\,$5{\times}10^{-3}$, ep\,=\,5 \\
 &  & SalUn & sp\,=\,0.1, lr\,=\,$10^{-4}$, ep\,=\,3 $\times$ & sp\,=\,0.9, lr\,=\,$10^{-4}$, ep\,=\,3 $\times$ & sp\,=\,0.3, lr\,=\,$10^{-4}$, ep\,=\,3 $\times$ \\
 &  & UniCLUN & lr\,=\,$10^{-4}$, ep\,=\,5, buf\,=\,250 $\times$ & lr\,=\,$10^{-2}$, ep\,=\,5, buf\,=\,250 $\times$ & lr\,=\,$10^{-4}$, ep\,=\,3, buf\,=\,250 $\times$ \\
 &  & LwU & $q_f$\,=\,0.99, margin\,=\,1, d.\,ep\,=\,10, d.\,lr\,=\,$10^{-3}$, align\,=\,off, mem\,=\,\texttt{m01\_s025\_c08} & $q_f$\,=\,0.99, margin\,=\,1, d.\,ep\,=\,30, d.\,lr\,=\,$10^{-3}$, align\,=\,off, mem\,=\,\texttt{m10\_s100\_c08} & $q_f$\,=\,0.99, margin\,=\,1, d.\,ep\,=\,10, d.\,lr\,=\,$10^{-2}$, align\,=\,off, mem\,=\,\texttt{m01\_s025\_c08} \\
\midrule
 & instance & Fine-tune & lr\,=\,$10^{-4}$, ep\,=\,10 & lr\,=\,$10^{-4}$, ep\,=\,10 & lr\,=\,$10^{-4}$, ep\,=\,5 \\
 &  & Bad Teacher & $T$\,=\,4, ep\,=\,10, lr\,=\,$10^{-3}$ & $T$\,=\,1, ep\,=\,1, lr\,=\,$10^{-4}$ & $T$\,=\,2, ep\,=\,1, lr\,=\,$10^{-4}$ \\
 &  & UNSIR & imp.\,ep\,=\,2, rep.\,ep\,=\,3, imp.\,lr\,=\,$10^{-2}$, rep.\,lr\,=\,$10^{-2}$ & imp.\,ep\,=\,2, rep.\,ep\,=\,1, imp.\,lr\,=\,$10^{-3}$, rep.\,lr\,=\,$10^{-2}$ & imp.\,ep\,=\,2, rep.\,ep\,=\,1, imp.\,lr\,=\,$10^{-3}$, rep.\,lr\,=\,$10^{-2}$ \\
 &  & SSD & $\alpha$\,=\,5, $\lambda$\,=\,0.1 & $\alpha$\,=\,5, $\lambda$\,=\,0.5 & $\alpha$\,=\,5, $\lambda$\,=\,0.1 \\
 &  & SCRUB & max\,=\,2, lr\,=\,$5{\times}10^{-3}$, ep\,=\,10 & max\,=\,2, lr\,=\,$5{\times}10^{-3}$, ep\,=\,10 & max\,=\,1, lr\,=\,$5{\times}10^{-3}$, ep\,=\,5 \\
 &  & SalUn & sp\,=\,0.9, lr\,=\,$10^{-4}$, ep\,=\,3 $\times$ & sp\,=\,0.3, lr\,=\,$10^{-2}$, ep\,=\,10 $\times$ & sp\,=\,0.5, lr\,=\,$10^{-2}$, ep\,=\,10 $\times$ \\
 &  & UniCLUN & lr\,=\,$10^{-2}$, ep\,=\,10, buf\,=\,250 $\times$ & lr\,=\,$10^{-2}$, ep\,=\,10, buf\,=\,250 $\times$ & lr\,=\,$10^{-2}$, ep\,=\,10, buf\,=\,250 $\times$ \\
 &  & LwU & $q_f$\,=\,0.9, margin\,=\,3, d.\,ep\,=\,10, d.\,lr\,=\,$10^{-2}$, align\,=\,on, mem\,=\,\texttt{m01\_s025\_c08} & $q_f$\,=\,0.9, margin\,=\,3, d.\,ep\,=\,10, d.\,lr\,=\,$10^{-2}$, align\,=\,off, mem\,=\,\texttt{m01\_s025\_c08} & $q_f$\,=\,0.9, margin\,=\,3, d.\,ep\,=\,10, d.\,lr\,=\,$10^{-2}$, align\,=\,off, mem\,=\,\texttt{m10\_s100\_c08} \\
\midrule
ResNet-18 & class & Fine-tune & lr\,=\,$10^{-4}$, ep\,=\,10 & lr\,=\,$10^{-4}$, ep\,=\,10 & lr\,=\,$10^{-4}$, ep\,=\,10 \\
 &  & Bad Teacher & $T$\,=\,4, ep\,=\,10, lr\,=\,$10^{-2}$ & $T$\,=\,4, ep\,=\,10, lr\,=\,$10^{-2}$ & $T$\,=\,4, ep\,=\,3, lr\,=\,$10^{-2}$ \\
 &  & UNSIR & imp.\,ep\,=\,1, rep.\,ep\,=\,3, imp.\,lr\,=\,$10^{-2}$, rep.\,lr\,=\,$10^{-2}$ & imp.\,ep\,=\,1, rep.\,ep\,=\,2, imp.\,lr\,=\,$10^{-2}$, rep.\,lr\,=\,$10^{-2}$ & imp.\,ep\,=\,1, rep.\,ep\,=\,3, imp.\,lr\,=\,$10^{-2}$, rep.\,lr\,=\,$10^{-2}$ \\
 &  & SSD & $\alpha$\,=\,50, $\lambda$\,=\,0.1 & $\alpha$\,=\,10, $\lambda$\,=\,5 & $\alpha$\,=\,50, $\lambda$\,=\,5 \\
 &  & SCRUB & max\,=\,3, lr\,=\,$5{\times}10^{-3}$, ep\,=\,5 & max\,=\,3, lr\,=\,$5{\times}10^{-3}$, ep\,=\,5 & max\,=\,3, lr\,=\,$5{\times}10^{-3}$, ep\,=\,5 \\
 &  & SalUn & sp\,=\,0.3, lr\,=\,$10^{-4}$, ep\,=\,3 $\times$ & sp\,=\,0.9, lr\,=\,$10^{-4}$, ep\,=\,3 $\times$ & sp\,=\,0.9, lr\,=\,$10^{-4}$, ep\,=\,3 $\times$ \\
 &  & UniCLUN & lr\,=\,$10^{-4}$, ep\,=\,3, buf\,=\,250 $\times$ & lr\,=\,$10^{-4}$, ep\,=\,3, buf\,=\,250 $\times$ & lr\,=\,$10^{-4}$, ep\,=\,3, buf\,=\,250 $\times$ \\
 &  & LwU & margin\,=\,3, d.\,ep\,=\,30, d.\,lr\,=\,$10^{-3}$, align\,=\,off, mem\,=\,\texttt{m01\_s025\_c08} & margin\,=\,3, d.\,ep\,=\,30, d.\,lr\,=\,$10^{-2}$, align\,=\,off, mem\,=\,\texttt{m01\_s025\_c08} & margin\,=\,3, d.\,ep\,=\,30, d.\,lr\,=\,$10^{-2}$, align\,=\,off, mem\,=\,\texttt{m01\_s025\_c08} \\
\midrule
 & subclass & Fine-tune & lr\,=\,$10^{-4}$, ep\,=\,1 & lr\,=\,$10^{-4}$, ep\,=\,3 & lr\,=\,$10^{-4}$, ep\,=\,10 \\
 &  & Bad Teacher & $T$\,=\,4, ep\,=\,10, lr\,=\,$10^{-2}$ & $T$\,=\,4, ep\,=\,5, lr\,=\,$10^{-2}$ & $T$\,=\,4, ep\,=\,10, lr\,=\,$10^{-2}$ \\
 &  & UNSIR & imp.\,ep\,=\,2, rep.\,ep\,=\,2, imp.\,lr\,=\,$10^{-3}$, rep.\,lr\,=\,$10^{-2}$ & imp.\,ep\,=\,1, rep.\,ep\,=\,1, imp.\,lr\,=\,$10^{-2}$, rep.\,lr\,=\,$10^{-2}$ & imp.\,ep\,=\,1, rep.\,ep\,=\,3, imp.\,lr\,=\,$10^{-3}$, rep.\,lr\,=\,$10^{-2}$ \\
 &  & SSD & $\alpha$\,=\,50, $\lambda$\,=\,1 & $\alpha$\,=\,5, $\lambda$\,=\,0.1 & $\alpha$\,=\,50, $\lambda$\,=\,5 \\
 &  & SCRUB & max\,=\,2, lr\,=\,$5{\times}10^{-3}$, ep\,=\,10 & max\,=\,3, lr\,=\,$5{\times}10^{-3}$, ep\,=\,10 & max\,=\,3, lr\,=\,$10^{-3}$, ep\,=\,10 \\
 &  & SalUn & sp\,=\,0.9, lr\,=\,$10^{-4}$, ep\,=\,3 $\times$ & sp\,=\,0.3, lr\,=\,$10^{-4}$, ep\,=\,3 $\times$ & sp\,=\,0.1, lr\,=\,$10^{-4}$, ep\,=\,3 $\times$ \\
 &  & UniCLUN & lr\,=\,$10^{-4}$, ep\,=\,5, buf\,=\,250 $\times$ & lr\,=\,$10^{-4}$, ep\,=\,3, buf\,=\,250 $\times$ & lr\,=\,$10^{-4}$, ep\,=\,3, buf\,=\,250 $\times$ \\
 &  & LwU & $q_f$\,=\,0.99, margin\,=\,1, d.\,ep\,=\,10, d.\,lr\,=\,$10^{-3}$, align\,=\,off, mem\,=\,\texttt{m01\_s025\_c08} & $q_f$\,=\,0.99, margin\,=\,3, d.\,ep\,=\,30, d.\,lr\,=\,$10^{-2}$, align\,=\,off, mem\,=\,\texttt{m10\_s100\_c08} & $q_f$\,=\,0.99, margin\,=\,1, d.\,ep\,=\,10, d.\,lr\,=\,$10^{-3}$, align\,=\,off, mem\,=\,\texttt{m01\_s025\_c08} \\
\midrule
 & instance & Fine-tune & lr\,=\,$10^{-4}$, ep\,=\,3 & lr\,=\,$10^{-4}$, ep\,=\,1 & lr\,=\,$10^{-4}$, ep\,=\,10 \\
 &  & Bad Teacher & $T$\,=\,2, ep\,=\,1, lr\,=\,$10^{-4}$ & $T$\,=\,1, ep\,=\,1, lr\,=\,$10^{-4}$ & $T$\,=\,4, ep\,=\,1, lr\,=\,$10^{-4}$ \\
 &  & UNSIR & imp.\,ep\,=\,2, rep.\,ep\,=\,3, imp.\,lr\,=\,$10^{-2}$, rep.\,lr\,=\,$10^{-2}$ & imp.\,ep\,=\,1, rep.\,ep\,=\,3, imp.\,lr\,=\,$10^{-3}$, rep.\,lr\,=\,$10^{-3}$ & imp.\,ep\,=\,1, rep.\,ep\,=\,1, imp.\,lr\,=\,$10^{-3}$, rep.\,lr\,=\,$10^{-2}$ \\
 &  & SSD & $\alpha$\,=\,10, $\lambda$\,=\,0.5 & $\alpha$\,=\,5, $\lambda$\,=\,1 & $\alpha$\,=\,5, $\lambda$\,=\,0.1 \\
 &  & SCRUB & max\,=\,3, lr\,=\,$5{\times}10^{-3}$, ep\,=\,5 & max\,=\,2, lr\,=\,$5{\times}10^{-3}$, ep\,=\,5 & max\,=\,3, lr\,=\,$5{\times}10^{-3}$, ep\,=\,5 \\
 &  & SalUn & sp\,=\,0.9, lr\,=\,$10^{-2}$, ep\,=\,10 $\times$ & sp\,=\,0.9, lr\,=\,$10^{-2}$, ep\,=\,10 $\times$ & sp\,=\,0.7, lr\,=\,$10^{-2}$, ep\,=\,10 $\times$ \\
 &  & UniCLUN & lr\,=\,$10^{-2}$, ep\,=\,10, buf\,=\,250 $\times$ & lr\,=\,$10^{-2}$, ep\,=\,10, buf\,=\,250 $\times$ & lr\,=\,$10^{-2}$, ep\,=\,10, buf\,=\,250 $\times$ \\
 &  & LwU & $q_f$\,=\,0.99, margin\,=\,1, d.\,ep\,=\,10, d.\,lr\,=\,$10^{-3}$, align\,=\,off, mem\,=\,\texttt{m01\_s025\_c08} & $q_f$\,=\,0.9, margin\,=\,3, d.\,ep\,=\,30, d.\,lr\,=\,$10^{-2}$, align\,=\,off, mem\,=\,\texttt{m01\_s025\_c08} & $q_f$\,=\,0.99, margin\,=\,1, d.\,ep\,=\,10, d.\,lr\,=\,$10^{-3}$, align\,=\,on, mem\,=\,\texttt{m01\_s025\_c08} \\
\end{longtable}
}
\clearpage
\twocolumn


\begin{table*}[tp]
\centering
\caption{Hyperparameter search space and selected configurations on Seq-CIFAR-100.}
\label{tab:cifar100-hyperparameters}
\begin{adjustbox}{max width=\textwidth,max totalheight=0.8\textheight}
\begin{tabular}{lp{5cm}rp{3cm}rp{3cm}}
\toprule
& & \multicolumn{2}{c}{\textbf{TIL Scenario}} & \multicolumn{2}{c}{\textbf{CIL Scenario}} \\
\cmidrule(lr){3-4} \cmidrule(lr){5-6}
Method & Search space & Trial & Selected configuration & Trial & Selected configuration \\
\midrule
EWC & \texttt{lr} $\sim \text{LogUniform}(10^{-4}, 10^{-2})$ \newline \texttt{ewc\_gamma} $\sim \text{Uniform}(0.9, 1.0)$ \newline \texttt{ewc\_lambda} $\sim \text{LogUniform}(1, 10^4)$ & 4 & \texttt{lr=0.000706}, \texttt{ewc\_gamma=0.912}, \texttt{ewc\_lambda=2029.6} & 23 & \texttt{lr=0.00213}, \texttt{ewc\_gamma=0.970}, \texttt{ewc\_lambda=1.749} \\
\midrule
SI & \texttt{lr} $\sim \text{LogUniform}(10^{-4}, 10^{-2})$ \newline \texttt{si\_c} $\sim \text{LogUniform}(10^{-2}, 10)$ \newline \texttt{si\_xi} $\sim \text{LogUniform}(10^{-4}, 10^{-2})$ & 2 & \texttt{lr=0.000131}, \texttt{si\_c=0.333}, \texttt{si\_xi=0.000119} & 0 & \texttt{lr=0.000444}, \texttt{si\_c=0.0284}, \texttt{si\_xi=0.00200} \\
\midrule
LwF & Candidate list (30 configs) \newline \texttt{lr} $\in [1.1\text{e-}4, 8.4\text{e-}3]$ \newline \texttt{lwf\_lambda} $\in [0.14, 9.13]$ \newline \texttt{temperature} $\in \{1, 2, 3, 4\}$ & 13 & \texttt{lr=0.00173}, \texttt{lwf\_lambda=0.984}, \texttt{temperature=4} & 27 & \texttt{lr=0.000111}, \texttt{lwf\_lambda=0.838}, \texttt{temperature=2} \\
\midrule
WSN & Candidate list \newline \texttt{lr} $\in [1.2\text{e-}4, 7.9\text{e-}3]$ \newline \texttt{wsn\_density} $\in \{0.1, 0.3, 0.5, 0.7\}$ & 15 & \texttt{lr=0.00132}, \texttt{wsn\_density=0.3} & -- & -- \\
\midrule
PEC & \texttt{lr} $\sim \text{LogUniform}(10^{-4}, 10^{-2})$ & -- & -- & 3 & \texttt{lr=0.000140} \\
\midrule
SpaceNet & \texttt{lr} $\sim \text{LogUniform}(10^{-4}, 10^{-2})$ \newline \texttt{density\_factor} $\in \{0.5, 1.0, 1.5\}$ \newline \texttt{rewire\_fraction} $\in \{0.1, 0.2, 0.3\}$ & 8 & \texttt{lr=0.00147}, \texttt{density\_factor=0.5}, \texttt{rewire\_fraction=0.3} & 11 & \texttt{lr=0.000689}, \texttt{density\_factor=1.5}, \texttt{rewire\_fraction=0.1} \\
\midrule
NISPA & \texttt{lr} $\sim \text{LogUniform}(10^{-4}, 10^{-2})$ \newline \texttt{nispa\_prune\_perc} $\sim \text{Uniform}(70, 95)$ \newline \texttt{nispa\_recovery\_perc} $\sim \text{Uniform}(1, 5)$ & 10 & \texttt{lr=0.00190}, \texttt{prune\_perc=79.3}, \texttt{recovery\_perc=3.19} & 6 & \texttt{lr=0.00143}, \texttt{prune\_perc=79.9}, \texttt{recovery\_perc=4.91} \\
\midrule
SNV & Candidate list (23 of the 30 draws of \newline \texttt{lr} $\sim \text{LogUniform}(10^{-4}, 10^{-2})$, keeping \texttt{lr} $\geq 3\text{e-}4$) \newline \texttt{truncation} $\in \{0.05, 0.1, 0.2\}$ & 14 & \texttt{lr=0.00130}, \texttt{truncation=0.05}, \texttt{rot\_aux=0} & 14 & \texttt{lr=0.00130}, \texttt{truncation=0.05}, \texttt{rot\_aux=1.0}, \texttt{routing=rot+energy} \\
\midrule
UniCLUN & \texttt{lr} $\sim \text{LogUniform}(10^{-4}, 10^{-2})$ \newline \texttt{alpha1, alpha2, alpha3} $\sim \text{LogUniform}(0.1, 10)$ & 8 & \texttt{lr=0.00125}, \texttt{alpha1=0.134}, \texttt{alpha2=0.132}, \texttt{alpha3=0.258} & 10 & \texttt{lr=0.000806}, \texttt{alpha1=0.398}, \texttt{alpha2=3.879}, \texttt{alpha3=2.500} \\
\midrule
\textbf{LwU (ours)} & \texttt{lr} $\sim \text{LogUniform}(10^{-4}, 10^{-2})$ \newline \texttt{ce\_weight} $\sim \text{LogUniform}(0.1, 10)$ \newline \texttt{memory\_conf} $\in \{0.0, 0.8\}$ \newline \texttt{memory\_lr} $\in \{0.01, 0.05, 0.1\}$ \newline \texttt{memory\_scale} $\in \{0.25, 0.5, 1.0\}$ \newline \texttt{retain\_quant} $\in \{0.5, 0.7, 0.9, 0.95, 0.99\}$ \newline \texttt{sd\_temp} $\in \{1, 2, 3, 4\}$ & 15 & \texttt{lr=0.000625}, \texttt{ce\_weight=0.215}, \texttt{memory\_conf=0.8}, \texttt{memory\_lr=0.05}, \texttt{memory\_scale=1}, \texttt{retain\_quant=0.95}, \texttt{sd\_temp=1} & 20 & \texttt{lr=0.000819}, \texttt{ce\_weight=0.625}, \texttt{memory\_conf=0.8}, \texttt{memory\_lr=0.05}, \texttt{memory\_scale=0.25}, \texttt{retain\_quant=0.99}, \texttt{sd\_temp=4} \\
\bottomrule
\end{tabular}
\end{adjustbox}
\end{table*}

\begin{table*}[tp]
\centering
\scriptsize
\setlength{\tabcolsep}{3.2pt}
\caption{Hyperparameter search space and selected configurations on Seq-Tiny-ImageNet.}
\label{tab:tiny-hyperparameters}
\resizebox{\textwidth}{!}{%
\begin{tabular}{lp{5.5cm}rp{3.7cm}rp{3.7cm}}
\toprule
& & \multicolumn{2}{c}{\textbf{TIL Scenario}} & \multicolumn{2}{c}{\textbf{CIL Scenario}} \\
\cmidrule(lr){3-4}\cmidrule(lr){5-6}
Method & Search space & Trial & Selected configuration & Trial & Selected configuration \\
\midrule
EWC & \texttt{lr} $\sim\mathrm{LogUniform}(10^{-4},10^{-2})$ \newline \texttt{ewc\_lambda} $\sim\mathrm{LogUniform}(1,10^4)$ \newline \texttt{ewc\_gamma} $\sim\mathrm{Uniform}(0.9,1.0)$ & 30 & \texttt{lr=0.00196905}, \newline \texttt{ewc\_lambda=9383.91}, \newline \texttt{ewc\_gamma=0.982192} & 30 & \texttt{lr=0.000444289}, \newline \texttt{ewc\_lambda=4.01233}, \newline \texttt{ewc\_gamma=0.965093} \\
\midrule
SI & \texttt{lr} $\sim\mathrm{LogUniform}(10^{-4},10^{-2})$ \newline \texttt{si\_c} $\sim\mathrm{LogUniform}(0.01,10)$ \newline \texttt{si\_xi} $\sim\mathrm{LogUniform}(10^{-4},10^{-2})$ & 30 & \texttt{lr=0.00553151}, \newline \texttt{si\_c=0.0174479}, \newline \texttt{si\_xi=0.000791361} & 30 & \texttt{lr=0.000444289}, \newline \texttt{si\_c=0.0283496}, \newline \texttt{si\_xi=0.00200387} \\
\midrule
LwF & \texttt{lr} $\sim\mathrm{LogUniform}(10^{-4},10^{-2})$ \newline \texttt{lwf\_lambda} $\sim\mathrm{LogUniform}(0.1,10)$ \newline \texttt{temperature} $\in\{1,2,3,4\}$ & 30 & \texttt{lr=0.00112303}, \newline \texttt{lwf\_lambda=5.62698}, \newline \texttt{temperature=4} & 30 & \texttt{lr=0.00011095}, \newline \texttt{lwf\_lambda=0.838283}, \newline \texttt{temperature=2} \\
\midrule
WSN & \texttt{lr} $\sim\mathrm{LogUniform}(10^{-4},10^{-2})$ \newline \texttt{wsn\_density} $\in\{0.1,0.3,0.5,0.7\}$ & $15^{\dagger}$ & \texttt{lr=0.0013198195}, \newline \texttt{wsn\_density=0.3} & -- & -- \\
\midrule
PEC & \texttt{lr} $\sim \text{LogUniform}(10^{-4}, 10^{-2})$ & -- & -- & $3^{\dagger}$ & \texttt{lr=0.000140} \\
\midrule
SpaceNet & \texttt{lr} $\sim \text{LogUniform}(10^{-4}, 10^{-2})$ \newline \texttt{density\_factor} $\in \{0.5, 1.0, 1.5\}$ \newline \texttt{rewire\_fraction} $\in \{0.1, 0.2, 0.3\}$ & $8^{\dagger}$ & \texttt{lr=0.0014655269}, \newline \texttt{density\_factor=0.5}, \newline \texttt{rewire\_fraction=0.3} & $11^{\dagger}$ & \texttt{lr=0.0006890934}, \newline \texttt{density\_factor=1.5}, \newline \texttt{rewire\_fraction=0.1} \\
\midrule
NISPA & \texttt{lr} $\sim \text{LogUniform}(10^{-4}, 10^{-2})$ \newline \texttt{nispa\_prune\_perc} $\sim \text{Uniform}(70, 95)$ \newline \texttt{nispa\_recovery\_perc} $\sim \text{Uniform}(1, 5)$ & $10^{\dagger}$ & \texttt{lr=0.0018959503}, \newline \texttt{nispa\_prune\_perc=79.310}, \newline \texttt{nispa\_recovery\_perc=3.191} & $6^{\dagger}$ & \texttt{lr=0.0014262836}, \newline \texttt{nispa\_prune\_perc=79.917}, \newline \texttt{nispa\_recovery\_perc=4.905} \\
\midrule
SNV & \texttt{lr} $\sim\mathrm{LogUniform}(10^{-4},10^{-2})$ \newline \texttt{truncation} $\in\{0.05,0.1,0.2\}$ & $14^{\dagger}$ & \texttt{lr=0.0012981647}, \newline \texttt{truncation=0.05} & $14^{\dagger}$ & \texttt{lr=0.0012981647}, \newline \texttt{truncation=0.05} \\
\midrule
UniCLUN & \texttt{lr} $\sim \text{LogUniform}(10^{-4}, 10^{-2})$ \newline \texttt{alpha1, alpha2, alpha3} $\sim \text{LogUniform}(0.1, 10)$ & $8^{\dagger}$ & \texttt{lr=0.0012459165}, \newline \texttt{alpha1=0.1335}, \newline \texttt{alpha2=0.1316}, \newline \texttt{alpha3=0.2582} & $10^{\dagger}$ & \texttt{lr=0.0008060626}, \newline \texttt{alpha1=0.3977}, \newline \texttt{alpha2=3.8793}, \newline \texttt{alpha3=2.5003} \\
\midrule
\textbf{LwU (ours)} & \texttt{lr} $\sim\mathrm{LogUniform}(10^{-4},10^{-2})$ \newline \texttt{retain\_quantile} $\in\{0.5,0.7,0.9,0.95,0.99\}$ \newline \texttt{sd\_temperature} $\in\{1,2,3,4\}$ \newline \texttt{ce\_weight} $\sim\mathrm{LogUniform}(0.1,10)$ \newline \texttt{memory\_lr} $\in\{0.01,0.05,0.1\}$ \newline \texttt{memory\_scale} $\in\{0.25,0.5,1.0\}$ \newline \texttt{memory\_confidence} $\in\{0.0,0.8\}$ & 30 & \texttt{lr=0.000302919}, \newline \texttt{retain\_quantile=0.99}, \newline \texttt{sd\_temperature=4}, \newline \texttt{ce\_weight=0.131287}, \newline \texttt{memory\_lr=0.1}, \newline \texttt{memory\_scale=0.25}, \newline \texttt{memory\_confidence=0} & 30 & \texttt{lr=0.000105712}, \newline \texttt{retain\_quantile=0.99}, \newline \texttt{sd\_temperature=2}, \newline \texttt{ce\_weight=0.335346}, \newline \texttt{memory\_lr=0.01}, \newline \texttt{memory\_scale=0.25}, \newline \texttt{memory\_confidence=0.8} \\
\midrule
SGD (lower bound) & \texttt{lr} $\sim\mathrm{LogUniform}(10^{-4},10^{-2})$ & 30 & \texttt{lr=0.00785992} & 30 & \texttt{lr=0.000413965} \\
\bottomrule
\end{tabular}%
}
\end{table*}
\begin{table*}[tp]
\centering
\scriptsize
\setlength{\tabcolsep}{3.2pt}
\caption{Hyperparameter search space and selected configurations on RTI.}
\label{tab:rti-hyperparameters}
\resizebox{\textwidth}{!}{%
\begin{tabular}{lp{5cm}rp{3cm}rp{3cm}}
\toprule
& & \multicolumn{2}{c}{\textbf{TIL Scenario}} & \multicolumn{2}{c}{\textbf{CIL Scenario}} \\
\cmidrule(lr){3-4} \cmidrule(lr){5-6}
Method & Search space & Trial & Selected configuration & Trial & Selected configuration \\
\midrule
Joint (upper bound)
& \texttt{lr} $\sim \mathrm{LogUniform}(10^{-4},10^{-2})$
& 7 & \texttt{lr=0.00103}
& 9 & \texttt{lr=0.000737} \\
\midrule
EWC
& \texttt{lr} $\sim \mathrm{LogUniform}(10^{-4},10^{-2})$ \newline
\texttt{ewc\_gamma} $\sim \mathrm{Uniform}(0.9,1.0)$ \newline
\texttt{ewc\_lambda} $\sim \mathrm{LogUniform}(1,10^4)$
& 4 & \texttt{lr=0.000706}, \texttt{ewc\_gamma=0.912}, \texttt{ewc\_lambda=2030}
& 23 & \texttt{lr=0.00213}, \texttt{ewc\_gamma=0.970}, \texttt{ewc\_lambda=1.75} \\
\midrule
SI
& \texttt{lr} $\sim \mathrm{LogUniform}(10^{-4},10^{-2})$ \newline
\texttt{si\_c} $\sim \mathrm{LogUniform}(10^{-2},10)$ \newline
\texttt{si\_xi} $\sim \mathrm{LogUniform}(10^{-4},10^{-2})$
& 2 & \texttt{lr=0.000131}, \texttt{si\_c=0.333}, \texttt{si\_xi=0.000119}
& 27 & \texttt{lr=0.000171}, \texttt{si\_c=0.015}, \texttt{si\_xi=0.00344} \\
\midrule
LwF
& \texttt{lr} $\sim \mathrm{LogUniform}(10^{-4},10^{-2})$ \newline
\texttt{lwf\_lambda} $\sim \mathrm{LogUniform}(0.1,10)$ \newline
\texttt{temperature} $\in\{1,2,3,4\}$
& 11 & \texttt{lr=0.00146}, \texttt{lwf\_lambda=1.90}, \texttt{temperature=3}
& 23 & \texttt{lr=0.00144}, \texttt{lwf\_lambda=0.817}, \texttt{temperature=1} \\
\midrule
WSN
& \texttt{lr} $\sim \mathrm{LogUniform}(10^{-4},10^{-2})$ \newline
\texttt{wsn\_density} $\in\{0.1,0.3,0.5,0.7\}$
& 25 & \texttt{lr=0.00112}, \texttt{wsn\_density=0.5}
& -- & -- \\
\midrule
PEC
& \texttt{lr} $\sim \mathrm{LogUniform}(10^{-4},10^{-2})$
& -- & --
& 11 & \texttt{lr=0.000152} \\
\midrule
SpaceNet
& \texttt{lr} $\sim \mathrm{LogUniform}(10^{-4},10^{-2})$ \newline
\texttt{density\_factor} $\in\{0.5,1.0,1.5\}$ \newline
\texttt{rewire\_fraction} $\in\{0.1,0.2,0.3\}$
& 8 & \texttt{lr=0.00147}, \texttt{density\_factor=0.5}, \texttt{rewire\_fraction=0.3}
& 14 & \texttt{lr=0.00190}, \texttt{density\_factor=1.0}, \texttt{rewire\_fraction=0.1} \\
\midrule
NISPA
& \texttt{lr} $\sim \mathrm{LogUniform}(10^{-4},10^{-2})$ \newline
\texttt{nispa\_prune\_perc} $\sim \mathrm{Uniform}(70,95)$ \newline
\texttt{nispa\_recovery\_perc} $\sim \mathrm{Uniform}(1,5)$
& 3 & \texttt{lr=0.000737}, \texttt{nispa\_prune\_perc=71.7}, \texttt{nispa\_recovery\_perc}\newline\texttt{=1.36}
& 3 & \texttt{lr=0.000737}, \texttt{nispa\_prune\_perc=71.7}, \texttt{nispa\_recovery\_perc}\newline\texttt{=1.36} \\
\midrule
UniCLUN
& \texttt{lr} $\sim \mathrm{LogUniform}(10^{-4},10^{-2})$ \newline
\texttt{alpha1, alpha2, alpha3} \newline
$\sim \mathrm{LogUniform}(0.1,10)$ each
& 26 & \texttt{lr=0.0015074642}, \texttt{alpha1=0.335346084}, \texttt{alpha2=0.101903056}, \texttt{alpha3=0.688482653}
& 27 & \texttt{lr=0.000548}, \texttt{alpha1=1.36}, \texttt{alpha2=8.06}, \texttt{alpha3=2.40} \\
\midrule
\textbf{LwU (ours)}
& \texttt{lr} $\sim \mathrm{LogUniform}(10^{-4},10^{-2})$ \newline
\texttt{tau\_r} $\in\{0,10^{-8},10^{-7},3\!\times\!10^{-7},$ \newline
$10^{-6},3\!\times\!10^{-6},10^{-5},3\!\times\!10^{-5},$ \newline
$10^{-4},3\!\times\!10^{-4},10^{-3}\}$ \newline
\texttt{momentum\_decay} $\in\{0.9,0.95,0.99\}$ \newline
\texttt{adaptation\_rate} $\in\{0.001,0.01,0.05\}$
& 12 & \texttt{lr=0.00125}, \texttt{tau\_r=1e-8}, \texttt{momentum\_decay=0.99}, \texttt{adaptation\_rate=0.001}
& 18 & \texttt{lr=0.000790}, \texttt{tau\_r=3e-4}, \texttt{momentum\_decay=0.9}, \texttt{adaptation\_rate=0.001} \\
\midrule
SGD (lower bound)
& \texttt{lr} $\sim \mathrm{LogUniform}(10^{-4},10^{-2})$
& 10 & \texttt{lr=0.000138}
& 9 & \texttt{lr=0.000737} \\
\bottomrule
\end{tabular}%
}
\end{table*}
\end{document}